\documentclass[twoside]{article}

\pdfoutput=1
\usepackage{ecj,palatino,epsfig,latexsym,natbib}
\usepackage{amsmath}
\usepackage{amssymb}
\usepackage{mathtools}
\usepackage{url}
\usepackage[color=red,colorinlistoftodos]{todonotes}

\usepackage{tikz}
\usetikzlibrary{patterns}
\usetikzlibrary{arrows.meta}
\usetikzlibrary{math}
\usetikzlibrary{calc}
\usetikzlibrary{quotes}
\usetikzlibrary{angles}

\DeclareMathOperator*{\argmin}{arg\,min}

\usepackage{algorithm}
\usepackage{algpseudocode}
\algnewcommand\algorithmicswitch{\textbf{switch}}
\algnewcommand\algorithmiccase{\textbf{case}}
\algdef{SE}[SWITCH]{Switch}{EndSwitch}[1]{\algorithmicswitch\ #1\ \algorithmicdo}{\algorithmicend\ \algorithmicswitch}%
\algdef{SE}[CASE]{Case}{EndCase}[1]{\algorithmiccase\ #1}{\algorithmicend\ \algorithmiccase}%
\algtext*{EndCase}
\algrenewcommand\Return{\State \algorithmicreturn{}}
\algrenewcommand\textproc{}

\usepackage{cleveref}

\providecommand{\sppatemp}{}

\begin{document}

\newcommand{\sppatitle}
           {Analysis of the $(\mu/\mu_I,\lambda)$-CSA-ES
             with Repair by Projection Applied to a Conically
             Constrained Problem}
\newcommand{\sppaauthor}
           {\name{\bf Patrick Spettel} \hfill \addr{patrick.spettel@fhv.at}\\
             \addr{Research Center Process and Product Engineering,
               Vorarlberg University of Applied Sciences, Dornbirn,
               6850, Austria}
             \AND
             \name{\bf Hans-Georg Beyer} \hfill \addr{hans-georg.beyer@fhv.at}\\
             \addr{Department of Computer Science,
               Research Center Process and Product Engineering,
               Vorarlberg University of Applied Sciences, Dornbirn,
               6850, Austria}}

\ecjHeader{x}{x}{xxx-xxx}{201X}
          {CSA-ES with Repair on a Constrained Problem}
          {P. Spettel and H.-G. Beyer}
\title{\sppatitle}

\author{\sppaauthor}

{\maketitle}

\begin{abstract}
  Theoretical analyses of evolution strategies are indispensable for gaining
  a deep understanding of their inner workings. For constrained problems,
  rather simple problems are of interest in the current research. This work
  presents a theoretical analysis of a multi-recombinative evolution strategy
  with cumulative step size adaptation applied to a conically constrained
  linear optimization problem. The state of the strategy is modeled by
  random variables and a stochastic iterative mapping is introduced.
  For the analytical treatment, fluctuations are neglected and the mean
  value iterative system is considered. Non-linear difference equations
  are derived based on one-generation progress rates. Based on that,
  expressions for the steady state of the mean value iterative system
  are derived. By comparison with real algorithm runs, it is shown that for
  the considered assumptions, the theoretical derivations are able to
  predict the dynamics and the steady state values of the real runs.
\end{abstract}

\begin{keywords}
Evolution strategy,
constraint handling,
repair by projection,
cumulative step size adaptation,
conically constrained problem.
\end{keywords}

\section{Introduction}
\label{sec:intro}
Thorough theoretical investigations of evolution strategies (ESs) are necessary
for gaining a deep understanding of how they work. A lot of research
has been done for analyzing ESs applied to unconstrained problems.
For the constrained setting, there are still aspects for which
a deep theoretical understanding is missing. As a step
in that direction, this work theoretically analyzes a $(\mu/\mu_I,\lambda)$-ES
with cumulative step size adaptation (CSA)
applied to a conically constrained linear problem.

Regarding related work, a $(1, \lambda)$-ES with constraint handling
by discarding infeasible offspring has been analyzed
by~\citet{Arnold2011Behaviour} for a single linear constraint.
Repair by projection has been considered~\citep{Arnold2011Analysis}
and a comparison with repair by reflection and repair by
truncation has been performed by~\citet{Hellwig2016Comparison}.
Based on Lagrangian constraint handling,~\citet{Arnold2015Lagrangian}
presented a $(1+1)$-ES applied to a single
linear inequality constraint with the sphere model.
The one-generation behavior has been analyzed in that work.

A theoretical investigation based on Markov chains for
a multi-recombinative variant with Lagrangian constraint handling
has been presented by \cite{Atamna2016LagrangianES}. Investigation
of a single linear constraint in that work has been extended to multiple
linear constraints~\citep{Atamna2017LagrangianES}.

\citet{Arnold2013Behaviour} has considered a conically constrained
problem. In that work, a $(1, \lambda)$-ES is applied to the
problem by discarding infeasible offspring.
\citet{SpettelBeyer2018SigmaSaEsConeTCS} have considered
the same problem and have analyzed a $(1,\lambda)$-$\sigma$-Self-Adaptation
ES ($\sigma$SA-ES). It has been extended to the multi-recombinative
$(\mu/\mu_I,\lambda)$ variant~\citep{SpettelBeyer2018SigmaSaEsConeMulti}.
The contribution of this paper is the analysis
considering CSA instead of $\sigma$SA for the mutation strength
control mechanism.

The remainder of the paper is organized as follows.
\Cref{sec:problemalgorithm} introduces the optimization problem
under consideration and describes the algorithm that is analyzed.
\Cref{sec:theoreticalanalysis} concerns the theoretical analysis.
First, a mean value iterative system that models the dynamics of
the ES is derived in
\Cref{sec:theoreticalanalysis:subsec:iterativesystem}.
Second, steady state considerations are shown
in \Cref{sec:theoreticalanalysis:subsec:steadystate}.
For the theoretical considerations, plots comparing
them to results of real ES runs are presented for showing
the approximation quality. Finally, \Cref{sec:conclusions}
discusses the results and concludes the paper.

\section{Problem and Algorithm}
\label{sec:problemalgorithm}
Minimization of
\begin{equation}
    \label{sec:problemalgorithm:eq:optfun}
    f(\mathbf{x}) = x_1
\end{equation}
subject to constraints
\begin{align}
  x_1^2 - \xi \sum_{k=2}^N x_k^2 &\ge 0
  \label{sec:problemalgorithm:eq:coneconstraint}\\
  x_1 &\ge 0
\end{align}
is considered in this work
($\mathbf{x} = (x_1, \ldots, x_N)^T \in \mathbb{R}^N$ and $\xi > 0$).

\begin{figure}
    \centering
    \includegraphics{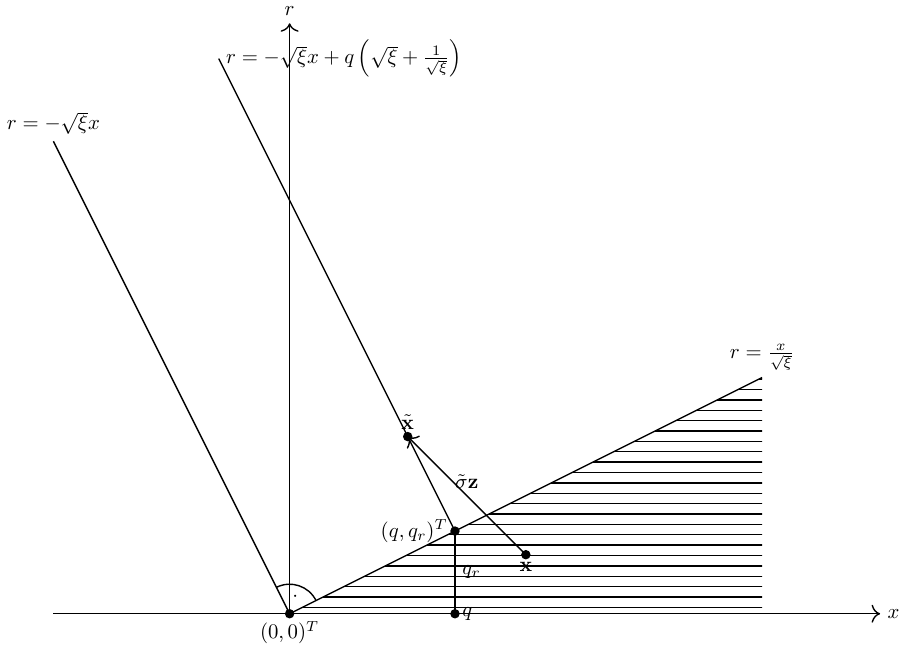}
    \caption[The conically constrained optimization problem in $N$ dimensions
             shown in the $(x,r)^T$-space.]
            {The conically constrained optimization problem in $N$ dimensions
              shown in the $(x,r)^T$-space. As shown in the picture, the
              offspring individual $\tilde{\mathbf{x}}$ is infeasible and
              therefore projected onto the cone boundary at $(q,q_r)^T$.
            }
    \label{sec:problemalgorithm:fig:conicconstraintnd}
\end{figure}

The state of an ES individual can be uniquely described in the
$(x,r)^T$-space. It consists of $x$, the distance from $0$ in
$x_1$-direction (cone axis), and $r$, the distance from the
cone axis. Because isotropic mutations are considered in
the ES, the coordinate system can be rotated (w.l.o.g.)
such that $(\tilde{x},\tilde{r})^T$ corresponds to
$(\tilde{x},\tilde{r}, 0, \ldots, 0)^T$ in the parameter space.
\Cref{sec:problemalgorithm:fig:conicconstraintnd} visualizes the problem.
The equation for the cone boundary is $r=\frac{x}{\sqrt{\xi}}$,
which follows from
\Cref{sec:problemalgorithm:eq:coneconstraint}.
The projection line can be derived
using the cone direction vector $\left(1, \frac{1}{\sqrt{\xi}}\right)^T$ and
its counterclockwise rotation by 90 degrees
$\left(-\frac{1}{\sqrt{\xi}}, 1\right)^T$ yielding
$r = -\sqrt{\xi}x + q\left(\sqrt{\xi} + \frac{1}{\sqrt{\xi}}\right)$.
The values $\mathbf{x}$ and $\tilde{\mathbf{x}}$ denote
a parental individual and an offspring individual, respectively.
The corresponding mutation is indicated as $\tilde{\sigma}\mathbf{z}$.
The values of $x$ and $r$ after projection are denoted by
$q$ and $q_r$, respectively.

The algorithm to be analyzed is a $(\mu/\mu_I,\lambda)$-CSA-ES
with repair by projection applied to the problem introduced
above. Its pseudo code is shown in \Cref{sec:problemalgorithm:alg:es}.
In the beginning, the parameters are initialized
(\Crefrange{sec:problemalgorithm:alg:es:init}
{sec:problemalgorithm:alg:es:ginit}).
In the generational loop, $\lambda$ offspring are created
(\Crefrange{sec:problemalgorithm:alg:es:forloopbegin}
{sec:problemalgorithm:alg:es:forloopend}).
Each offspring's parameter vector is sampled
from a multivariate normal distribution with mean $\mathbf{x}^{(g)}$ and
standard deviation $\sigma^{(g)}$
in \Cref{sec:problemalgorithm:alg:es:generateoffspringz,%
sec:problemalgorithm:alg:es:generateoffspring}.
If the generated offspring is infeasible
($\mathrm{isFeasible}(\mathbf{x}) =
x_1 \ge 0 \land x_1^2 - \xi \sum_{k=2}^N x_k^2 \ge 0$),
its parameter vector is projected onto the point on the boundary of the feasible
region that minimizes the Euclidean distance to the offspring point.
The corresponding mutation vector leading to this repaired point
is calculated back
(\Crefrange{sec:problemalgorithm:alg:es:feasibilitycheck}
{sec:problemalgorithm:alg:es:feasibilitycheckend}).
Projection means solving the optimization problem
\begin{align}
\label{sec:problemalgorithm:alg:coneprojection}
\begin{split}
  \mathbf{\hat{x}} = \argmin_{\mathbf{x'}}
    \lVert\mathbf{x'} - \mathbf{x}\rVert^2\\
  \text{s.t. }{x'_1}^2 - \xi \sum_{k=2}^N {x'_k}^2 \ge 0\\
  x'_1 \ge 0
\end{split}
\end{align}
where $\mathbf{x}$ is the individual to be projected.
The function
\begin{equation}
  \label{sec:problemalgorithm:alg:coneprojectionhelper}
  \mathbf{\hat{x}} = \text{projectOntoCone}(\mathbf{x})
\end{equation}
is introduced, which returns $\mathbf{\hat{x}}$ of the
problem~\eqref{sec:problemalgorithm:alg:coneprojection}.
Appendix A in the supplementary material
of~\cite{SpettelBeyer2018SigmaSaEsConeMulti}
shows a geometrical approach for deriving a closed-form solution
to the projection optimization
problem~\eqref{sec:problemalgorithm:alg:coneprojection}.
Given an infeasible individual $\mathbf{x}$, it reads
\begin{equation}
    \mathbf{\hat{x}} =
    \begin{cases}
      \frac{\xi}{\xi + 1}\left(x_1 +
      \frac{||\mathbf{r}||}
           {\sqrt{\xi}}\right)
           \left(1,
           \frac{x_2}{\sqrt{\xi}||\mathbf{r}||},
           \ldots,
           \frac{x_N}{\sqrt{\xi}||\mathbf{r}||}
           \right)^T
           &\text{ if }
           \sqrt{\frac{\xi}{\xi + 1}}\left(x_1 +
           \frac{||\mathbf{r}||}{\sqrt{\xi}}\right)
           > 0\\
    \mathbf{0} & \text{ otherwise}
    \end{cases}
\end{equation}
where $||\mathbf{r}|| = \sqrt{\sum_{k=2}^N x_k^2}$.
After possible repair, the offspring's fitness is determined
in \Cref{sec:problemalgorithm:alg:es:offspringf}.
The next generation's parental individual $\mathbf{x}^{(g + 1)}$
(\Cref{sec:problemalgorithm:alg:es:replaceparent})
and the
next generation's mutation strength $\sigma^{(g + 1)}$
(\Cref{sec:problemalgorithm:alg:es:replacesigma})
are computed next. The next generation's parental parameter vector is set to the
mean of the $\mu$ best (w.r.t. fitness) offspring parameter
vectors\footnote{Note that the order statistic notation $m;\lambda$
  is used to denote the $m$-th best (w.r.t. fitness) out of $\lambda$
  values. The notation $(\mathbf{x})_k$ is used to denote the $k$-th
  element of a vector $\mathbf{x}$. It is equivalent to writing
  $x_k$.}.
For the mutation strength update, first the cumulative $\mathbf{s}$-vector
is updated. The cumulation parameter $c$ determines the fading strength.
The mutation strength is then updated using this
$\mathbf{s}$-vector. The parameter $D$ acts as a damping factor.
If the squared length of the $\mathbf{s}$-vector
is smaller than $N$, the step size is decreased. Otherwise, the step
size is increased. Intuitively, this means that multiple correlated steps
allow a larger step size and vice versa.
The update of the generation counter
ends one iteration of the generation loop.
The values $x^{(g)}$, $r^{(g)}$, $q_l$, $\mathbf{q'}_l$,
$\langle q \rangle$, and $\langle q_r \rangle$ are only needed
in the theoretical analysis and can be removed in practical
applications of the ES.
They are indicated in the algorithm in
\Cref{sec:problemalgorithm:alg:es:x,%
sec:problemalgorithm:alg:es:r,%
sec:problemalgorithm:alg:es:q,%
sec:problemalgorithm:alg:es:qr,%
sec:problemalgorithm:alg:es:bestq,%
sec:problemalgorithm:alg:es:bestqr},
respectively.

\begin{algorithm}[H]
  \caption{Pseudo-code of the $(\mu/\mu_I,\lambda)$-CSA-ES with
    repair by projection applied to the conically constrained problem.}
  \label{sec:problemalgorithm:alg:es}
  \begin{algorithmic}[1]
    \State{Initialize $\mathbf{x}^{(0)}$, $\mathbf{s}^{(0)}$,
      $\sigma^{(0)}$, $c$, $D$, $\lambda$, $\mu$}
    \label{sec:problemalgorithm:alg:es:init}
    \State{$g \gets 0$}
    \label{sec:problemalgorithm:alg:es:ginit}
    \Repeat
    \label{sec:problemalgorithm:alg:es:repeatbegin}
    \State{$x^{(g)} = (\mathbf{x}^{(g)})_1$}
    \label{sec:problemalgorithm:alg:es:x}
    \State{$r^{(g)} = \sqrt{\sum_{k=2}^N(\mathbf{x}^{(g)})_k^2}$}
    \label{sec:problemalgorithm:alg:es:r}
    \For{$l \gets 1 \textbf{ to } \lambda$}
    \label{sec:problemalgorithm:alg:es:forloopbegin}
    \State{$\tilde{\mathbf{z}}_l \gets \mathcal{N}_l(\mathbf{0}, \mathbf{I})$}
    \Comment{sample from normal distribution}
    \label{sec:problemalgorithm:alg:es:generateoffspringz}
    \State{$\tilde{\mathbf{x}}_l \gets \mathbf{x}^{(g)} +
      \sigma^{(g)}\tilde{\mathbf{z}}_l$}
    \label{sec:problemalgorithm:alg:es:generateoffspring}
    \If{\textbf{not} \Call{isFeasible}{$\tilde{\mathbf{x}}_l$}}
    \label{sec:problemalgorithm:alg:es:feasibilitycheck}
    \State{$\tilde{\mathbf{x}}_l \gets$
      \Call{projectOntoCone}{$\tilde{\mathbf{x}}_l$}}
    \Comment{see
      \Cref{sec:problemalgorithm:alg:coneprojection,%
        sec:problemalgorithm:alg:coneprojectionhelper}}
    \label{sec:problemalgorithm:alg:es:projectionw}
    \State{$\tilde{\mathbf{z}}_l \gets
      (\tilde{\mathbf{x}}_l - \mathbf{x}^{(g)}) / \sigma^{(g)}$}
    \label{sec:problemalgorithm:alg:es:projectionz}
    \EndIf
    \label{sec:problemalgorithm:alg:es:feasibilitycheckend}
    \State{$\tilde{f}_l \gets f(\tilde{\mathbf{x}}_l)=(\tilde{\mathbf{x}}_l)_1$}
    \Comment{determine fitness of the offspring}
    \label{sec:problemalgorithm:alg:es:offspringf}
    \State{$q_l = (\tilde{\mathbf{x}}_l)_1$}
    \label{sec:problemalgorithm:alg:es:q}
    \State{$\mathbf{q'}_l = \tilde{\mathbf{x}}_l$}
    \label{sec:problemalgorithm:alg:es:qr}
    \EndFor
    \label{sec:problemalgorithm:alg:es:forloopend}
    \State{Sort offspring according to $\tilde{f}_l$ in ascending order}
    \label{sec:problemalgorithm:alg:es:sortoffspring}
    \State{$\mathbf{x}^{(g + 1)} \gets
      \frac{1}{\mu}\sum_{m=1}^\mu\tilde{\mathbf{x}}_{m;\lambda}$}
    \Comment{compute centroid of the $\mu$ best offspring}
    \label{sec:problemalgorithm:alg:es:replaceparent}
    \State{$\mathbf{s}^{(g + 1)} \gets (1-c)\mathbf{s}^{(g)} +
      \sqrt{\mu c(2-c)} \underbrace{\left(\frac{1}{\mu}\sum_{m=1}^\mu
        \tilde{\mathbf{z}}_{m;\lambda}\right)}_{\langle
        \tilde{\mathbf{z}} \rangle}$}
    \Comment{compute next $\mathbf{s}$}
    \label{sec:problemalgorithm:alg:es:replaces}
    \State{$\sigma^{(g + 1)} \gets
      \sigma^{(g)}\exp\left(\frac{||\mathbf{s}||^2 - N}{2DN}\right)$}
    \Comment{compute next $\sigma$}
    \label{sec:problemalgorithm:alg:es:replacesigma}
    \State{$\langle q \rangle = (\mathbf{x}^{(g+1)})_1$}
    \label{sec:problemalgorithm:alg:es:bestq}
    \State{$\langle q_r \rangle = \sqrt{\sum_{k=2}^N(\mathbf{x}^{(g+1)})_k^2}$}
    \label{sec:problemalgorithm:alg:es:bestqr}
    \State{$g \gets g + 1$}
    \label{sec:problemalgorithm:alg:es:gupdate}
    \Until{termination criteria are met}
    \label{sec:problemalgorithm:alg:es:repeatend}
  \end{algorithmic}
\end{algorithm}

\begin{figure}
  \centering
  \begin{tabular}{@{\hspace{-0.0\textwidth}}c@{\hspace{-0.0\textwidth}}c}
    \includegraphics[width=0.46\textwidth]{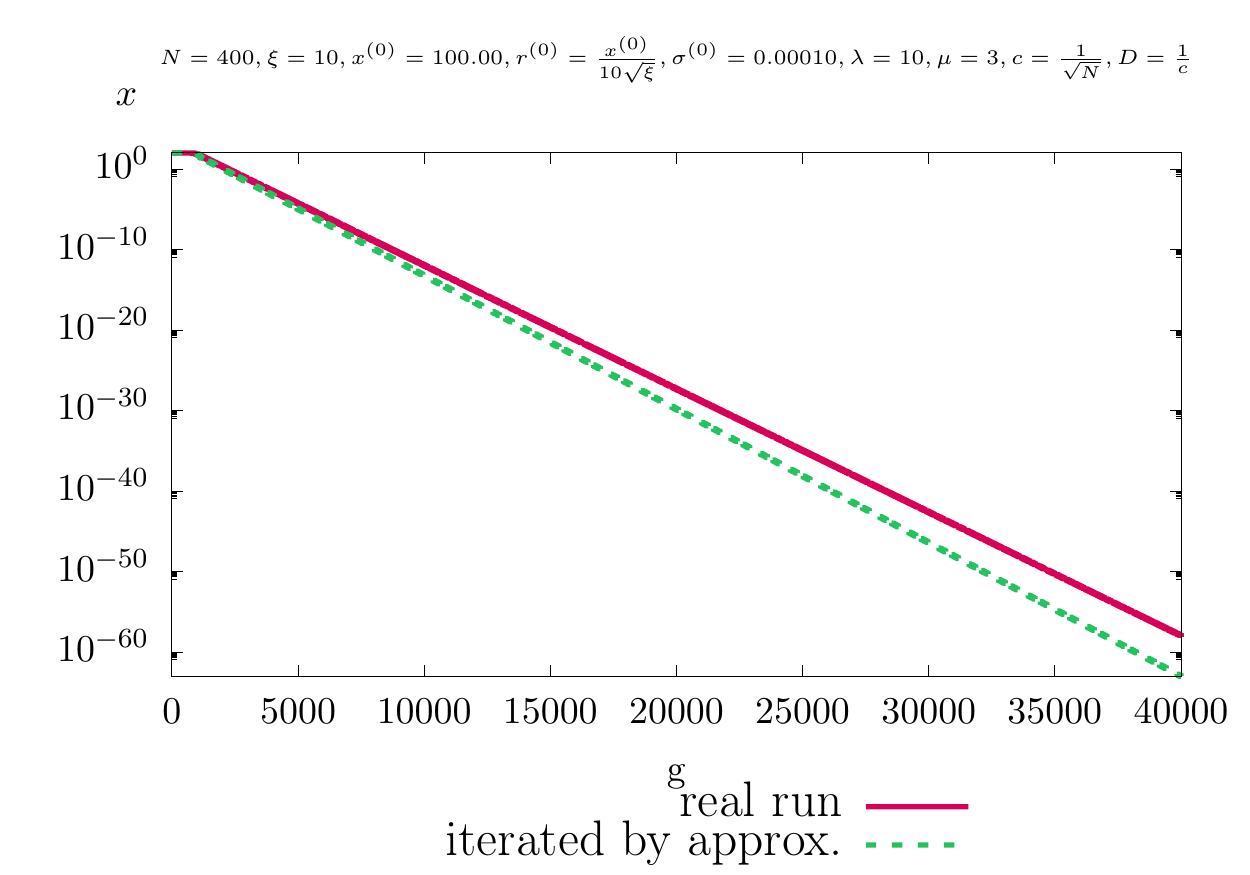}
    \includegraphics[width=0.46\textwidth]{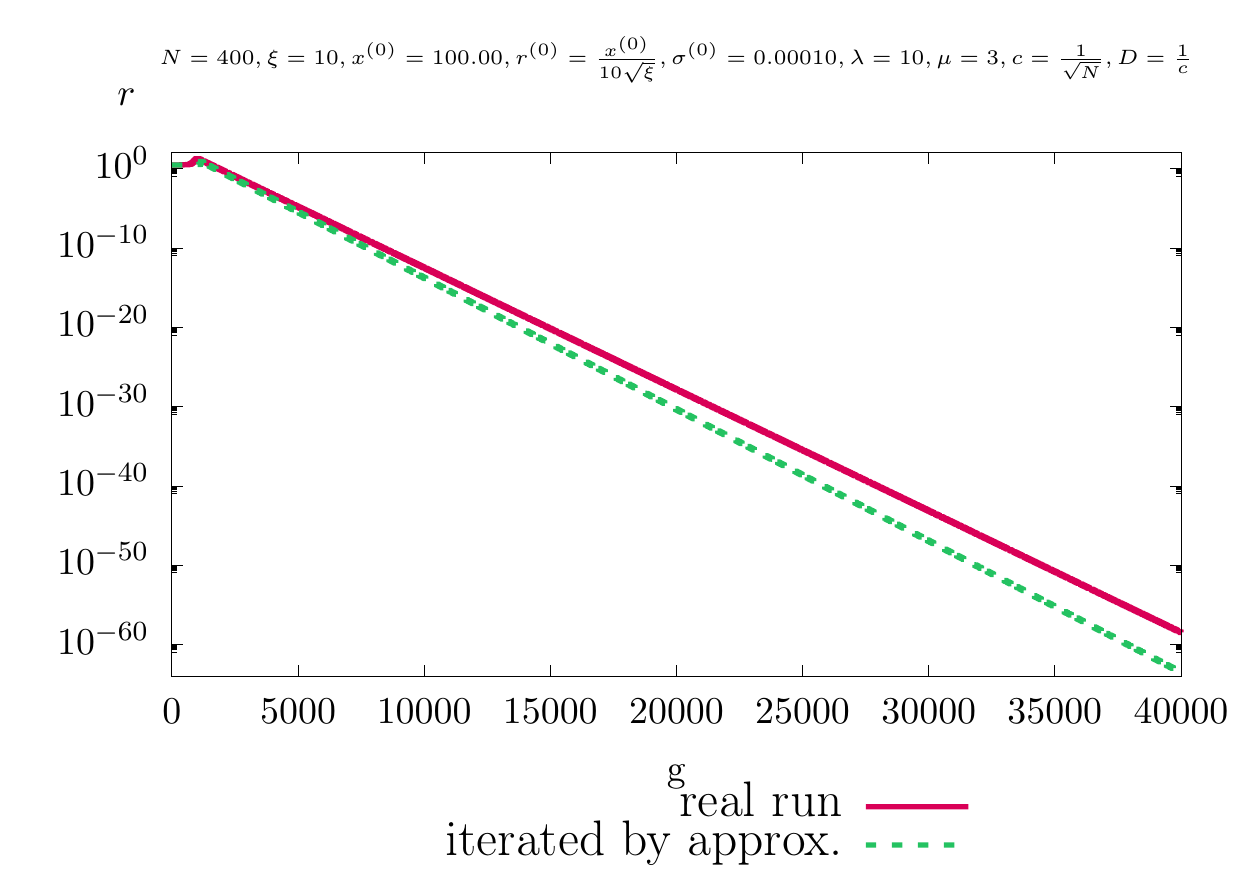}
  \end{tabular}
  \caption{Real $(3/3_I, 10)$-CSA-ES run mean value dynamics
    (solid line) in comparison
    to the iteration of the closed-form (approximate)
    iterative system (dotted line).}
  \label{sec:problemalgorithm:fig:dynamicsexample}
\end{figure}

\Cref{sec:problemalgorithm:fig:dynamicsexample}
shows an example of the $x$- and $r$-dynamics of
\Cref{sec:problemalgorithm:alg:es} (solid line)
in comparison with results of the closed-form approximate
iterative system (dotted line) that is derived in the
sections that follow. As one can see, the real
dynamics are predicted satisfactorily by the theoretical
considerations for the case shown.

\section{Theoretical Analysis}
\label{sec:theoreticalanalysis}
To completely describe the state of the ES, the random variables
$\sigma$, $\mathbf{s}$, and the squared length
$||\mathbf{s}||^2$ need to be modeled in addition
to the variables for the position in the parameter space,
$x$ and $r$. The random vector $\mathbf{s}$ is decomposed
into its magnitude along the cone axis $s_1^{(g)}$ and its magnitude
in direction of the parental individual's $2..N$ components
\begin{equation}
  \label{sec:theoreticalanalysis:eq:sodotdefinition}
  s_{\odot}^{(g)} :=
  \frac{1}{r^{(g)}}\sum_{k=2}^N(\mathbf{x}^{(g)})_k(\mathbf{s}^{(g)})_k.
\end{equation}
This leads to a stochastic iterative system of the form
\begin{equation}
  \left(\begin{array}{c}
    x^{(g+1)}\\
    r^{(g+1)}\\
    s_1^{(g+1)}\\
    s_\odot^{(g+1)}\\
    ||\mathbf{s}^{(g+1)}||^2\\
    \sigma^{(g+1)}
  \end{array}\right)
  \gets
  \left(\begin{array}{c}
    x^{(g)}\\
    r^{(g)}\\
    s_1^{(g)}\\
    s_\odot^{(g)}\\
    ||\mathbf{s}^{(g)}||^2\\
    \sigma^{(g)}
  \end{array}\right).
\end{equation}

\subsection{Derivation of a Mean Value Iterative System for Modeling the
  Dynamics of the ES}
\label{sec:theoreticalanalysis:subsec:iterativesystem}
Similar to the analysis in Section~IV
of~\cite{SpettelBeyer2018SigmaSaEsConeMulti},
fluctuation terms are neglected and
deterministic evolution equations under asymptotic assumptions
are derived. This allows predicting the mean value dynamics of
the ES. To make the distinction between the random variable
and its mean value in the iterative system clear,
$\overline{z} := \mathrm{E}[z]$ is used to denote the expected
value of a random variate $z$.
Thus, the mean value iterative system is represented as
\begin{equation}
  \label{sec:theoreticalanalysis:eq:meanvalueiterativesystemintro}
  \left(\begin{array}{c}
    \overline{x^{(g+1)}}\\
    \overline{r^{(g+1)}}\\
    \overline{s_1^{(g+1)}}\\
    \overline{s_\odot^{(g+1)}}\\
    \overline{||\mathbf{s}^{(g+1)}||^2}\\
    \overline{\sigma^{(g+1)}}
  \end{array}\right)
  \gets
  \left(\begin{array}{c}
    \overline{x^{(g)}}\\
    \overline{r^{(g)}}\\
    \overline{s_1^{(g)}}\\
    \overline{s_\odot^{(g)}}\\
    \overline{||\mathbf{s}^{(g)}||^2}\\
    \overline{\sigma^{(g)}}
  \end{array}\right).
\end{equation}
This section presents derivations of difference equations for
the system~\eqref{sec:theoreticalanalysis:eq:meanvalueiterativesystemintro}.
In \Cref{sec:theoreticalanalysis:subsec:iterativesystem:diffeqxr},
difference equations are presented for expressing
$\overline{x^{(g+1)}}$ with $\overline{x^{(g)}}$ and
$\overline{r^{(g+1)}}$ with $\overline{r^{(g)}}$
by using the respective local progress rates.
\Cref{sec:theoreticalanalysis:subsec:iterativesystem:diffeqs1},
\Cref{sec:theoreticalanalysis:subsec:iterativesystem:diffeqsodot},
and
\Cref{sec:theoreticalanalysis:subsec:iterativesystem:diffeqssquaredlength}
deal with the derivation of difference equations for
$\overline{s_1}$, $\overline{s_\odot}$, and $\overline{||\mathbf{s}^{(g)}||^2}$,
respectively.
They are derived from the corresponding steps of
\Cref{sec:problemalgorithm:alg:es}
and they also make use of the local progress rates.
Finally, the difference equation for
$\overline{\sigma}$ is stated in
\Cref{sec:theoreticalanalysis:subsec:iterativesystem:diffeqsigma},
the derived system of equations is summarized, and it is
compared to real ES runs in
\Cref{sec:theoreticalanalysis:subsec:evolutionequationssummary}.

\subsubsection{Derivation of Mean Value Difference Equations for $x$
  and $r$}
\label{sec:theoreticalanalysis:subsec:iterativesystem:diffeqxr}
The starting points for the derivation of mean value difference equations
for $x$ and $r$ are the progress rates in $x$ and $r$ direction.
Their definitions read
\begin{equation}
  \label{sec:theoreticalanalysis:eq:varphix}
  \varphi_{x}(\overline{x^{(g)}},
  \overline{r^{(g)}}, \overline{\sigma^{(g)}}) :=
  \mathrm{E}[\overline{x^{(g)}} - x^{(g + 1)}\,|\,
    \overline{x^{(g)}}, \overline{r^{(g)}}, \overline{\sigma^{(g)}}]
\end{equation}
\begin{equation}
  \label{sec:theoreticalanalysis:eq:varphir}
  \varphi_{r}(\overline{x^{(g)}},
  \overline{r^{(g)}}, \overline{\sigma^{(g)}}) :=
  \mathrm{E}[\overline{r^{(g)}} - r^{(g + 1)}\,|\,
    \overline{x^{(g)}}, \overline{r^{(g)}}, \overline{\sigma^{(g)}}].
\end{equation}
They describe the one-generation expected change in
the parameter space. The normalizations
\begin{equation}
  \label{sec:theoreticalanalysis:eq:varphixnormalized}
  \varphi_{x}^*(\cdot) := \frac{N\varphi_{x}(\cdot)}{\overline{x^{(g)}}},
\end{equation}
\begin{equation}
  \label{sec:theoreticalanalysis:eq:varphirnormalized}
  \varphi_{r}^*(\cdot) := \frac{N\varphi_{r}(\cdot)}{\overline{r^{(g)}}},
\end{equation}
and
\begin{equation}
  \label{sec:theoreticalanalysis:eq:sigmanormalized}
  \sigma^* := \frac{N\sigma}{\overline{r^{(g)}}}
\end{equation}
are introduced in order to have quantities that
are independent of the position in the search space.
Using
\Cref{sec:theoreticalanalysis:eq:varphix} with
\Cref{sec:theoreticalanalysis:eq:varphixnormalized}
and
\Cref{sec:theoreticalanalysis:eq:varphir} with
\Cref{sec:theoreticalanalysis:eq:varphirnormalized},
the equations
\begin{align}
  \overline{x^{(g + 1)}} &= \overline{x^{(g)}} -
  \frac{\overline{x^{(g)}}{\varphi^{(g)}_{x}}^*}{N}
  =\overline{x^{(g)}}\left(1-\frac{{\varphi^{(g)}_{x}}^*}{N}\right)
  \label{sec:theoreticalanalysis:eq:meanvalueiterativesystemx}\\
  \overline{r^{(g + 1)}} &= \overline{r^{(g)}} -
  \frac{\overline{r^{(g)}}{\varphi^{(g)}_r}^*}{N}
  =\overline{r^{(g)}}\left(1-\frac{{\varphi^{(g)}_r}^*}{N}\right)
  \label{sec:theoreticalanalysis:eq:meanvalueiterativesystemr}
\end{align}
follow.
Approximations for
${\varphi^{(g)}_{x}}^*$ and ${\varphi^{(g)}_{r}}^*$ have already been
derived by~\citet[Equations (37) and (38)]{SpettelBeyer2018SigmaSaEsConeMulti}.
In that work, expressions for ${\varphi^{(g)}_{x}}^*$
and ${\varphi^{(g)}_{r}}^*$ have been derived
under the asymptotic assumptions of sufficiently large values of $\xi$
and $N$. In those derivations, two cases have been distinguished.
If one considers the ES being far from the cone boundary, offspring
are feasible with overwhelming probability. The opposite case
of being in the vicinity of the cone boundary results in infeasible
offspring almost surely. These observations allow simplifications
for the former case because the projection can be ignored.
Both cases are combined into single equations by weighting
the feasible and infeasible cases with an approximation for the
offspring feasibility and offspring infeasibility probability,
respectively. The $r$-distribution in those derivations
has been approximated by a normal distribution
$\mathcal{N}(\bar{r}, \sigma_r^2)$ where
\begin{equation}
  \label{sec:theoreticalanalysis:eq:approximatedrmean}
  \bar{r}=
      {\overline{r^{(g)}}} \sqrt{1 +
        \frac{\overline{{{\sigma^{(g)}}^*}}^2}{N}\left(1 - \frac{1}{N}\right)}
\end{equation}
and
\begin{equation}
  \label{sec:theoreticalanalysis:eq:approximatedrstd}
  \sigma_r=
        {\overline{r^{(g)}}}\frac{\overline{{\sigma^{(g)}}^*}}{N}
        \sqrt{\frac{1 + \frac{{\overline{{\sigma^{(g)}}^*}}^2}
            {2N}\left(1 - \frac{1}{N}\right)}
          {1 +
            \frac{{\overline{{\sigma^{(g)}}^*}}^2}{N}
            \left(1 - \frac{1}{N}\right)}}
\end{equation}
(it is referred to Appendix B in the supplementary material
of~\cite{SpettelBeyer2018SigmaSaEsConeMulti} for the detailed derivation).
The results that build the basis for the following CSA analysis are
briefly recapped here.\footnote{In the further considerations, the
symbols ``$\simeq$'' and ``$\approx$'' are used.
Expressions in the form of $\mathrm{lhs} \simeq \mathrm{rhs}$ denote
that $\mathrm{lhs}$ is asymptotically equal to $\mathrm{rhs}$
for given asymptotical assumptions (e.g.
$N \rightarrow \infty$). The particular assumptions are
stated explicitly for every use of ``$\simeq$''. That is,
in the limit case of the given assumptions, $\mathrm{lhs}$
is equal to $\mathrm{rhs}$.
The form $\mathrm{lhs} \approx \mathrm{rhs}$ is used for cases
where $\mathrm{rhs}$ is an approximation for $\mathrm{lhs}$
with given assumptions that are not of asymptotical nature.
In this sense, ``$\approx$'' is weaker than ``$\simeq$''.}
The expression for ${\varphi^{(g)}_{x}}^*$ has been derived as
\begin{equation}
  \label{sec:theoreticalanalysis:eq:varphixnormalizedcombined}
  \begin{multlined}
  {\varphi^{(g)}_{x}}^* \approx
  P_{\text{feas}}(\overline{x^{(g)}}, \overline{r^{(g)}},
  \overline{\sigma^{(g)}})
  \left[\frac{\overline{r^{(g)}}}{\overline{x^{(g)}}}
    \overline{{\sigma^{(g)}}^*}c_{\mu/\mu,\lambda}\right]
  + [1 - P_{\text{feas}}(\overline{x^{(g)}}, \overline{r^{(g)}},
    \overline{\sigma^{(g)}})]\\\times
  \underbrace{
  \left[\frac{N}{1+\xi}
  \left(1-\frac{\sqrt{\xi}\overline{r^{(g)}}}{\overline{x^{(g)}}}
    \sqrt{1+\frac{{\overline{{\sigma^{(g)}}^*}}^2}{N}}\right)
    +\frac{\sqrt{\xi}}{1+\xi}
    \frac{\sqrt{\xi}\overline{r^{(g)}}}{\overline{x^{(g)}}}
    \overline{{\sigma^{(g)}}^*} c_{\mu/\mu,\lambda}\sqrt{
      1+
      \frac{1}{\xi}
      \frac{1+\frac{{\overline{{\sigma^{(g)}}^*}}^2}{2N}}
           {1+\frac{{\overline{{\sigma^{(g)}}^*}}^2}{N}}}\right]}_
  {=: {\varphi_x^*}_{\text{infeas}}^{(g)}}
  \end{multlined}
\end{equation}
and the one for ${\varphi^{(g)}_{r}}^*$ reads
\begin{align}
  &\begin{multlined}
  {\varphi^{(g)}_r}^*
  \approx
  P_{\text{feas}}(\overline{x^{(g)}}, \overline{r^{(g)}},
  \overline{\sigma^{(g)}})
  N\left(1-\sqrt{1 + \frac{{\overline{{\sigma^{(g)}}^*}}^2}{\mu N}}\right)\\
  + [1 - P_{\text{feas}}(\overline{x^{(g)}}, \overline{r^{(g)}},
    \overline{\sigma^{(g)}})]
  N\left(
  1 -
  \frac
  {\overline{x^{(g)}}}
  {\sqrt{\xi}\overline{r^{(g)}}}
  \left(1-\frac{{\varphi_x^*}_{\text{infeas}}^{(g)}}{N}\right)
  \sqrt{\frac{1+\frac{{\overline{{\sigma^{(g)}}^*}}^2}{\mu N}}
             {1+\frac{{\overline{{\sigma^{(g)}}^*}}^2}{N}}}
  \right).
  \end{multlined}
  \label{sec:theoreticalanalysis:eq:varphirnormalizedcombinedmaintext}
\end{align}
The approximate offspring feasibility probability writes
\begin{align}
  P_{\text{feas}}(\overline{x^{(g)}}, \overline{r^{(g)}},
  \overline{\sigma^{(g)}})
  \simeq\Phi\left[
    \frac{1}{\overline{\sigma^{(g)}}}
    \left(
    \frac{\overline{x^{(g)}}}{\sqrt{\xi}}-\bar{r}
    \right)
    \right]
  \label{sec:theoreticalanalysis:eq:Pfeasapprox1}
\end{align}
where $\Phi(\cdot)$ denotes the cumulative distribution
function of the standard normal distribution.
${\varphi_x^*}_{\text{infeas}}^{(g)}$ denotes the infeasible part of
\Cref{sec:theoreticalanalysis:eq:varphixnormalizedcombined}. The constant
$c_{\mu/\mu,\lambda}$ is a so-called progress coefficient. A definition
is given in~\cite[Eq. 6.102, p. 247]{Beyer2001}. It reads
\begin{equation}
  \label{sec:theoreticalanalysis:eq:cmumulambda}
  c_{\mu/\mu,\lambda} :=
  \frac{\lambda-\mu}{2\pi}\binom{\lambda}{\mu}
  \int_{t=-\infty}^{t=\infty}
  e^{-t^2}
  [\Phi(t)]^{\lambda-\mu-1}[1-\Phi(t)]^{\mu-1}
  \,\mathrm{d}t.
\end{equation}

\subsubsection{Derivation of a Mean Value Difference Equation for $s_1$}
\label{sec:theoreticalanalysis:subsec:iterativesystem:diffeqs1}
For $s_1$, a mean value difference equation can be derived using
the update rule from \Cref{sec:problemalgorithm:alg:es:replaces} of
\Cref{sec:problemalgorithm:alg:es}.
Computation of the expected value with
$\overline{s_1^{(g+1)}}:=\mathrm{E}[s_1^{(g+1)}]$ directly yields
\begin{align}
  \overline{s_1^{(g+1)}} &= (1-c)\overline{s_1^{(g)}} +
  \sqrt{\mu c(2-c)}\mathrm{E}[({\langle \tilde{\mathbf{z}}^{(g)} \rangle})_1].
\end{align}
$\mathrm{E}[(\langle \tilde{\mathbf{z}}^{(g)} \rangle)_1]$
can be expressed with the progress
rate in $x$-direction $\varphi_x$.
From the definition of the $x$ progress rate,
\begin{align}
  \varphi_x^{(g)}
  = \mathrm{E}[x^{(g)} - x^{(g+1)}]
  = \mathrm{E}[x^{(g)} - (x^{(g)}+
    \sigma^{(g)}(\langle \tilde{\mathbf{z}}^{(g)} \rangle)_1)]
  = -\overline{\sigma^{(g)}}\mathrm{E}
         [(\langle \tilde{\mathbf{z}}^{(g)} \rangle)_1]
\end{align}
follows.
Therefore,
\begin{align}
  \mathrm{E}[(\langle \tilde{\mathbf{z}}^{(g)} \rangle)_1]
  &= -\frac{\varphi_x^{(g)}}{\overline{\sigma^{(g)}}}
  \label{sec:theoreticalanalysis:eq:z1}
\end{align}
holds. Using
\Cref{sec:theoreticalanalysis:eq:z1}
and
\Cref{sec:theoreticalanalysis:eq:sigmanormalized},
\begin{align}
  \overline{s_1^{(g+1)}}
  &=(1-c)\overline{s_1^{(g)}} +
  \sqrt{\mu c(2-c)}
  \left(-\frac{N\varphi_x^{(g)}}
       {\overline{{\sigma^{(g)}}^*}\,\overline{r^{(g)}}}\right)
\end{align}
follows.

\subsubsection{Derivation of a Mean Value Difference Equation for $s_\odot$}
\label{sec:theoreticalanalysis:subsec:iterativesystem:diffeqsodot}
For $s_\odot$, a mean value difference equation can be derived using
the update rule from \Cref{sec:problemalgorithm:alg:es:replaces} of
\Cref{sec:problemalgorithm:alg:es} and considering
\Cref{sec:theoreticalanalysis:eq:sodotdefinition}.
To begin with,
\begin{align}
  &\begin{multlined}
     s_{\odot}^{(g+1)}=
     \frac{1}{r^{(g+1)}}
     \sum_{k=2}^N(\mathbf{x}^{(g+1)})_k(\mathbf{s}^{(g+1)})_k
   \end{multlined}\\
  &\begin{multlined}
     \phantom{s_{\odot}^{(g+1)}}=
     \frac{1}{r^{(g+1)}}
     \sum_{k=2}^N
     \left[(\mathbf{x}^{(g)})_k+
       \sigma^{(g)}({\langle \tilde{\mathbf{z}}^{(g)} \rangle})_k\right]
     \left[(1-c)(\mathbf{s}^{(g)})_k + \sqrt{\mu c(2-c)}
       ({\langle \tilde{\mathbf{z}}^{(g)} \rangle})_k\right]
   \end{multlined}\\
  &\begin{multlined}
     \phantom{s_{\odot}^{(g+1)}}=
     \frac{1}{r^{(g+1)}}
     \sum_{k=2}^N
     (1-c)\left[(\mathbf{x}^{(g)})_k(\mathbf{s}^{(g)})_k
       +\sigma^{(g)}({\langle \tilde{\mathbf{z}}^{(g)} \rangle})_k
       (\mathbf{s}^{(g)})_k\right]\\
     \hspace{2cm}+ \frac{1}{r^{(g+1)}}\sum_{k=2}^N
     \sqrt{\mu c(2-c)}\left[(\mathbf{x}^{(g)})_k
       ({\langle \tilde{\mathbf{z}}^{(g)} \rangle})_k
       +\sigma^{(g)}({\langle \tilde{\mathbf{z}}^{(g)} \rangle})_k
       ({\langle \tilde{\mathbf{z}}^{(g)} \rangle})_k\right]
   \end{multlined}
   \label{sec:theoreticalanalysis:eq:sodotnextg1}
\end{align}
can be derived.
\Cref{sec:theoreticalanalysis:eq:sodotnextg1}
can further be rewritten by the introduction of
$z_{\odot}^{(g)} :=
\frac{1}{r^{(g)}}\sum_{k=2}^N(\mathbf{x}^{(g)})_k
({\langle \tilde{\mathbf{z}}^{(g)} \rangle})_k$
(similar to \Cref{sec:theoreticalanalysis:eq:sodotdefinition})
and use of \Cref{sec:theoreticalanalysis:eq:sigmanormalized}
resulting in
\begin{align}
  &\begin{multlined}
     s_{\odot}^{(g+1)}=
     \frac{r^{(g)}}{r^{(g+1)}}
     (1-c)\left[s_\odot^{(g)}
       +\frac{{\sigma^{(g)}}^*}{N}({\langle \tilde{\mathbf{z}}^{(g)}
         \rangle})_{2..N}^T(\mathbf{s}^{(g)})_{2..N}\right]\\
     \hspace{1cm}+ \frac{r^{(g)}}{r^{(g+1)}}\sqrt{\mu c(2-c)}\left[z_\odot^{(g)}
       +\frac{{\sigma^{(g)}}^*}{N}||
       ({\langle \tilde{\mathbf{z}}^{(g)} \rangle})_{2..N}||^2\right].
   \end{multlined}
   \label{sec:theoreticalanalysis:eq:sodotnextg2}
\end{align}
For the fraction $r^{(g)}/r^{(g+1)}$, $r^{(g+1)}$ has to be derived.
From the offspring generation and selection steps it follows that
\begin{align}
  r^{(g+1)}&=\sqrt{\sum_{k=2}^N((\mathbf{x}^{(g)})_k+\sigma^{(g)}
    ({\langle \tilde{\mathbf{z}}^{(g)} \rangle})_k)^2}\\
  &=\sqrt{\sum_{k=2}^N\left((\mathbf{x}^{(g)})_k^2+
    2\sigma^{(g)}(\mathbf{x}^{(g)})_k({\langle \tilde{\mathbf{z}}^{(g)}
      \rangle})_k+{\sigma^{(g)}}^2({\langle \tilde{\mathbf{z}}^{(g)}
      \rangle})_k^2\right)}\\
  &=\sqrt{{r^{(g)}}^2+2\frac{{\sigma^{(g)}}^*}{N}{r^{(g)}}^2z_\odot^{(g)}
    +\frac{{{\sigma^{(g)}}^*}^2}{N^2}{r^{(g)}}^2||
    ({\langle \tilde{\mathbf{z}}^{(g)} \rangle})_{2..N}||^2}
  \label{sec:theoreticalanalysis:eq:rnextg}
\end{align}
holds.
Using the result from
\Cref{sec:theoreticalanalysis:eq:rnextg},
\begin{equation}
  \label{sec:theoreticalanalysis:eq:rfraction1}
  \frac{r^{(g)}}{r^{(g+1)}} = \sqrt{\frac{1}{1
      + \frac{2{{\sigma^{(g)}}^*}}{N}z_\odot^{(g)}
      + \frac{{{\sigma^{(g)}}^*}^2}{N^2}
      ||({\langle \tilde{\mathbf{z}}^{(g)} \rangle})_{2..N}||^2}}
\end{equation}
can be derived.
For further simplification of
\Cref{sec:theoreticalanalysis:eq:rfraction1},
asymptotic assumptions are made for $N \rightarrow \infty$.
Because the mutation vector is corrected in case of projection
(\Cref{sec:problemalgorithm:alg:es:projectionz}
in \Cref{sec:problemalgorithm:alg:es}),
${\langle \tilde{\mathbf{z}}^{(g)} \rangle}$ denotes the
centroid of the $\mu$ best (w.r.t. fitness) offspring mutation vectors
after the projection step. Approximation of
${\langle \tilde{\mathbf{z}}^{(g)} \rangle}$ for the asymptotic case
by its value before projection and selection
yields a normal distribution for
$({\langle \tilde{\mathbf{z}}^{(g)} \rangle})_k
=\frac{1}{\mu}\sum_{m=1}^\mu(\tilde{\mathbf{z}}_{m;\lambda})_k
\sim \mathcal{N}(0, \frac{1}{\mu})=\frac{1}{\sqrt{\mu}}\mathcal{N}(0, 1)$,
which follows by the properties of a sum
of normal distributed random variables.

Hence, $||({\langle \tilde{\mathbf{z}}^{(g)}
  \rangle})_{2..N}||^2$ can be approximated
by a $\chi^2$ distribution with $N-1$ degrees of freedom.
As the expected value of the $\chi^2$ distribution corresponds to
its number of degrees of freedom,
\begin{equation}
  \label{sec:theoreticalanalysis:eq:mutationlln}
  \frac{||({\langle \tilde{\mathbf{z}}^{(g)} \rangle})_{2..N}||^2}{N}
  \simeq \frac{1}{\mu}
\end{equation}
follows for $N \rightarrow \infty$ by the law of large numbers.
With
\Cref{sec:theoreticalanalysis:eq:mutationlln}
and the assumptions $N \gg 2{\sigma^{(g)}}^*z^{(g)}_\odot$
and $\mu N \gg {{\sigma^{(g)}}^*}^2$,
\begin{equation}
  \label{sec:theoreticalanalysis:eq:rfraction2}
  \frac{r^{(g)}}{r^{(g+1)}} \simeq 1
\end{equation}
follows.
Making use of
\Cref{sec:theoreticalanalysis:eq:rfraction2}
and
\Cref{sec:theoreticalanalysis:eq:mutationlln},
\Cref{sec:theoreticalanalysis:eq:sodotnextg2}
can be simplified for the asymptotic case $N \rightarrow \infty$ yielding
\begin{align}
  &\begin{multlined}
     s_{\odot}^{(g+1)}\simeq
     (1-c)s_\odot^{(g)}
     +(1-c)\frac{{\sigma^{(g)}}^*}{N}({\langle \tilde{\mathbf{z}}^{(g)}
       \rangle})_{2..N}^T(\mathbf{s}^{(g)})_{2..N}\\
     +\sqrt{\mu c(2-c)}z_\odot^{(g)}+\sqrt{\mu c(2-c)}
     \frac{{\sigma^{(g)}}^*}{\mu}.
   \end{multlined}
   \label{sec:theoreticalanalysis:eq:sodotnextg3}
\end{align}
Taking expected values of
\Cref{sec:theoreticalanalysis:eq:sodotnextg3}
with
$\mathrm{E}[s_{\odot}^{(g+1)}] := \overline{s_{\odot}^{(g+1)}}$
results in
\begin{align}
   &\begin{multlined}
     \overline{s_{\odot}^{(g+1)}}\simeq
     (1-c)\overline{s_\odot^{(g)}}
     +(1-c)\frac{\overline{{\sigma^{(g)}}^*}}{N}\mathrm{E}
     [({\langle \tilde{\mathbf{z}}^{(g)} \rangle})_{2..N}^T
       (\mathbf{s}^{(g)})_{2..N}]\\
     +\sqrt{\mu c(2-c)}\mathrm{E}[{z_\odot^{(g)}}]+
     \sqrt{\mu c(2-c)}\frac{\overline{{\sigma^{(g)}}^*}}{\mu}.
   \end{multlined}
   \label{sec:theoreticalanalysis:eq:sodotnextgexpectation}
\end{align}
To treat \Cref{sec:theoreticalanalysis:eq:sodotnextgexpectation} further,
$\mathrm{E}[({\langle \tilde{\mathbf{z}}^{(g)} \rangle})_{2..N}^T
  (\mathbf{s}^{(g)})_{2..N}]$
and $\mathrm{E}[{z_\odot^{(g)}}]$ need to be derived.

For
$\mathrm{E}[({\langle \tilde{\mathbf{z}}^{(g)} \rangle})_{2..N}^T
  (\mathbf{s}^{(g)})_{2..N}]$,
$({\langle \tilde{\mathbf{z}}^{(g)} \rangle})_{2..N}$
is decomposed into a vector
in direction of the parental individual's $2..N$ components
$\mathbf{e}^{(g)}_\odot$
and in a direction $\mathbf{e}^{(g)}_\ominus$ that is orthogonal to
$\mathbf{e}^{(g)}_\odot$, i.e.,
${\mathbf{e}^{(g)}_\odot}^T\mathbf{e}^{(g)}_\ominus = 0$. Further,
in the following the assumption is made that those direction vectors are
unit vectors, i.e., $||\mathbf{e}^{(g)}_\odot|| = 1$ and
$||\mathbf{e}^{(g)}_\ominus|| = 1$.
Therefore,
$({\langle \tilde{\mathbf{z}}^{(g)} \rangle})_{2..N}$ can be written as
\begin{equation}
  \label{sec:theoreticalanalysis:eq:zdecomposed}
  ({\langle \tilde{\mathbf{z}}^{(g)} \rangle})_{2..N} =
  z_\odot^{(g)}\mathbf{e}^{(g)}_\odot + z_\ominus^{(g)}\mathbf{e}^{(g)}_\ominus,
\end{equation}
where $z_\odot^{(g)}$ and $z_\ominus^{(g)}$ are the projections
of the mutation vector in direction of $\mathbf{e}^{(g)}_\odot$
and $\mathbf{e}^{(g)}_\ominus$, respectively.
Using
\Cref{sec:theoreticalanalysis:eq:zdecomposed},
\begin{align}
  ({\langle \tilde{\mathbf{z}}^{(g)} \rangle})_{2..N}^T(\mathbf{s}^{(g)})_{2..N}
  &= z_\odot^{(g)}\underbrace{{\mathbf{e}^{(g)}_\odot}^T
    (\mathbf{s}^{(g)})_{2..N}}_{=s_\odot^{(g)}}
  + z_\ominus^{(g)}{\mathbf{e}^{(g)}_\ominus}^T(\mathbf{s}^{(g)})_{2..N}
\end{align}
follows. Note that ${\mathbf{e}^{(g)}_\odot}^T(\mathbf{s}^{(g)})_{2..N}$
corresponds to the definition in
\Cref{sec:theoreticalanalysis:eq:sodotdefinition}.
Taking into account the statistical independence of
the cumulated path vector and the mutation
in the current generation, taking expectation results in
\begin{align}
  \mathrm{E}[({\langle \tilde{\mathbf{z}}^{(g)} \rangle})_{2..N}^T
    (\mathbf{s}^{(g)})_{2..N}]
  &= \mathrm{E}[z_\odot^{(g)}]\mathrm{E}[s_\odot^{(g)}]
  + \mathrm{E}[z_\ominus^{(g)}]\mathrm{E}[{\mathbf{e}^{(g)}_\ominus}^T
    (\mathbf{s}^{(g)})_{2..N}]
  \label{sec:theoreticalanalysis:eq:zsmixterm0}\\
  &= \mathrm{E}[z_\odot^{(g)}]\overline{s_\odot^{(g)}}.
  \label{sec:theoreticalanalysis:eq:zsmixterm}
\end{align}
Note that $\mathrm{E}[z_\ominus^{(g)}]$ vanishes because the mutations in
direction $\mathbf{e}^{(g)}_\ominus$ are isotropic and selectively neutral.
Hence, the second summand of
\Cref{sec:theoreticalanalysis:eq:zsmixterm0}
is $0$ in expectation. To investigate the behavior of
$\mathrm{E}[{\mathbf{e}^{(g)}_\ominus}^T(\mathbf{s}^{(g)})_{2..N}]$, the
dynamics of ${\mathbf{e}^{(g)}_\ominus}^T(\mathbf{s}^{(g)})_{2..N}$
have been empirically determined for different parameter configurations in real
ES runs (not shown here). It turned out
that ${\mathbf{e}^{(g)}_\ominus}^T(\mathbf{s}^{(g)})_{2..N}$
fluctuates around $0$ (with the empirical mean being approximately $0$),
which further justifies the step from
\Cref{sec:theoreticalanalysis:eq:zsmixterm0} to
\Cref{sec:theoreticalanalysis:eq:zsmixterm}.

$\mathrm{E}[{z_\odot^{(g)}}]$ can be calculated from the progress rate of the
quadratic distance from the cone axis. It writes
\begin{align}
  \varphi_{r^2}^{(g)}
  &:= \mathrm{E}[{r^{(g)}}^2 - {r^{(g+1)}}^2]
  = {r^{(g)}}^2 - \mathrm{E}[{r^{(g+1)}}^2]
  = {r^{(g)}}^2 - \mathrm{E}[{{\langle q_r \rangle}^2}]
  \label{sec:theoreticalanalysis:eq:progressratersquared-1}\\
  &\approx {r^{(g)}}^2 -
  \left\{P_{\text{feas}}(x^{(g)}, r^{(g)}, \sigma^{(g)})
  \mathrm{E}[{{\langle q_r \rangle}_{\text{feas}}^2}]
  +[1-P_{\text{feas}}(x^{(g)}, r^{(g)}, \sigma^{(g)})]
  \mathrm{E}[{{\langle q_r \rangle}_{\text{infeas}}^2}]\right\}
  \label{sec:theoreticalanalysis:eq:progressratersquared0}
\end{align}
where $\langle q_r \rangle$ denotes the distance from
the cone boundary of the centroid after projection
(cf. \Cref{sec:problemalgorithm:alg:es:bestqr}
of \Cref{sec:problemalgorithm:alg:es}).
Expressions for
$\mathrm{E}[{{\langle q_r \rangle}_{\text{feas}}^2}]$
and
$\mathrm{E}[{{\langle q_r \rangle}_{\text{infeas}}^2}]$
have already been derived
in Appendix D in the supplementary material
of~\cite{SpettelBeyer2018SigmaSaEsConeMulti}.
The used Taylor approximation in Equation (D.157) of that
work allows using the square of Equation (D.165) for the
feasible case yielding
\begin{equation}
  \mathrm{E}[{{\langle q_r \rangle}_{\text{feas}}^2}]
  \approx
  {r^{(g)}}^2 + \frac{{\sigma^{(g)}}^2}{\mu}(N-1).
  \label{sec:theoreticalanalysis:eq:expqrcentroidfeas}
\end{equation}
Similarly, the Taylor expansion used in Equation (D.172)
of that work allows using the square of Equation (D.217) as an approximation
for the infeasible case. It reads
\begin{equation}
  \mathrm{E}[{{\langle q_r \rangle}_{\text{infeas}}^2}]
  \approx
  \frac
      {\mathrm{E}\left[{\langle q \rangle}_{\text{infeas}}\right]^2}
      {\xi}
      \left(\frac{1+\frac{{{\sigma^{(g)}}^*}^2}{\mu N}}
           {1+\frac{{{\sigma^{(g)}}^*}^2}{N}}\right).
  \label{sec:theoreticalanalysis:eq:progressinfeashelper}
\end{equation}
Using
\Cref{sec:theoreticalanalysis:eq:expqrcentroidfeas}
and
\Cref{sec:theoreticalanalysis:eq:progressinfeashelper},
\begin{equation}
  \begin{multlined}
    \varphi_{r^2}^{(g)}
    \approx {r^{(g)}}^2 -
    \left\{P_{\text{feas}}(x^{(g)}, r^{(g)}, \sigma^{(g)})
    \left[{r^{(g)}}^2 + \frac{{\sigma^{(g)}}^2}{\mu}(N-1)\right]\right.\\\left.
    +[1-P_{\text{feas}}(x^{(g)}, r^{(g)}, \sigma^{(g)})]
    \left[
      \frac
          {\mathrm{E}\left[{\langle q \rangle}_{\text{infeas}}\right]^2}
          {\xi}
          \left(\frac{1+\frac{{{\sigma^{(g)}}^*}^2}{\mu N}}
               {1+\frac{{{\sigma^{(g)}}^*}^2}{N}}\right)\right]\right\}
  \end{multlined}
  \label{sec:theoreticalanalysis:eq:progressratersquared}
\end{equation}
follows, where a closed-form approximation
\begin{align}
  \mathrm{E}[{\langle q \rangle}_{\text{infeas}}]
  &\approx
  \frac{\xi}{1+\xi}\left(x^{(g)} + \bar{r}/\sqrt{\xi}\right)
  - \frac{\xi}{1+\xi}\sqrt{{\sigma^{(g)}}^2+\sigma_r^2/\xi}
  c_{\mu/\mu,\lambda}
  \label{sec:theoreticalanalysis:eq:qcentroidinfeasinsertedfinal}
\end{align}
has been derived
in~\cite{SpettelBeyer2018SigmaSaEsConeMulti} as well
(refer to the derivations leading to Equation (C.149)
in Appendix C in the supplementary of that work for the
details).
With \Cref{sec:theoreticalanalysis:eq:progressratersquared} and
\Cref{sec:theoreticalanalysis:eq:qcentroidinfeasinsertedfinal},
$\varphi_{r^2}^{(g)}$ can be computed for a given state
of the system. The goal is now to express $\varphi_{r^2}^{(g)}$ in terms of
$\mathrm{E}[{z_\odot^{(g)}}]$. Subsequently solving for
$\mathrm{E}[{z_\odot^{(g)}}]$ allows then to compute its value.
Using
\Cref{sec:theoreticalanalysis:eq:progressratersquared-1}
and
\Cref{sec:theoreticalanalysis:eq:rnextg}
with \Cref{sec:theoreticalanalysis:eq:mutationlln},
$\varphi_{r^2}^{(g)}$ can alternatively be written as
\begin{align}
  \varphi_{r^2}^{(g)}
  &= \overline{{r^{(g)}}^2} - \mathrm{E}\left[\overline{{r^{(g)}}^2}+
    2{\overline{\sigma^{(g)}}}\,{\overline{r^{(g)}}}z_\odot^{(g)}+
    {\overline{\sigma^{(g)}}}^2||({\langle \tilde{\mathbf{z}}^{(g)}
      \rangle})_{2..N}||^2\right]
  \label{sec:theoreticalanalysis:eq:varphirsquared1}\\
  &\simeq -2{\overline{\sigma^{(g)}}}\,
        {\overline{r^{(g)}}}\mathrm{E}\left[z_\odot^{(g)}\right]-
        {\overline{\sigma^{(g)}}}^2\frac{N}{\mu}.
  \label{sec:theoreticalanalysis:eq:varphirsquared2}
\end{align}
\Cref{sec:theoreticalanalysis:eq:varphirsquared2}
can be solved for $\mathrm{E}[z_\odot^{(g)}]$ yielding
\begin{align}
  \mathrm{E}[z_\odot^{(g)}]
  \simeq-\left(\frac{\varphi_{r^2}^{(g)}
    +{\overline{\sigma^{(g)}}}^2\frac{N}{\mu}}
         {2{\overline{\sigma^{(g)}}}\,{\overline{r^{(g)}}}}\right)
         &=-\frac{\varphi_{r^2}^{(g)}}
         {2{\overline{\sigma^{(g)}}}\,{\overline{r^{(g)}}}}
         -\frac{{\overline{\sigma^{(g)}}}^2N}{2{\overline{\sigma^{(g)}}}\,
           {\overline{r^{(g)}}}\mu}
         =-\frac{N\varphi_{r^2}^{(g)}}{2{\overline{{\sigma^{(g)}}^*}}\,
           {{\overline{r^{(g)}}}^2}}
    -\frac{{\overline{{\sigma^{(g)}}^*}}}{2\mu}.
  \label{sec:theoreticalanalysis:eq:zodotexpectation}
\end{align}
Reinsertion of
\Cref{sec:theoreticalanalysis:eq:zsmixterm}
and
\Cref{sec:theoreticalanalysis:eq:zodotexpectation}
into
\Cref{sec:theoreticalanalysis:eq:sodotnextgexpectation}
yields
\begin{align}
  &\begin{multlined}
     \overline{s_{\odot}^{(g+1)}}\simeq
     (1-c)\left(\overline{s_\odot^{(g)}}
     +\frac{\overline{{\sigma^{(g)}}^*}}{N}
     \mathrm{E}[{z_\odot^{(g)}}]\overline{s_\odot^{(g)}}\right)
     +\sqrt{\mu c(2-c)}\left(\mathrm{E}[{z_\odot^{(g)}}]+
     \frac{\overline{{\sigma^{(g)}}^*}}{\mu}\right)
  \end{multlined}\\
  &\begin{multlined}
     \phantom{\overline{s_{\odot}^{(g+1)}}}\simeq
     (1-c)\left(1
     +\frac{\overline{{\sigma^{(g)}}^*}}{N}
     \left(-\frac{N\varphi_{r^2}^{(g)}}
          {2{\overline{{\sigma^{(g)}}^*}}\,{\overline{{r^{(g)}}}^2}}
    -\frac{\overline{{\sigma^{(g)}}^*}}{2\mu}\right)
     \right)\overline{s_\odot^{(g)}}\\
     +\sqrt{\mu c(2-c)}\left(
     -\frac{N\varphi_{r^2}^{(g)}}
     {2{\overline{{\sigma^{(g)}}^*}}\,{\overline{{r^{(g)}}}^2}}
     +\frac{\overline{{\sigma^{(g)}}^*}}{2\mu}\right).
  \end{multlined}
\end{align}

\subsubsection{Derivation of a Mean Value Difference
  Equation for $||\mathbf{s}||^2$}
\label{sec:theoreticalanalysis:subsec:iterativesystem:diffeqssquaredlength}
Using the update rule from
\Cref{sec:problemalgorithm:alg:es:replaces}
of
\Cref{sec:problemalgorithm:alg:es},
\begin{align}
  &\begin{multlined}
     ||\mathbf{s}^{(g+1)}||^2=
     ||(1-c)\mathbf{s}^{(g)}
     +\sqrt{\mu c(2-c)}{\langle \tilde{\mathbf{z}}^{(g)} \rangle}||^2
  \end{multlined}\\
  &\begin{multlined}
     \phantom{||\mathbf{s}^{(g+1)}||^2}=
     (1-c)^2||\mathbf{s}^{(g)}||^2
     +2(1-c)\sqrt{\mu c(2-c)}{\mathbf{s}^{(g)}}^T
     {\langle \tilde{\mathbf{z}}^{(g)} \rangle}
     +\mu c(2-c)||{\langle \tilde{\mathbf{z}}^{(g)} \rangle}||^2
  \end{multlined}
  \label{sec:theoreticalanalysis:eq:ssquaredlength1}
\end{align}
can be derived.
For treating ${\mathbf{s}^{(g)}}^T{\langle \tilde{\mathbf{z}}^{(g)} \rangle}$,
the vector $\mathbf{s}^{(g)}$ can be decomposed into a sum of vectors
in direction of the cone axis $\mathbf{e}^{(g)}_1$,
in direction of the parental individual's $2..N$ components
$\mathbf{e}^{(g)}_\odot$,
and in a direction $\mathbf{e}^{(g)}_\ominus$
that is orthogonal to $\mathbf{e}^{(g)}_1$ and $\mathbf{e}^{(g)}_\odot$.
Formally, this can be written as
\begin{equation}
  \mathbf{s}^{(g)}=
  s_1^{(g)}\mathbf{e}^{(g)}_1
  + s_\odot^{(g)}\mathbf{e}^{(g)}_\odot
  + s_\ominus^{(g)}\mathbf{e}^{(g)}_\ominus
\end{equation}
where
$||\mathbf{e}^{(g)}_1|| = ||\mathbf{e}^{(g)}_\odot|| =
||\mathbf{e}^{(g)}_\ominus|| = 1$
and
${\mathbf{e}^{(g)}_1}^T\mathbf{e}^{(g)}_\odot =
{\mathbf{e}^{(g)}_1}^T\mathbf{e}^{(g)}_\ominus
= {\mathbf{e}^{(g)}_\odot}^T\mathbf{e}^{(g)}_\ominus = 0$.
$s_1^{(g)}$, $s_\odot^{(g)}$, and $s_\ominus^{(g)}$ denote
the corresponding projections in those directions.
Consequently,
\begin{equation}
  {\mathbf{s}^{(g)}}^T{\langle \tilde{\mathbf{z}}^{(g)} \rangle}=
  s_1^{(g)}({\langle \tilde{\mathbf{z}}^{(g)} \rangle})_1
  + s_\odot^{(g)}z^{(g)}_\odot
  + s_\ominus^{(g)}{\mathbf{e}^{(g)}_\ominus}^T
  {\langle \tilde{\mathbf{z}}^{(g)} \rangle}
\end{equation}
and subsequently
\begin{equation}
  \mathrm{E}[{\mathbf{s}^{(g)}}^T{\langle \tilde{\mathbf{z}}^{(g)} \rangle}]=
  \mathrm{E}[s_1^{(g)}({\langle \tilde{\mathbf{z}}^{(g)} \rangle})_1]
  + \mathrm{E}[s_\odot^{(g)}z^{(g)}_\odot]
  + \mathrm{E}[s_\ominus^{(g)}{\mathbf{e}^{(g)}_\ominus}^T
    {\langle \tilde{\mathbf{z}}^{(g)} \rangle}]
\end{equation}
follow. Taking into account the
statistical independence between the cumulation path
and a particular generation's mutations allows writing
\begin{align}
  \mathrm{E}[{\mathbf{s}^{(g)}}^T{\langle \tilde{\mathbf{z}}^{(g)} \rangle}]&=
  \mathrm{E}[s_1^{(g)}]
  \mathrm{E}[({\langle \tilde{\mathbf{z}}^{(g)} \rangle})_1]
  + \mathrm{E}[s_\odot^{(g)}]\mathrm{E}[z^{(g)}_\odot]
  + \mathrm{E}[s_\ominus^{(g)}]\mathrm{E}[{\mathbf{e}^{(g)}_\ominus}^T
    {\langle \tilde{\mathbf{z}}^{(g)} \rangle}]\\
  &\simeq \mathrm{E}[s_1^{(g)}]\mathrm{E}[({\langle
      \tilde{\mathbf{z}}^{(g)} \rangle})_1]
  + \mathrm{E}[s_\odot^{(g)}]\mathrm{E}[z^{(g)}_\odot].
  \label{sec:theoreticalanalysis:eq:ssquaredlength2}
\end{align}
Again,
$\mathrm{E}[{\mathbf{e}^{(g)}_\ominus}^T
  {\langle \tilde{\mathbf{z}}^{(g)} \rangle}]$
vanishes because those mutations are selectively neutral and isotropic.
Taking expectation of
\Cref{sec:theoreticalanalysis:eq:ssquaredlength1},
considering
\Cref{sec:theoreticalanalysis:eq:ssquaredlength2,%
sec:theoreticalanalysis:eq:z1,%
sec:theoreticalanalysis:eq:zodotexpectation,%
sec:theoreticalanalysis:eq:sigmanormalized},
and using
$\mathrm{E}[||{\langle \tilde{\mathbf{z}}^{(g)}
    \rangle}||^2] \simeq \frac{N}{\mu}$,
\begin{align}
  &\begin{multlined}
     \overline{||\mathbf{s}^{(g+1)}||^2}\simeq
     (1-c)^2\overline{||\mathbf{s}^{(g)}||^2}
     +2(1-c)\sqrt{\mu c(2-c)}\\
     \hspace{2cm}\times\left(\overline{s_1^{(g)}}
     \mathrm{E}[({\langle \tilde{\mathbf{z}}^{(g)} \rangle})_1]
     + \overline{s_\odot^{(g)}}\mathrm{E}[z^{(g)}_\odot]\right)
     +c(2-c)N
  \end{multlined}\\
  &\begin{multlined}
     \phantom{\overline{||\mathbf{s}^{(g+1)}||^2}}\simeq
     (1-c)^2\overline{||\mathbf{s}^{(g)}||^2}
     +2(1-c)\sqrt{\mu c(2-c)}\\
     \hspace{0.5cm}\times\left(\overline{s_1^{(g)}}
     \left(-\frac{N\varphi_x^{(g)}}{\overline{{\sigma^{(g)}}^*}\,
       \overline{r^{(g)}}}\right)
     + \overline{s_\odot^{(g)}}
     \left(-\frac{N\varphi_{r^2}^{(g)}}
          {2{\overline{{\sigma^{(g)}}^*}}\,{{\overline{r^{(g)}}}^2}}
    -\frac{\overline{{\sigma^{(g)}}^*}}{2\mu}\right)\right)
     +c(2-c)N
  \end{multlined}
\end{align}
follows.

\subsubsection{Derivation of a Mean Value Difference
  Equation for $\sigma$}
\label{sec:theoreticalanalysis:subsec:iterativesystem:diffeqsigma}
From the update rule of $\sigma$ in
\Cref{sec:problemalgorithm:alg:es:replacesigma}
of
\Cref{sec:problemalgorithm:alg:es},
${\sigma^{(g+1)}} = {\sigma^{(g)}}
\exp\left(\frac{||\mathbf{s}^{(g+1)}||^2-N}{2DN}\right)$
follows for the update of the mutation strength.
Taking expected values and knowing that ${\sigma^{(g)}}$
is constant w.r.t. $||\mathbf{s}^{(g+1)}||^2$, this writes
$\overline{{\sigma^{(g+1)}}} =
\overline{{\sigma^{(g)}}}
\mathrm{E}\left[\exp\left(
\frac{||\mathbf{s}^{(g+1)}||^2-N}{2DN}\right)\right]$.
Assuming that the fluctuations of $||\mathbf{s}^{(g+1)}||^2$ around its
expected value are sufficiently small, the expected value can
be pulled into the exponential function yielding
\begin{equation}
  \overline{{\sigma^{(g+1)}}} \simeq
  \overline{{\sigma^{(g)}}}
  \exp\left(\frac{\overline{||\mathbf{s}^{(g+1)}||^2}-N}{2DN}\right).
\end{equation}

\subsubsection{Summary of the Mean Value Difference Equations}
\label{sec:theoreticalanalysis:subsec:evolutionequationssummary}
\begin{align}
  &\begin{multlined}
    \overline{x^{(g + 1)}}
    =\overline{x^{(g)}}\left(1-\frac{{\varphi^{(g)}_{x}}^*}{N}\right)
  \end{multlined}
  \label{sec:theoreticalanalysis:eq:summaryx}\\
  &\begin{multlined}
    \overline{r^{(g + 1)}}
    =\overline{r^{(g)}}\left(1-\frac{{\varphi^{(g)}_r}^*}{N}\right)
  \end{multlined}
  \label{sec:theoreticalanalysis:eq:summaryr}\\
  &\begin{multlined}
    \overline{s_1^{(g+1)}}
    \simeq(1-c)\overline{s_1^{(g)}} +
    \sqrt{\mu c(2-c)}\left(-\frac{N\varphi_x^{(g)}}
         {\overline{{\sigma^{(g)}}^*}\,\overline{r^{(g)}}}\right)
  \end{multlined}
  \label{sec:theoreticalanalysis:eq:summarys1}\\
  &\begin{multlined}
     \overline{s_{\odot}^{(g+1)}}\simeq
     (1-c)\left(1
     +\frac{\overline{{\sigma^{(g)}}^*}}{N}
     \left(-\frac{N\varphi_{r^2}^{(g)}}
          {2{\overline{{\sigma^{(g)}}^*}}\,{{\overline{r^{(g)}}}^2}}
    -\frac{\overline{{\sigma^{(g)}}^*}}{2 \mu}\right)
     \right)\overline{s_\odot^{(g)}}\\
     +\sqrt{\mu c(2-c)}\left(
     -\frac{N\varphi_{r^2}^{(g)}}
          {2{\overline{{\sigma^{(g)}}^*}}\,{{\overline{r^{(g)}}}^2}}
     +\frac{\overline{{\sigma^{(g)}}^*}}{2\mu}\right)
  \end{multlined}
  \label{sec:theoreticalanalysis:eq:summarysr}\\
  &\begin{multlined}
     \overline{||\mathbf{s}^{(g+1)}||^2}\simeq
     (1-c)^2\overline{||\mathbf{s}^{(g)}||^2}
     +2(1-c)\sqrt{\mu c(2-c)}\\
     \hspace{2cm}\times\left(\overline{s_1^{(g)}}
     \left(-\frac{N\varphi_x^{(g)}}
          {\overline{{\sigma^{(g)}}^*}\,\overline{r^{(g)}}}\right)
     + \overline{s_\odot^{(g)}}
     \left(-\frac{N\varphi_{r^2}^{(g)}}
          {2{\overline{{\sigma^{(g)}}^*}}\,{{\overline{r^{(g)}}}^2}}
    -\frac{\overline{{\sigma^{(g)}}^*}}{2 \mu}\right)\right)
     +c(2-c)N
  \end{multlined}
  \label{sec:theoreticalanalysis:eq:summaryssquared}\\
  &\begin{multlined}
    \overline{{\sigma^{(g+1)}}} \simeq
    \overline{{\sigma^{(g)}}}
    \exp\left(\frac{\overline{||\mathbf{s}^{(g+1)}||^2}-N}{2DN}\right)
  \end{multlined}
  \label{sec:theoreticalanalysis:eq:summarysigma}\\
  &\begin{multlined}
    \overline{{\sigma^{(g+1)}}^*} =
    \frac{N\overline{{\sigma^{(g+1)}}}}{\overline{r^{(g+1)}}}
  \end{multlined}
  \label{sec:theoreticalanalysis:eq:summarysigmanormalized}
\end{align}

\begin{figure}
  \centering
  \begin{tabular}{@{\hspace{-0.0\textwidth}}c@{\hspace{-0.0\textwidth}}c}
    \includegraphics[width=0.46\textwidth]{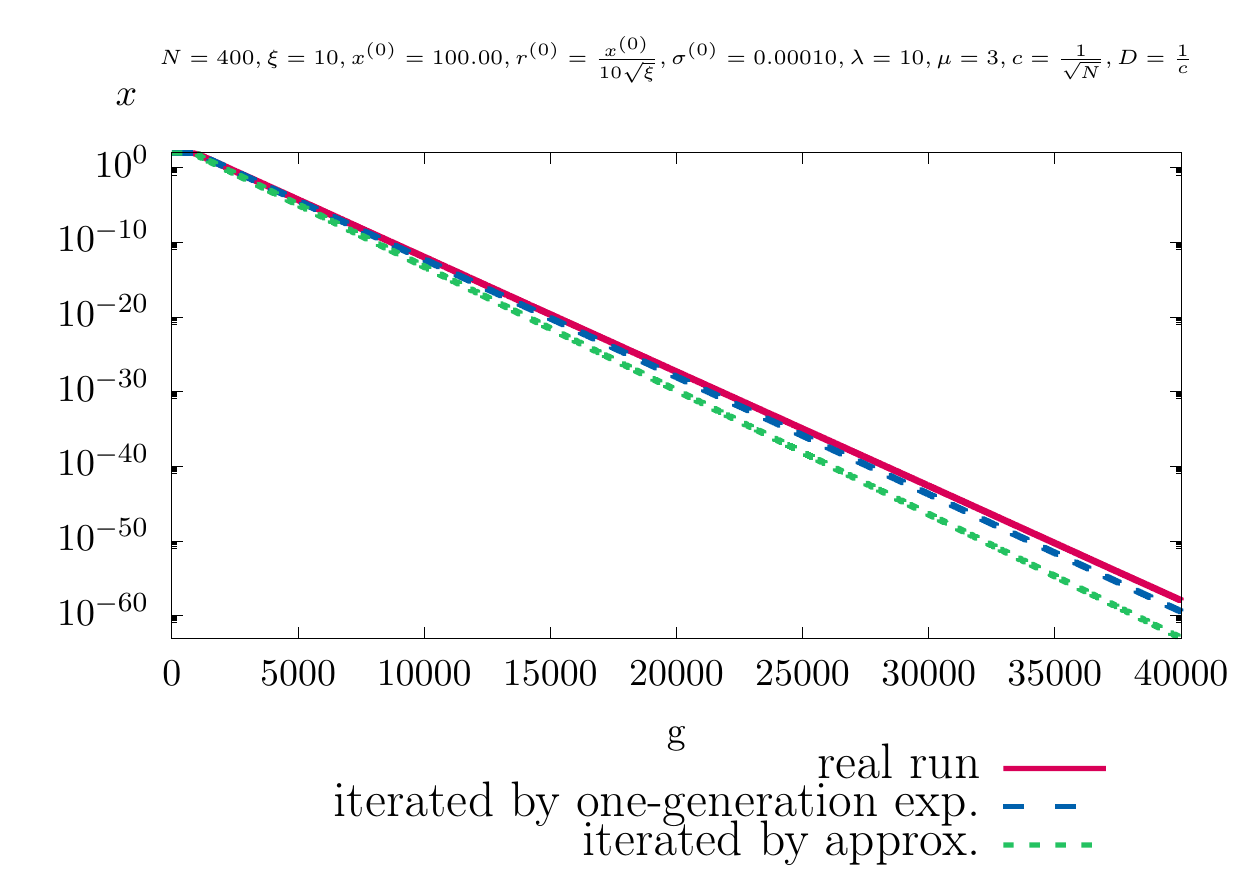}&
    \includegraphics[width=0.46\textwidth]{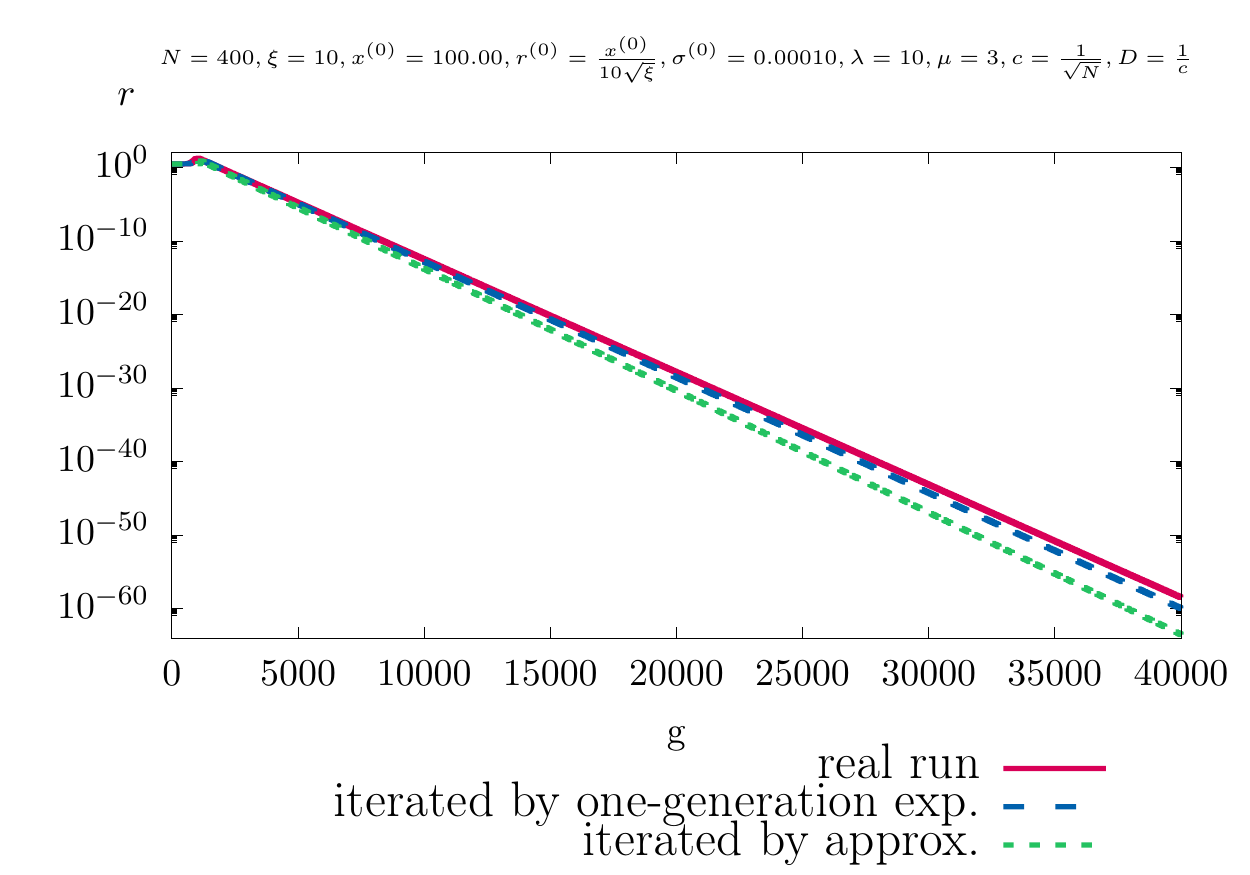}\\
    \includegraphics[width=0.46\textwidth]{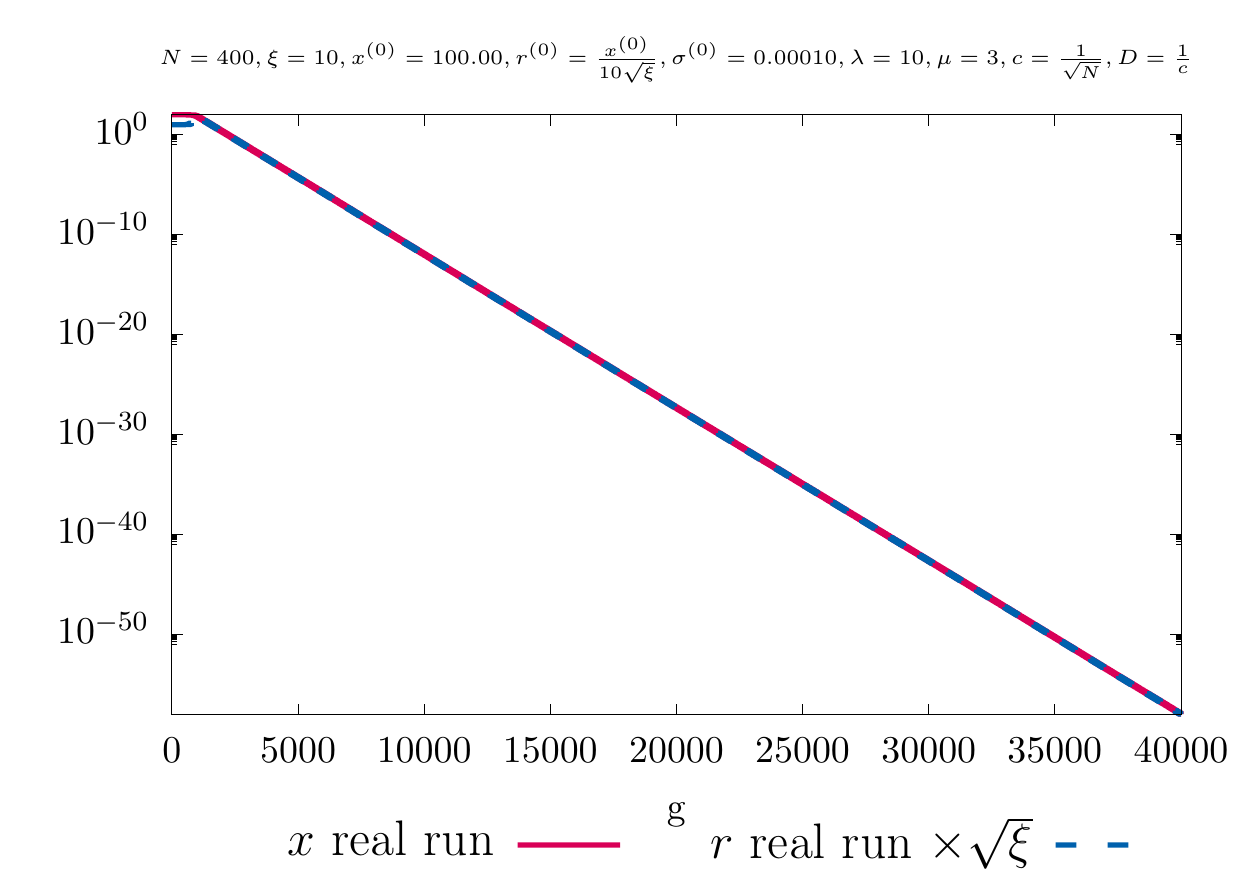}&
    \includegraphics[width=0.46\textwidth]{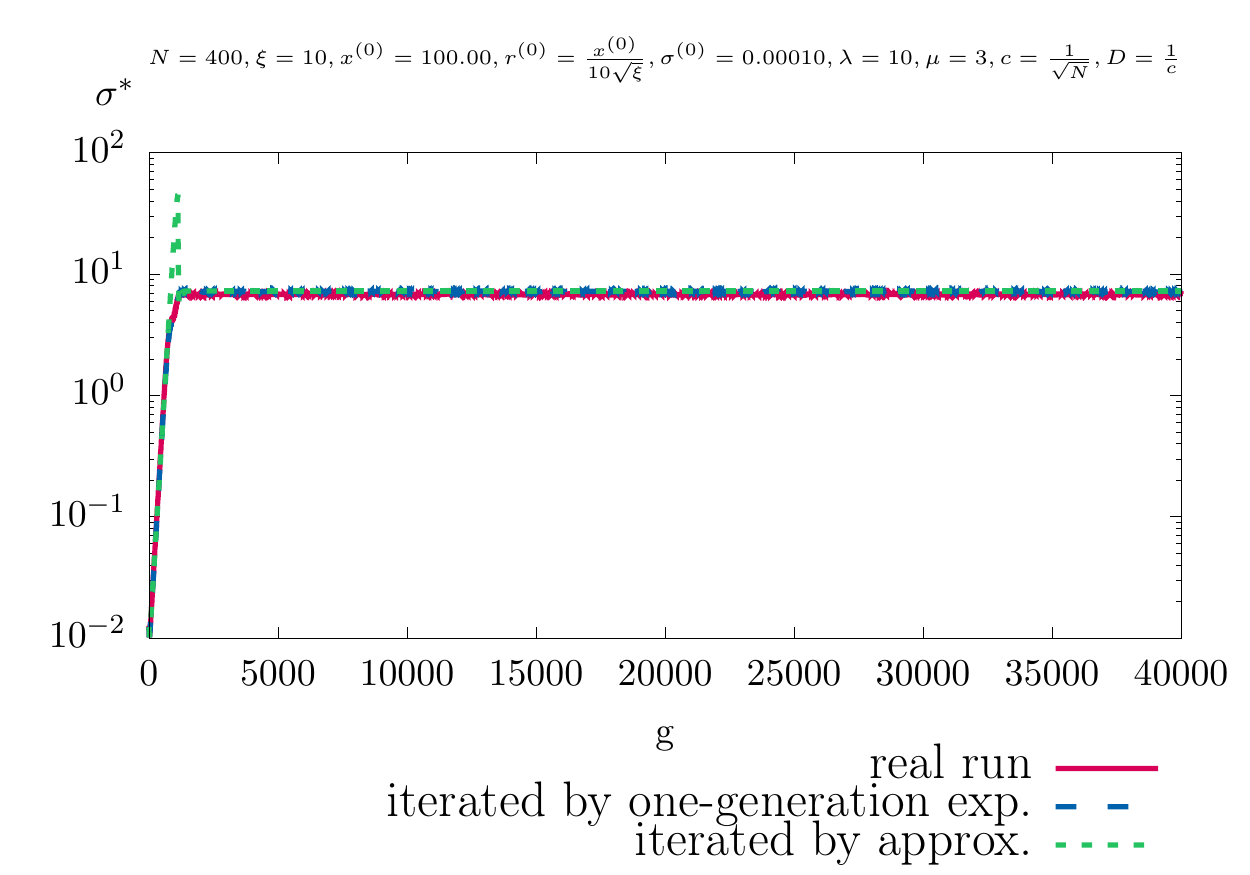}
  \end{tabular}
  \caption{Real run and approximation comparison
    of the $(3/3_I,10)$-CSA-ES mean value dynamics ($N=400$, $\xi=10$).
    The agreement of the iteration with the
    theoretically derived expressions and the real ES runs
    is satisfactory. In addition, the iteration with
    the one-generation experiments for the local progress
    rates is very similar to the mean value dynamics of the real ES
    runs. Consequently, the modeling of the system with
    \Crefrange{sec:theoreticalanalysis:eq:summaryx}
              {sec:theoreticalanalysis:eq:summarysigmanormalized} is
    appropriate.}
  \label{sec:theoreticalanalysis:fig:dynamics}
\end{figure}

The mean value dynamics of the $(3/3_I,10)$-CSA-ES
on the conically constrained problem
are shown in
\Cref{sec:theoreticalanalysis:fig:dynamics}
for $N=400$, $\xi=10$, $c=\frac{1}{\sqrt{N}}$, and $D=\frac{1}{c}$.
The agreement of the simulations and the derived expressions
is satisfactory. In particular, one observes that
the lines of the iteration with one-generation experiments
are very similar to the lines generated by real ES runs.
Consequently, the modeling of the system with
\Crefrange{sec:theoreticalanalysis:eq:summaryx}
{sec:theoreticalanalysis:eq:summarysigmanormalized} is
appropriate and the deviations for the theoretically derived
expressions are mainly due to approximations in the derivations
of the local progress rates. For this, it is referred to the
additional figures provided in the supplementary material
(\Cref{sec:appendix:additionalfigures}). They show a larger
deviation for smaller values of $\xi$ and smaller values
of $N$. But notice that in those figures the
iteration with one-generation experiments
for the local progress measures coincides well
with the results of real ES runs. This again shows
the appropriateness of the modeling
in \Crefrange{sec:theoreticalanalysis:eq:summaryx}
{sec:theoreticalanalysis:eq:summarysigmanormalized}.
The deviations for small $N$ stem from
asymptotic assumptions using $N \rightarrow \infty$.
They help simplifying expressions resulting in a theoretical analysis
that is tractable.
The deviations for small $\xi$ are due to approximations
in the derivation of the offspring
cumulative distribution function after the projection step in $x_1$-direction
$P_Q(q)$ (for the details, it is referred to Section
3.1.2.1.2.3 in~\cite{SpettelBeyer2018SigmaSaEsCone},
in particular to the step from Equation (3.73) to Equation (3.74)).

For the figures, results of $100$ real runs of the ES have been averaged
for generating the solid lines.
The lines for the iteration by approximation have been
computed by iterating the mean value iterative system
(\Crefrange{sec:theoreticalanalysis:eq:summaryx}
{sec:theoreticalanalysis:eq:summarysigmanormalized}) with
\Cref{sec:theoreticalanalysis:eq:varphixnormalizedcombined,%
  sec:theoreticalanalysis:eq:varphirnormalizedcombinedmaintext,%
  sec:theoreticalanalysis:eq:progressratersquared}
for ${\varphi^{(g)}_{x}}$ (and ${\varphi^{(g)}_{x}}^*$),
${\varphi^{(g)}_{r}}$ (and ${\varphi^{(g)}_{r}}^*$), and
${\varphi^{(g)}_{r^2}}$, respectively. The lines for the
iteration with one-generation experiments have been
generated by iterating the system
(\Crefrange{sec:theoreticalanalysis:eq:summaryx}
{sec:theoreticalanalysis:eq:summarysigmanormalized})
and simulating ${\varphi^{(g)}_{x}}$ (and ${\varphi^{(g)}_{x}}^*$),
${\varphi^{(g)}_{r}}$ (and ${\varphi^{(g)}_{r}}^*$), and
${\varphi^{(g)}_{r^2}}$.
It can happen that in a generation of iterating the system
(\Crefrange{sec:theoreticalanalysis:eq:summaryx}
{sec:theoreticalanalysis:eq:summarysigmanormalized}),
infeasible $(x^{(g)},r^{(g)})^T$ are created.
In such circumstances, the corresponding $(x^{(g)},r^{(g)})^T$
have been projected back.

\subsection{Behavior of the ES in the Steady State}
\label{sec:theoreticalanalysis:subsec:steadystate}
The goal of this section is to derive approximate closed-form expressions
for the steady state values of the mean value iterative system that is
summarized in
\Cref{sec:theoreticalanalysis:subsec:evolutionequationssummary}.
A working ES should steadily decrease $x$ and $r$
(\Cref{sec:theoreticalanalysis:eq:summaryx}
and
\Cref{sec:theoreticalanalysis:eq:summaryr},
respectively)
in order to move towards
the optimizer. For determining the steady state normalized mutation strength
value, the fixed point of the system of non-linear equations
(\Crefrange{sec:theoreticalanalysis:eq:summarys1}
{sec:theoreticalanalysis:eq:summarysigmanormalized})
is to be computed.

\subsubsection{Derivations Towards Closed-Form Steady State Expressions}
\label{sec:theoreticalanalysis:subsubsec:steadystatenumerical}
This section comprises a first step towards closed-form approximations
for the steady state values of the system summarized in
\Cref{sec:theoreticalanalysis:subsec:evolutionequationssummary}.
Expressions are derived that finally lead to a steady state equation
for the normalized mutation strength. A closed form solution of
this equation is not apparent. Hence, further assumptions for
different cases are considered in the following sections.

To compute the fixed point of the system described by
\Crefrange{sec:theoreticalanalysis:eq:summarys1}
{sec:theoreticalanalysis:eq:summarysigmanormalized},
stationary state expressions ${\varphi_x}_{ss}^*$,
${\varphi_r}_{ss}^*$, $\left(-\frac{N\varphi_x^{(g)}}
{\overline{{\sigma^{(g)}}^*}\,\overline{r^{(g)}}}\right)_{ss}$, and
$\left(-\frac{N\varphi_{r^2}^{(g)}}
{2{\overline{{\sigma^{(g)}}^*}}\,{{\overline{r^{(g)}}}^2}}\right)_{ss}$
for ${\varphi_x^{(g)}}^*$, ${\varphi_r^{(g)}}^*$,
$-\frac{N\varphi_x^{(g)}}
{\overline{{\sigma^{(g)}}^*}\,\overline{r^{(g)}}}$, and
$-\frac{N\varphi_{r^2}^{(g)}}
{2{\overline{{\sigma^{(g)}}^*}}\,{{\overline{r^{(g)}}}^2}}$,
respectively, need to be derived first because they
are dependent on the position in the parameter space.
The bottom left subplot of
\Cref{sec:theoreticalanalysis:fig:dynamics}
shows
that the ES moves in the vicinity of the cone
boundary in the steady state. This can be seen because
the dynamics of $x$ and $r$ are plotted by converting
them into each other for the cone boundary case. Notice
that those lines coincide in the steady state.
In this situation, $P_{\text{feas}} \simeq 0$ for
$N \rightarrow \infty$. This follows from
\Cref{sec:theoreticalanalysis:eq:Pfeasapprox1}. By the cone boundary
equation (\Cref{sec:problemalgorithm:eq:coneconstraint}),
a parental individual $(x^{(g)},r^{(g)})^T$ is on the cone boundary for
$r^{(g)}=x^{(g)}/\sqrt{\xi}$. Using this together with
\Cref{sec:theoreticalanalysis:eq:sigmanormalized} and
\Cref{sec:theoreticalanalysis:eq:Pfeasapprox1} yields
\begin{equation}
  P_{\text{feas}} \simeq \Phi\left[
    N\left(\frac{{\sigma^{(g)}}^*}{r^{(g)}}
    (r^{(g)}-\bar{r})
    \right)
    \right].
\end{equation}
By taking into account \Cref{sec:theoreticalanalysis:eq:approximatedrmean},
\begin{equation}
  P_{\text{feas}} \simeq \Phi\left[
    N{\sigma^{(g)}}^*\left(
    1-\sqrt{1 +
        \frac{{{{\sigma^{(g)}}^*}}^2}{N}\left(1 - \frac{1}{N}\right)}
    \right)
    \right]
\end{equation}
follows. If $N{\sigma^{(g)}}^*$ is sufficiently large,
$P_{\text{feas}} \simeq 0$.

For the distance ratio $\frac{r^{(g)}}{x^{(g)}}$, one observes
that it approaches a stationary state value
$\left(\frac{r}{x}\right)_{ss} :=
\lim_{g \rightarrow \infty}\frac{r^{(g)}}{x^{(g)}}$.
This can be expressed
with the condition
$\frac{r^{(g)}}{x^{(g)}}
= \frac{r^{(g+1)}}{x^{(g+1)}}
= \left(\frac{r}{x}\right)_{ss}$
for sufficiently large values of $g$.
Making use of the progress rates
(\Crefrange{sec:theoreticalanalysis:eq:varphix}
           {sec:theoreticalanalysis:eq:varphirnormalized}),
$\left(\frac{r}{x}\right)_{ss} =
\left(\frac{r}{x}\right)_{ss}
\frac{\left(1-\frac{{\varphi_r}_{ss}^*}{N}\right)}
     {\left(1-\frac{{\varphi_x}_{ss}^*}{N}\right)}$
follows, which implies
\begin{equation}
  {\varphi_r}_{ss}^*={\varphi_x}_{ss}^*.
  \label{sec:theoreticalanalysis:eq:steadystatedist1}
\end{equation}

The normalized mutation strength should be constant on
average in the steady state for a
continuous decrease towards the optimizer. That is,
the definition of the steady state normalized mutation strength reads
$\sigma_{ss}^* := \lim_{g \rightarrow \infty} {\sigma^{(g)}}^*$.
Expressed as a condition, it can be stated as
${\sigma^{(g)}}^* = {\sigma^{(g + 1)}}^* = \sigma_{ss}^*$.

Considering the case of $P_{\text{feas}} \simeq 0$, use of the
infeasible case approximations
(the infeasible part of
\Cref{sec:theoreticalanalysis:eq:varphixnormalizedcombined}
and the infeasible part
\Cref{sec:theoreticalanalysis:eq:varphirnormalizedcombinedmaintext})
for handling
\Cref{sec:theoreticalanalysis:eq:steadystatedist1}, results in
\begin{align}
  &\begin{multlined}
  N\bigg(
  1 -
  \left(\frac
  {x}
  {\sqrt{\xi}r}\right)_{ss}
  \left(1-\frac{{{\varphi_x^*}_{ss}}_{\text{infeas}}}{N}\right)
  \sqrt{\frac{1+\frac{{{\sigma_{ss}^*}}^2}{\mu N}}
             {1+\frac{{{\sigma_{ss}^*}}^2}{N}}}
  \bigg)
  ={{\varphi_x^*}_{ss}}_{\text{infeas}}.
  \end{multlined}
\end{align}
This can subsequently be rewritten to
\begin{align}
  &\begin{multlined}
  \left(\frac
  {x}
  {\sqrt{\xi}r}\right)_{ss}
  =
  \frac{1}{\sqrt{\frac{1+\frac{{{\sigma_{ss}^*}}^2}{\mu N}}
      {1+\frac{{{\sigma_{ss}^*}}^2}{N}}}}.
  \end{multlined}
  \label{sec:theoreticalanalysis:eq:steadystatedist2}
\end{align}
For $P_{\text{feas}} \simeq 0$, the infeasible case
approximations can be used. Insertion of
\Cref{sec:theoreticalanalysis:eq:steadystatedist2}
into the infeasible part of
\Cref{sec:theoreticalanalysis:eq:varphixnormalizedcombined}
assuming the expected $\sigma_{ss}^*$ steady state
together with
\Cref{sec:theoreticalanalysis:eq:steadystatedist1}
and $\frac{1}{\xi}\sqrt{
\frac{1+\frac{{{\sigma_{ss}^*}}^2}{2N}}
     {1+\frac{{{\sigma_{ss}^*}}^2}{N}}} \simeq \frac{1}{\xi}$
yields
\begin{align}
  &\begin{multlined}
  {\varphi_r^*}_{ss} = {\varphi_x^*}_{ss}
  \approx
  \frac{N}{1+\xi}
  \left(1-\sqrt{\frac{1+\frac{{{\sigma_{ss}^*}}^2}{\mu N}}
                     {1+\frac{{{\sigma_{ss}^*}}^2}{N}}}
    \sqrt{1+\frac{{{\sigma_{ss}^*}}^2}{N}}\right)
    +\frac{{\sigma_{ss}^*} c_{\mu/\mu,\lambda}}{\sqrt{1+\xi}}
    \sqrt{\frac{1+\frac{{{\sigma_{ss}^*}}^2}{\mu N}}
               {1+\frac{{{\sigma_{ss}^*}}^2}{N}}}
  \end{multlined}
  \label{sec:theoreticalanalysis:eq:varphixnormalizedss1}\\
  &\begin{multlined}
  \phantom{{\varphi_x^*}_{ss}}
  =
  \frac{N}{1+\xi}
  \left(1-\sqrt{1+\frac{{{\sigma_{ss}^*}}^2}{\mu N}}\right)
    +\frac{{\sigma_{ss}^*} c_{\mu/\mu,\lambda}}{\sqrt{1+\xi}}
    \sqrt{\frac{1+\frac{{{\sigma_{ss}^*}}^2}{\mu N}}
               {1+\frac{{{\sigma_{ss}^*}}^2}{N}}}
  \end{multlined}
  \label{sec:theoreticalanalysis:eq:varphixnormalizedss2}\\
  &\begin{multlined}
  \phantom{{\varphi_x^*}_{ss}}
  \simeq
  \frac{N}{1+\xi}
  \left(1-\left(1+\frac{{{\sigma_{ss}^*}}^2}{2 \mu N}\right)\right)
    +\frac{{\sigma_{ss}^*} c_{\mu/\mu,\lambda}}{\sqrt{1+\xi}}
  \end{multlined}
  \label{sec:theoreticalanalysis:eq:varphixnormalizedss3}\\
  &\begin{multlined}
  \phantom{{\varphi_x^*}_{ss}}
  =
  \frac{N}{1+\xi}
  \left(-\frac{{{\sigma_{ss}^*}}^2}{2 \mu N}\right)
  +\frac{{\sigma_{ss}^*} c_{\mu/\mu,\lambda}}{\sqrt{1+\xi}}.
  \end{multlined}
  \label{sec:theoreticalanalysis:eq:varphixnormalizedss4}
\end{align}
$\sqrt{\frac{1+\frac{{{\sigma_{ss}^*}}^2}{\mu N}}
            {1+\frac{{{\sigma_{ss}^*}}^2}{N}}} \simeq 1$
for $N \gg {\sigma_{ss}^*}^2$ has been used from
\Cref{sec:theoreticalanalysis:eq:varphixnormalizedss2}
to
\Cref{sec:theoreticalanalysis:eq:varphixnormalizedss3}.
In addition, a Taylor
expansion with cut-off after the linear term
has been applied to $\sqrt{1+\frac{{{\sigma_{ss}^*}}^2}{\mu N}}$.

A steady state expression for
$-\frac{N\varphi_x^{(g)}}{\overline{{\sigma^{(g)}}^*}\,\overline{r^{(g)}}}$
is derived next. With
\Cref{sec:theoreticalanalysis:eq:varphixnormalized}
and
\Cref{sec:theoreticalanalysis:eq:varphixnormalizedss4},
\begin{equation}
  \left(-\frac{N\varphi_x}
       {{\sigma}^* r}\right)_{ss}
       =
       -\left(\frac{x}{r}\right)_{ss}
       \frac{{\varphi_x^*}_{ss}}{{\sigma_{ss}^*}}
\end{equation}
can be derived.
Use of
\Cref{sec:theoreticalanalysis:eq:steadystatedist2}
for the fraction $\left(\frac{x}{r}\right)_{ss}$
results in
\newcommand{\sppassxterm}
           {-\sqrt{\frac{\xi}{\frac{1+\frac{{{\sigma_{ss}^*}}^2}{\mu N}}
                 {1+\frac{{{\sigma_{ss}^*}}^2}{N}}}}
             \frac{{\varphi_x^*}_{ss}}{{\sigma_{ss}^*}}}
\begin{equation}
  \left(-\frac{N\varphi_x}
       {\sigma^*r}\right)_{ss}
       =
       \sppassxterm{}.
  \label{sec:theoreticalanalysis:eq:ssxterm}
\end{equation}

Similarly, a steady state expression for
$-\frac{N\varphi_{r^2}^{(g)}}
{2{\overline{{\sigma^{(g)}}^*}}\,{{\overline{r^{(g)}}}^2}}$
can be derived. Considering the infeasible case (because in the steady state
$P_{\text{feas}} \simeq 0$) of
\Cref{sec:theoreticalanalysis:eq:progressratersquared},
we have
\begin{equation}
  \left(-\frac{N{\varphi_{r^2}}}{2{\sigma^*}{{r}^2}}\right)_{ss}
  =
  -\frac{N}{2\sigma_{ss}^*}
  \left[1-
  \left(\frac{1}{\xi r^2}\right)_{ss}
  \mathrm{E}\left[{\langle q \rangle}_{\text{infeas}}\right]^2
  \left(\frac{1+\frac{{\sigma_{ss}^*}^2}{\mu N}}
       {1+\frac{{\sigma_{ss}^*}^2}{N}}\right)
  \right].
  \label{sec:theoreticalanalysis:eq:ssrsquaredtermhelper}
\end{equation}
According to
\Cref{sec:problemalgorithm:alg:es:bestq,%
sec:problemalgorithm:alg:es:x}
of
\Cref{sec:problemalgorithm:alg:es},
$\mathrm{E}\left[{\langle q \rangle}\right]
=\mathrm{E}\left[x^{(g+1)}\right]
=\overline{x^{(g+1)}}$.
Hence,
\Cref{sec:theoreticalanalysis:eq:ssrsquaredtermhelper}
can be rewritten using
\Cref{sec:theoreticalanalysis:eq:meanvalueiterativesystemx}
for the infeasible $\langle q \rangle$ case
$\mathrm{E}\left[{\langle q \rangle}_{\text{infeas}}\right]$
and
\Cref{sec:theoreticalanalysis:eq:steadystatedist2}
for $\left(\frac{x^2}{\xi r^2}\right)_{ss}$,
resulting in
\newcommand{\sppassrsquaredterm}
           {-\frac{N}{2\sigma_{ss}^*}
             \left[1-
             \left(1-\frac{{\varphi_x^*}_{ss}}{N}\right)^2\right]}
\begin{equation}
  \left(-\frac{N{\varphi_{r^2}}}{2{\sigma^*}{{r}^2}}\right)_{ss}
  =
  \sppassrsquaredterm{}.
  \label{sec:theoreticalanalysis:eq:ssrsquaredterm}
\end{equation}

Using
\Cref{sec:theoreticalanalysis:eq:ssxterm}
and
\Cref{sec:theoreticalanalysis:eq:ssrsquaredterm},
steady state expressions for
\Crefrange{sec:theoreticalanalysis:eq:summarys1}
{sec:theoreticalanalysis:eq:summarysigmanormalized}
can be derived. Requiring
$\overline{s_1^{(g+1)}}=\overline{s_1^{(g)}}={s_1}_{ss}$
in
\Cref{sec:theoreticalanalysis:eq:summarys1}
using
\Cref{sec:theoreticalanalysis:eq:ssxterm}
yields
\begin{equation}
  {s_1}_{ss}=-\frac{\sqrt{\mu c(2-c)}}{c}
  \left(
  \sqrt{\frac{\xi}{\frac{1+\frac{{\sigma_{ss}^*}^2}{\mu N}}
      {1+\frac{{\sigma_{ss}^*}^2}{N}}}}
  \frac{{\varphi_x^*}_{ss}}{\sigma_{ss}^*}
  \right).
  \label{sec:theoreticalanalysis:eq:s1ss}
\end{equation}
Analogously,
requiring
$\overline{s_{\odot}^{(g+1)}}=\overline{s_{\odot}^{(g)}}={s_{\odot}}_{ss}$
in
\Cref{sec:theoreticalanalysis:eq:summarysr}
using
\Cref{sec:theoreticalanalysis:eq:ssrsquaredterm}
results in
\begin{equation}
  {s_{\odot}}_{ss}=
  \frac{\sqrt{\mu c(2-c)}
    \left(
      \sppassrsquaredterm{} + \frac{\sigma_{ss}^*}{2\mu}
      \right)}
       {c - (1-c)\frac{\sigma_{ss}^*}{N}
         \left[\sppassrsquaredterm{} - \frac{\sigma_{ss}^*}{2\mu}
           \right]}.
  \label{sec:theoreticalanalysis:eq:srss}
\end{equation}
In the same way, setting
$\overline{||\mathbf{s}^{(g+1)}||^2}=
\overline{||\mathbf{s}^{(g)}||^2}=||\mathbf{s}||^2_{ss}$ in
\Cref{sec:theoreticalanalysis:eq:summaryssquared}
using
\Cref{sec:theoreticalanalysis:eq:ssxterm}
and
\Cref{sec:theoreticalanalysis:eq:ssrsquaredterm}
gives
\begin{equation}
  \begin{multlined}
    ||\mathbf{s}||^2_{ss}=
    N
    -\frac{2(1-c)\sqrt{\mu c(2-c)}}{-2c+c^2}\\\times
    \left[
      {s_1}_{ss}\left(\sppassxterm{}\right)
      +{s_{\odot}}_{ss}\left(\sppassrsquaredterm{}
      -\frac{\sigma_{ss}^*}{2\mu}\right)
    \right].
  \end{multlined}
  \label{sec:theoreticalanalysis:eq:ssquaredlengthss}
\end{equation}
For the mutation strength,
\begin{equation}
  \frac{{\sigma^{(g+1)}}^*r^{(g+1)}}{N}
  =
  \frac{{\sigma^{(g)}}^*r^{(g)}}{N}
  \exp\left(\frac{||\mathbf{s}^{(g+1)}||^2 - N}{2DN}\right)
  \label{sec:theoreticalanalysis:eq:sigmass1}
\end{equation}
follows from
\Cref{sec:problemalgorithm:alg:es:replacesigma}
of
\Cref{sec:problemalgorithm:alg:es}
with the use of
\Cref{sec:theoreticalanalysis:eq:sigmanormalized}.
Rewriting
\Cref{sec:theoreticalanalysis:eq:sigmass1}
and using
\Cref{sec:theoreticalanalysis:eq:rfraction1}
together with
\Cref{sec:theoreticalanalysis:eq:mutationlln}
for the fraction $r^{(g)}/r^{(g+1)}$, we have
\begin{align}
  {\sigma^{(g+1)}}^*
  \simeq
  {\sigma^{(g)}}^*
  &\frac{1}{\sqrt{1
      + \frac{2{{\sigma^{(g)}}^*}}{N}z_\odot^{(g)}
      + \frac{{{\sigma^{(g)}}^*}^2}{\mu N}}}
  \exp\left(\frac{||\mathbf{s}^{(g+1)}||^2 - N}{2DN}\right)\\
  {{\sigma^{(g+1)}}^*}^2
  \left(1
  + \frac{2{{\sigma^{(g)}}^*}}{N}z_\odot^{(g)}
  + \frac{{{\sigma^{(g)}}^*}^2}{\mu N}\right)
  &\simeq
  {{\sigma^{(g)}}^*}^2
  \exp\left(\frac{||\mathbf{s}^{(g+1)}||^2 - N}{DN}\right).
  \label{sec:theoreticalanalysis:eq:sigmass2}
\end{align}
Use of the Taylor expansion $\exp(x) \simeq 1 + x$
(around zero and neglecting terms of
quadratic and higher order) results in
\begin{equation}
  {{\sigma^{(g+1)}}^*}^2
  \left(1
  + \frac{2{{\sigma^{(g)}}^*}}{N}z_\odot^{(g)}
  + \frac{{{\sigma^{(g)}}^*}^2}{\mu N}\right)
  \simeq
  {{\sigma^{(g)}}^*}^2
  \left(1 + \frac{||\mathbf{s}^{(g+1)}||^2 - N}{DN}\right).
  \label{sec:theoreticalanalysis:eq:sigmass3}
\end{equation}
Computing the expectation of
\Cref{sec:theoreticalanalysis:eq:sigmass3}
and requiring
$\overline{{\sigma^{(g+1)}}^*}=\overline{{\sigma^{(g)}}^*}=\sigma_{ss}^*$,
we get
\begin{align}
  {\sigma_{ss}^*}^2
  \left[\frac{1}{2} + \frac{1}{N}
  \left(\sigma_{ss}^*\mathrm{E}[z_\odot]
  +\frac{{\sigma_{ss}^*}^2}{2\mu}\right)\right]
  &=
  {\sigma_{ss}^*}^2
  \left(
  \frac{1}{2} +\frac{||\mathbf{s}||^2_{ss} - N}{2DN}\right)\\
  \sigma_{ss}^*\mathrm{E}[z_\odot]+\frac{{\sigma_{ss}^*}^2}{2\mu}
  &=
  \frac{||\mathbf{s}||^2_{ss} - N}{2D}.
  \label{sec:theoreticalanalysis:eq:sigmass4}
\end{align}
Usage of
\Cref{sec:theoreticalanalysis:eq:zodotexpectation}
together with the steady state expression derived in
\Cref{sec:theoreticalanalysis:eq:ssrsquaredterm}
for $\mathrm{E}[z_\odot]$ results in
\begin{align}
  \sigma_{ss}^*
  \left(\sppassrsquaredterm{}-\frac{{\sigma_{ss}^*}}{2\mu}\right)
  +\frac{{\sigma_{ss}^*}^2}{2\mu}
  =
  \frac{||\mathbf{s}||^2_{ss} - N}{2D}.
  \label{sec:theoreticalanalysis:eq:sigmass}
\end{align}
Consideration of
\Cref{sec:theoreticalanalysis:eq:varphixnormalizedss4,%
sec:theoreticalanalysis:eq:s1ss,%
sec:theoreticalanalysis:eq:srss,%
sec:theoreticalanalysis:eq:ssquaredlengthss}
allows numerically solving
\Cref{sec:theoreticalanalysis:eq:sigmass}
for $\sigma_{ss}^*$.

\subsubsection{Derivation of Closed-Form Approximations for the Steady State
  with the Assumptions $c = O\left(\frac{1}{\sqrt{N}}\right)$
  and $N \rightarrow \infty$}
\label{sec:theoreticalanalysis:subsubsec:steadystateclosedform}
The goal of this section is to simplify the expressions derived in
\Cref{sec:theoreticalanalysis:subsubsec:steadystatenumerical}
further using additional asymptotic assumptions in order to arrive at
closed-form steady state approximations.

The expression derived for
$\left(-\frac{N{\varphi_{r^2}}}{2{\sigma^*}{{r}^2}}\right)_{ss}$ as
\Cref{sec:theoreticalanalysis:eq:ssrsquaredterm}
is simplified further yielding
\begin{align}
  \left(-\frac{N{\varphi_{r^2}}}{2{\sigma^*}{{r}^2}}\right)_{ss}
  &=
  \sppassrsquaredterm{}
  =
  -\frac{N}{2{\sigma^*}_{ss}}
  \left(2\frac{{\varphi_x^*}_{ss}}{N}-
  \frac{{{\varphi_x^*}_{ss}}^2}{N^2}\right)
  \label{sec:theoreticalanalysis:eq:ssrsquaredterm2_2}\\
  &=
  -\frac{{\varphi_x^*}_{ss}}{{\sigma^*}_{ss}}+
  \frac{{{\varphi_x^*}_{ss}}^2}{2{\sigma^*}_{ss}N}
  \simeq
  -\frac{{\varphi_x^*}_{ss}}{{\sigma^*}_{ss}}.
  \label{sec:theoreticalanalysis:eq:ssrsquaredterm2_4}
\end{align}
In
\Cref{sec:theoreticalanalysis:eq:ssrsquaredterm2_4},
${\sigma^*}_{ss}N \gg {{\varphi_x^*}_{ss}}^2$ has been assumed
and therefore the second summand has been neglected.

Insertion of
\Cref{sec:theoreticalanalysis:eq:ssrsquaredterm2_4}
into
\Cref{sec:theoreticalanalysis:eq:sigmass}
replacing $\sppassrsquaredterm{}$
yields (after simplification)
\begin{align}
  -{\varphi_x^*}_{ss}
  =
  \frac{||\mathbf{s}||^2_{ss} - N}{2D}
  \label{sec:theoreticalanalysis:eq:sigmassinserted}
\end{align}
for the steady state mutation strength equation.
\Cref{sec:theoreticalanalysis:eq:ssrsquaredterm2_4}
can also be inserted into
\Cref{sec:theoreticalanalysis:eq:srss}
replacing $\sppassrsquaredterm{}$.
This results in
\begin{align}
  {s_{\odot}}_{ss}&\simeq
  \frac{\sqrt{\mu c(2-c)}
    \left[
      -\frac{{\varphi_x^*}_{ss}}{{\sigma^*}_{ss}}
      + \frac{\sigma_{ss}^*}{2\mu}
      \right]}
       {c - (1-c)\frac{\sigma_{ss}^*}{N}
         \left[-\frac{{\varphi_x^*}_{ss}}{{\sigma^*}_{ss}}
           - \frac{\sigma_{ss}^*}{2\mu}
           \right]}
  =
  \frac{\sqrt{\mu c(2-c)}
      \left(-\frac{{\varphi_x^*}_{ss}}{{\sigma}^*_{ss}} +
      \frac{\sigma_{ss}^*}{2\mu}\right)}
       {c - (1-c)\left(-\frac{{\varphi_x^*}_{ss}}{N}
           - \frac{{\sigma_{ss}^*}^2}{2\mu N}
           \right)}.
 \label{sec:theoreticalanalysis:eq:srss2_2}\\
 &=
 \frac{\sqrt{\mu c(2-c)}
   \left(-\frac{{\varphi_x^*}_{ss}}{{\sigma}^*_{ss}} +
   \frac{\sigma_{ss}^*}{2\mu}\right)}
      {c + \frac{{\varphi_x^*}_{ss}}{N}
        + \frac{{\sigma_{ss}^*}^2}{2\mu N}
        - \frac{c{\varphi_x^*}_{ss}}{N}
        - \frac{c{\sigma_{ss}^*}^2}{2\mu N}}.
  \label{sec:theoreticalanalysis:eq:srss2_3}
\end{align}
With the assumptions $N \rightarrow \infty$ and
$c = O\left(\frac{1}{\sqrt{N}}\right)$,
the expression
$\frac{{\varphi_x^*}_{ss}}{N}
+ \frac{{\sigma_{ss}^*}^2}{2\mu N}
- \frac{c{\varphi_x^*}_{ss}}{N}
- \frac{c{\sigma_{ss}^*}^2}{2\mu N}$
is an order of magnitude smaller than c
and can therefore be neglected w.r.t. c. Hence,
\Cref{sec:theoreticalanalysis:eq:srss2_3}
simplifies to
\begin{align}
  {s_{\odot}}_{ss}&\simeq
  \frac{\sqrt{\mu c(2-c)}}{c}
      \left(-\frac{{\varphi_x^*}_{ss}}{\sigma^*_{ss}} +
      \frac{\sigma_{ss}^*}{2\mu}\right).
  \label{sec:theoreticalanalysis:eq:srss2_4}
\end{align}
Similarly,
\Cref{sec:theoreticalanalysis:eq:ssrsquaredterm2_4}
inserted into
\Cref{sec:theoreticalanalysis:eq:ssquaredlengthss}
replacing $\sppassrsquaredterm{}$
results in
\begin{equation}
  \begin{multlined}
    ||\mathbf{s}||^2_{ss}=
    N
    -\frac{2(1-c)\sqrt{\mu c(2-c)}}{-2c+c^2}
    \left[
      {s_1}_{ss}\left(\sppassxterm{}\right)
      +{s_{\odot}}_{ss}\left(-\frac{{\varphi_x^*}_{ss}}{\sigma_{ss}}
      -\frac{\sigma_{ss}^*}{2\mu}\right)
    \right].
  \end{multlined}
  \label{sec:theoreticalanalysis:eq:ssquaredlengthssinserted}
\end{equation}
Insertion of
\Cref{sec:theoreticalanalysis:eq:srss2_4}
and
\Cref{sec:theoreticalanalysis:eq:s1ss}
into
\Cref{sec:theoreticalanalysis:eq:ssquaredlengthssinserted}
yields (after straight-forward simplification)
\begin{align}
  ||\mathbf{s}||^2_{ss}
  &\simeq
  N + \frac{2(1-c)\mu}{c}
  \left[
    \left(
    \xi\left(
    \frac{1+\frac{{\sigma_{ss}^*}^2}{N}}
         {1+\frac{{\sigma_{ss}^*}^2}{\mu N}}\right)
         + 1\right)
    \frac{{\varphi_x^*}_{ss}^2}{{\sigma^*_{ss}}^2}
    -\frac{{\sigma_{ss}^*}^2}{4\mu^2}
    \right].
  \label{sec:theoreticalanalysis:eq:ssquaredlengthss2_2}
\end{align}
$\xi\left(\frac{1+\frac{{\sigma_{ss}^*}^2}{N}}
{1+\frac{{\sigma_{ss}^*}^2}{\mu N}}\right) \simeq \xi$
for $N \rightarrow \infty$ allows writing
\begin{align}
  ||\mathbf{s}||^2_{ss}
  &\simeq
  N + \frac{2(1-c)\mu}{c}
  \left[
    \left(\xi + 1\right)
    \frac{{\varphi_x^*}_{ss}^2}{{{\sigma}^*_{ss}}^2}
    -\frac{{\sigma_{ss}^*}^2}{4\mu^2}
    \right]
  \label{sec:theoreticalanalysis:eq:ssquaredlengthss2_3}\\
  &=
  N + \frac{2(1-c)\mu}{c}
  \left[
    \left(
    c_{\mu/\mu,\lambda}^2
    -\frac{{\sigma_{ss}^*} c_{\mu/\mu,\lambda}}{\mu\sqrt{1+\xi}}
    +\frac{{\sigma_{ss}^*}^2}{(1 + \xi) 4 \mu^2}
    \right)
    -\frac{{\sigma_{ss}^*}^2}{4\mu^2}
    \right].
  \label{sec:theoreticalanalysis:eq:ssquaredlengthss2_4}
\end{align}
From
\Cref{sec:theoreticalanalysis:eq:ssquaredlengthss2_3}
to
\Cref{sec:theoreticalanalysis:eq:ssquaredlengthss2_4},
${\varphi_x^*}_{ss}$ has been substituted by
\Cref{sec:theoreticalanalysis:eq:varphixnormalizedss4},
its square has been calculated,
and the resulting expression has been simplified.

With this, insertion of
\Cref{sec:theoreticalanalysis:eq:varphixnormalizedss4}
and
\Cref{sec:theoreticalanalysis:eq:ssquaredlengthss2_4}
into \Cref{sec:theoreticalanalysis:eq:sigmassinserted}
yields the quadratic equation
\begin{equation}
  \frac{{\sigma_{ss}^*}^2}{(1 + \xi) 2 \mu}
  -\frac{{\sigma_{ss}^*} c_{\mu/\mu,\lambda}}{\sqrt{1+\xi}}
  =
  \frac{2(1-c)\mu}{2 c D}
  \left(
    c_{\mu/\mu,\lambda}^2
    -\frac{{\sigma_{ss}^*} c_{\mu/\mu,\lambda}}{\mu\sqrt{1+\xi}}
    +\frac{{\sigma_{ss}^*}^2}{(1 + \xi) 4 \mu^2}
    -\frac{{\sigma_{ss}^*}^2}{4\mu^2}
    \right)
  \label{sec:theoreticalanalysis:eq:sigmassquadraticequation}
\end{equation}
for the steady state normalized mutation strength equation.
By solving
\Cref{sec:theoreticalanalysis:eq:sigmassquadraticequation}
for the positive root (because $\sigma_{ss}^* > 0$) with subsequent
simplification of the result we get
\begin{equation}
  \label{sec:theoreticalanalysis:eq:sigmassquadraticequationsolution}
  \begin{multlined}
    \sigma_{ss}^*=
    \frac{2\mu \sqrt{\xi+1} c_{\mu/\mu,\lambda}\left(
      (c D + c - 1)
      +\sqrt{c^2 \left(D^2+\xi+1\right)-2 c (\xi+1)+\xi+1}
      \right)}{2 c D-c \xi+\xi}
  \end{multlined}
\end{equation}
as an asymptotic ($N \rightarrow \infty$)
closed-form expression for the steady state normalized mutation strength.
Insertion of
$c = 1/\sqrt{N}$ and $D = 1/c = \sqrt{N}$ into
\Cref{sec:theoreticalanalysis:eq:sigmassquadraticequationsolution}
results in the expression
\begin{equation}
  \label{sec:theoreticalanalysis:eq:sigmassquadraticequationsolutionsimplified1}
  \begin{multlined}
    \sigma_{ss}^*=
    \frac{2\mu\sqrt{\xi+1}c_{\mu/\mu,\lambda}
      \left(1 + \frac{1}{\sqrt{N}} - 1 +
      \sqrt{1 + \frac{\xi}{N} + \frac{1}{N} -
        \frac{2\xi}{\sqrt{N}} - \frac{2}{\sqrt{N}} + \xi + 1}\right)}
         {2 - \frac{\xi}{\sqrt{N}} + \xi}.
  \end{multlined}
\end{equation}
Assuming $N \rightarrow \infty$ and $\frac{\xi}{\sqrt{N}} \rightarrow 0$ allows
a further asymptotic simplification of
\Cref{sec:theoreticalanalysis:eq:sigmassquadraticequationsolutionsimplified1}
(neglecting $\frac{1}{\sqrt{N}}$, $\frac{\xi}{N}$, $\frac{1}{N}$,
$\frac{2\xi}{\sqrt{N}}$, and $\frac{\xi}{\sqrt{N}}$)
resulting in
\begin{equation}
  \label{sec:theoreticalanalysis:eq:sigmassquadraticequationsolutionsimplified2}
  \sigma_{ss}^*
  \simeq\frac{2\mu\sqrt{\xi+1}\sqrt{\xi+2}c_{\mu/\mu,\lambda}}{\xi+2}
  =\frac{2\mu\sqrt{\xi+1}c_{\mu/\mu,\lambda}}{\sqrt{\xi+2}}.
\end{equation}
For sufficiently large $\xi$, $\sqrt{\xi + 1} \simeq \sqrt{\xi + 2}$, and
\Cref{sec:theoreticalanalysis:eq:sigmassquadraticequationsolutionsimplified2}
writes $\sigma_{ss}^* \simeq 2 \mu c_{\mu/\mu,\lambda}$.
Back-insertion of
\Cref{sec:theoreticalanalysis:eq:sigmassquadraticequationsolution}
(or
\Cref{sec:theoreticalanalysis:eq:sigmassquadraticequationsolutionsimplified2})
into
\Cref{sec:theoreticalanalysis:eq:steadystatedist2,%
sec:theoreticalanalysis:eq:varphixnormalizedss4,%
sec:theoreticalanalysis:eq:s1ss,%
sec:theoreticalanalysis:eq:srss,%
sec:theoreticalanalysis:eq:ssquaredlengthss}
allows calculating
the steady state distance from the cone boundary,
the normalized steady state progress,
${s_1}_{ss}$,
${s_{\odot}}_{ss}$,
and $||\mathbf{s}||^2_{ss}$.

\begin{figure}
  \centering
  \begin{tabular}{@{\hspace{-0.000\textwidth}}c@{\hspace{-0.000\textwidth}}c@{\hspace{-0.000\textwidth}}c}
    \includegraphics[width=0.30\textwidth]{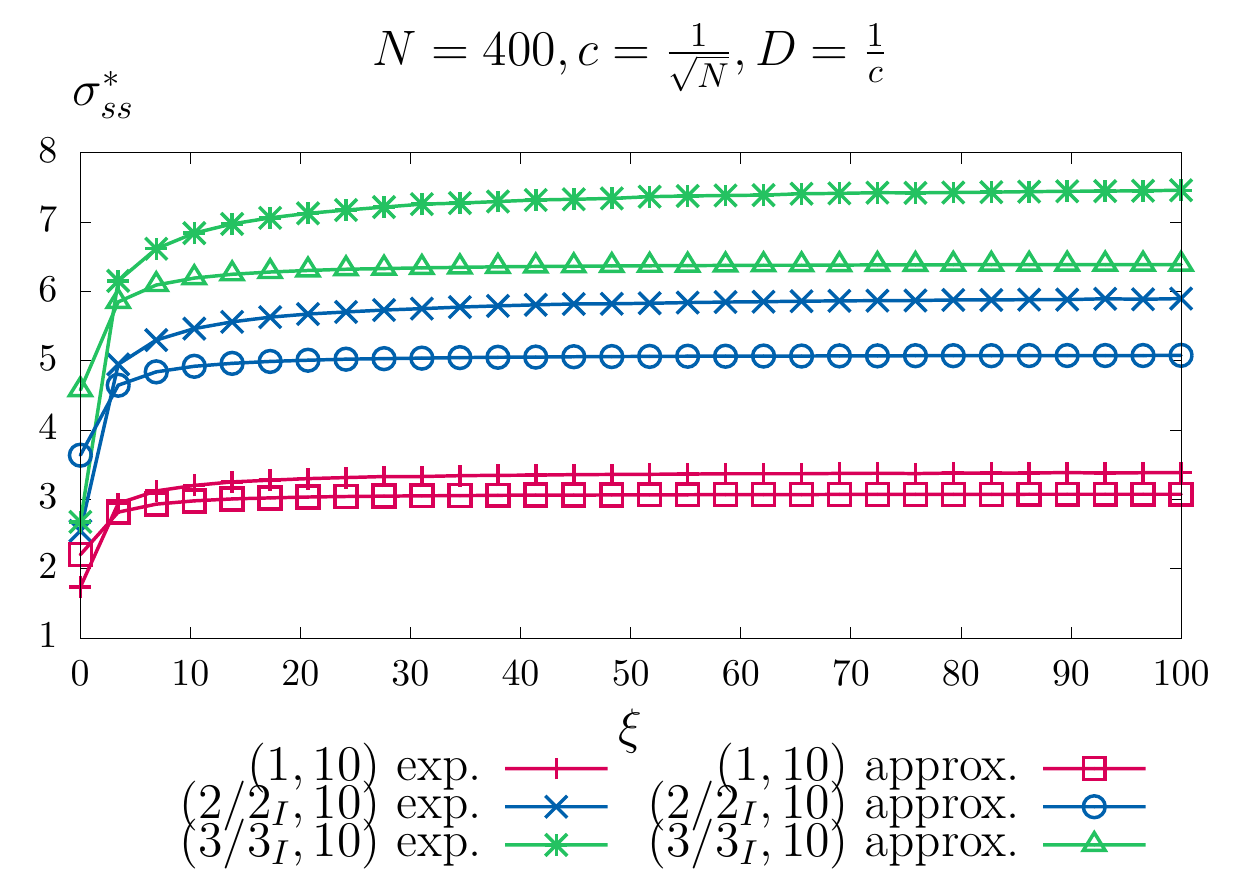}&
    \includegraphics[width=0.30\textwidth]{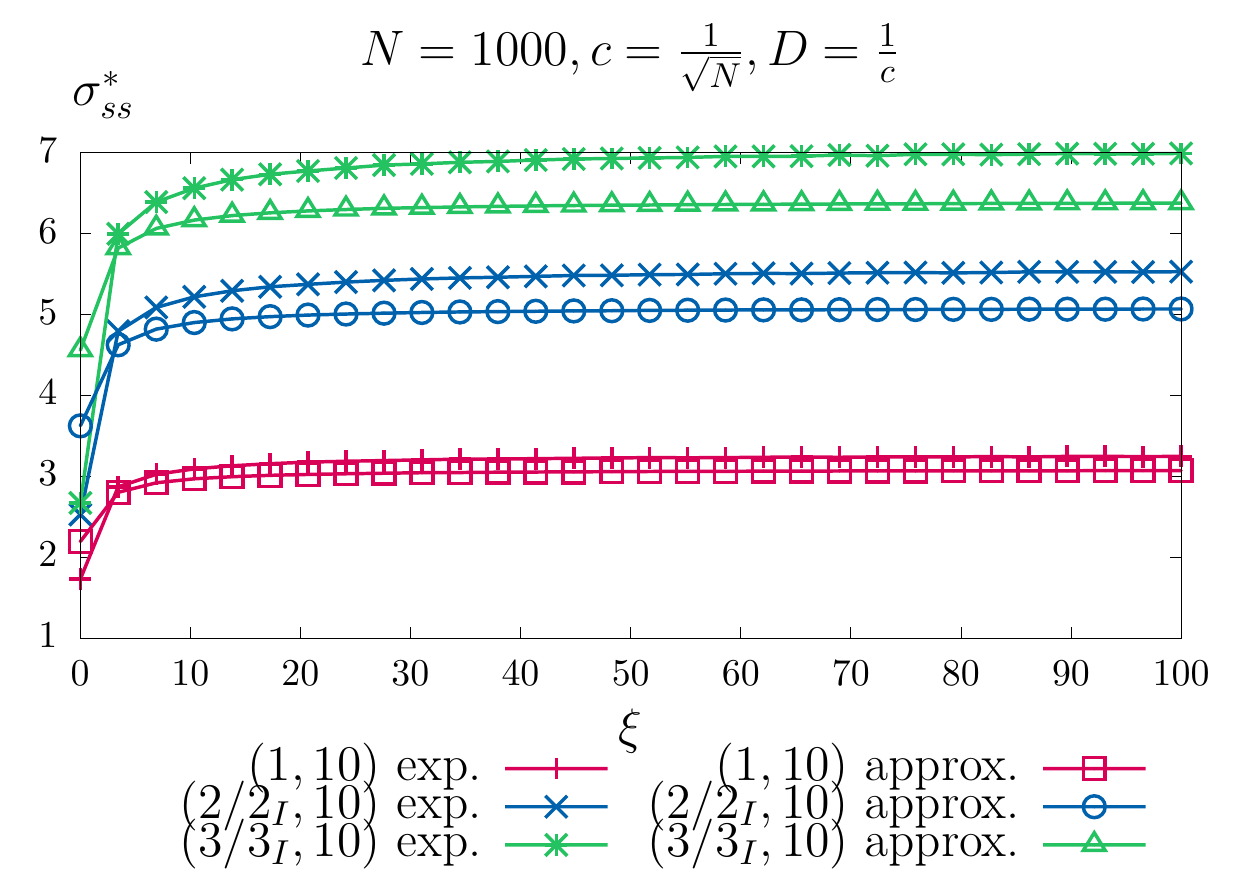}&
    \includegraphics[width=0.30\textwidth]{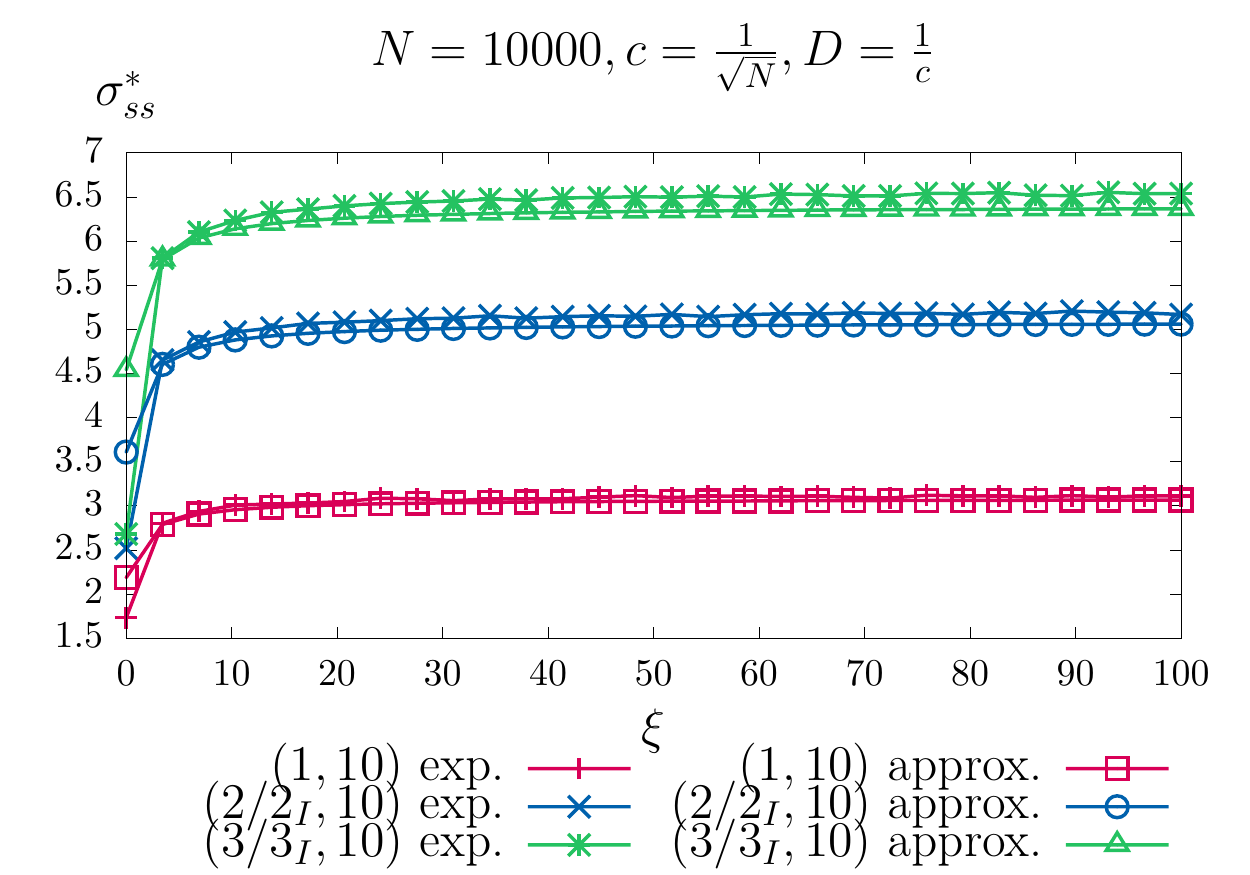}\\
    \includegraphics[width=0.30\textwidth]{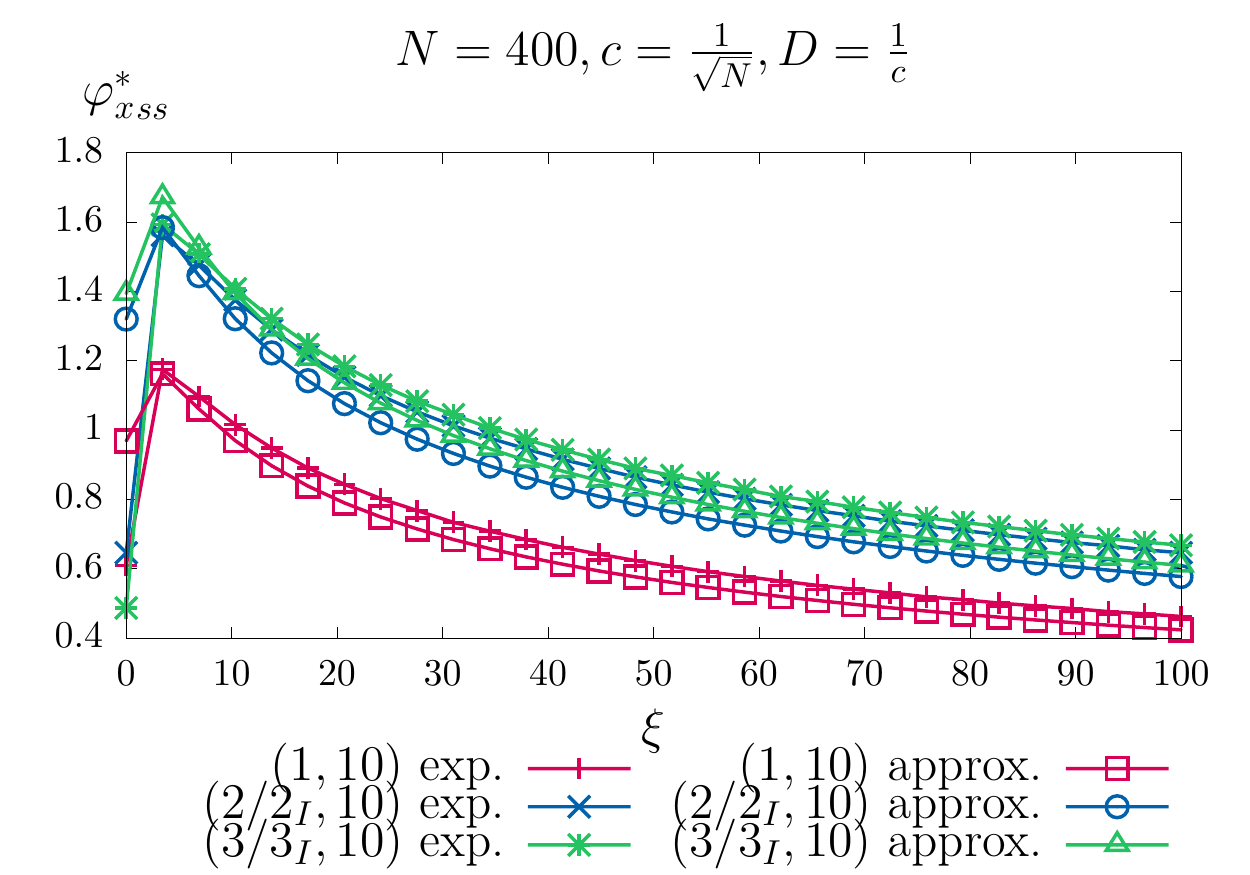}&
    \includegraphics[width=0.30\textwidth]{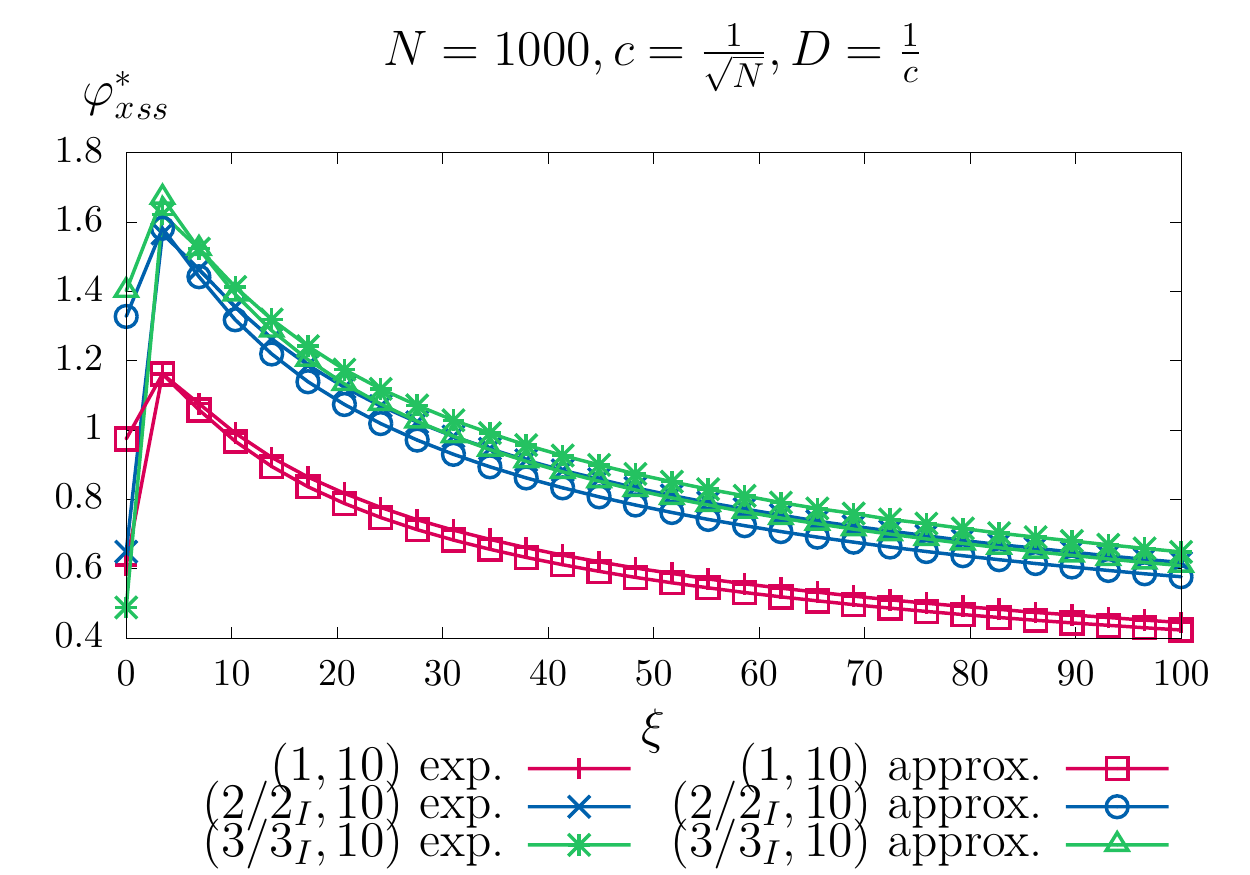}&
    \includegraphics[width=0.30\textwidth]{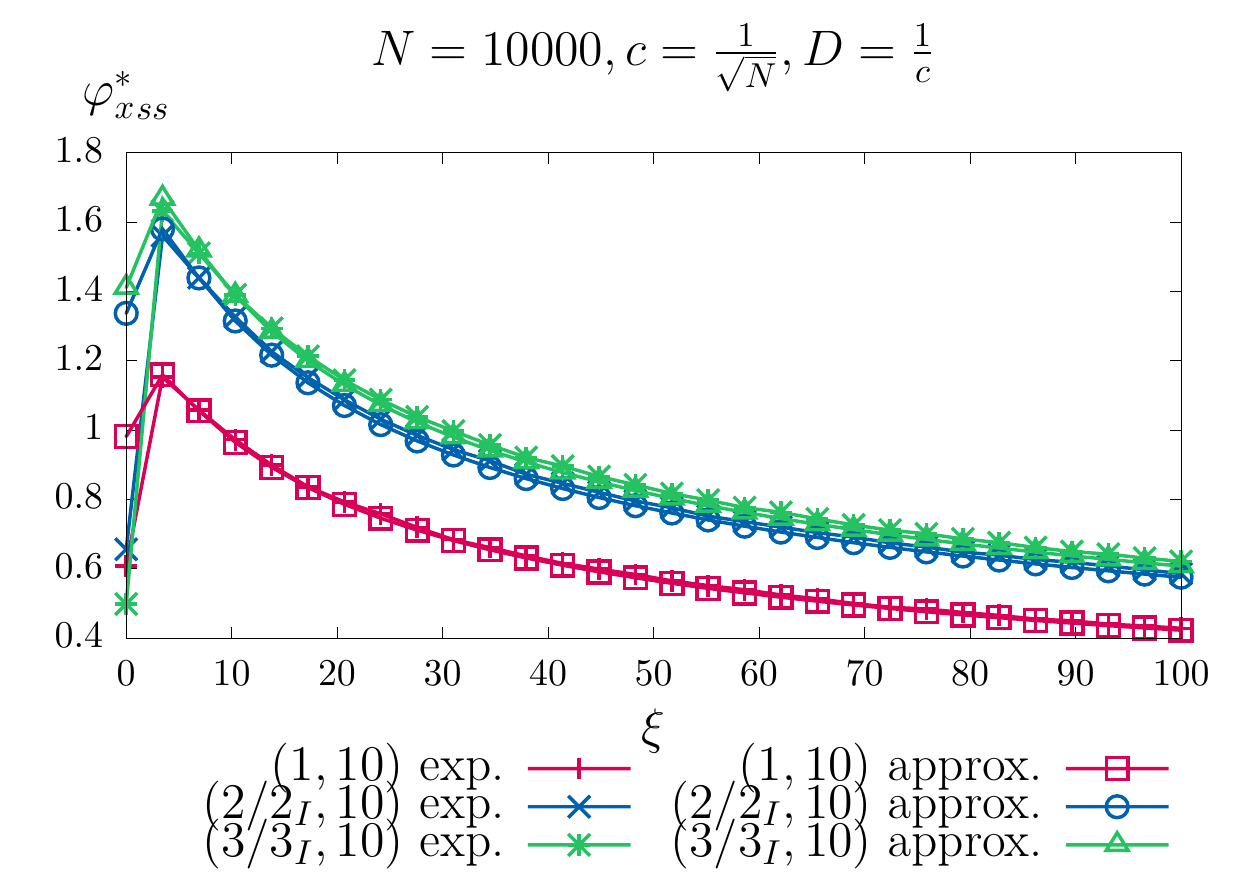}\\
    \includegraphics[width=0.30\textwidth]{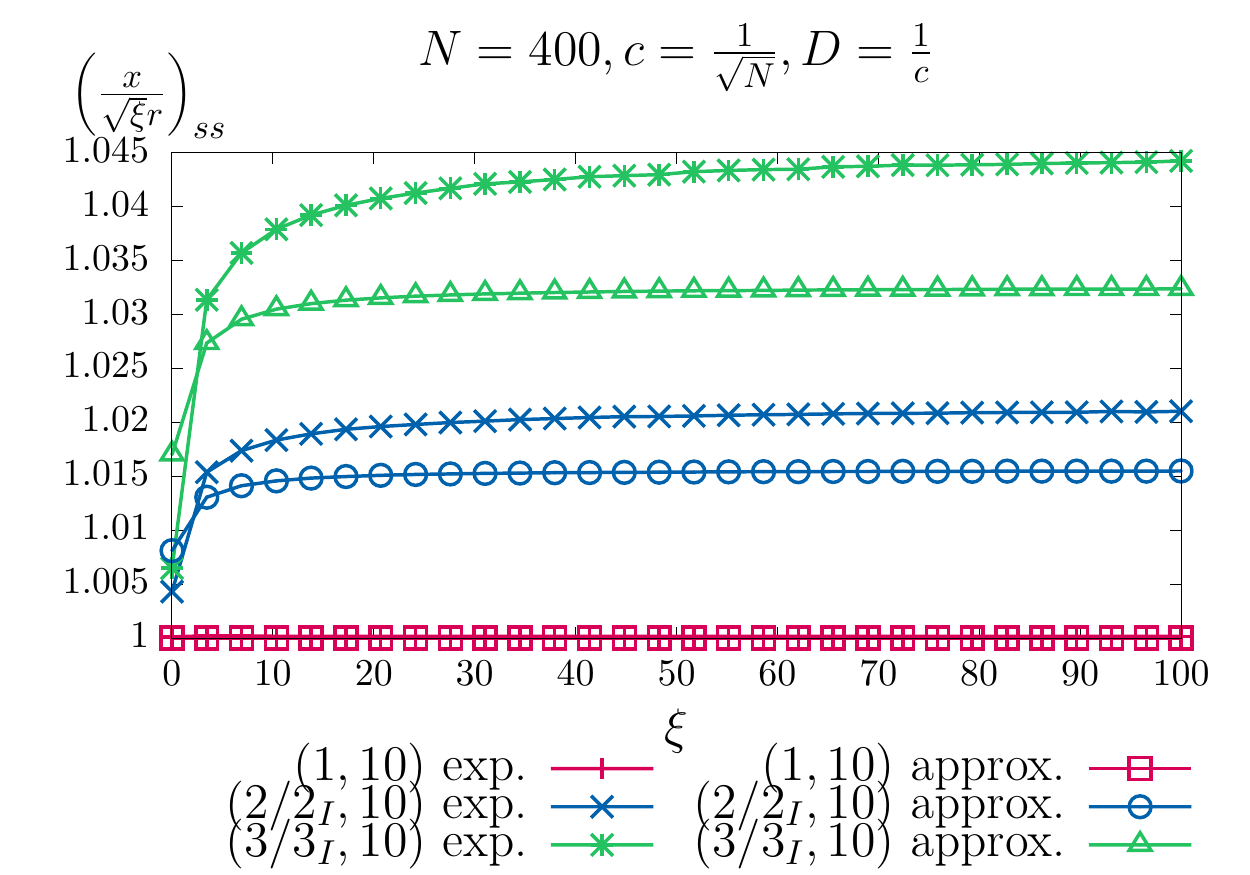}&
    \includegraphics[width=0.30\textwidth]{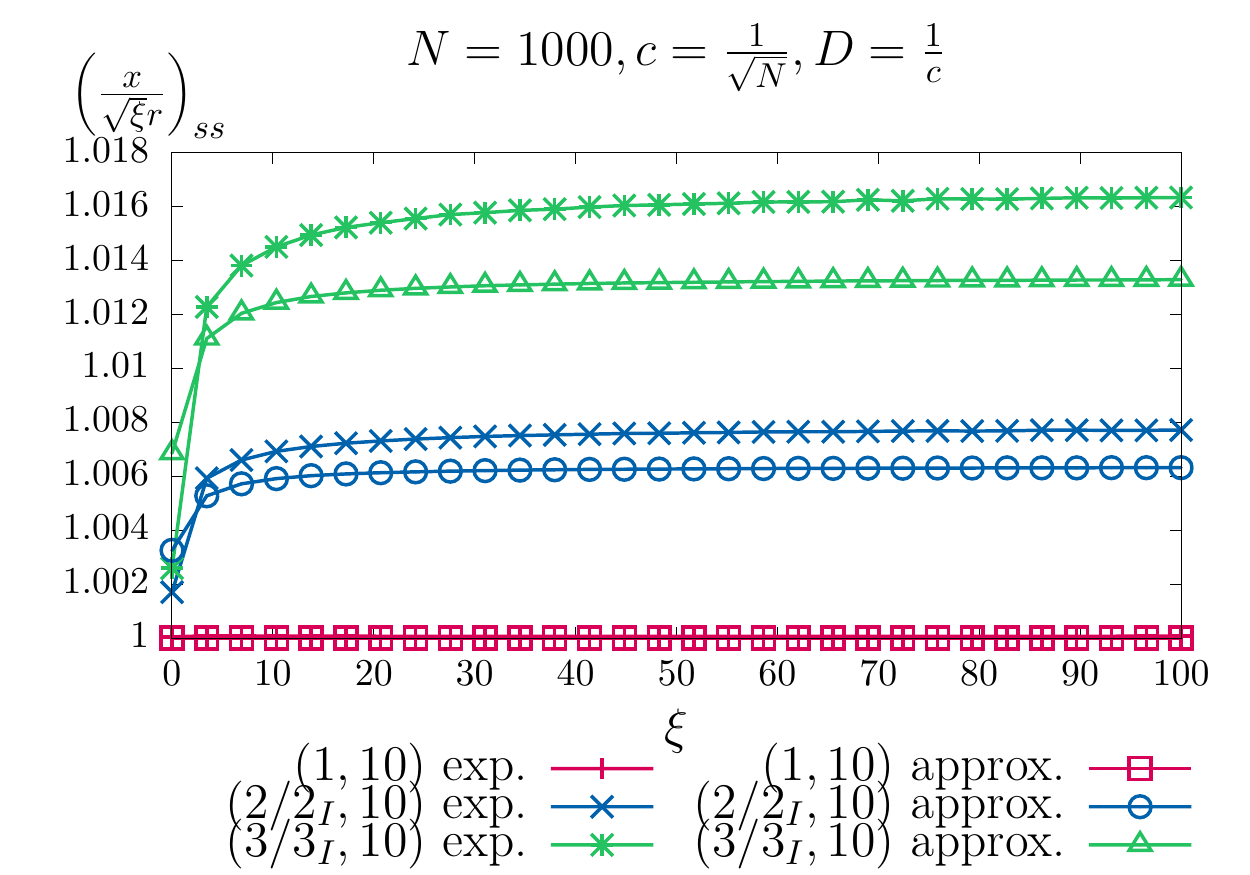}&
    \includegraphics[width=0.30\textwidth]{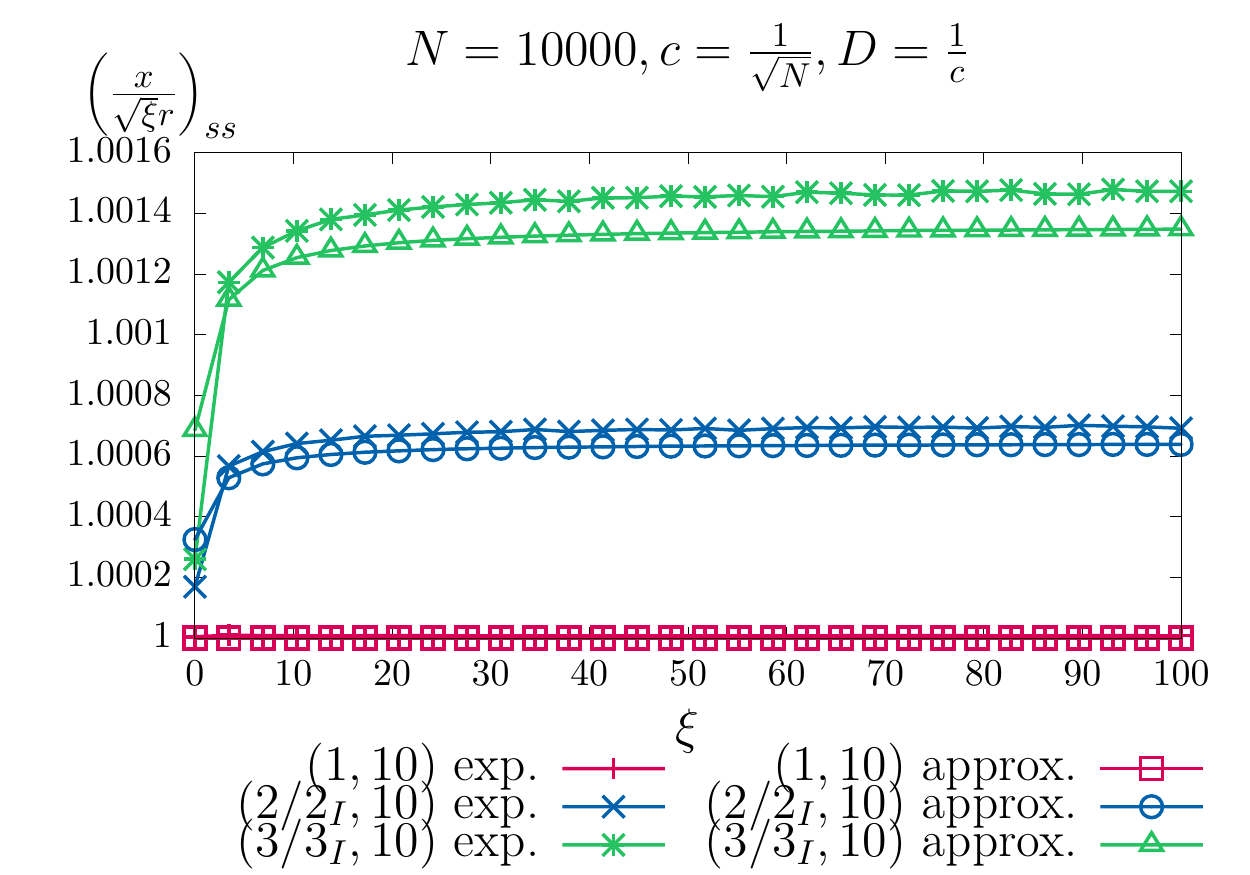}
  \end{tabular}
  \caption
  {Steady state closed-form approximation and real run
  comparison of the $(\mu/\mu_I,\lambda)$-CSA-ES
  with repair by projection
  applied to the conically constrained problem.}
  \label{sec:theoreticalanalysis:fig:steadystatecomparisonclosedform}
\end{figure}

\Cref{sec:theoreticalanalysis:fig:steadystatecomparisonclosedform}
shows plots of the steady state computations. Results computed by
\Cref{sec:theoreticalanalysis:eq:sigmassquadraticequationsolution}
have been compared to real ES runs.
The values for the points denoting the approximations
have been determined by computing the normalized steady state
mutation strength $\sigma_{ss}^*$ using
\Cref{sec:theoreticalanalysis:eq:sigmassquadraticequationsolution}
for different values of $\xi$. The results for
$\varphi_x^*$ and $\varphi_r^*$ have been determined by using the
computed steady state $\sigma_{ss}^*$ values with
\Cref{sec:theoreticalanalysis:eq:varphixnormalizedss4}.
The approximations for $\left(\frac{x}{\sqrt{\xi}r}\right)_{ss}$ have been
determined by evaluating
\Cref{sec:theoreticalanalysis:eq:steadystatedist2}.
The values for the points denoting the experiments have been determined by
computing the averages of the particular values in real ES runs.
The figures show that the derived expressions get better
for larger values of $\xi$ and $N$. Again, the deviations
for small $\xi$ are due to approximations
in the derivation of the local progress rates. The deviations
for small $N$ stem from the use of asymptotic assumptions
$N \rightarrow \infty$.

\subsubsection{Derivation of Closed-Form Approximations for the Steady State
  with the Assumptions $c = O\left(\frac{1}{N}\right)$
  and $N \rightarrow \infty$}
\label{sec:theoreticalanalysis:subsubsec:steadystateclosedform1divN}
In
\Cref{sec:theoreticalanalysis:subsubsec:steadystateclosedform}
it has been assumed that $c=O(\frac{1}{\sqrt{N}})$ from
\Cref{sec:theoreticalanalysis:eq:srss2_3}
to
\Cref{sec:theoreticalanalysis:eq:srss2_4}.
This section presents a derivation for the case $c = O(\frac{1}{N})$.
To this end,
\Cref{sec:theoreticalanalysis:eq:srss2_2}
is rewritten to
\begin{equation}
  {s_{\odot}}_{ss}\simeq
  \frac{\sqrt{\mu c(2-c)}}{c}
    \frac{1}{\sigma_{ss}^*}
    \frac{\left(-{\varphi_x^*}_{ss} +
      \frac{{\sigma_{ss}^*}^2}{2\mu}\right)}
         {\left(1 + \frac{1}{N}(\frac{1}{c}-1)\left({\varphi_x^*}_{ss}
           + \frac{{\sigma_{ss}^*}^2}{2\mu}
           \right)\right)}.
  \label{sec:theoreticalanalysis:eq:srss1divN_1}
\end{equation}
With $c = O(\frac{1}{N})$, we have $\frac{1}{c} \gg 1$. This together with
the assumption $|{\varphi_x^*}_{ss}| \ll \frac{{\sigma_{ss}^*}^2}{2\mu}$ allows
rewriting
\Cref{sec:theoreticalanalysis:eq:srss1divN_1}
to
\begin{equation}
  {s_{\odot}}_{ss}\simeq
  \frac{\sqrt{\mu c(2-c)}}{c}
    \frac{{\sigma_{ss}^*}}
         {2\mu + \frac{{\sigma_{ss}^*}^2}{cN}}.
  \label{sec:theoreticalanalysis:eq:srss1divN_2}
\end{equation}
With the additional assumption
$2\mu \ll \frac{{\sigma_{ss}^*}^2}{cN}$,
\Cref{sec:theoreticalanalysis:eq:srss1divN_2}
simplifies to
\begin{equation}
  {s_{\odot}}_{ss}\simeq
  \frac{\sqrt{\mu c(2-c)}}{c}
    \frac{cN}{{\sigma_{ss}^*}}.
  \label{sec:theoreticalanalysis:eq:srss1divN_3}
\end{equation}
Insertion of
\Cref{sec:theoreticalanalysis:eq:srss1divN_3}
and
\Cref{sec:theoreticalanalysis:eq:s1ss}
into
\Cref{sec:theoreticalanalysis:eq:ssquaredlengthssinserted}
yields
\begin{align}
  ||\mathbf{s}||^2_{ss}
  &\simeq
  N - \frac{2(1-c)\mu c(2-c)}{(-2c+c^2)c}
  \left[
    \xi\left(
    \frac{1+\frac{{\sigma_{ss}^*}^2}{N}}
            {1+\frac{{\sigma_{ss}^*}^2}{\mu N}}\right)
    \frac{{\varphi_x^*}_{ss}^2}{{\sigma^*_{ss}}^2}
    +\frac{cN}{{\sigma_{ss}^*}}
    \left(-\frac{{\varphi_x^*}_{ss}}{{\sigma^*_{ss}}}
    -\frac{{\sigma_{ss}^*}}{2\mu}\right)
    \right]
  \label{sec:theoreticalanalysis:eq:ssquaredlengthss1divN_1}\\
  &=
  N + \frac{2(1-c)\mu}{c}
  \left(
    \xi
    \frac{{\varphi_x^*}_{ss}^2}{{\sigma^*_{ss}}^2}
    -\frac{cN{\varphi_x^*}_{ss}}{{\sigma^*_{ss}}^2}
    -\frac{cN}{2\mu}
    \right).
  \label{sec:theoreticalanalysis:eq:ssquaredlengthss1divN_2}
\end{align}
In the step from
\Cref{sec:theoreticalanalysis:eq:ssquaredlengthss1divN_1}
to
\Cref{sec:theoreticalanalysis:eq:ssquaredlengthss1divN_2},
$\xi\left(
\frac{1+\frac{{\sigma_{ss}^*}^2}{N}}
     {1+\frac{{\sigma_{ss}^*}^2}{\mu N}}\right) \simeq \xi$ for
$N \rightarrow \infty$ has been used.

Insertion of
\Cref{sec:theoreticalanalysis:eq:varphixnormalizedss4}
and
\Cref{sec:theoreticalanalysis:eq:ssquaredlengthss1divN_2}
into \Cref{sec:theoreticalanalysis:eq:sigmassinserted}
yields
\begin{equation}
  \begin{multlined}
    \frac{{\sigma_{ss}^*}^2}{(1 + \xi) 2 \mu}
    -\frac{{\sigma_{ss}^*} c_{\mu/\mu,\lambda}}{\sqrt{1+\xi}}
    =
    \frac{2(1-c)\mu}{2 c D}
    \left[\xi\left(
      \frac{c_{\mu/\mu,\lambda}^2}{1+\xi}
      -\frac{{\sigma_{ss}^*} c_{\mu/\mu,\lambda}}{\mu\sqrt{1+\xi}(1+\xi)}
      +\frac{{\sigma_{ss}^*}^2}{(1 + \xi)^2 4 \mu^2}\right)\right.\\\left.
      -\frac{cNc_{\mu/\mu,\lambda}}{\sqrt{1+\xi}{\sigma_{ss}^*}}
      +\frac{cN}{(1+\xi)2\mu}
      -\frac{cN}{2\mu}
      \right]
  \end{multlined}
  \label{sec:theoreticalanalysis:eq:sigmass1divNequation1}
\end{equation}
for the steady state mutation strength equation.
By assuming $c \ll 1$,
\Cref{sec:theoreticalanalysis:eq:sigmass1divNequation1}
simplifies. Together with grouping the powers of $\sigma_{ss}^*$ it writes
\begin{equation}
  \begin{multlined}
    \frac{{\sigma_{ss}^*}^2}{(1 + \xi) 2 \mu}
    -\frac{\mu \xi {\sigma_{ss}^*}^2}{c D (1 + \xi)^2 4 \mu^2}
    -\frac{{\sigma_{ss}^*} c_{\mu/\mu,\lambda}}{\sqrt{1+\xi}}
    +\frac{\mu \xi {\sigma_{ss}^*}
      c_{\mu/\mu,\lambda}}{c D \mu\sqrt{1+\xi}(1+\xi)}
    -\frac{\mu \xi c_{\mu/\mu,\lambda}^2}{c D (1+\xi)}\\
    -\frac{cN}{2 c D (1+\xi)}
    +\frac{cN}{2 c D}
    +\frac{\mu cNc_{\mu/\mu,\lambda}}{c D \sqrt{1+\xi}{\sigma_{ss}^*}}
    =
    0.
  \end{multlined}
  \label{sec:theoreticalanalysis:eq:sigmass1divNequation2}
\end{equation}
Introducing common denominators allows rewriting
\Cref{sec:theoreticalanalysis:eq:sigmass1divNequation2}
to
\begin{equation}
  \begin{multlined}
    {\sigma_{ss}^*}^2\left(\frac{2cD(1+\xi) -
      \xi}{c D (1 + \xi)^2 4 \mu}\right)
    +{\sigma_{ss}^*}^1\left(\frac{\xi c_{\mu/\mu,\lambda} -
      c_{\mu/\mu,\lambda} c D (1 + \xi)}{c D \sqrt{1+\xi}(1+\xi)}
    \right)\\
    +{\sigma_{ss}^*}^0\left(\frac{-2\mu \xi c_{\mu/\mu,\lambda}^2 -
      cN + cN (1 + \xi)}{2 c D (1+\xi)}\right)
    +{\sigma_{ss}^*}^{-1}\left(\frac{\mu
      cNc_{\mu/\mu,\lambda}}{c D \sqrt{1+\xi}}\right)
    =
    0.
  \end{multlined}
  \label{sec:theoreticalanalysis:eq:sigmass1divNequation3}
\end{equation}
Simplification of
\Cref{sec:theoreticalanalysis:eq:sigmass1divNequation3}
using $cD = 1$ and $cN \simeq 1$ results in
\begin{equation}
  \begin{multlined}
    {\sigma_{ss}^*}^2\left(\frac{2+\xi}{(1 + \xi)^2 4 \mu}\right)
    +{\sigma_{ss}^*}\left(\frac{-c_{\mu/\mu,\lambda}}{\sqrt{1+\xi}(1+\xi)}
    \right)
    +\left(\frac{\xi-2\mu \xi c_{\mu/\mu,\lambda}^2}
    {2(1+\xi)}\right)
    +\frac{1}{\sigma_{ss}^*}\left(\frac{\mu
      c_{\mu/\mu,\lambda}}{\sqrt{1+\xi}}\right)
    =
    0.
  \end{multlined}
  \label{sec:theoreticalanalysis:eq:sigmass1divNequation4}
\end{equation}
Note that multiplying
\Cref{sec:theoreticalanalysis:eq:sigmass1divNequation4}
by $\sigma_{ss}^*$ results in a cubic equation that can be solved. However, the
expressions for the closed-form solutions are rather long. Hence, a quadratic
equation is aimed for. To this end,
\Cref{sec:theoreticalanalysis:eq:sigmass1divNequation4}
is approximated quadratically.
Neglecting
${\sigma_{ss}^*}^{-1}\left(\frac{\mu c_{\mu/\mu,\lambda}}{\sqrt{1+\xi}}\right)$
and
${\sigma_{ss}^*}\left(\frac{-c_{\mu/\mu,\lambda}}{\sqrt{1+\xi}(1+\xi)}\right)$
in
\Cref{sec:theoreticalanalysis:eq:sigmass1divNequation4}
results in\footnote{Plots of further approximations are presented
in the supplementary material (\Cref{sec:appendix:steadystate1divN}).}
\begin{equation}
  \begin{multlined}
    {\sigma_{ss}^*}^2\left(\frac{2+\xi}{(1 + \xi)^2 4 \mu}\right)
    +\left(\frac{\xi-2\mu \xi c_{\mu/\mu,\lambda}^2}
    {2(1+\xi)}\right)
    =
    0.
  \end{multlined}
  \label{sec:theoreticalanalysis:eq:sigmass1divNequation6}
\end{equation}
Solving
\Cref{sec:theoreticalanalysis:eq:sigmass1divNequation6}
for the positive root with subsequent simplification yields
\begin{equation}
  \begin{multlined}
    {\sigma_{ss}^*} \approx
    \sqrt{\frac{2\mu\xi(2\mu c_{\mu/\mu,\lambda}^2 - 1)(1 + \xi)}{2 + \xi}}.
  \end{multlined}
  \label{sec:theoreticalanalysis:eq:sigmass1divNequationapproxsolution2}
\end{equation}

\begin{figure}
  \centering
  \begin{tabular}{@{\hspace{-0.000\textwidth}}c@{\hspace{-0.000\textwidth}}c@{\hspace{-0.000\textwidth}}c}
    \includegraphics[width=0.30\textwidth]{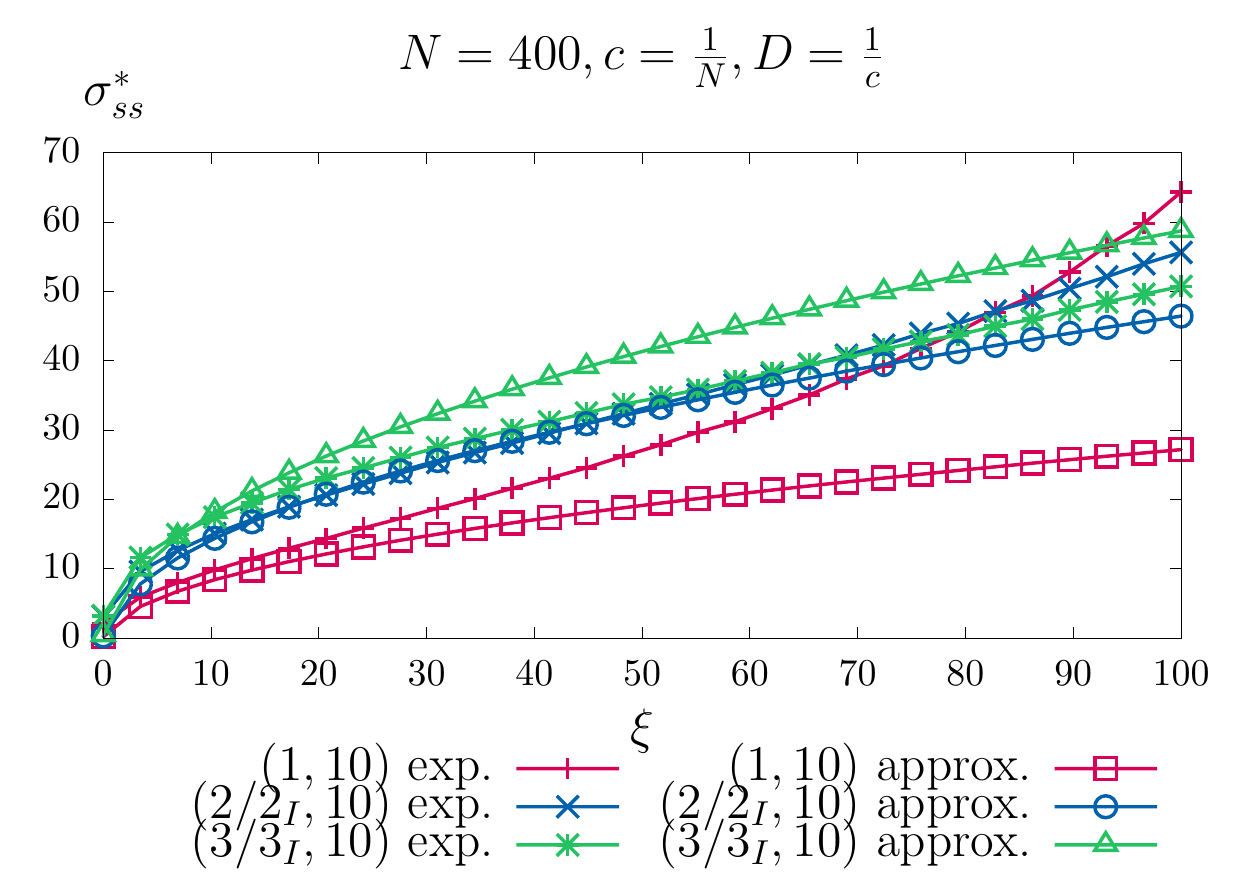}&
    \includegraphics[width=0.30\textwidth]{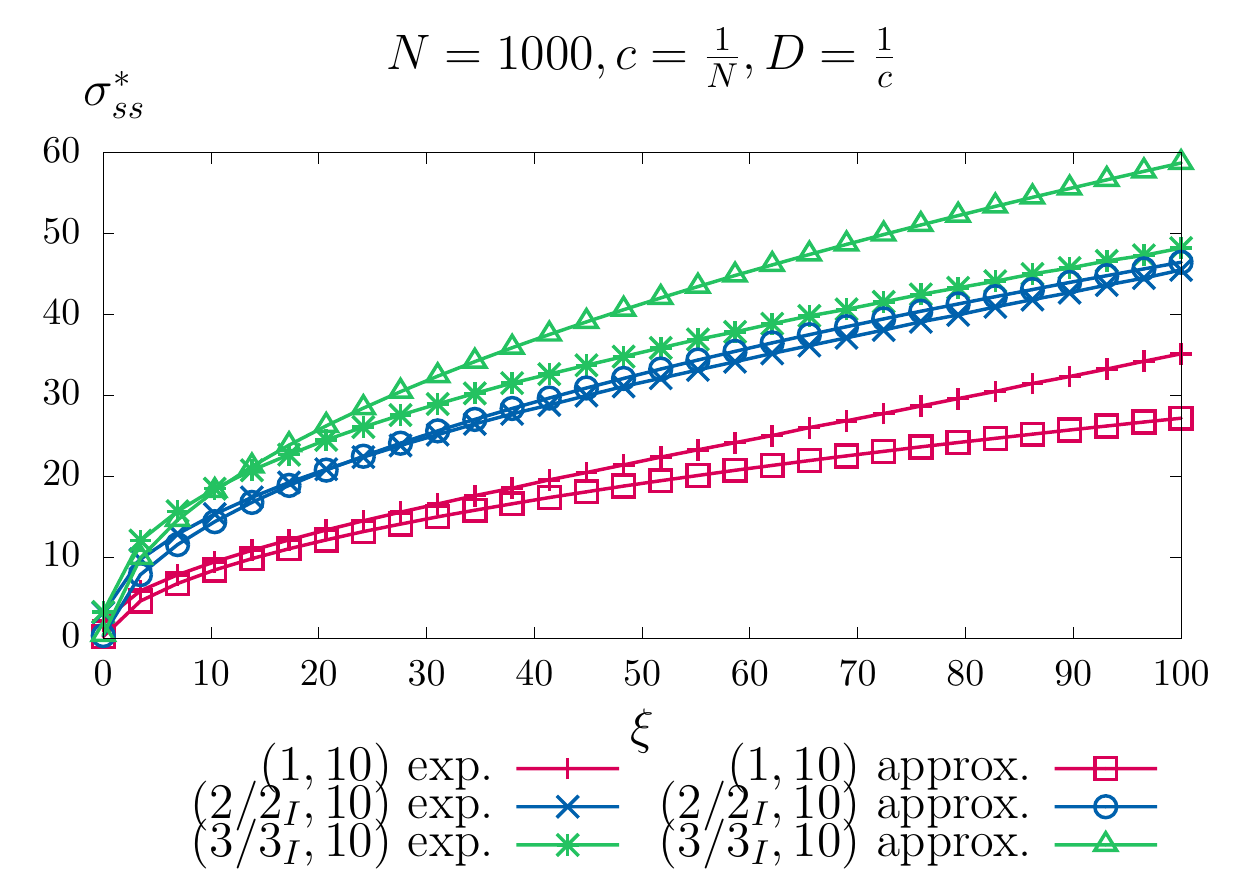}&
    \includegraphics[width=0.30\textwidth]{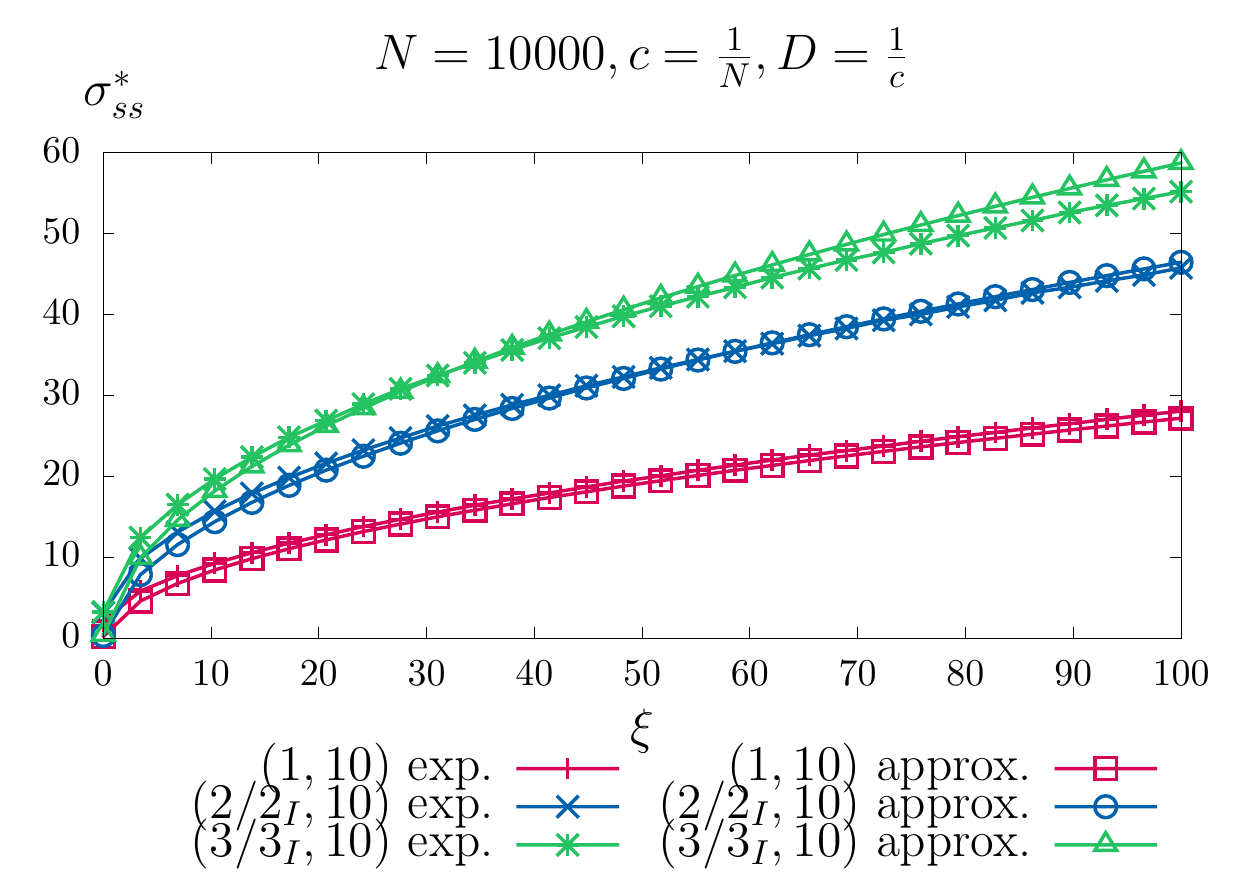}\\
    \includegraphics[width=0.30\textwidth]{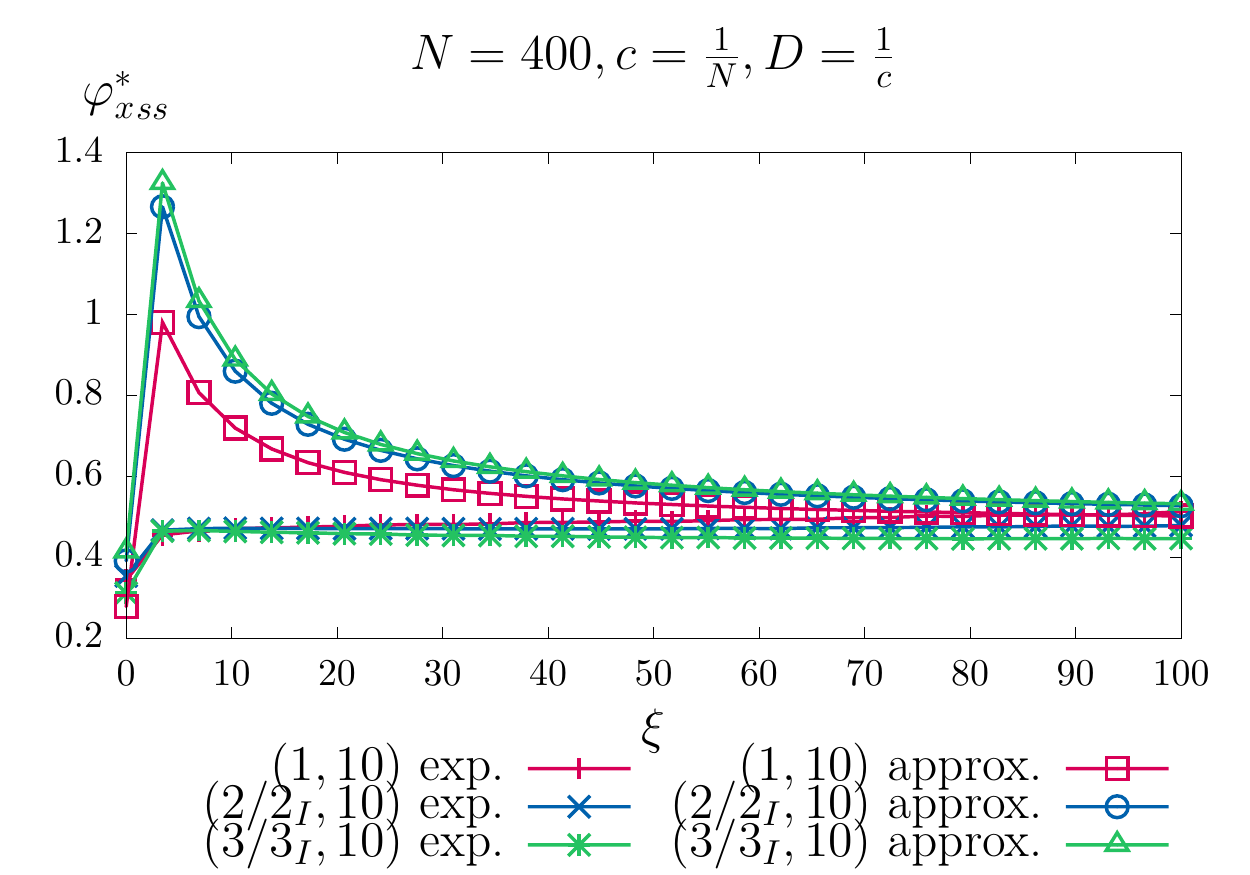}&
    \includegraphics[width=0.30\textwidth]{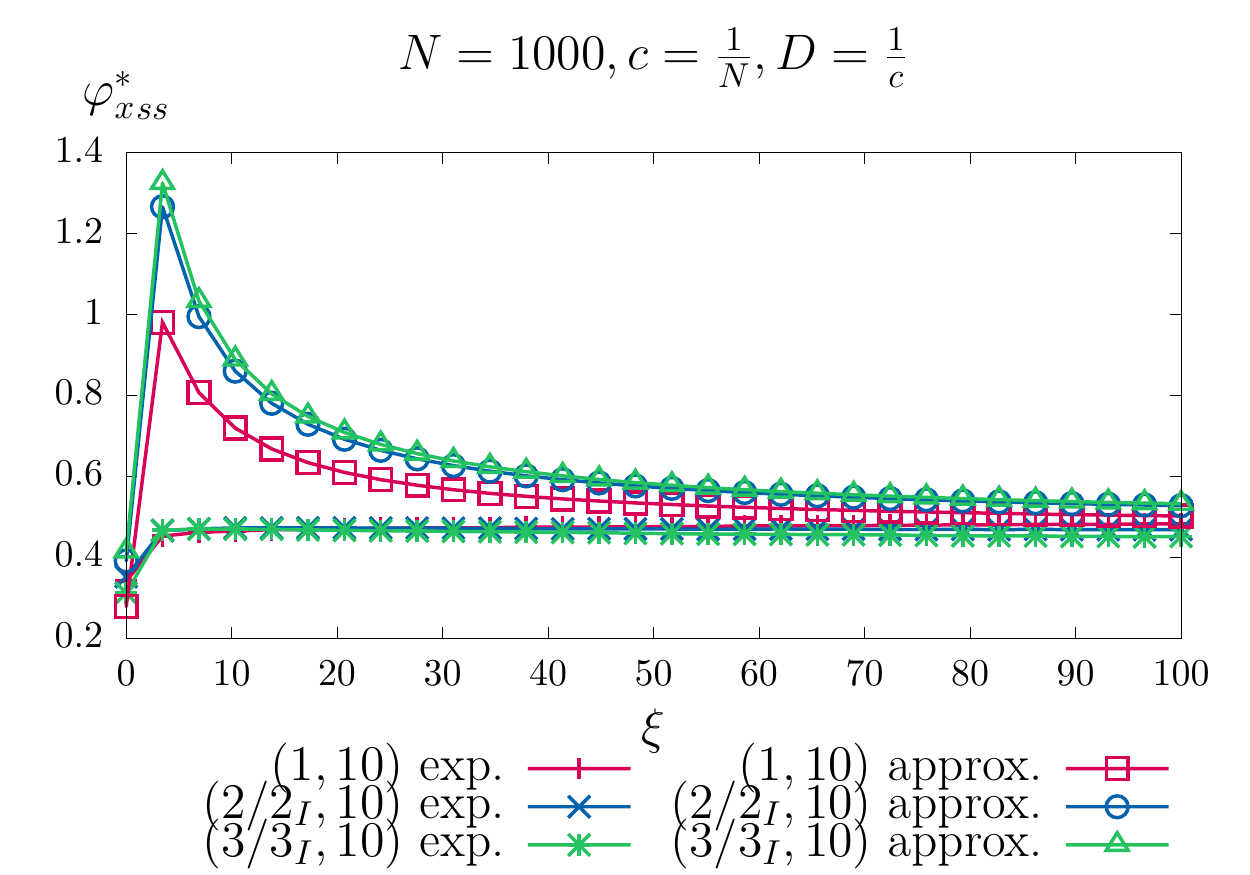}&
    \includegraphics[width=0.30\textwidth]{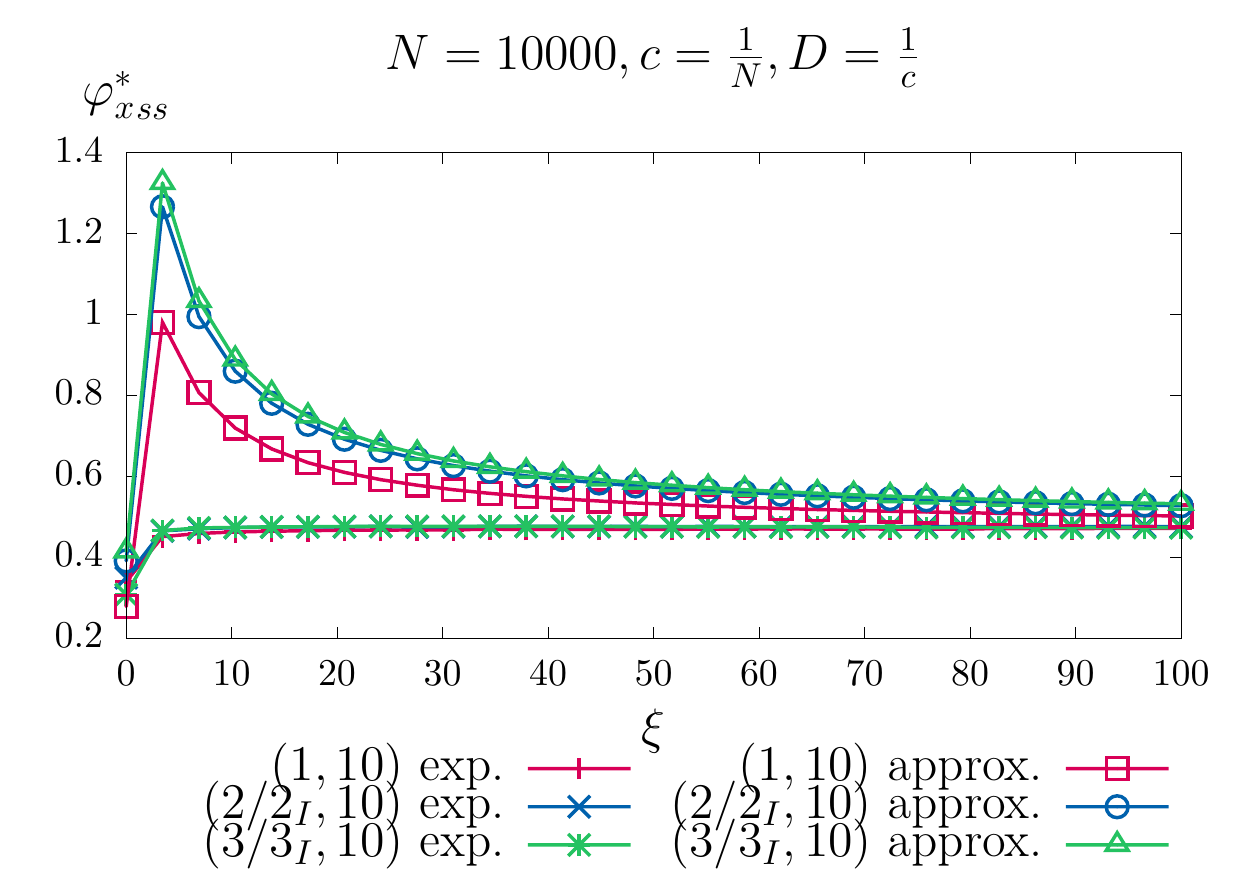}\\
    \includegraphics[width=0.30\textwidth]{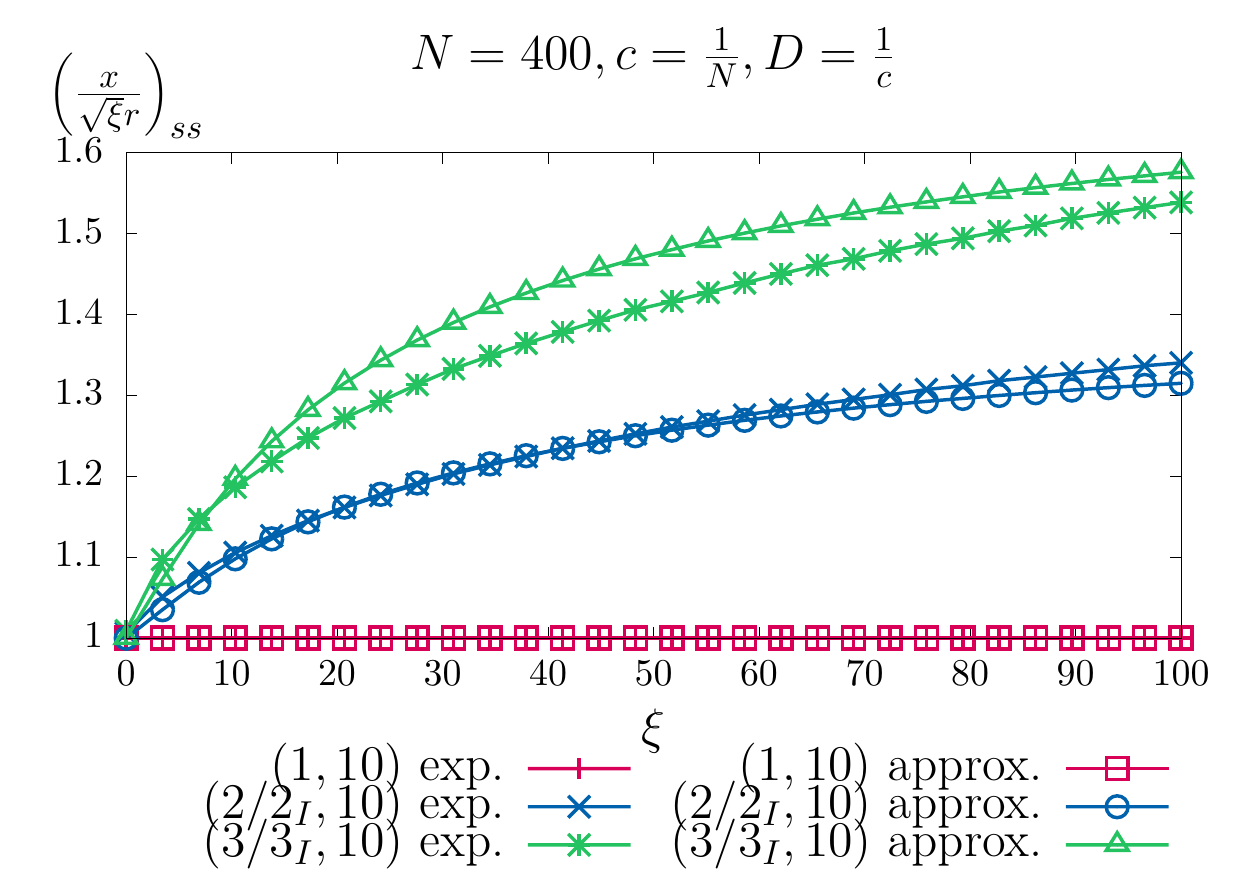}&
    \includegraphics[width=0.30\textwidth]{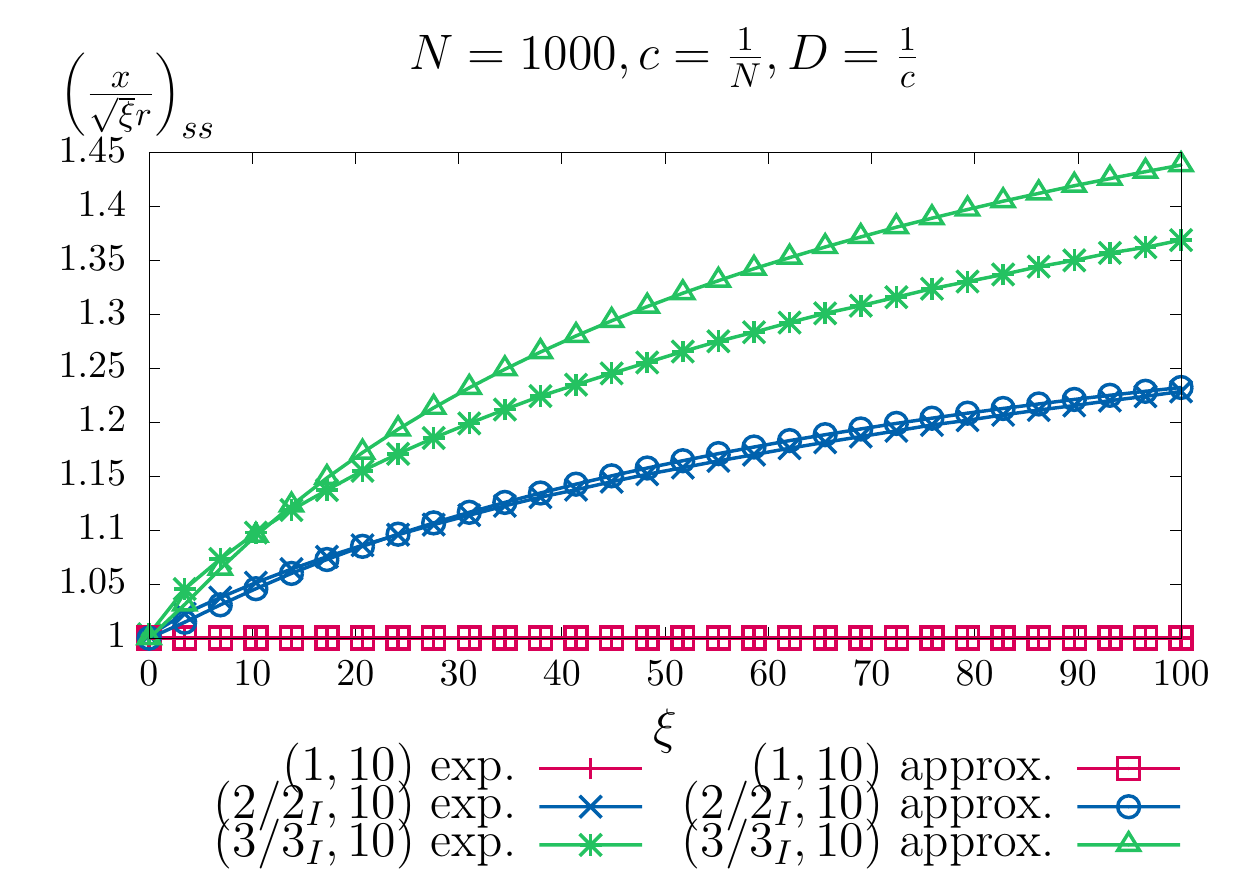}&
    \includegraphics[width=0.30\textwidth]{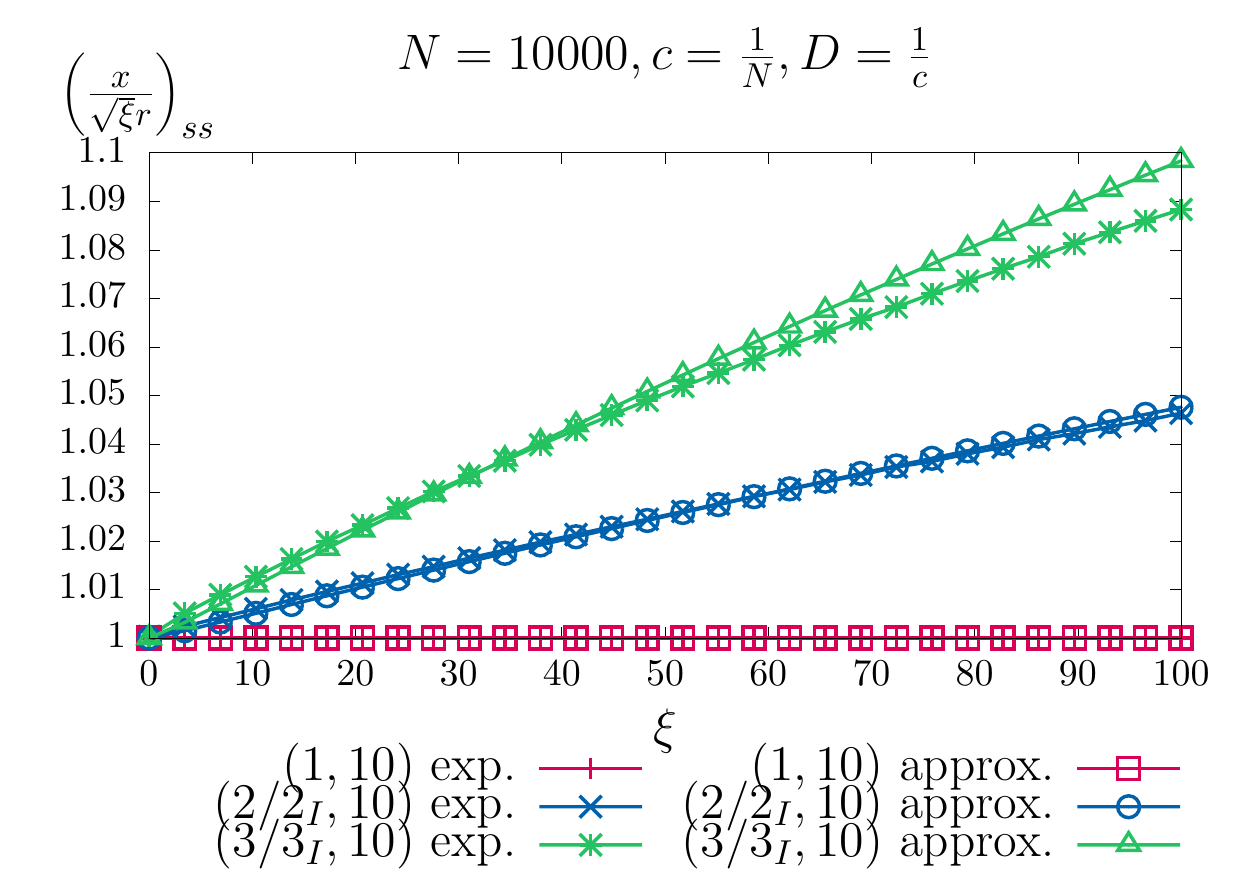}
  \end{tabular}
  \caption{Steady state closed-form approximation and real-run
  comparison of the $(\mu/\mu_I,\lambda)$-CSA-ES
  with repair by projection
  applied to the conically constrained problem.}
  \label{sec:theoreticalanalysis:fig:steadystatecomparisonclosedform1divN2}
\end{figure}

\Cref{sec:theoreticalanalysis:fig:steadystatecomparisonclosedform1divN2}
shows plots of the steady state computations. Results computed by
\Cref{sec:theoreticalanalysis:eq:sigmass1divNequationapproxsolution2}
have been compared to real ES runs.
\renewcommand{\sppatemp}{The values for the points denoting the approximations
have been determined by computing the normalized steady state
mutation strength $\sigma_{ss}^*$ using
\Cref{sec:theoreticalanalysis:eq:sigmass1divNequationapproxsolution2}
for different values of $\xi$. The results for
$\varphi_x^*$ and $\varphi_r^*$ have been determined by using the
computed steady state $\sigma_{ss}^*$ values with
\Cref{sec:theoreticalanalysis:eq:varphixnormalizedss4}.
The approximations for $\left(\frac{x}{\sqrt{\xi}r}\right)_{ss}$ have been
determined by evaluating
\Cref{sec:theoreticalanalysis:eq:steadystatedist2}.
The values for the points denoting the experiments have been determined by
computing the averages of the particular values in real ES
runs.}\sppatemp{}

\section{Conclusions}
\label{sec:conclusions}
In this work, the $(\mu/\mu_I,\lambda)$-CSA-ES has
been theoretically analyzed. For this, a mean value iterative
system has been introduced and compared to real ES runs.
Based on this derived system, steady state expressions have
been derived and compared to ES simulations.

The comparison of the mean value iterative system
summarized in
\Cref{sec:theoreticalanalysis:subsec:evolutionequationssummary}
with real ES runs
shows a satisfactory agreement of the theory and simulations
for large $\xi$ and large $N$
(see
\Cref{sec:theoreticalanalysis:fig:dynamics}).
The deviations for small $N$ are due to the
asymptotic assumptions $N \rightarrow \infty$ that are used in the derivations
of the microscopic and macroscopic aspects of the ES. They are used
to simplify the expressions and thus make a theoretical analysis tractable.
The deviations for small $\xi$ stem from the derivation of the offspring
cumulative distribution function after the projection step in $x_1$-direction
$P_Q(q)$ (for the details, it is referred to Section
3.1.2.1.2.3 in~\cite{SpettelBeyer2018SigmaSaEsCone},
in particular to the step from Equation (3.73) to Equation (3.74)).
The same observations regarding the deviations
can be made for the derived steady state expressions (see
\Cref{sec:theoreticalanalysis:fig:steadystatecomparisonclosedform,%
  sec:theoreticalanalysis:fig:steadystatecomparisonclosedform1divN2}).

For the steady state derivations,
it is of particular interest to compare the results obtained in this
work for the CSA-ES with the results obtained for the $\sigma$SA-ES.
The $(\mu/\mu_I,\lambda)$-$\sigma$SA-ES has been theoretically analyzed
by~\citet{SpettelBeyer2018SigmaSaEsConeMulti} applied to the same
conically constrained problem. In that work, the microscopic and
macroscopic aspects of the $(\mu/\mu_I,\lambda)$-$\sigma$SA-ES have
been investigated. For the microscopic aspects, expressions
for the local progress for $x$ and $r$ and the self-adaptation
response (SAR) function have been derived using
asymptotic assumptions. Those results have then been used for
the macroscopic analysis. The mean value dynamics generated
by iteration using those local measures have been compared to
real runs. In addition, steady state expressions have been derived
and discussed. They show that the $\sigma$SA-ES is able
to achieve sufficiently high mutation strengths to keep the progress
almost constant for increasing $\xi$. Surprisingly, for the CSA-ES, the
choice of the cumulation parameter $c$ has a qualitative influence on
the behavior.

Considering the choice of $c=1/\sqrt{N}$ proposed in early
publications on CMA-ES~\citep{Hansen1997},
the steady state mutation strengths attained flatten with
increasing $\xi$. As a consequence, the steady state progress decreases
with higher values of $\xi$. This can be seen by considering
\Cref{sec:theoreticalanalysis:eq:sigmassquadraticequationsolutionsimplified2}
that leads to $\sigma_{ss}^* \simeq 2 \mu c_{\mu/\mu,\lambda}$ for
sufficiently large $\xi$. For $\mu = 3$ and $\lambda = 10$, this results in
a steady state normalized mutation strength of approximately $6.39$.
Note that this value corresponds to the approximations for the larger
values of $\xi$ shown in
\Cref{sec:theoreticalanalysis:fig:steadystatecomparisonclosedform}
(right-most column, $N = 10000$).
\Cref{sec:theoreticalanalysis:eq:sigmassquadraticequationsolutionsimplified2}
can be inserted into the steady state progress rate
(\Cref{sec:theoreticalanalysis:eq:varphixnormalizedss4})
yielding
\begin{equation}
  {\varphi_r^*}_{ss}
  =
  {\varphi_x^*}_{ss}
  \approx
  \frac{2\mu\sqrt{\xi+1} c_{\mu/\mu,\lambda}^2}{\sqrt{\xi+1}\sqrt{\xi+2}}
  -\frac{4\mu^2 (\xi+1) c_{\mu/\mu,\lambda}^2}{(\xi + 2)(\xi + 1) 2 \mu}
  =
  \frac{2\mu c_{\mu/\mu,\lambda}^2}{\sqrt{\xi+2}}
  -\frac{2\mu c_{\mu/\mu,\lambda}^2}{\xi + 2}.
  \label{sec:theoreticalanalysis:eq:varphixnormalizedss4sigmainserted}
\end{equation}
From the simplified result of
\Cref{sec:theoreticalanalysis:eq:varphixnormalizedss4sigmainserted}
one immediately notices that $\varphi^*_{ss} \rightarrow 0$ for
$\xi \rightarrow \infty$ (respecting $\frac{\xi}{\sqrt{N}} \rightarrow 0$
that was used in the derivations leading to
\Cref{sec:theoreticalanalysis:eq:sigmassquadraticequationsolutionsimplified2}).
This is exactly what one sees in
\Cref{sec:theoreticalanalysis:fig:steadystatecomparisonclosedform},
the stationary state progress decreases with increasing $\xi$.

In contrast,
for the case $c = O\left(\frac{1}{N}\right)$ that is proposed
in newer publications ($\frac{4}{N + 4}$ by~\citet{Hansen2001} or
$\frac{\mu + 2}{N + \mu + 5}$ by~\citet{Hansen2016CMAESATutorial},
both of which are in $O\left(\frac{1}{N}\right)$ for $\mu \ll N$),
the steady state mutation strength
increases with increasing $\sigma_{ss}^*$. It is therefore able to
achieve a constant progress rate for increasing $\xi$. The steady
state progress is less than that of the $\sigma$SA-ES.
Due to the increase of $\xi$ with increasing $\sigma_{ss}^*$, the
increasing deviations of the approximation from the simulations can
be explained. In the derivations leading to
\Cref{sec:theoreticalanalysis:eq:rfraction2},
it has been assumed that $\mu N \gg {{\sigma^{(g)}}^*}^2$.
As the steady state $\sigma^*$ increases with $\xi$, $N$ must
be increased in order to have the same approximation quality
for higher values of $\xi$. This can be explained more formally.
Assuming large $\xi$, $(1+\xi)/(2+\xi) \simeq 1$ holds in
\Cref{sec:theoreticalanalysis:eq:sigmass1divNequationapproxsolution2}.
Hence,
\begin{equation}
  \sigma_{ss}^* \simeq \sqrt{\xi}\sqrt{2\mu(2\mu c_{\mu/\mu,\lambda}^2 - 1)}
\end{equation}
follows, which - for large $\xi$ - is of the same order as the one of the
$\sigma$SA-ES~(see Eq. (62) in \cite{SpettelBeyer2018SigmaSaEsConeMulti}).
While it is common practice to use $c=O\left(\frac{1}{N}\right)$ since
the seminal CMA-ES paper~\citep{Hansen2001}, this is the first theoretical
result that shows an advantage of the $O\left(\frac{1}{N}\right)$ choice.

A further aspect for discussion is the special case of the
non-recombinative $(1,\lambda)$-CSA-ES that is contained in
the derivations for the $(\mu/\mu_I,\lambda)$-CSA-ES.
The iterative system for the non-recombinative ($\mu = 1$) case
can be derived analogously to the multi-recombinative ($\mu > 1$)
case. The resulting equations differ in the expressions
for $\varphi_{x}^{(g)}$, $\varphi_{r}^{(g)}$, and $\varphi_{r^2}^{(g)}$.
It has been investigated further (not shown here) and
for $N^2(1+1/\xi) \gg {{\sigma^{(g)}}^*}^2$ the mean value iterative
systems of the $(1,\lambda)$-CSA-ES and the $(\mu/\mu_I,\lambda)$-CSA-ES
agree. An interesting observation between the case $\mu = 1$
and the case $\mu > 1$ is the evolution near the cone
boundary. Whereas the CSA-ES with $\mu = 1$ evolves on the boundary,
the CSA-ES with $\mu > 1$ attains a certain steady state
distance from the boundary (cf. the bottom subfigures of
\Cref{sec:theoreticalanalysis:fig:steadystatecomparisonclosedform}
and
\Cref{sec:theoreticalanalysis:fig:steadystatecomparisonclosedform1divN2}).
Considering a parental individual on the cone boundary, the offspring
are infeasible with overwhelming probability for sufficiently large
$N$. Hence, they are repaired by projection and are on the boundary
after projection. In particular, the best of them is on the boundary.
Therefore, for $\mu = 1$, the ES evolves on the boundary. For $\mu > 1$,
the centroid computation after projection results in offspring that
are inside the feasible region.

To conclude the paper, topics for future work are outlined. In
addition to the $\sigma$SA and the CSA for the $\sigma$ control
mechanism, it is of interest to investigate the behavior of Meta-ESs
applied to the conically constrained problem. Comparison of
the repair by projection approach with other repair methods is
another topic for further research. Analysis of ESs applied
to other constrained problems is another research direction
for the future.

\section*{Acknowledgments}
This work was supported by the Austrian Science Fund FWF under grant P29651-N32.

{\small
\bibliographystyle{apalike}
\bibliography{ms}
}

\newpage

\title{\sppatitle}

\author{
  \sppaauthor\\
  \vspace{1.0cm}
  \begin{center}\textsc{\Large Supplementary material}\end{center}
}

{\maketitle}

\appendix

\begin{appendices}

\setcounter{equation}{0}
\renewcommand{\theequation}{\thesection.\arabic{equation}}

This appendix contains material supplementing the main
text. Additional figures comparing the derived approximations with
simulations are presented in \Cref{sec:appendix:additionalfigures}.
In \Cref{sec:appendix:steadystate1divN}, results of further
investigations regarding the steady state approximations for
the case $c=O\left(\frac{1}{N}\right)$ are provided.

\section{Additional Results Comparing the Derived Approximations with
  Simulations}
\label{sec:appendix:additionalfigures}
\begin{figure}
  \centering
  \begin{tabular}{@{\hspace{-0.025\textwidth}}c@{\hspace{-0.025\textwidth}}c}
    \includegraphics[width=0.45\textwidth]{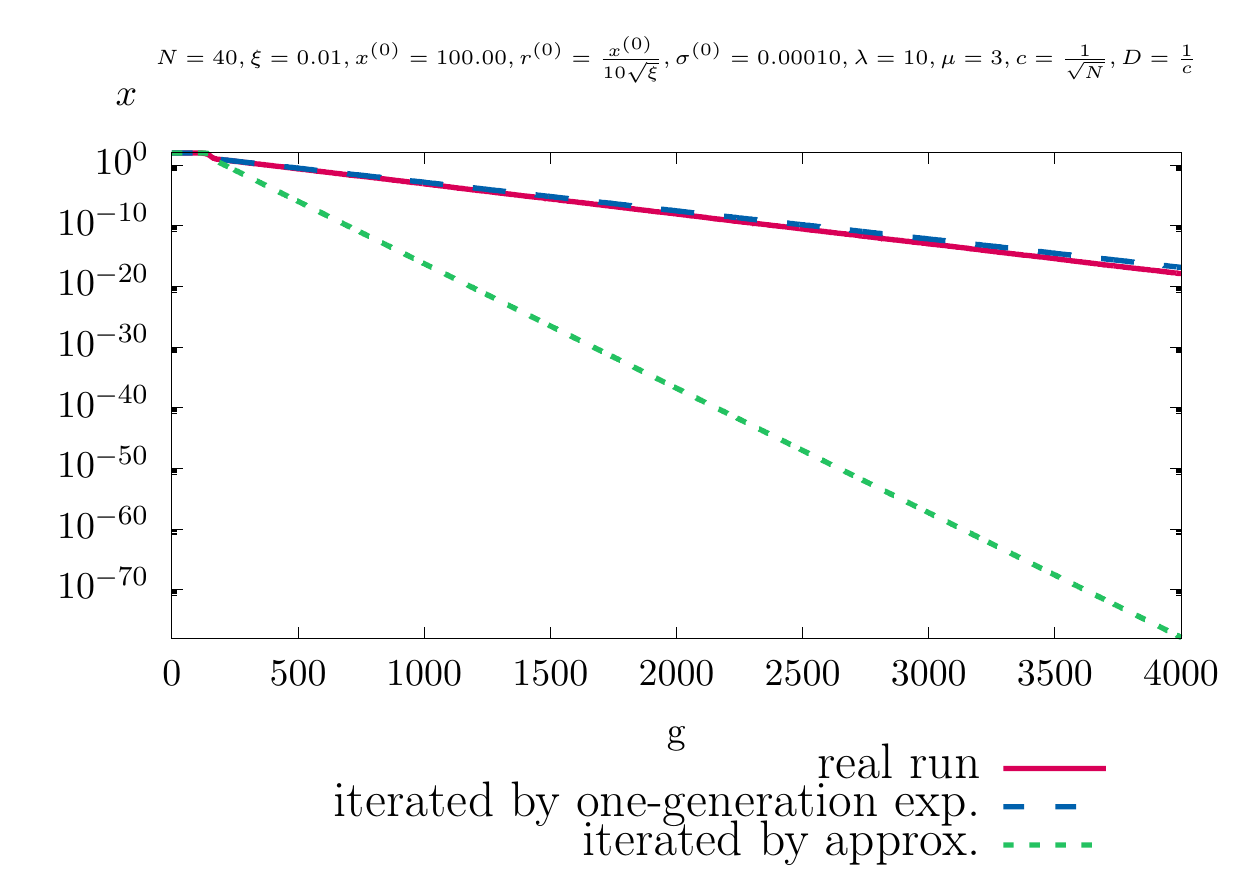}&
    \includegraphics[width=0.45\textwidth]{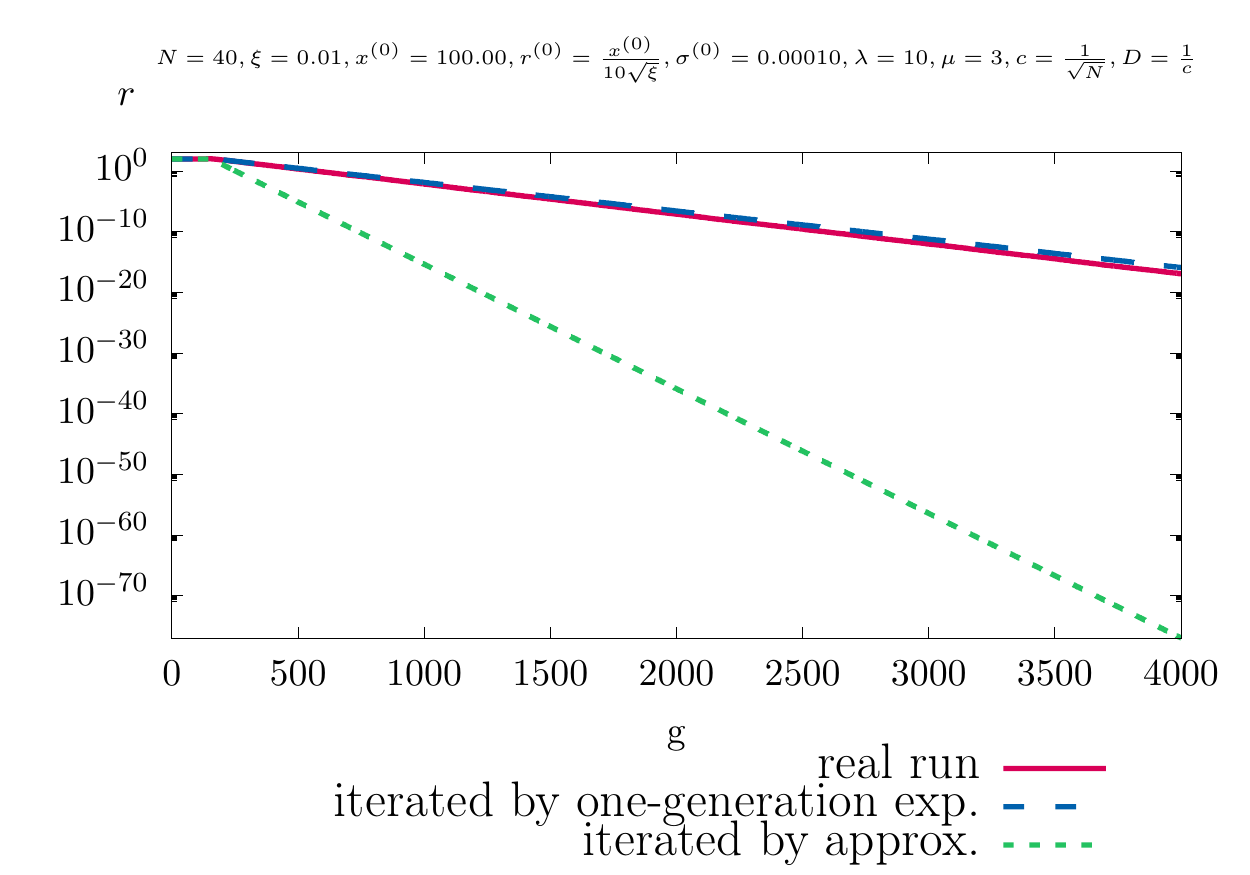}\\
    \includegraphics[width=0.45\textwidth]{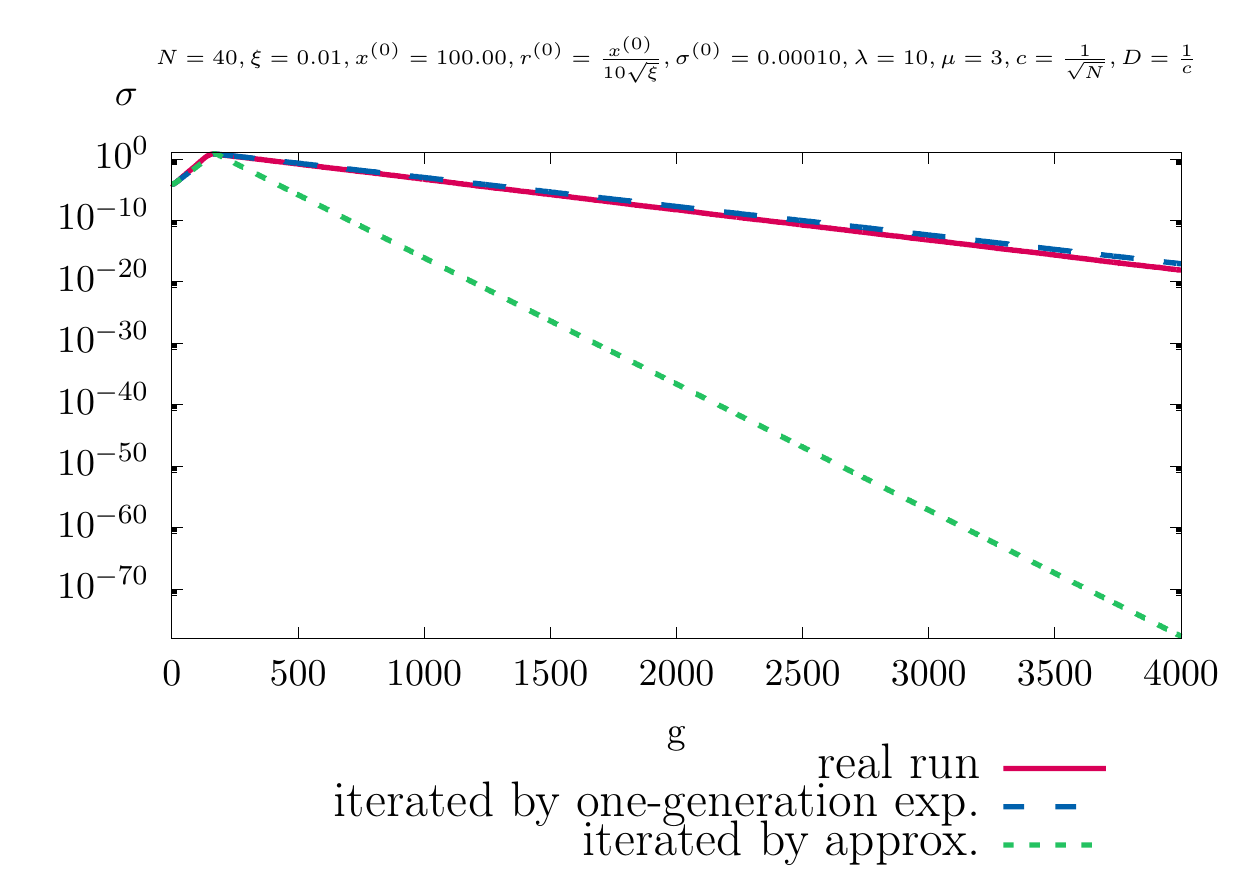}&
    \includegraphics[width=0.45\textwidth]{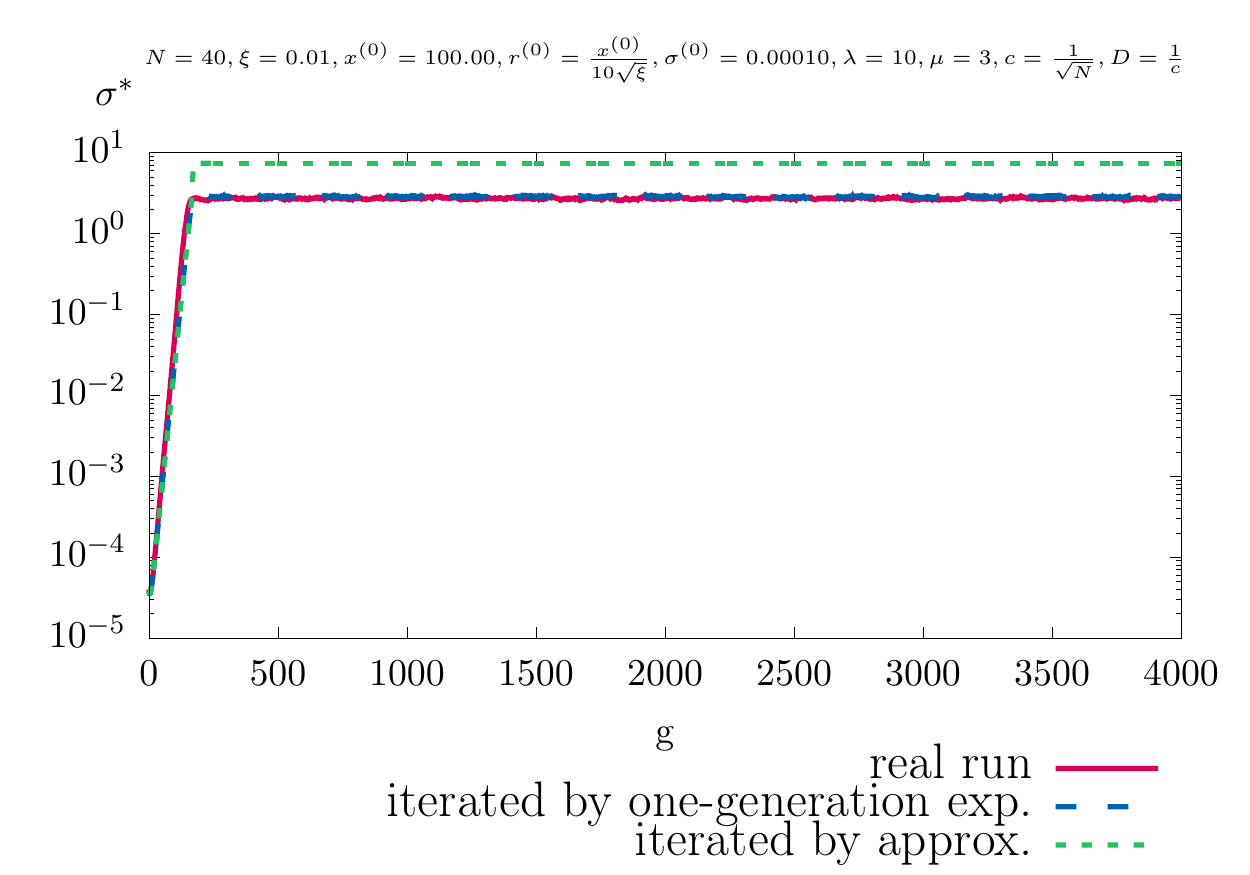}\\
    \includegraphics[width=0.45\textwidth]{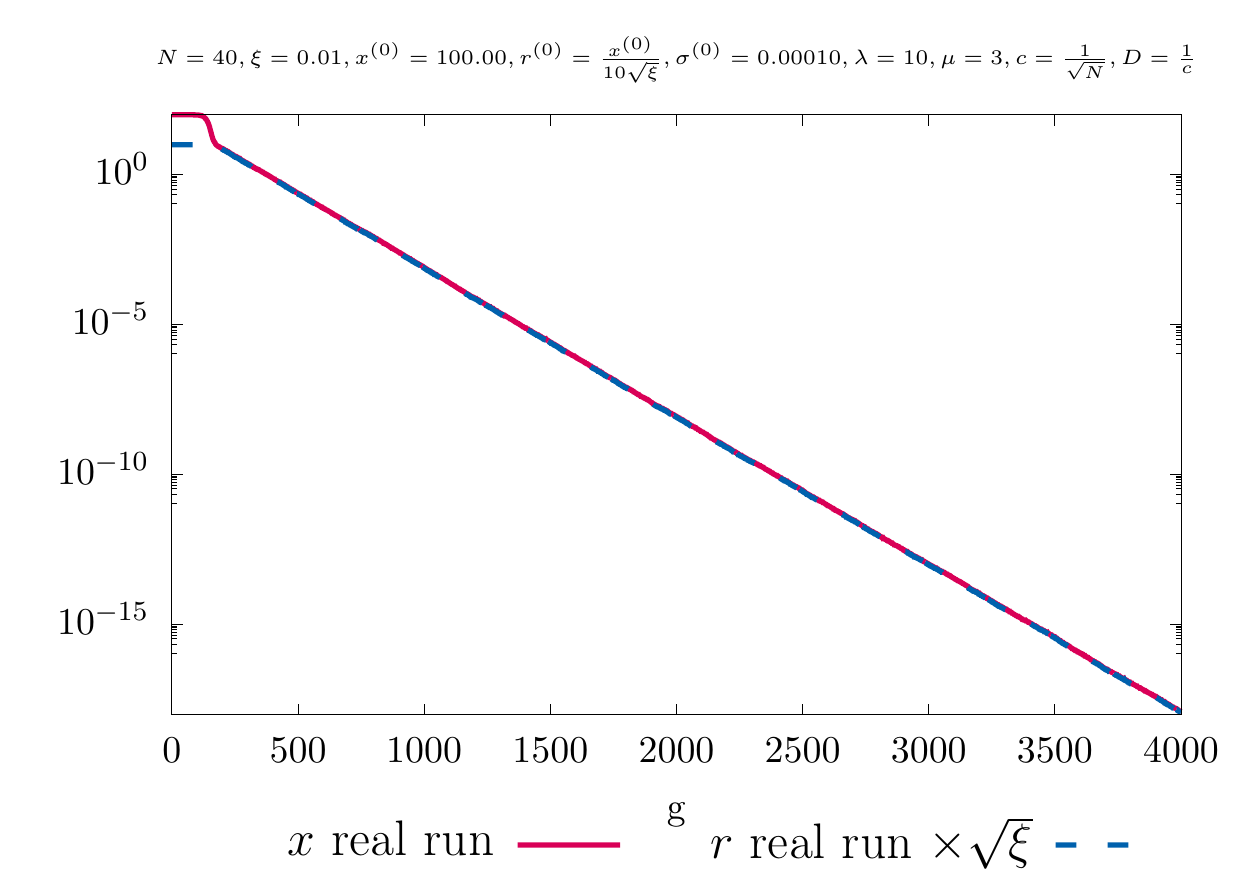}&
    \includegraphics[width=0.45\textwidth]{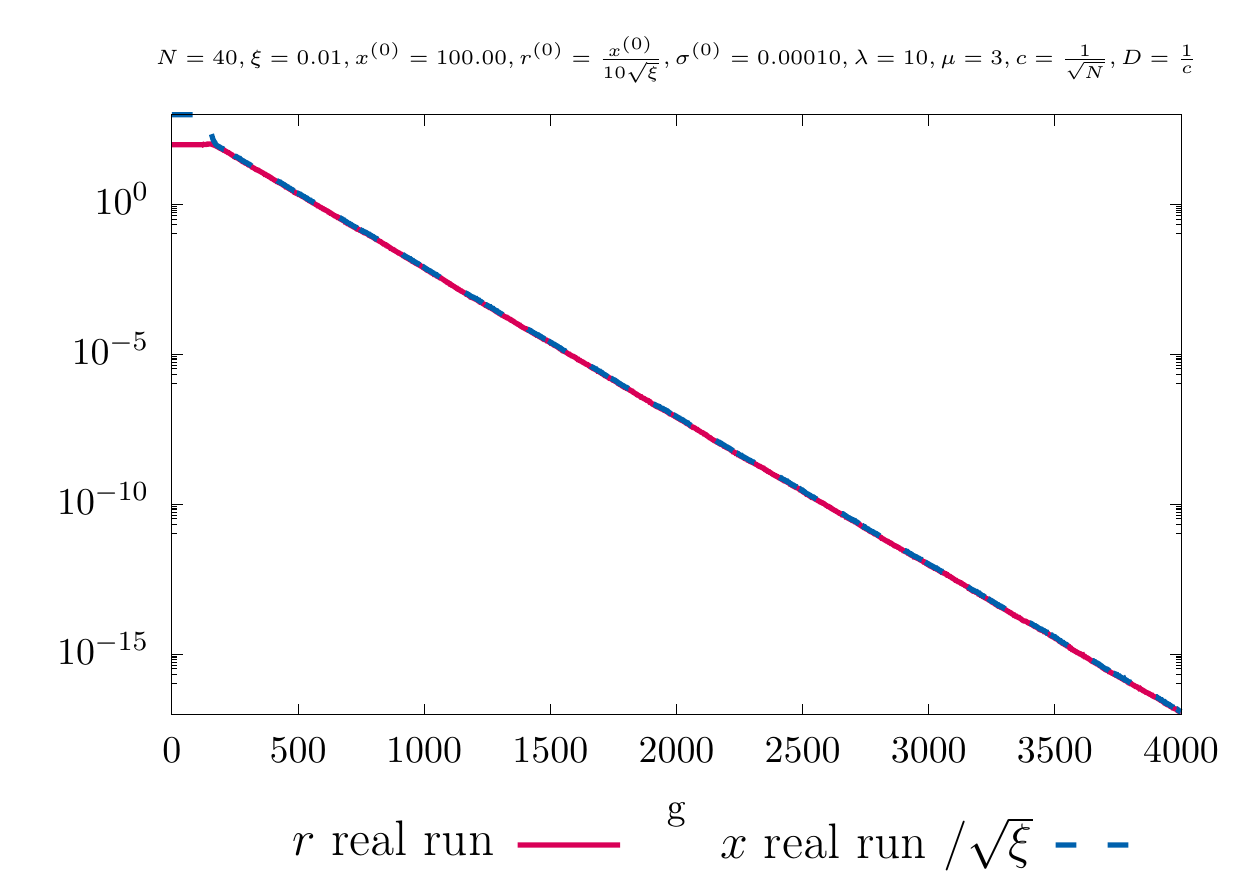}
  \end{tabular}
  \caption
  {Mean value dynamics closed-form approximation and real-run
  comparison of the $(3/3_I,10)$-CSA-ES
  with repair by projection
  applied to the conically constrained problem. (Part 1)}
  \label{sec:theoreticalanalysis:fig:dynamicsadditional1}
\end{figure}

\begin{figure}
  \centering
  \begin{tabular}{@{\hspace{-0.025\textwidth}}c@{\hspace{-0.025\textwidth}}c}
    \includegraphics[width=0.45\textwidth]{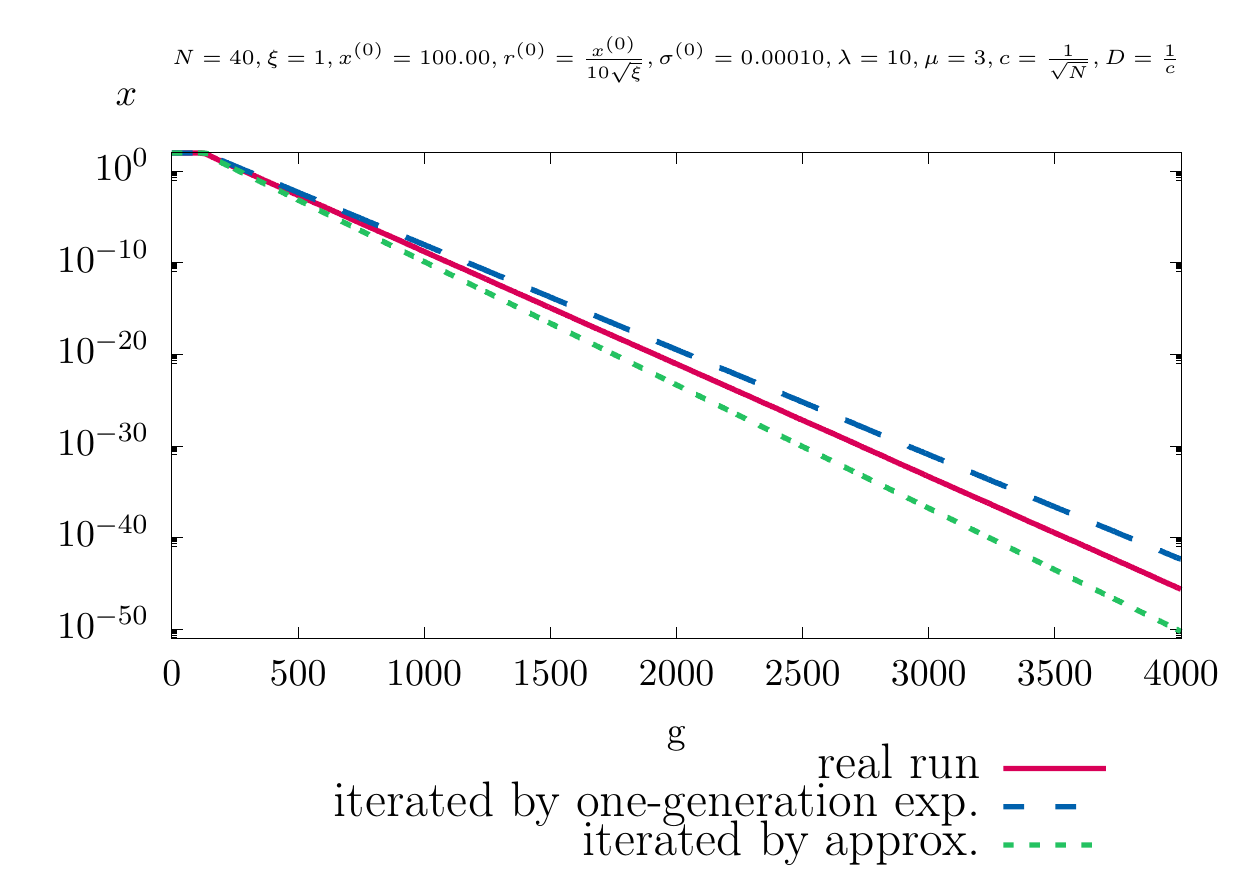}&
    \includegraphics[width=0.45\textwidth]{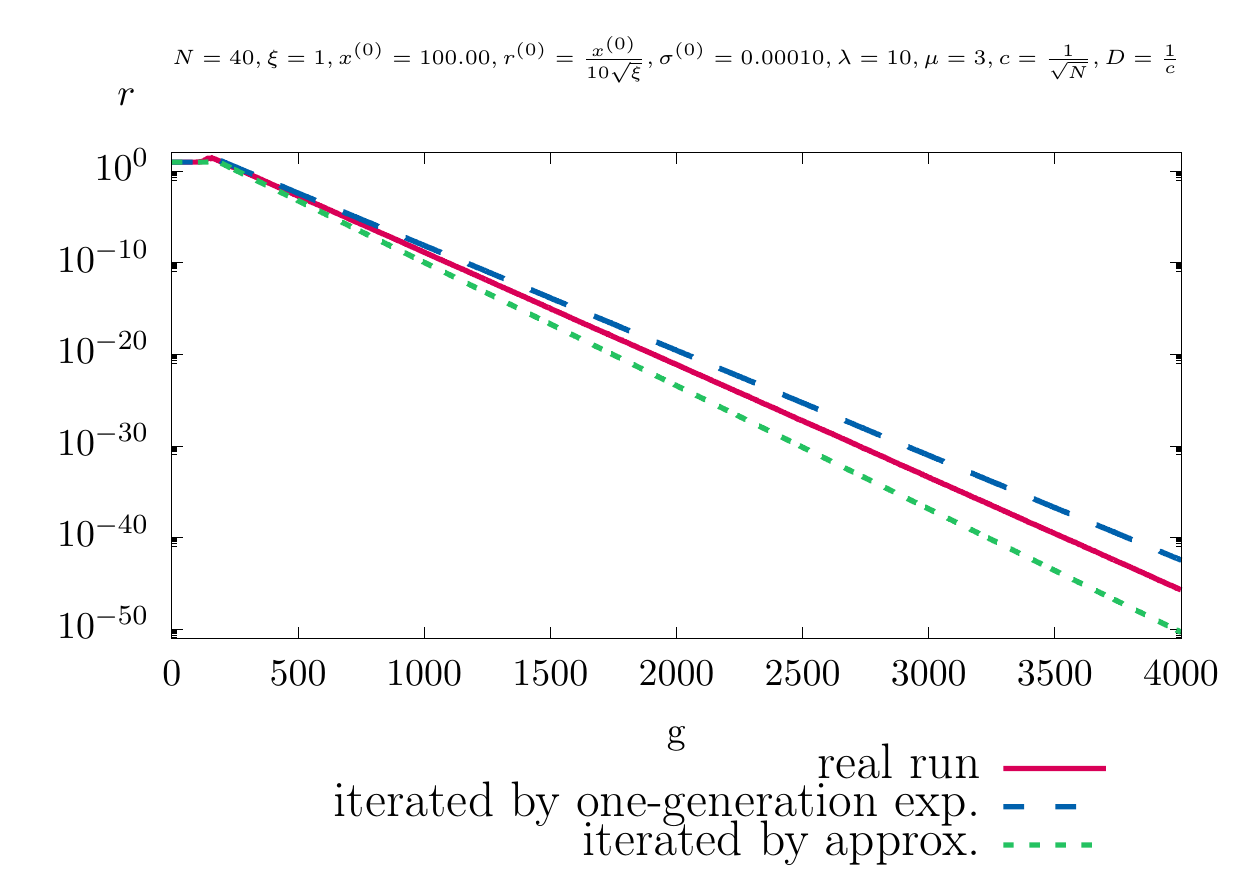}\\
    \includegraphics[width=0.45\textwidth]{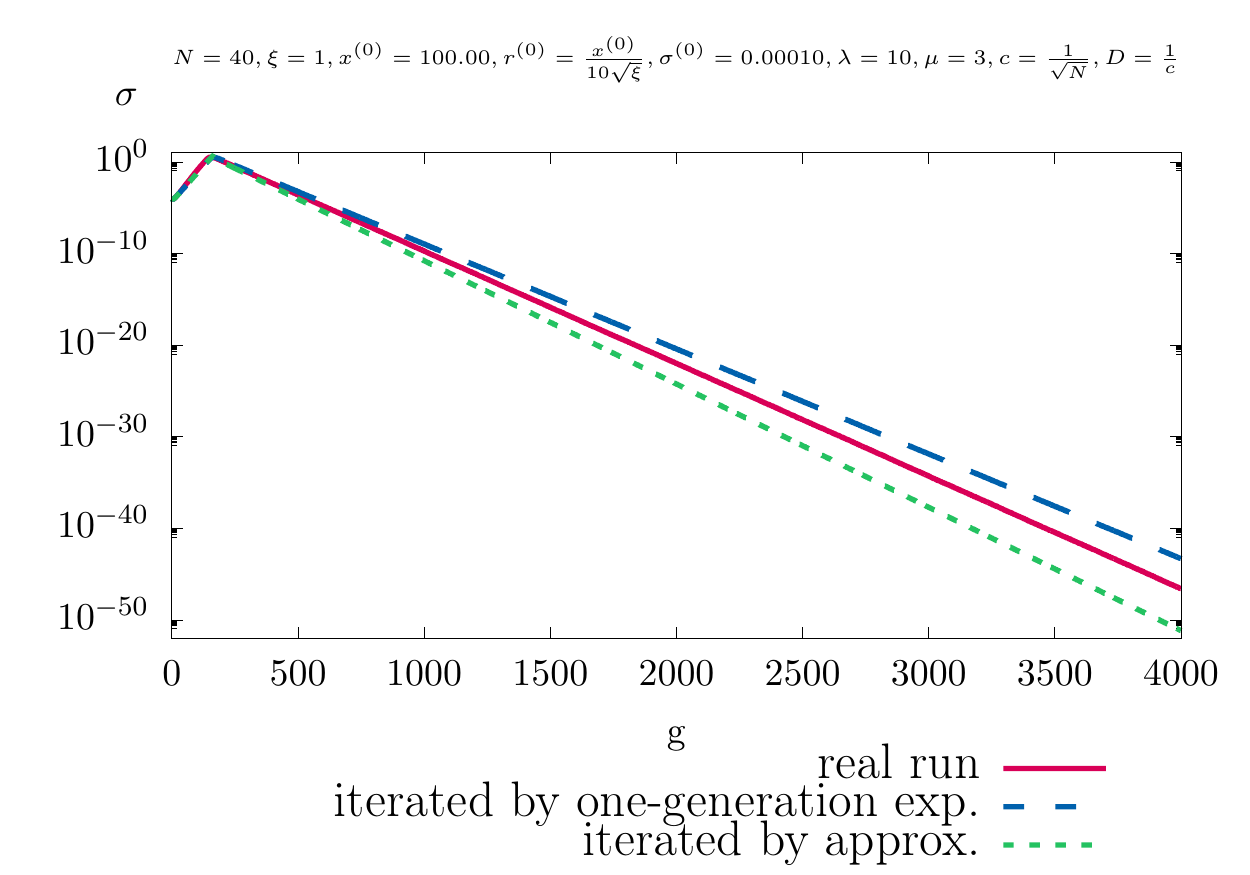}&
    \includegraphics[width=0.45\textwidth]{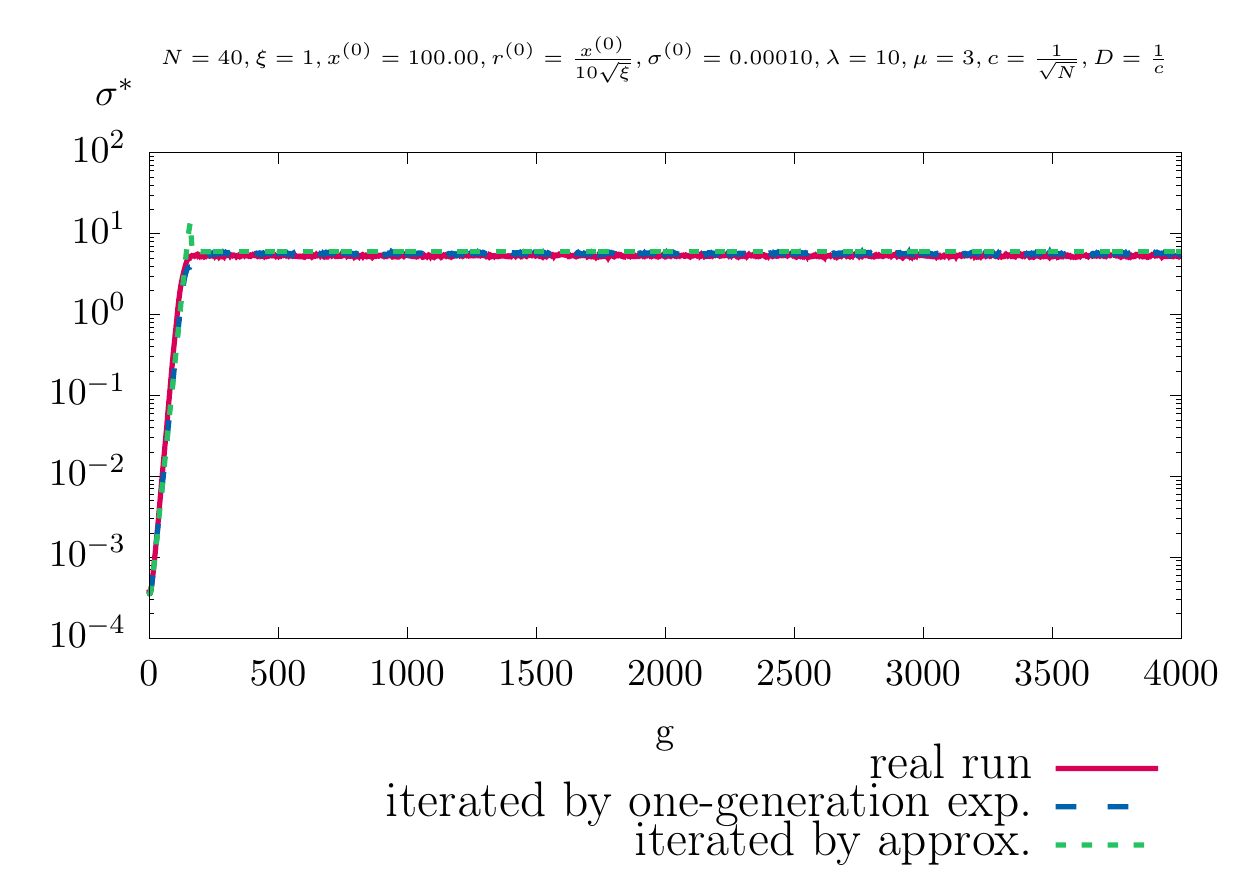}\\
    \includegraphics[width=0.45\textwidth]{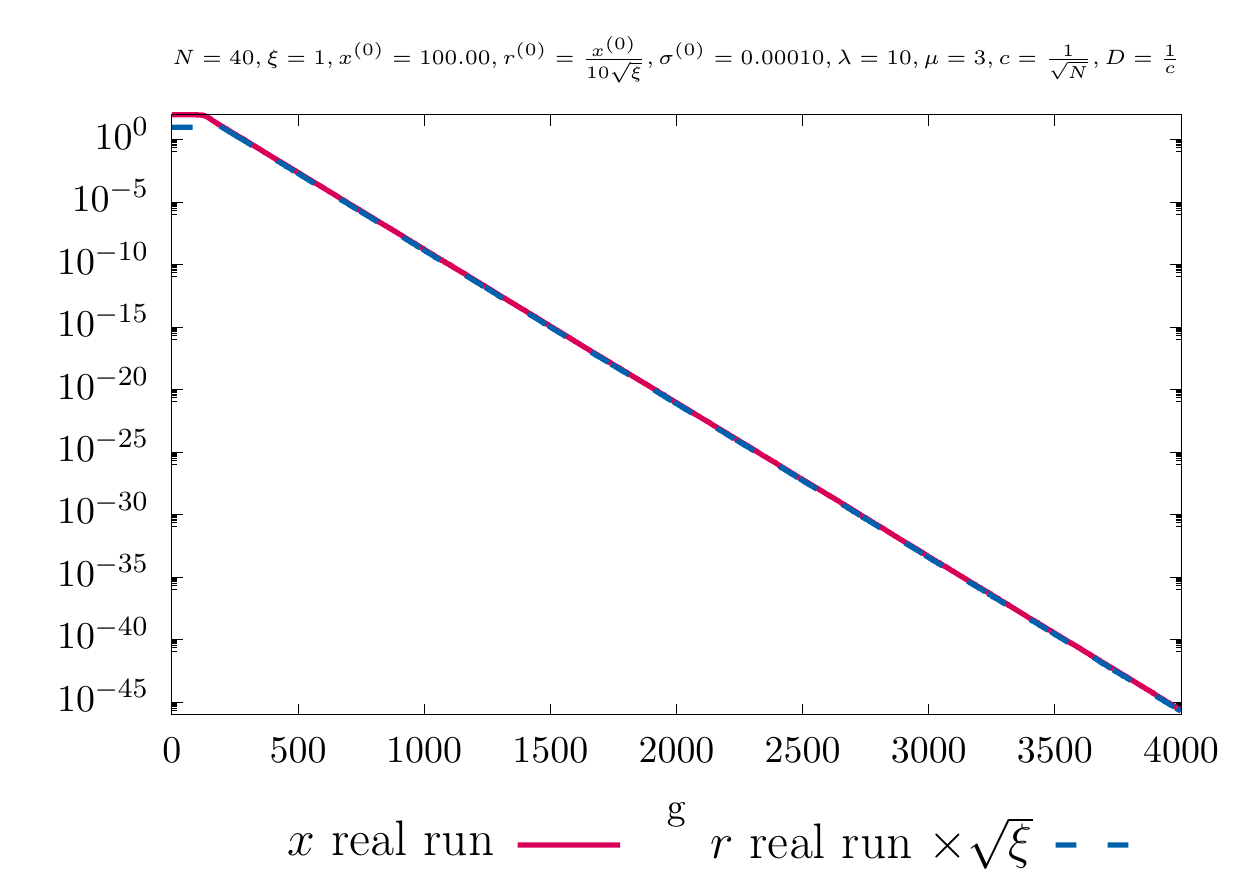}&
    \includegraphics[width=0.45\textwidth]{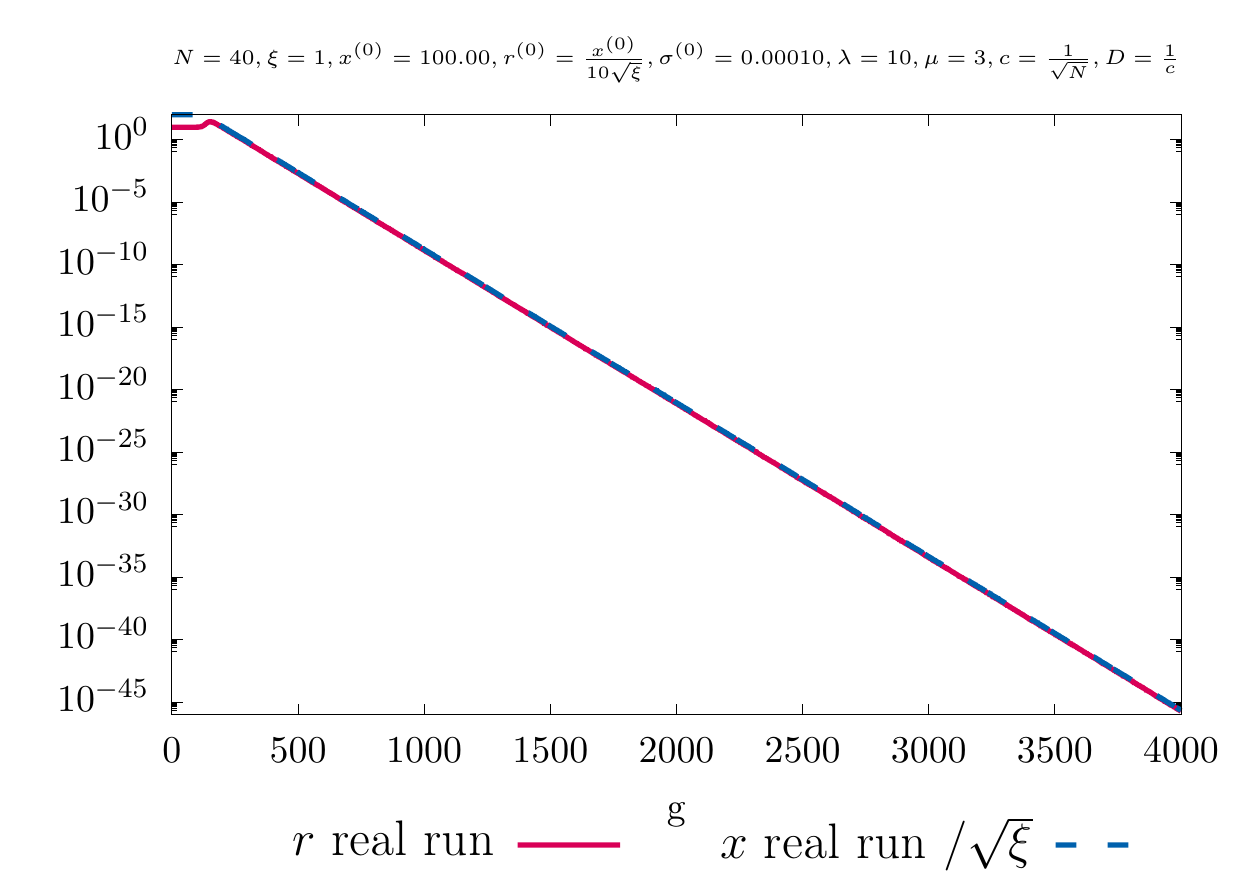}
  \end{tabular}
  \caption
  {Mean value dynamics closed-form approximation and real-run
  comparison of the $(3/3_I,10)$-CSA-ES
  with repair by projection
  applied to the conically constrained problem. (Part 2)}
  \label{sec:theoreticalanalysis:fig:dynamicsadditional2}
\end{figure}

\begin{figure}
  \centering
  \begin{tabular}{@{\hspace{-0.025\textwidth}}c@{\hspace{-0.025\textwidth}}c}
    \includegraphics[width=0.45\textwidth]{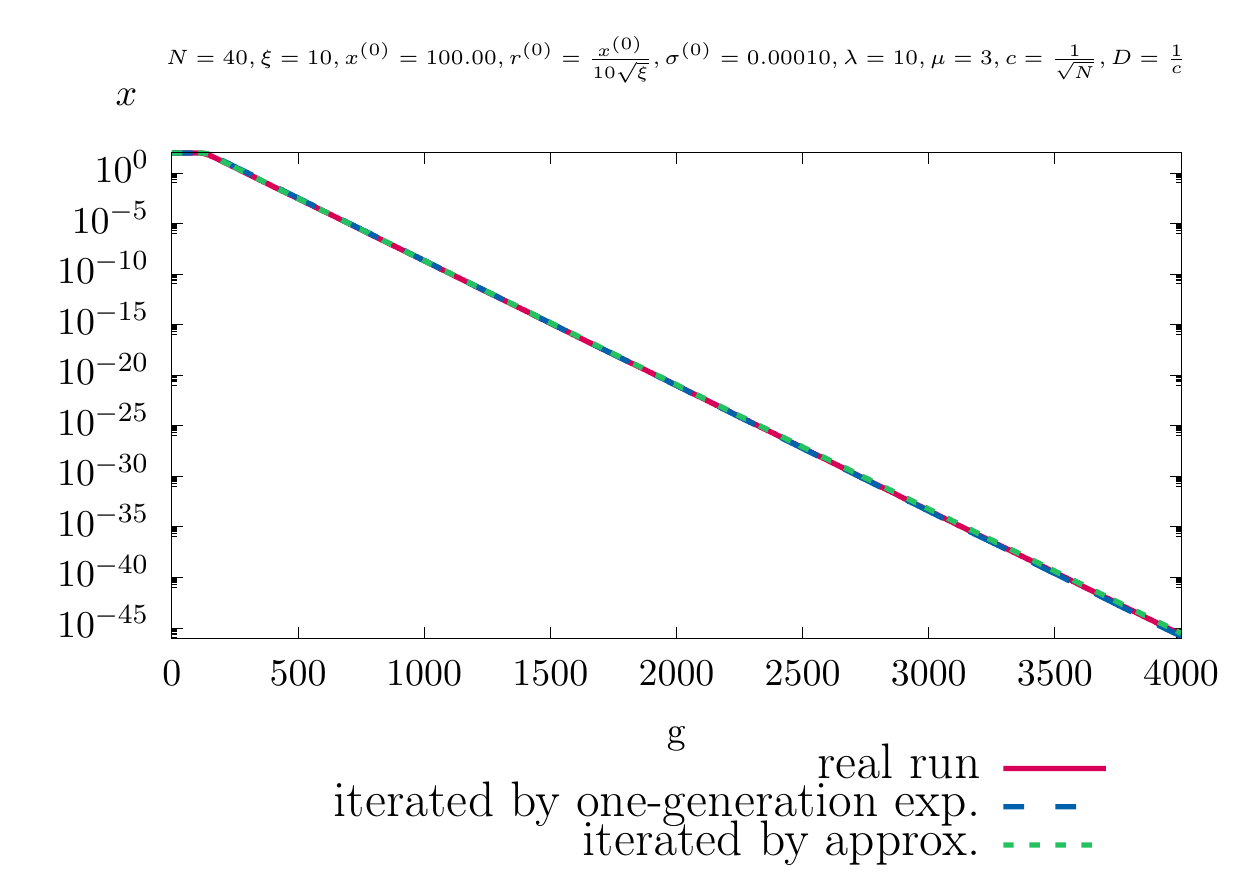}&
    \includegraphics[width=0.45\textwidth]{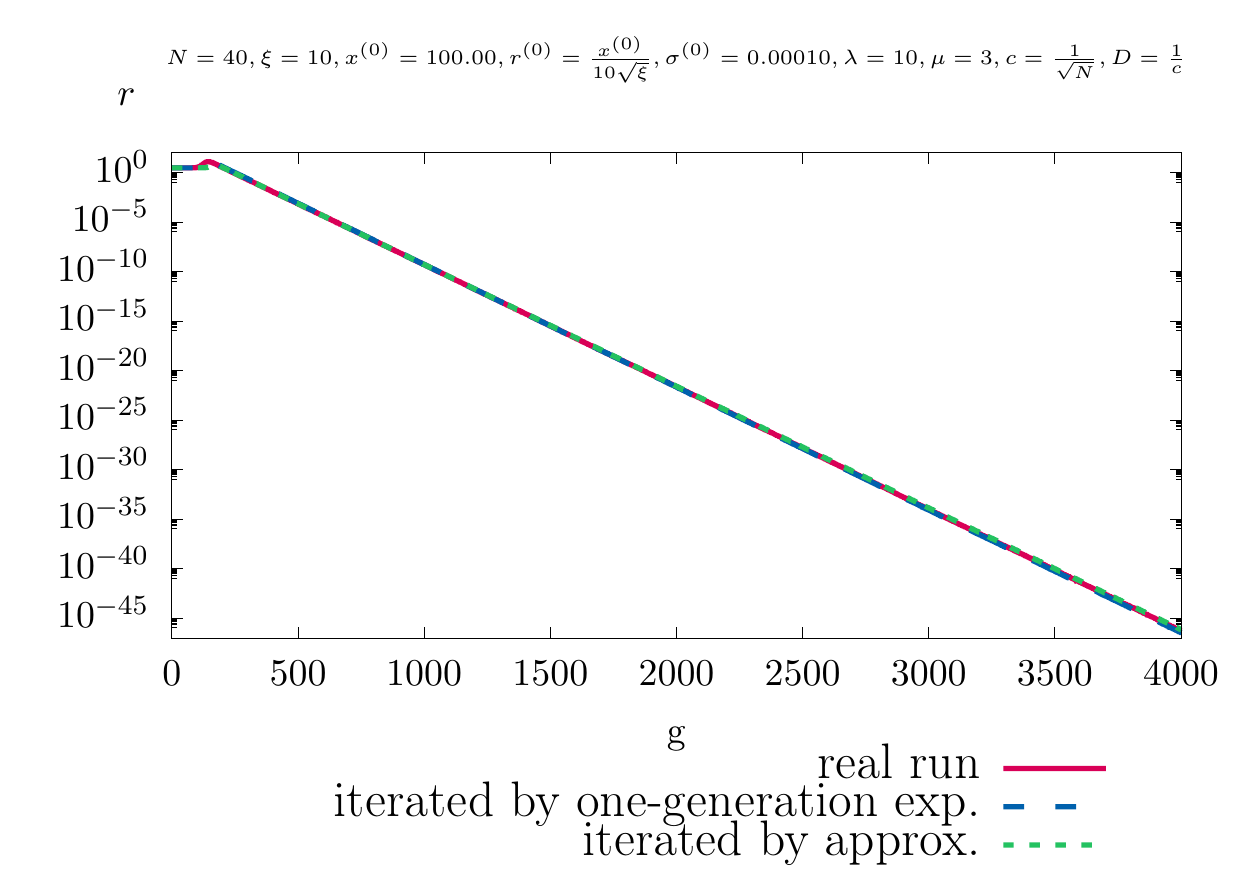}\\
    \includegraphics[width=0.45\textwidth]{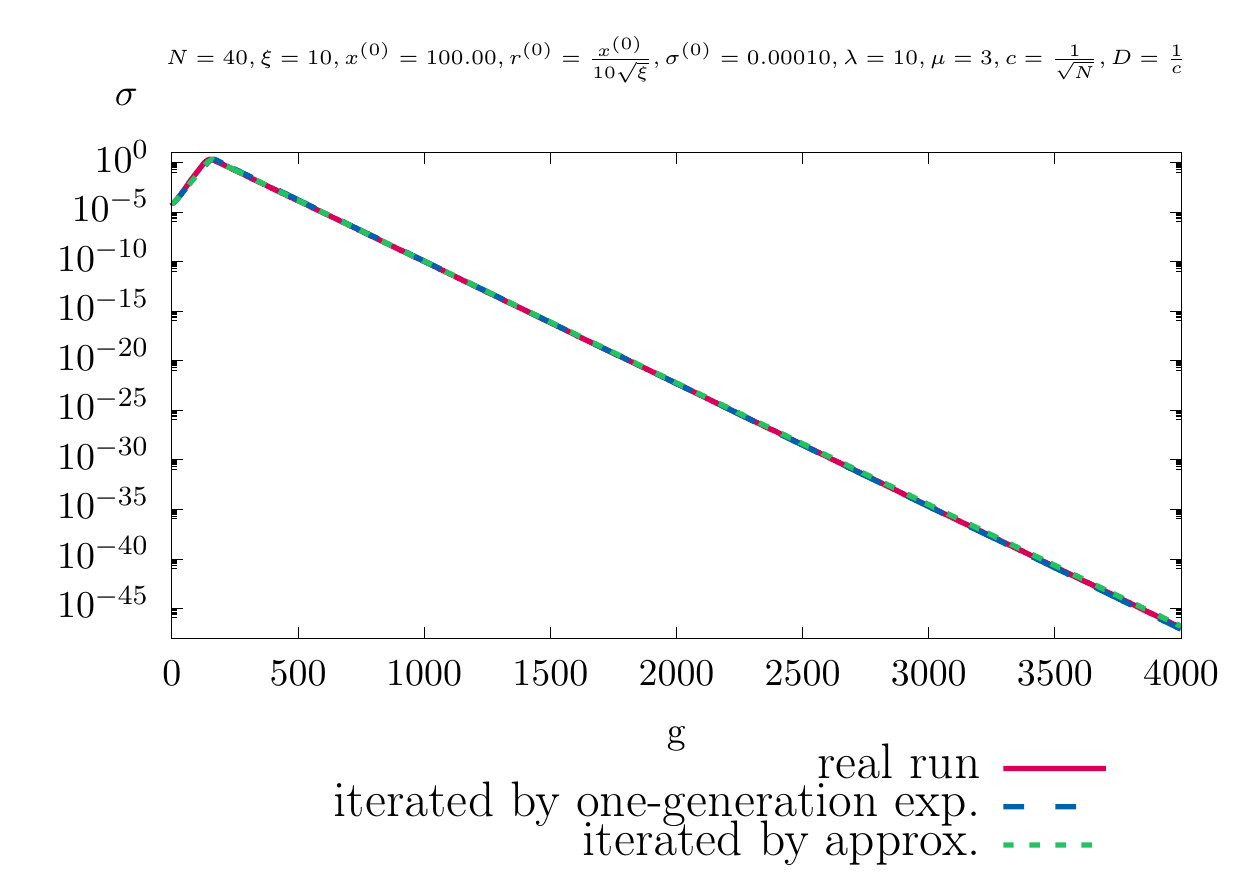}&
    \includegraphics[width=0.45\textwidth]{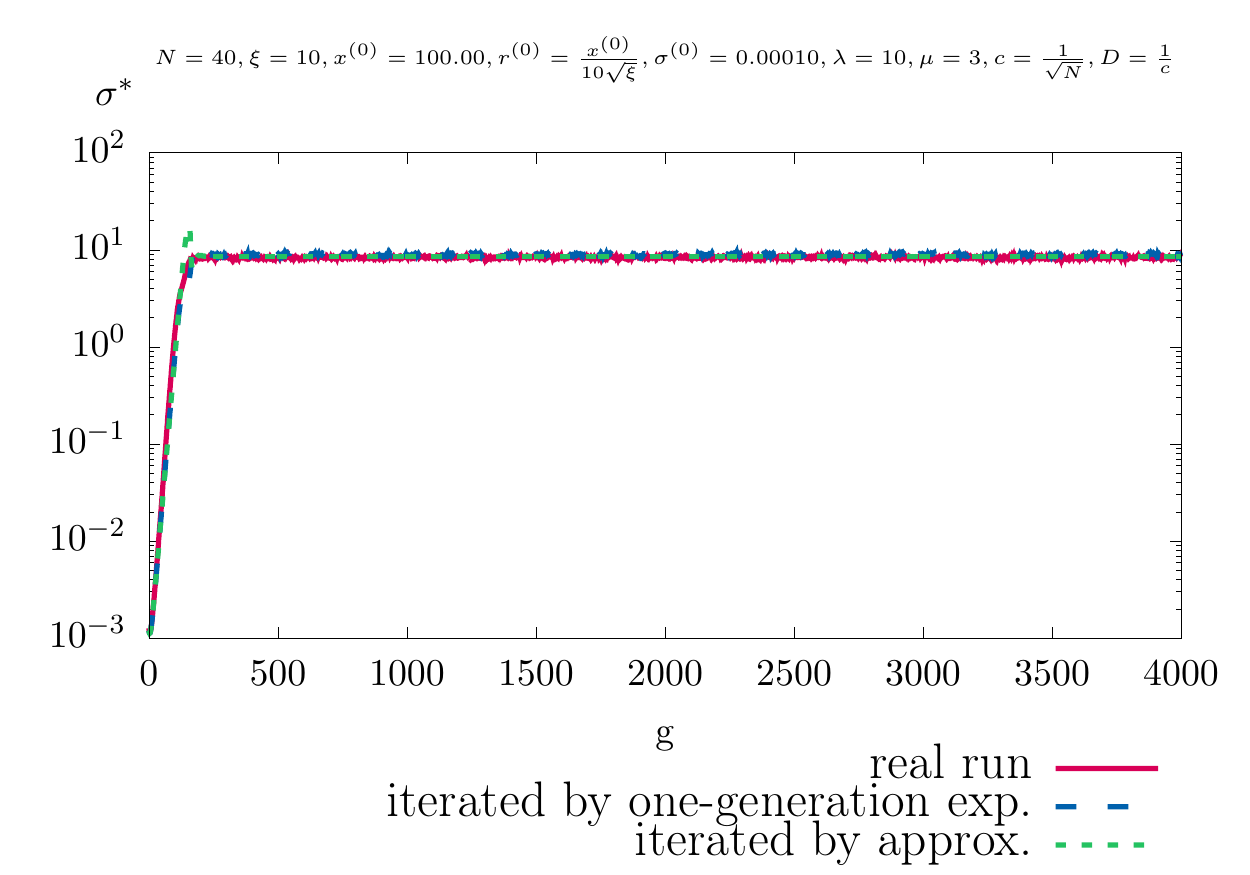}\\
    \includegraphics[width=0.45\textwidth]{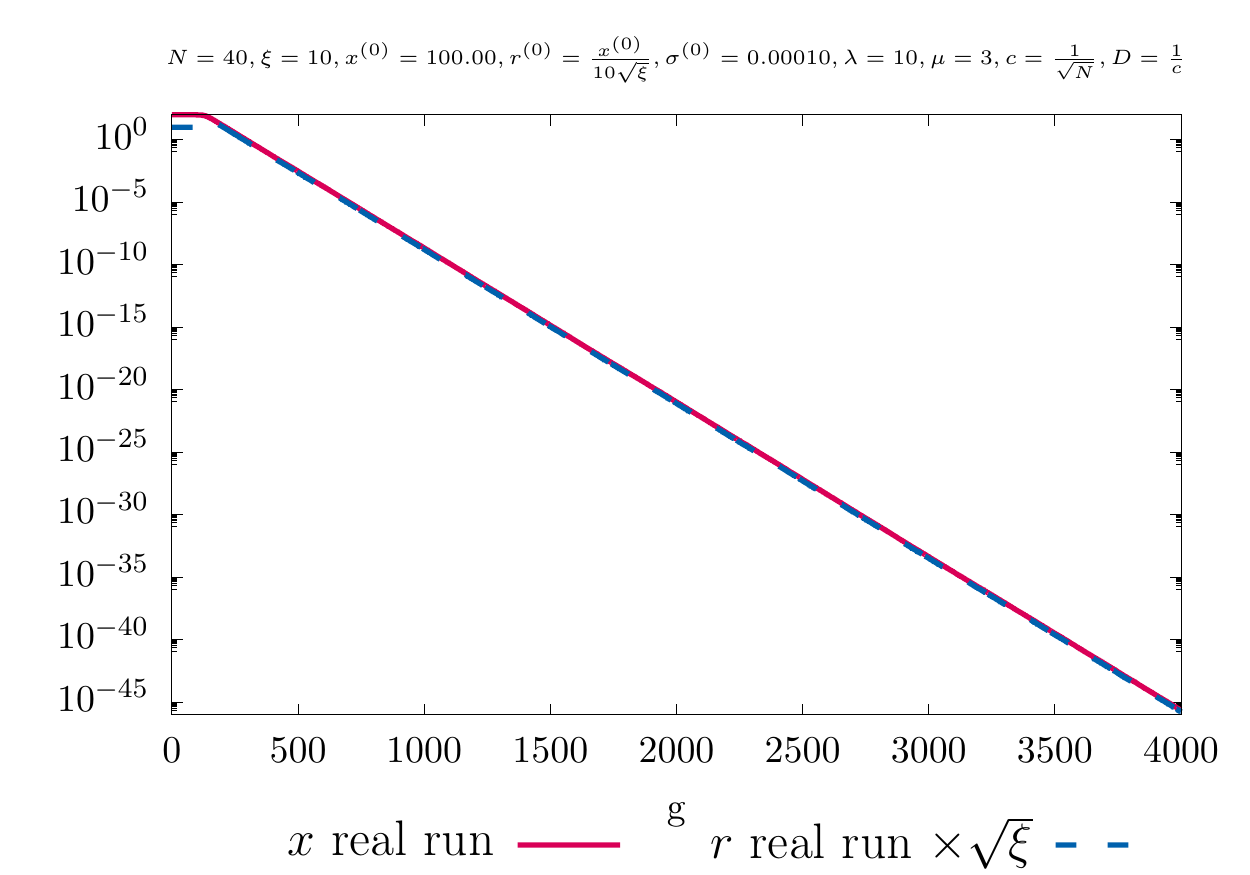}&
    \includegraphics[width=0.45\textwidth]{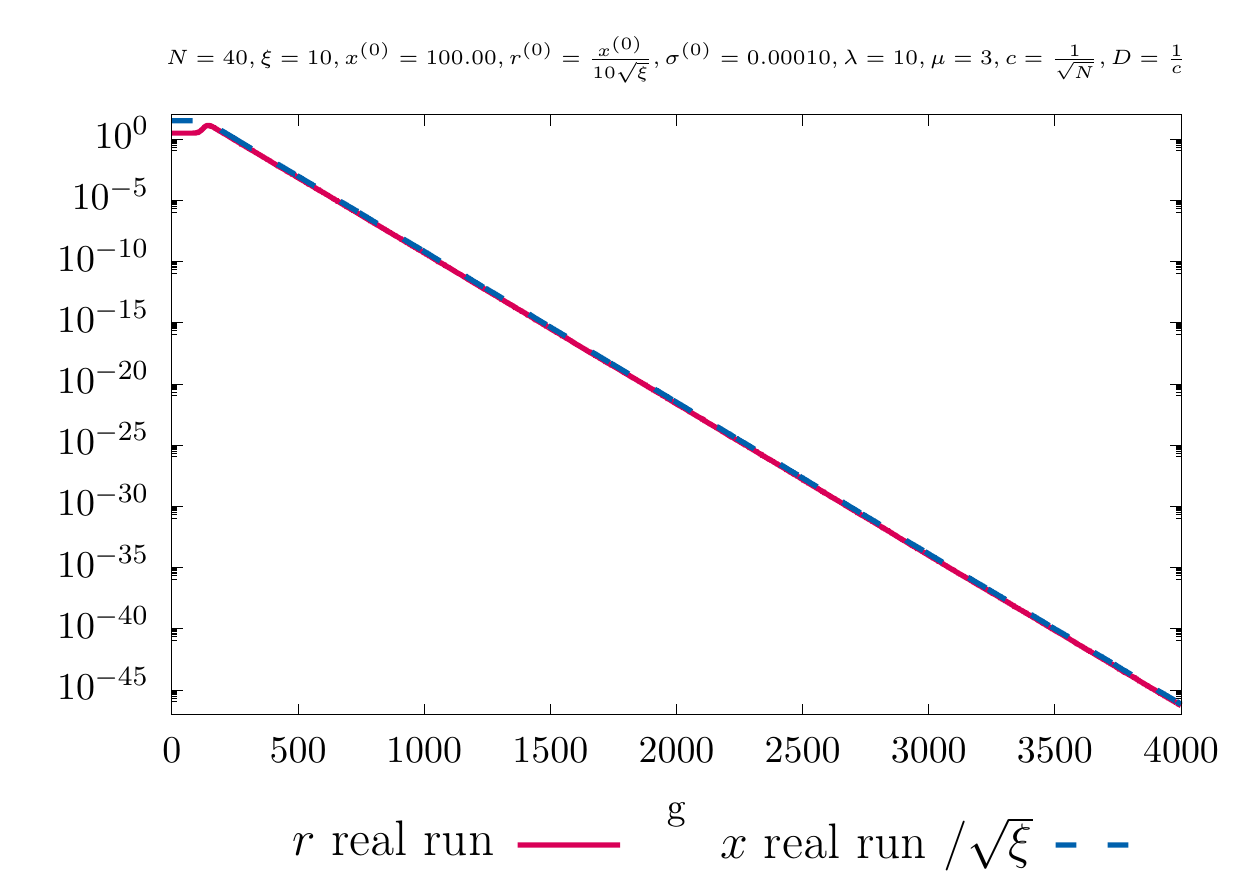}
  \end{tabular}
  \caption
  {Mean value dynamics closed-form approximation and real-run
  comparison of the $(3/3_I,10)$-CSA-ES
  with repair by projection
  applied to the conically constrained problem. (Part 3)}
  \label{sec:theoreticalanalysis:fig:dynamicsadditional3}
\end{figure}

\begin{figure}
  \centering
  \begin{tabular}{@{\hspace{-0.025\textwidth}}c@{\hspace{-0.025\textwidth}}c}
    \includegraphics[width=0.45\textwidth]{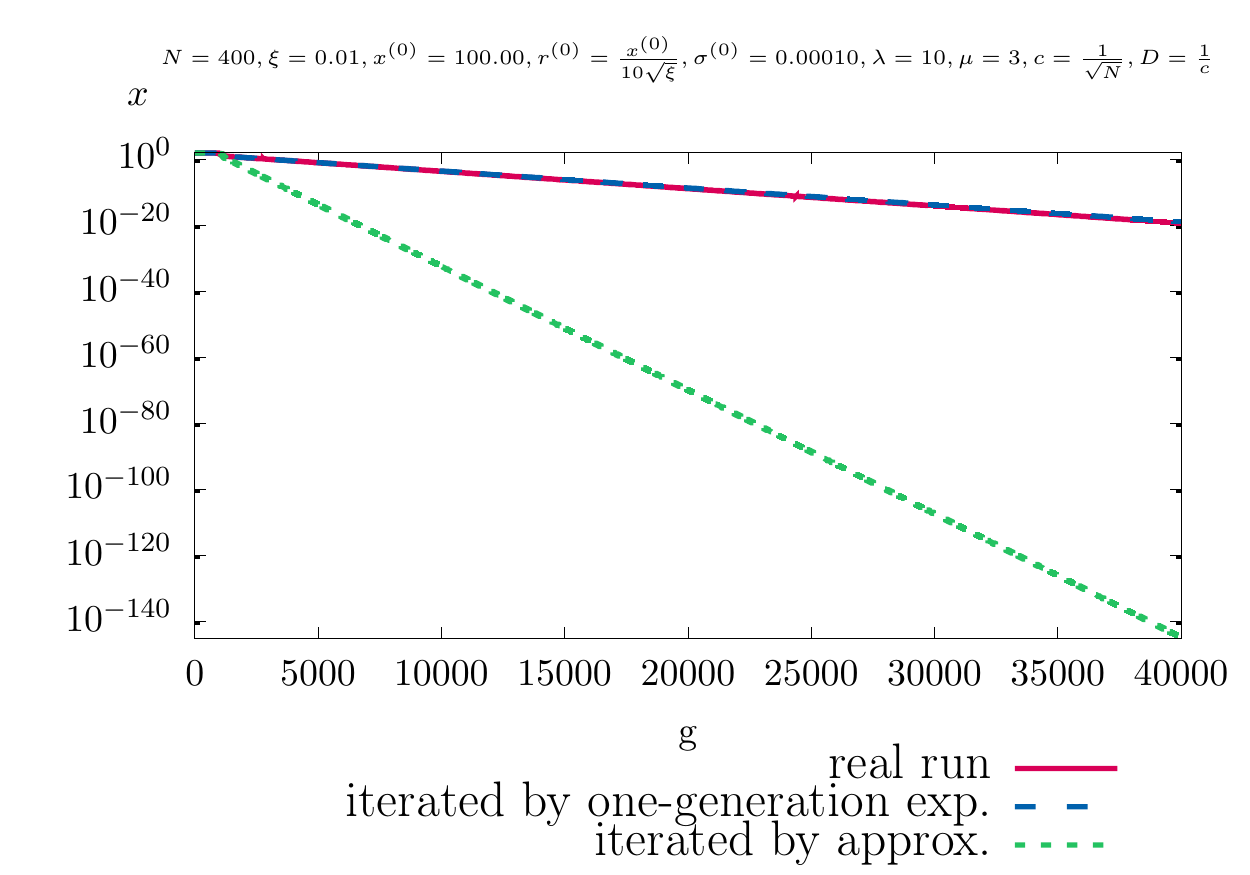}&
    \includegraphics[width=0.45\textwidth]{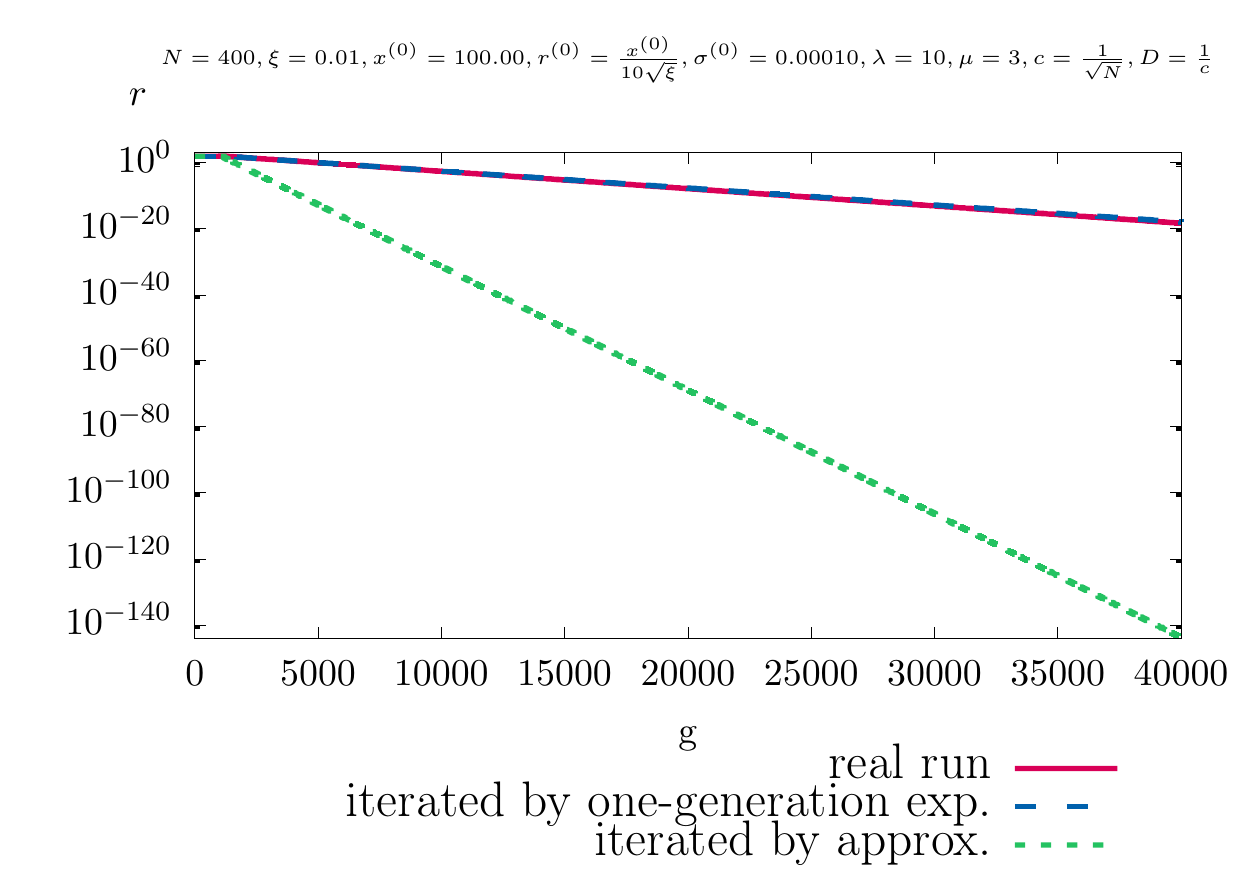}\\
    \includegraphics[width=0.45\textwidth]{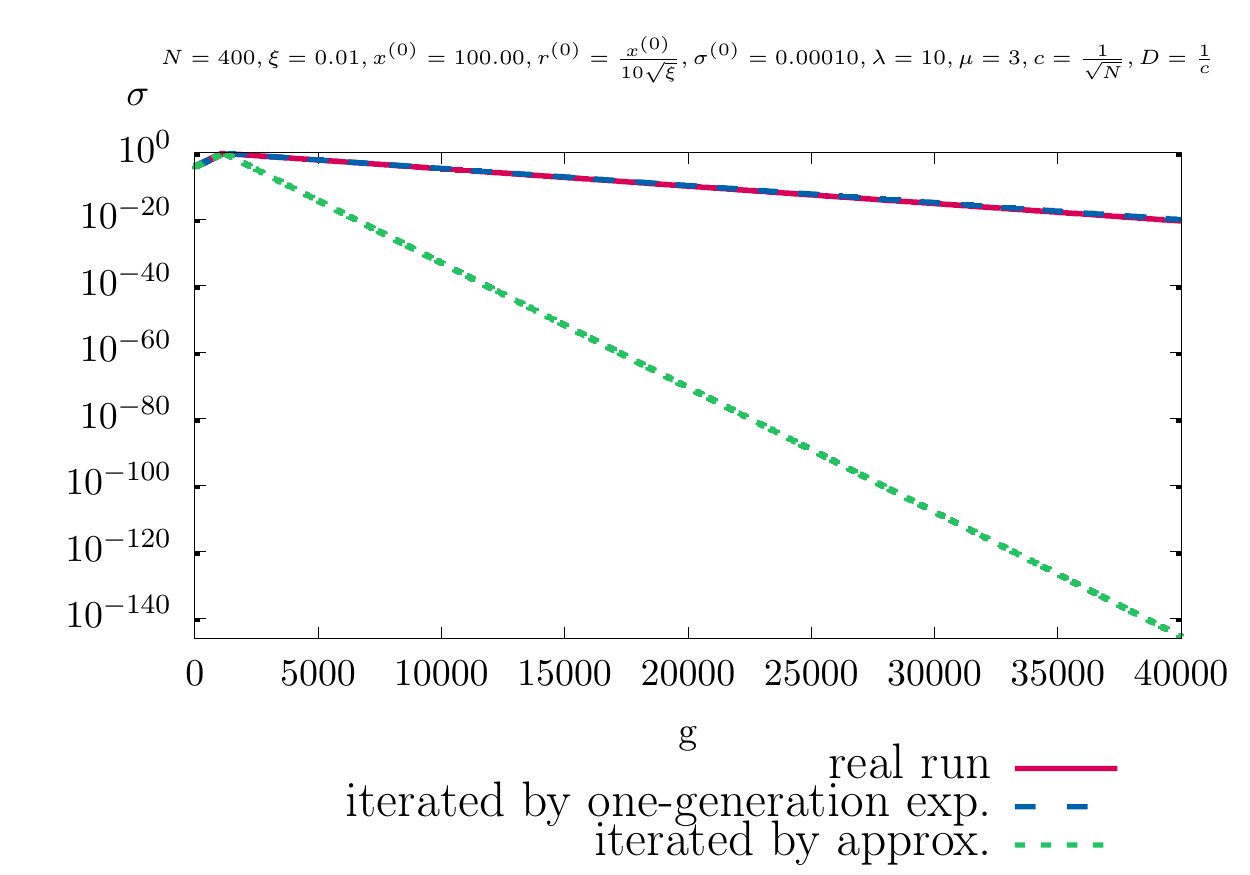}&
    \includegraphics[width=0.45\textwidth]{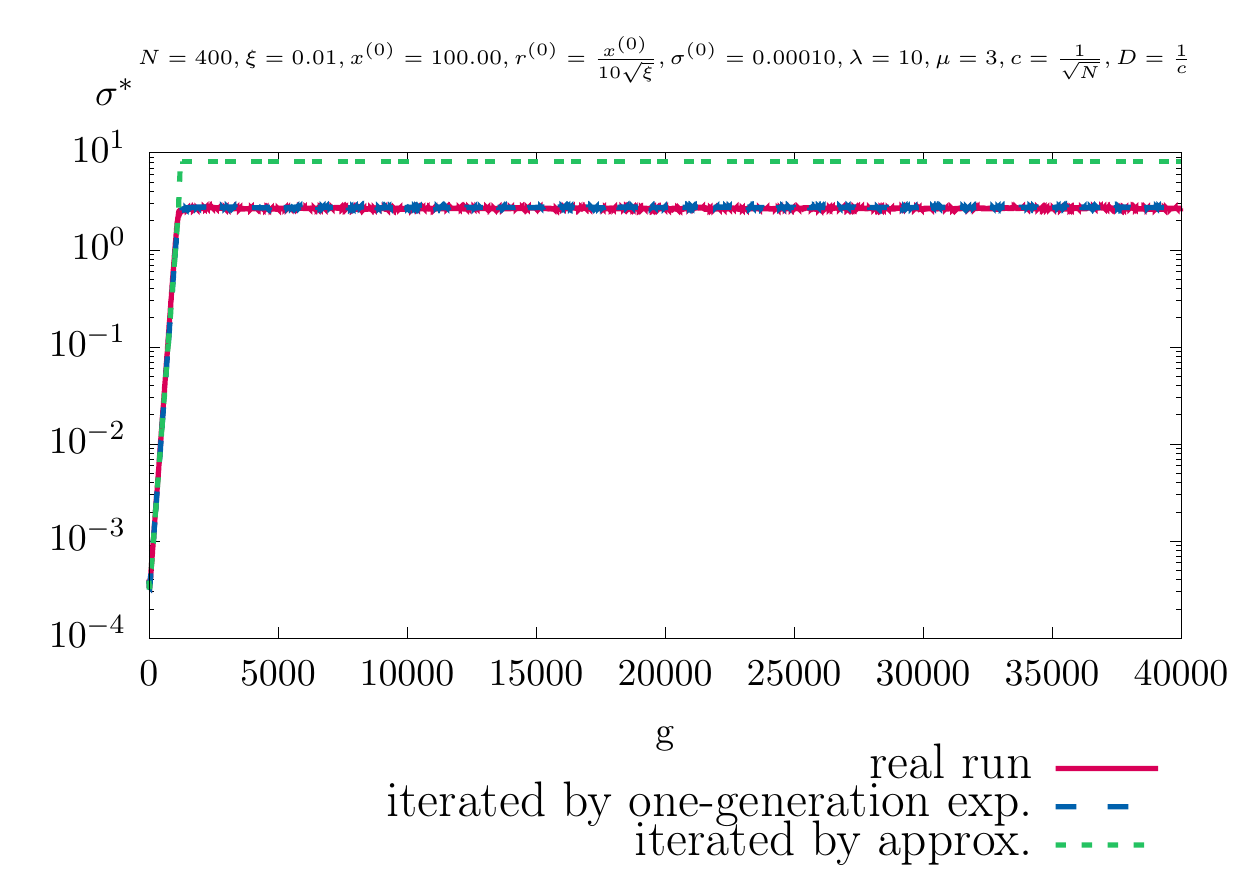}\\
    \includegraphics[width=0.45\textwidth]{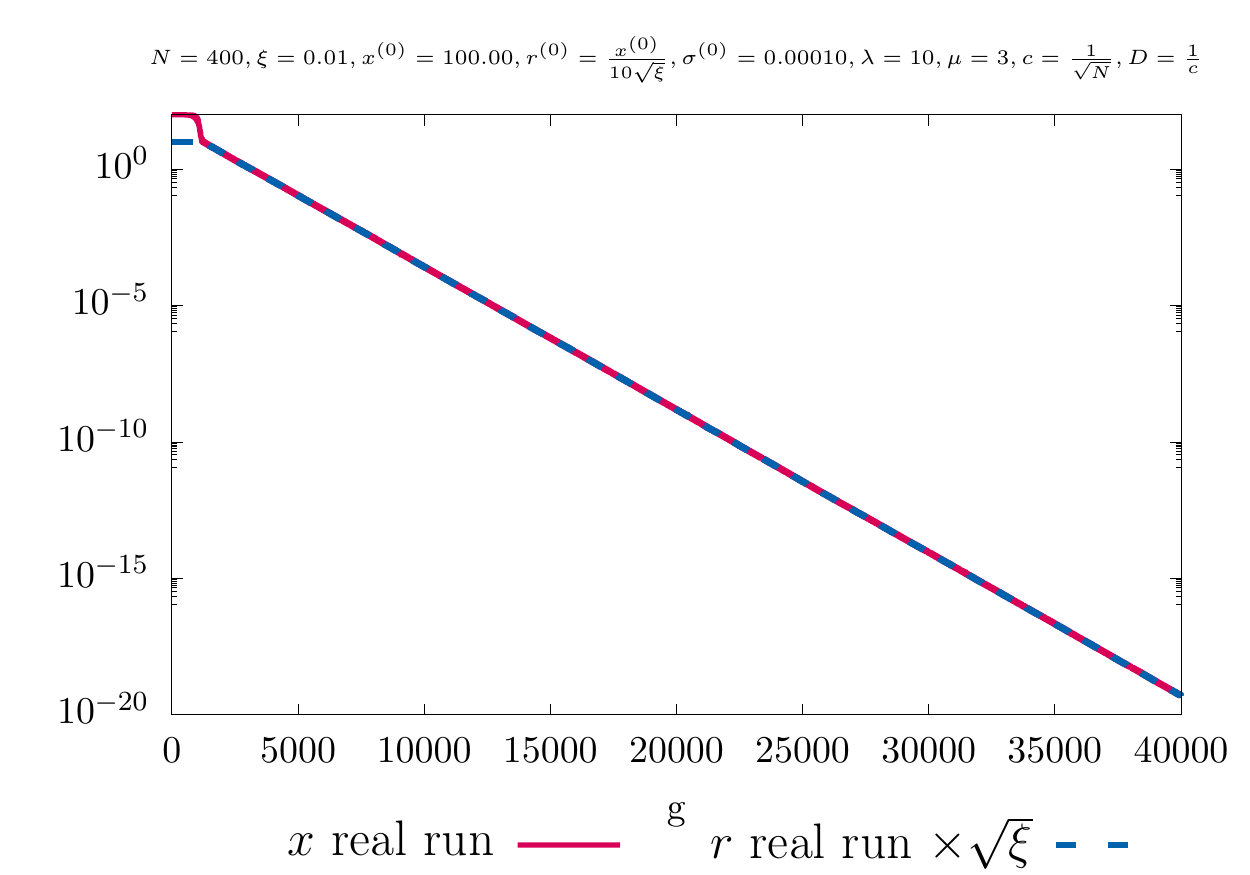}&
    \includegraphics[width=0.45\textwidth]{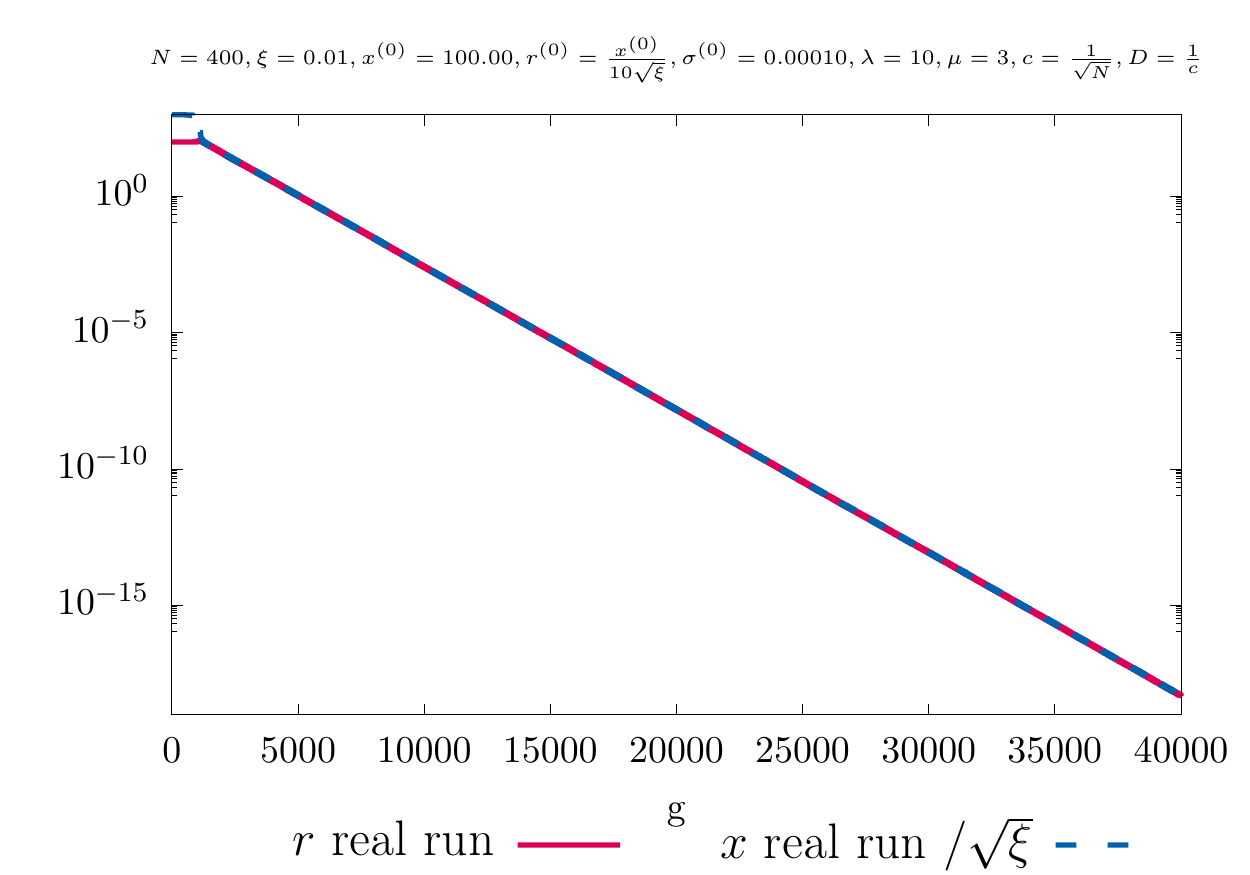}
  \end{tabular}
  \caption
  {Mean value dynamics closed-form approximation and real-run
  comparison of the $(3/3_I,10)$-CSA-ES
  with repair by projection
  applied to the conically constrained problem. (Part 4)}
  \label{sec:theoreticalanalysis:fig:dynamicsadditional4}
\end{figure}

\begin{figure}
  \centering
  \begin{tabular}{@{\hspace{-0.025\textwidth}}c@{\hspace{-0.025\textwidth}}c}
    \includegraphics[width=0.45\textwidth]{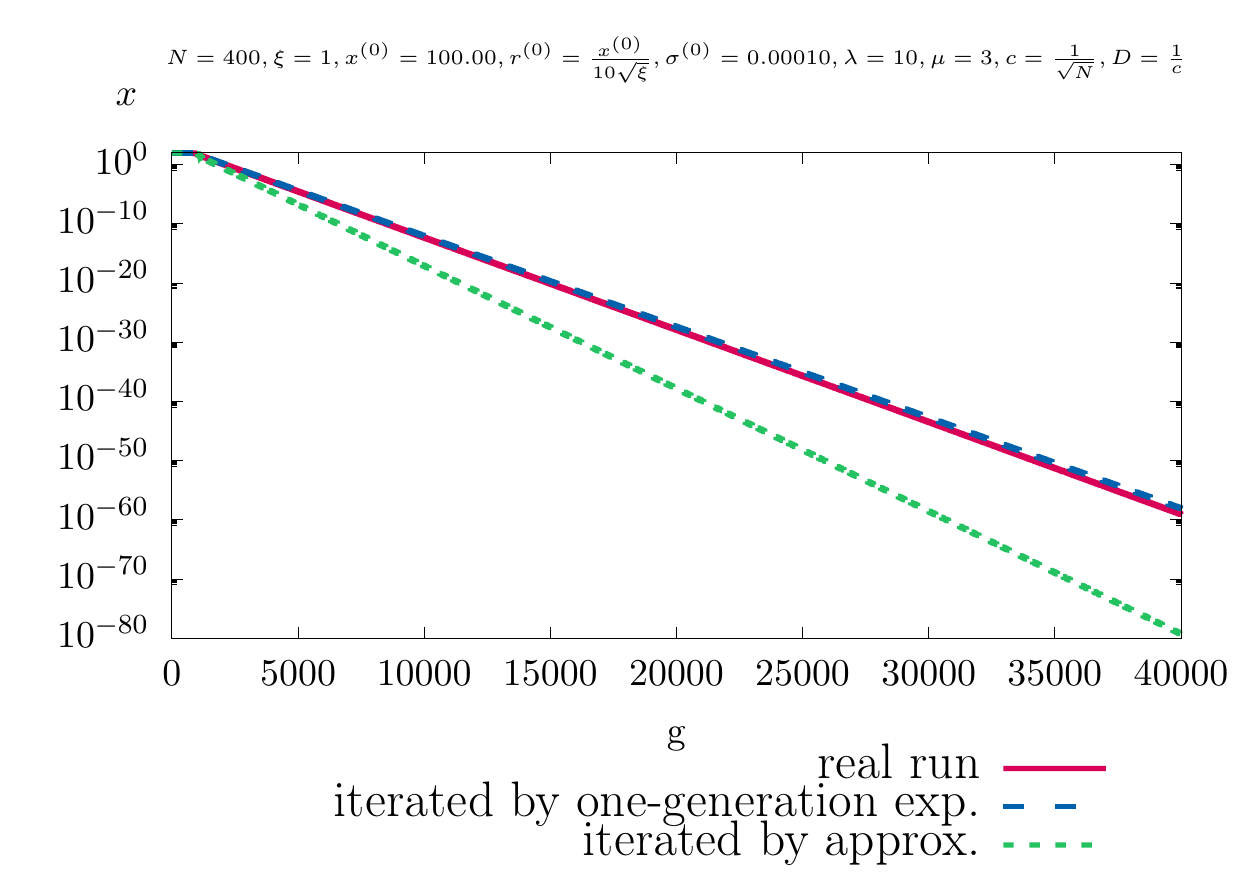}&
    \includegraphics[width=0.45\textwidth]{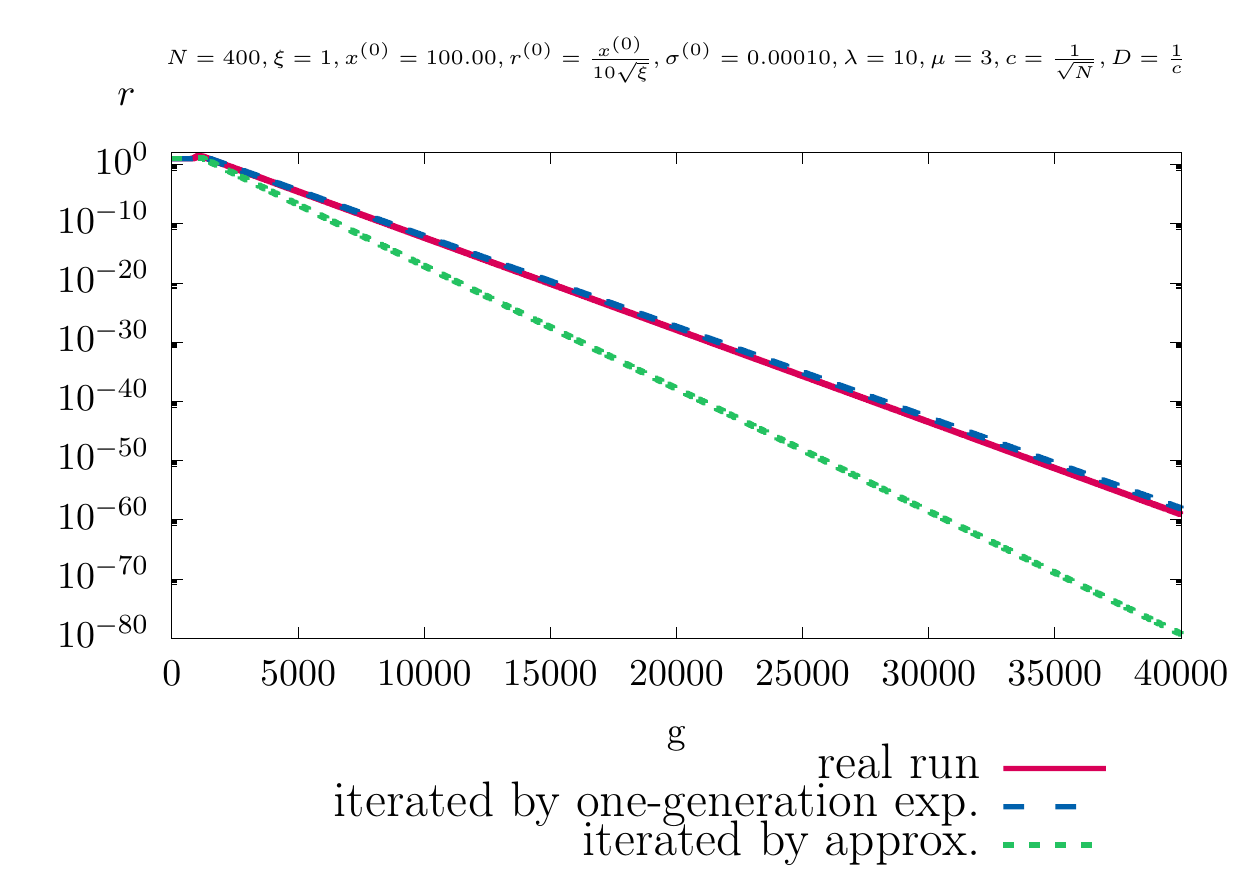}\\
    \includegraphics[width=0.45\textwidth]{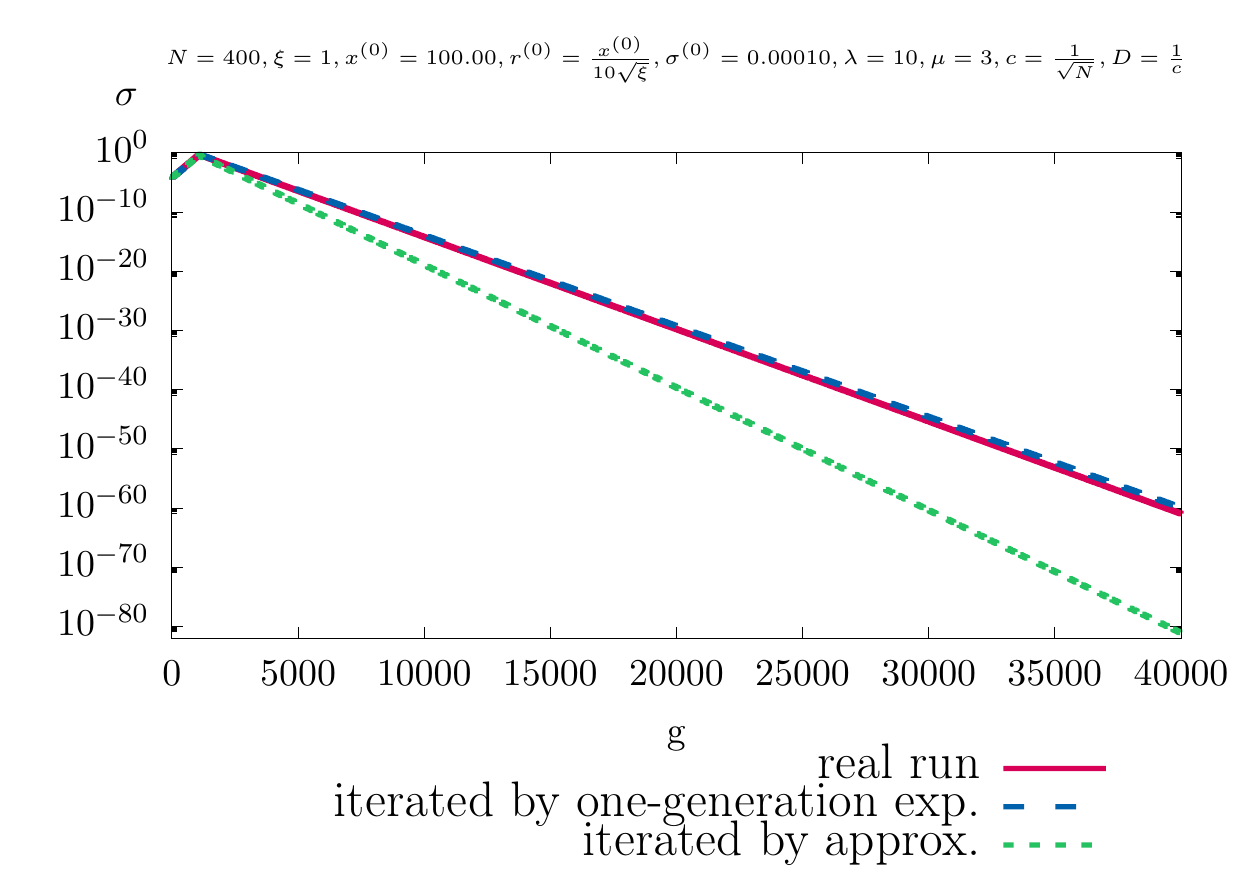}&
    \includegraphics[width=0.45\textwidth]{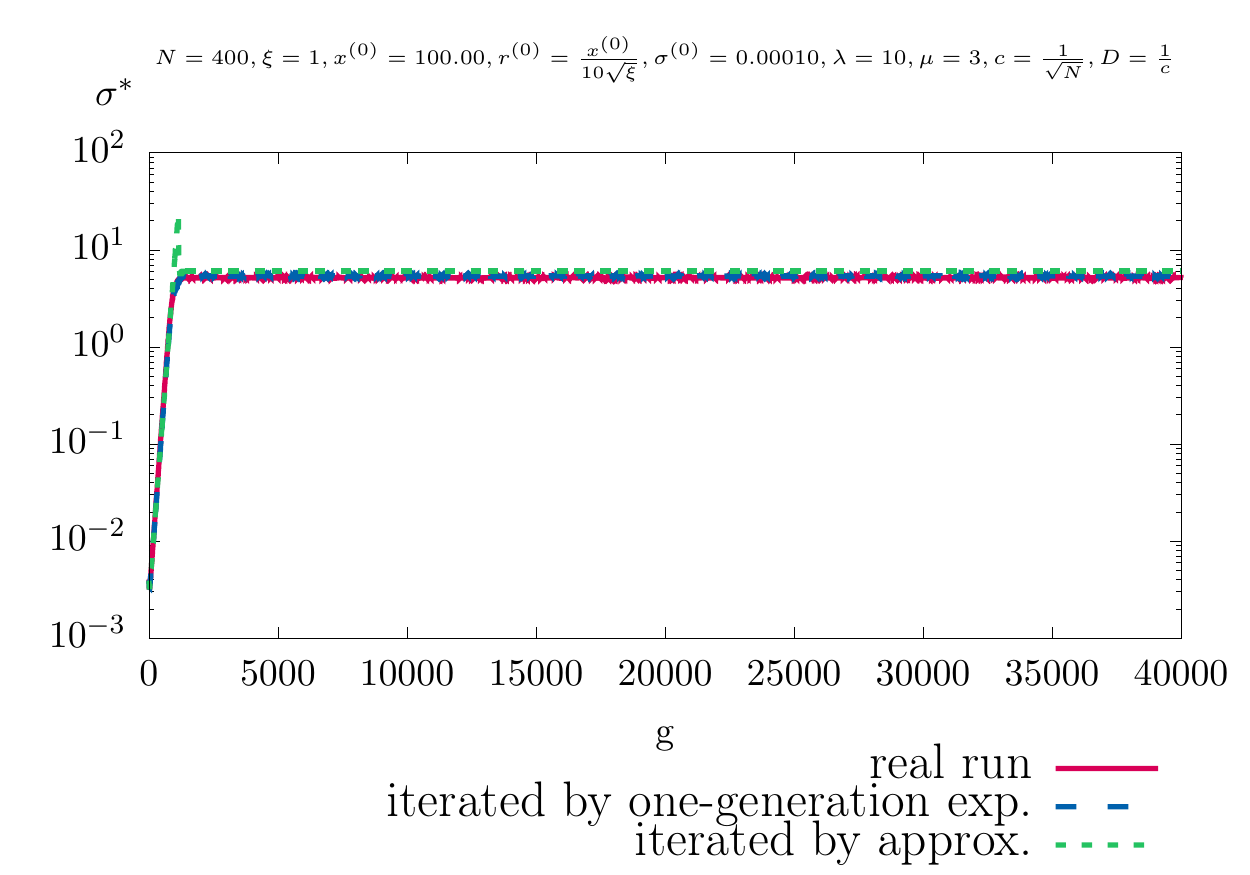}\\
    \includegraphics[width=0.45\textwidth]{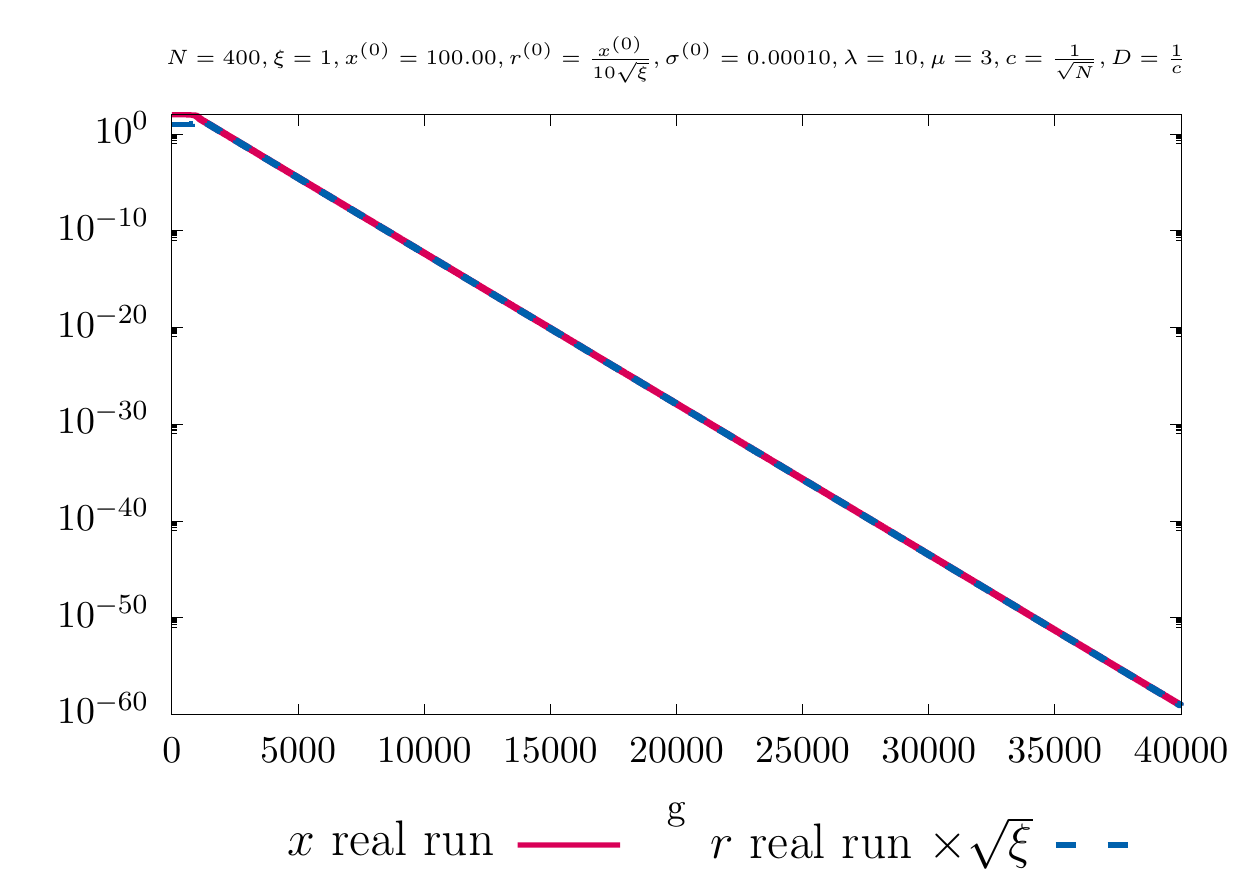}&
    \includegraphics[width=0.45\textwidth]{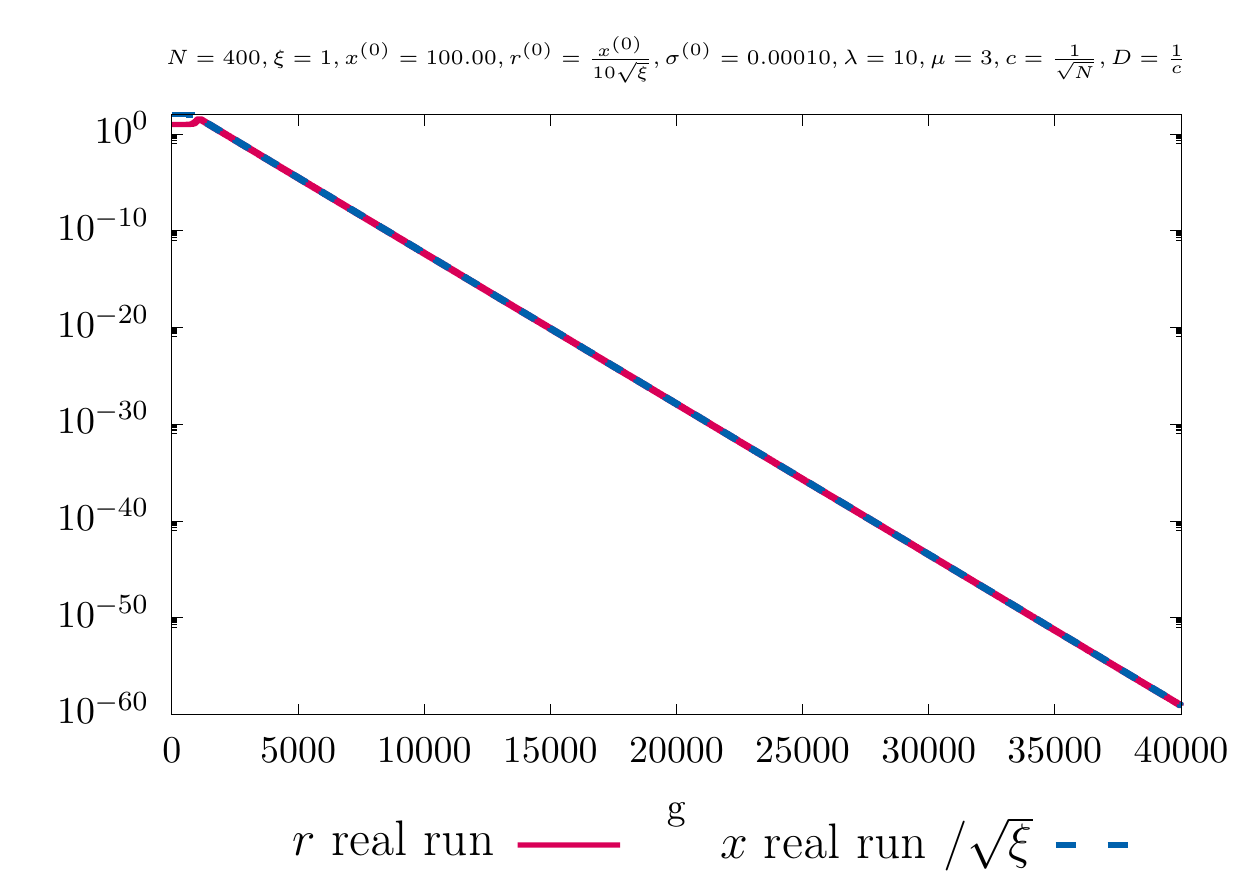}
  \end{tabular}
  \caption
  {Mean value dynamics closed-form approximation and real-run
  comparison of the $(3/3_I,10)$-CSA-ES
  with repair by projection
  applied to the conically constrained problem. (Part 5)}
  \label{sec:theoreticalanalysis:fig:dynamicsadditional5}
\end{figure}

\begin{figure}
  \centering
  \begin{tabular}{@{\hspace{-0.025\textwidth}}c@{\hspace{-0.025\textwidth}}c}
    \includegraphics[width=0.45\textwidth]{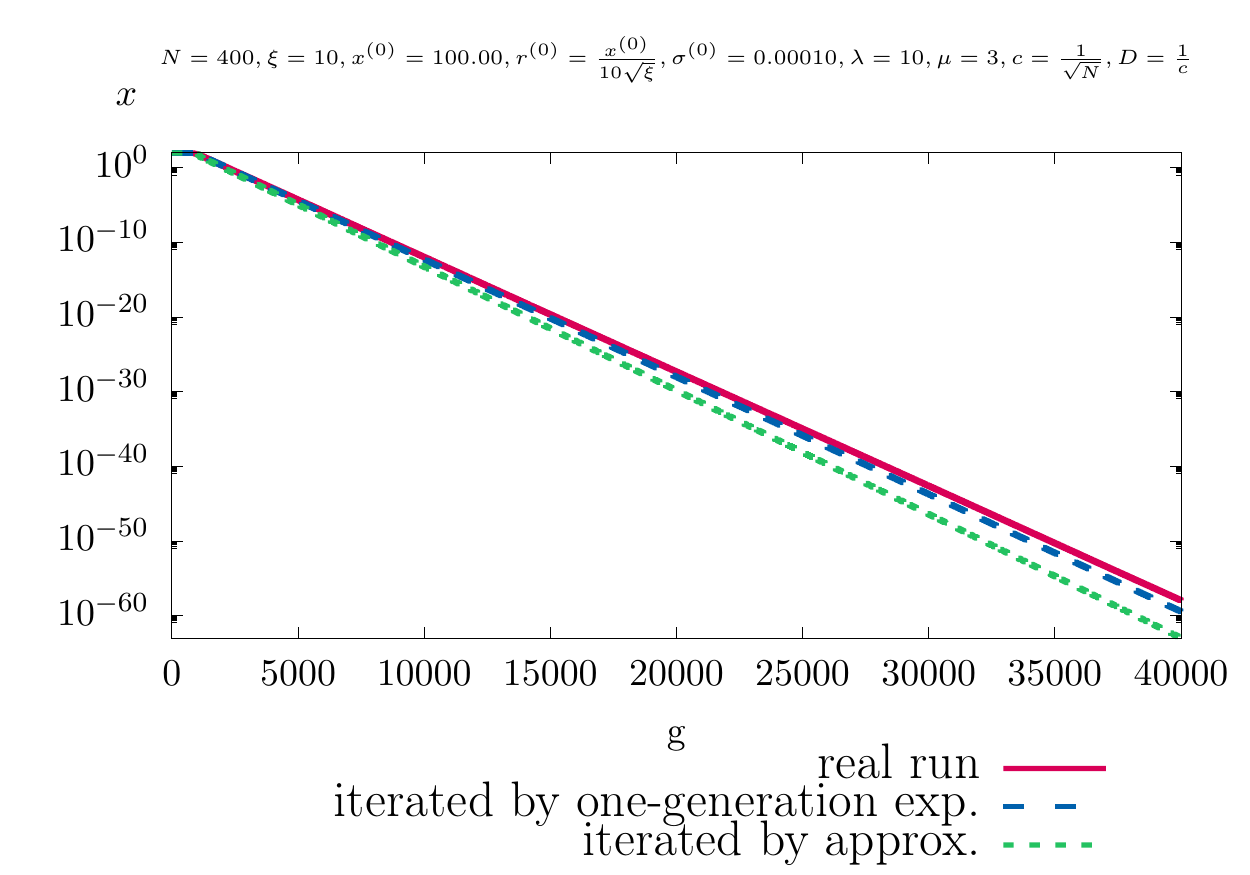}&
    \includegraphics[width=0.45\textwidth]{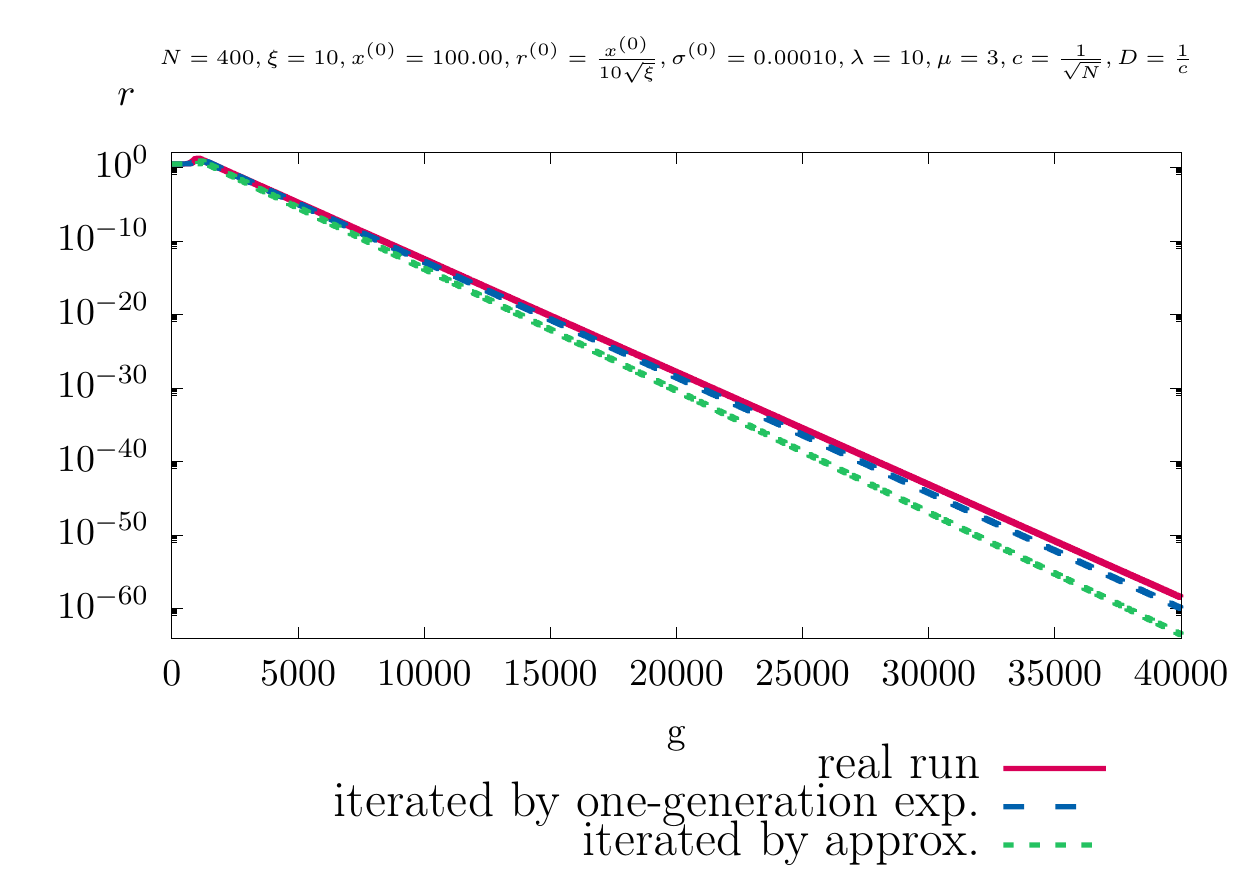}\\
    \includegraphics[width=0.45\textwidth]{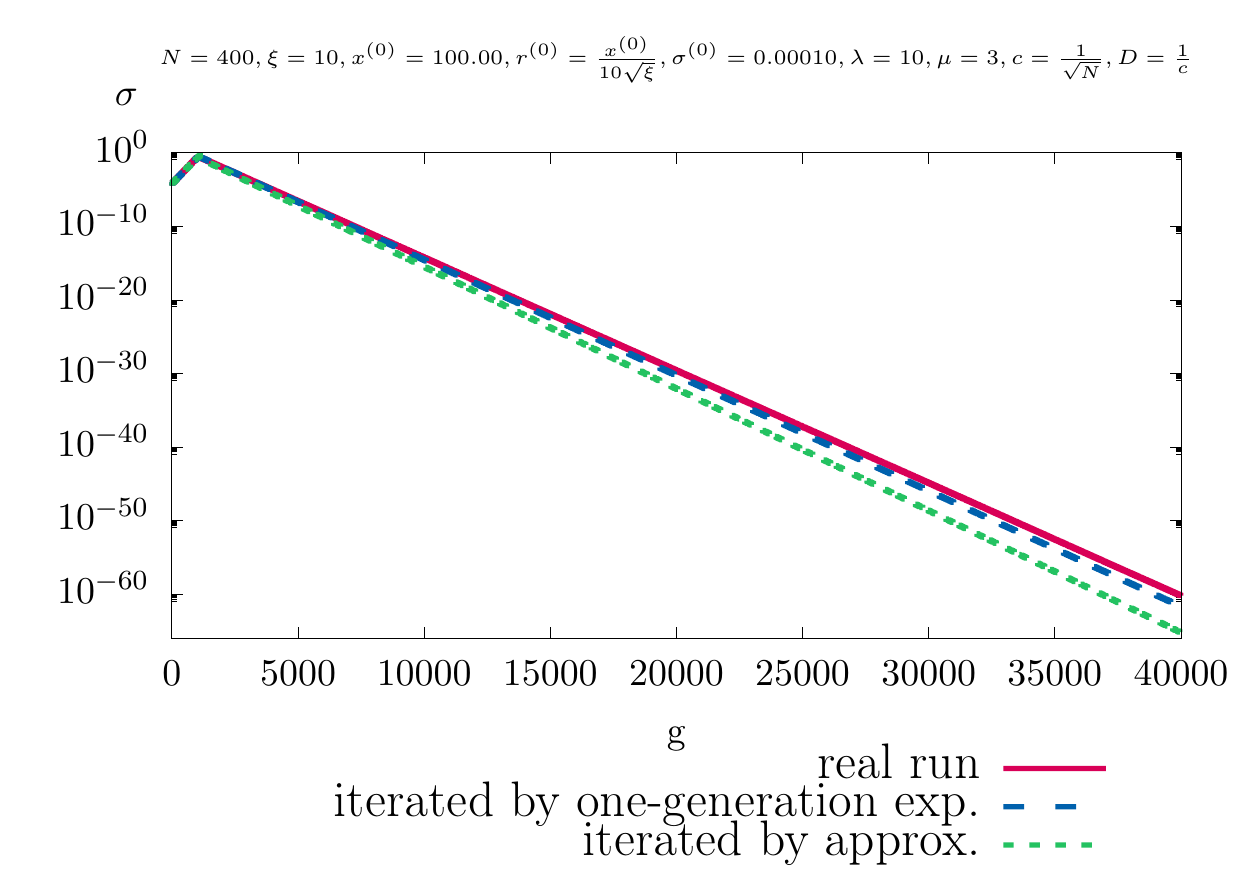}&
    \includegraphics[width=0.45\textwidth]{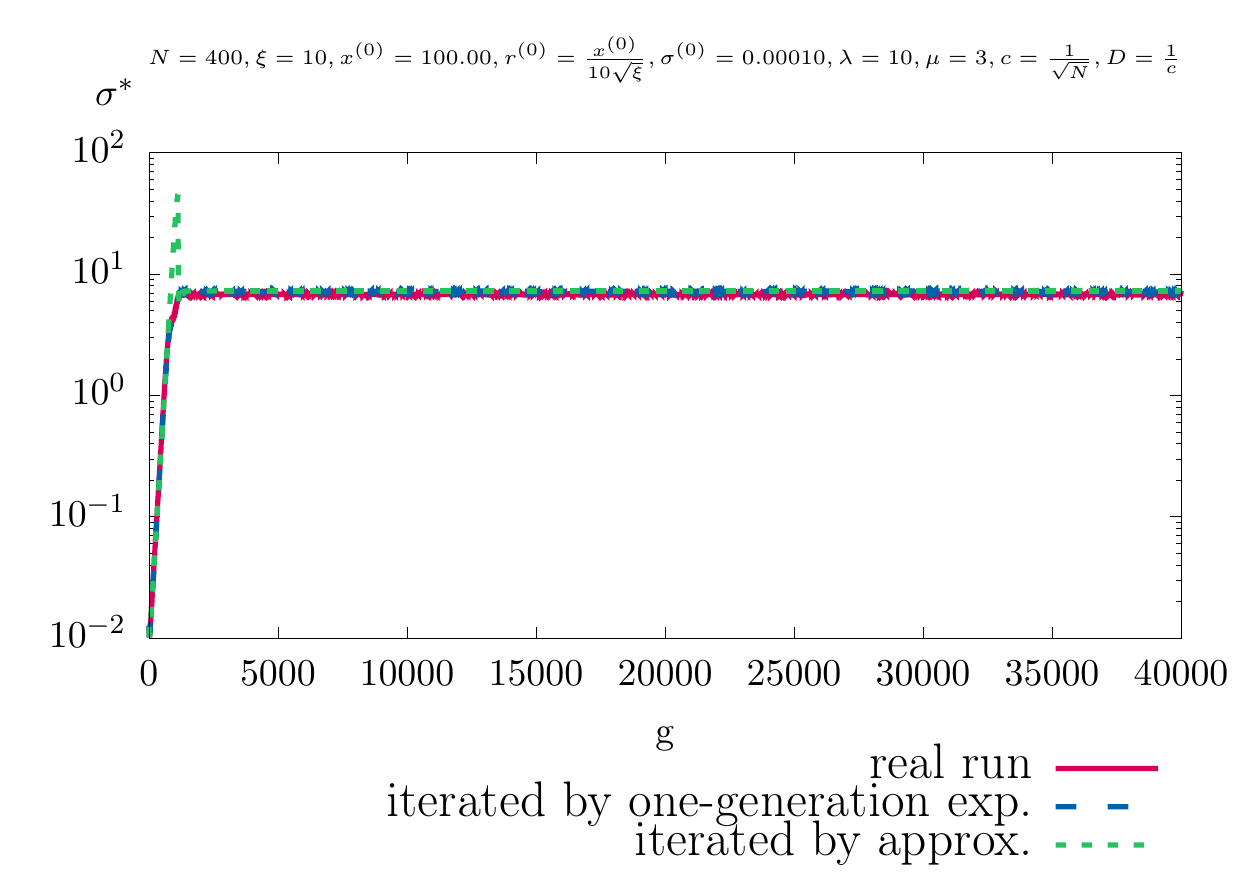}\\
    \includegraphics[width=0.45\textwidth]{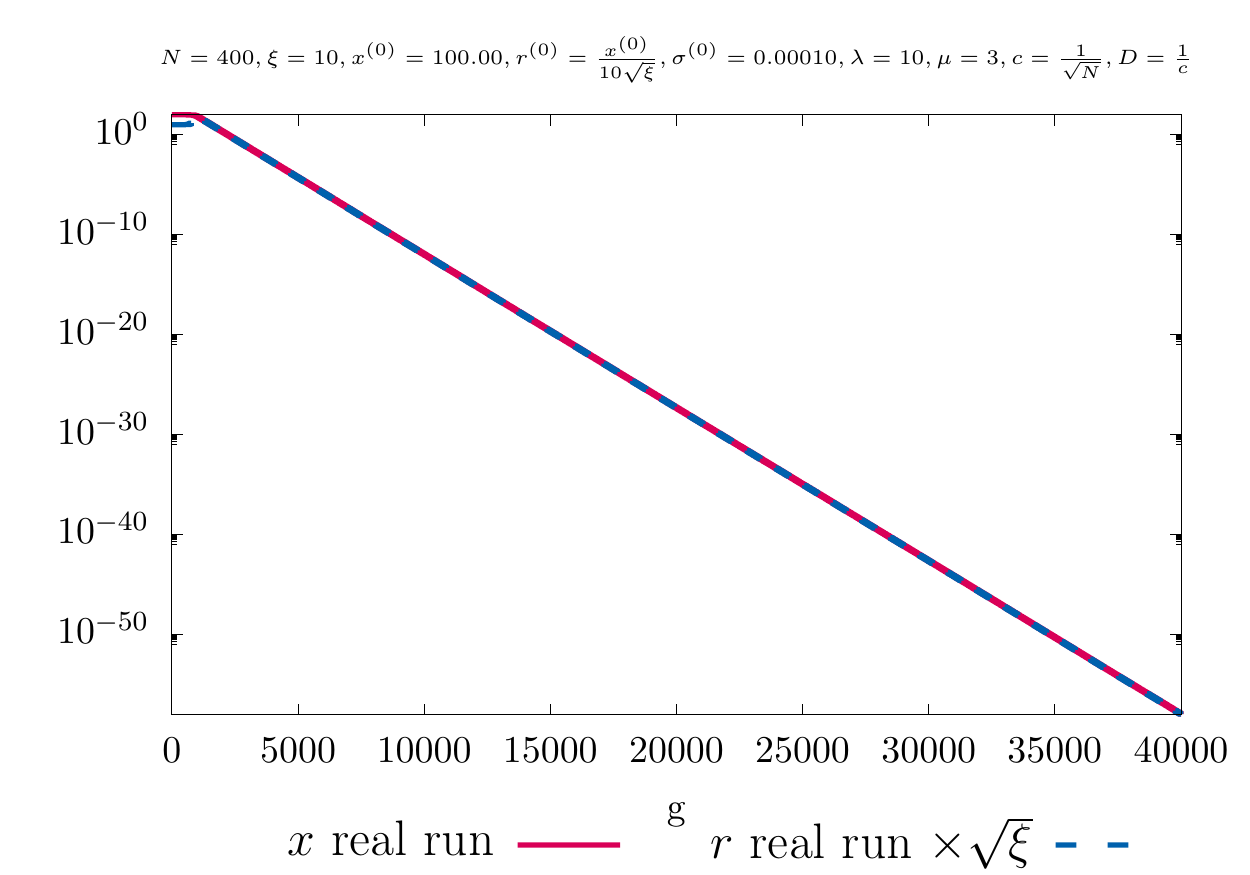}&
    \includegraphics[width=0.45\textwidth]{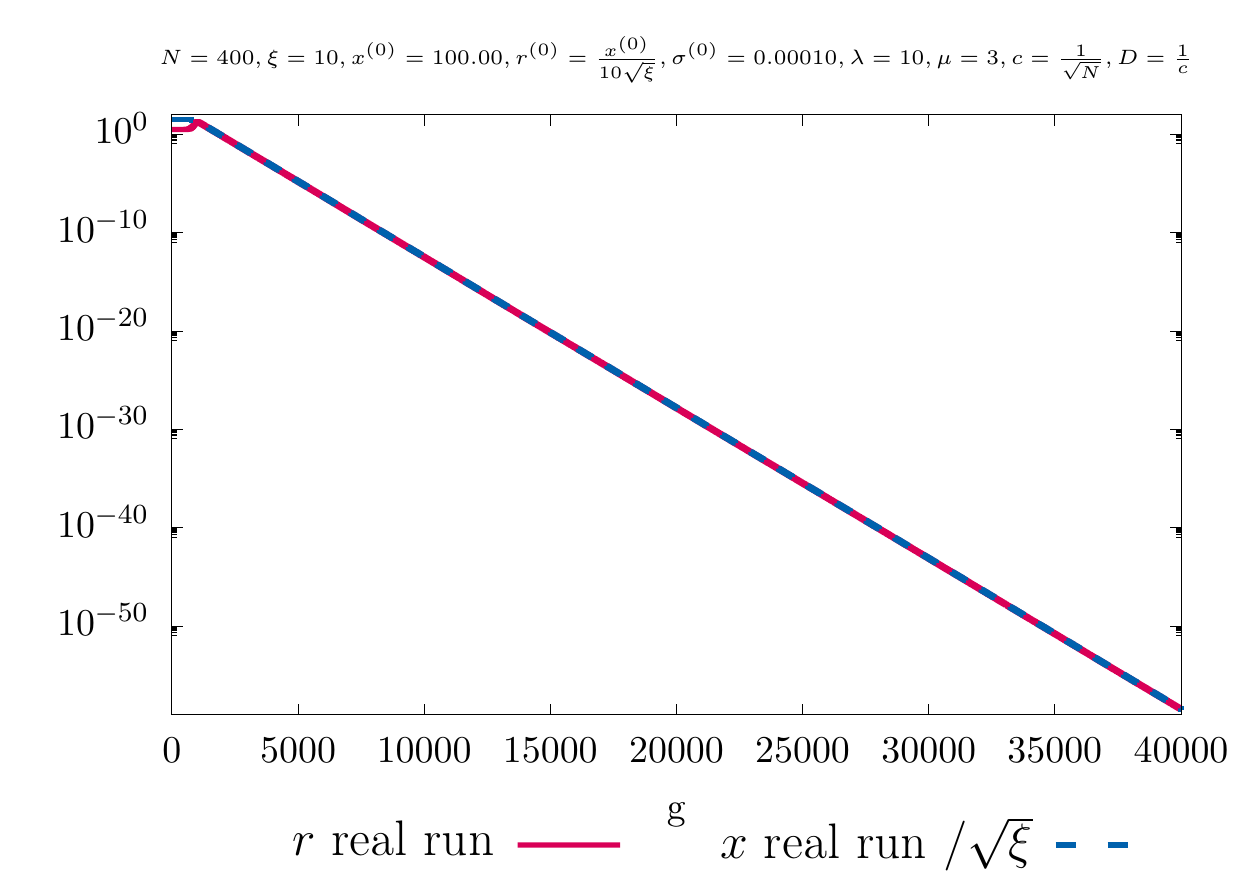}
  \end{tabular}
  \caption
  {Mean value dynamics closed-form approximation and real-run
  comparison of the $(3/3_I,10)$-CSA-ES
  with repair by projection
  applied to the conically constrained problem. (Part 6)}
  \label{sec:theoreticalanalysis:fig:dynamicsadditional6}
\end{figure}

\Crefrange{sec:theoreticalanalysis:fig:dynamicsadditional1}
{sec:theoreticalanalysis:fig:dynamicsadditional6}
show the mean value dynamics of the $(3/3_I,10)$-CSA-ES
applied to the conically constrained problem with different parameters
as indicated in the title of the subplots.
\renewcommand{\sppatemp}
{The plots are organized into
three rows and two columns. The first two rows show the
$x$ (first row, first column),
$r$ (first row, second column),
$\sigma$ (second row, first column),
and $\sigma^*$ (second row, second column) dynamics.
The third row shows $x$ and $r$ converted into each other by $\sqrt{\xi}$.
The third row shows that after some initial phase, the ES transitions into
a stationary state. In this steady state, the ES moves along the
cone boundary. This becomes clear in the plots because the equation for
the cone boundary is $r = x / \sqrt{\xi}$ or equivalently $x = r\sqrt{\xi}$.
In the first two rows, the solid line has been generated
by averaging $100$ real runs of the ES. The dashed line has been
determined by iterating the mean value iterative system of
\Cref{sec:theoreticalanalysis:subsec:evolutionequationssummary}
with one-generation experiments for
${\varphi^{(g)}_{x}}$, ${\varphi^{(g)}_{x}}^*$,
${\varphi^{(g)}_{r}}$, ${\varphi^{(g)}_{r}}^*$,
and $\varphi_{r^2}^{(g)}$.
The dotted lines have been
computed by iterating the mean value iterative system
with the derived approximations as indicated in the derivations
leading to the equations in
\Cref{sec:theoreticalanalysis:subsec:evolutionequationssummary}
for ${\varphi^{(g)}_{x}}$, ${\varphi^{(g)}_{x}}^*$,
${\varphi^{(g)}_{r}}$, ${\varphi^{(g)}_{r}}^*$,
and $\varphi_{r^2}^{(g)}$.
Due to the approximations used it is possible that in a generation $g$,
the iteration of the mean value iterative system yields infeasible
$(x^{(g)},r^{(g)})^T$. In such cases, the particular $(x^{(g)},r^{(g)})^T$
have been projected back and projected values have been used in the further
iterations.}\sppatemp{}


\section{Further Investigations Considering the Derivation of Closed-Form
  Approximations for the Steady State with the Assumptions
  $c = O\left(\frac{1}{N}\right)$ and $N \rightarrow \infty$}
\label{sec:appendix:steadystate1divN}
By plotting the left-hand side
of~\eqref{sec:theoreticalanalysis:eq:sigmass1divNequation4}
for different parameters, one observes that for the values of
interest (${\sigma_{ss}^*} > 0$), the function is quadratic.

Inspired by that, a Taylor expansion around $a > 0$ is performed
up to and including the quadratic term.
It results in a quadratic equation that can be solved
for $\sigma_{ss}^* > 0$.
\Cref{sec:theoreticalanalysis:fig:steadystatecomparisonclosedform1divN}
shows plots of the steady state computations with this approximation
compared to real ES runs.
\renewcommand{\sppatemp}{The values for the points denoting the approximations
have been determined by computing the normalized steady state
mutation strength $\sigma_{ss}^*$ using the solution of the
mentioned quadratic equation that has been derived by a Taylor expansion.
The results for
$\varphi_x^*$ and $\varphi_r^*$ have been determined by using the
computed steady state $\sigma_{ss}^*$ values with
\Cref{sec:theoreticalanalysis:eq:varphixnormalizedss4}.
The approximations for $\left(\frac{x}{\sqrt{\xi}r}\right)_{ss}$ have been
determined by evaluating
\Cref{sec:theoreticalanalysis:eq:steadystatedist2}.
The values for the points denoting the experiments have been determined by
computing the averages of the particular values in real ES
runs.}\sppatemp{}

\begin{figure}
  \centering
  \begin{tabular}{@{\hspace{-0.025\textwidth}}c@{\hspace{-0.025\textwidth}}c@{\hspace{-0.025\textwidth}}c}
    \includegraphics[width=0.35\textwidth]{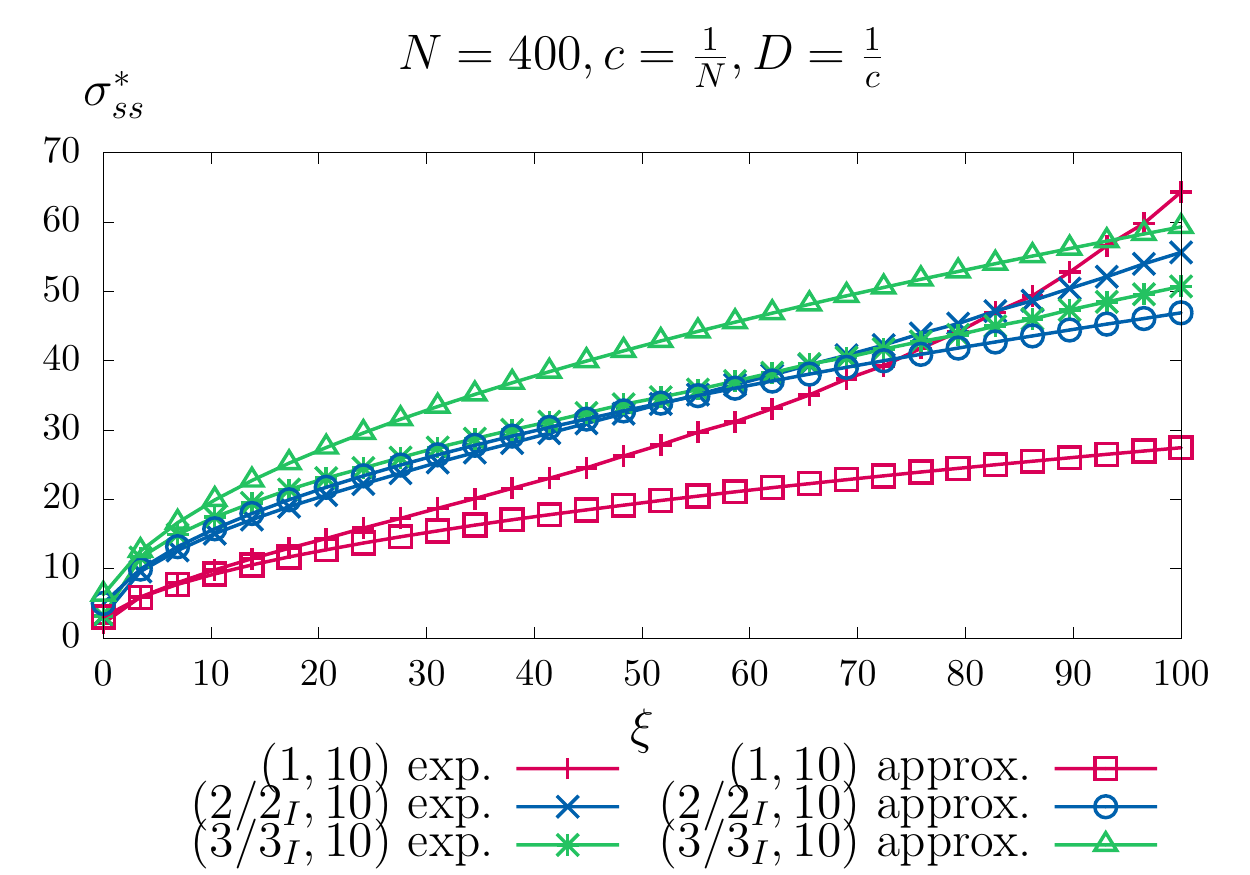}&
    \includegraphics[width=0.35\textwidth]{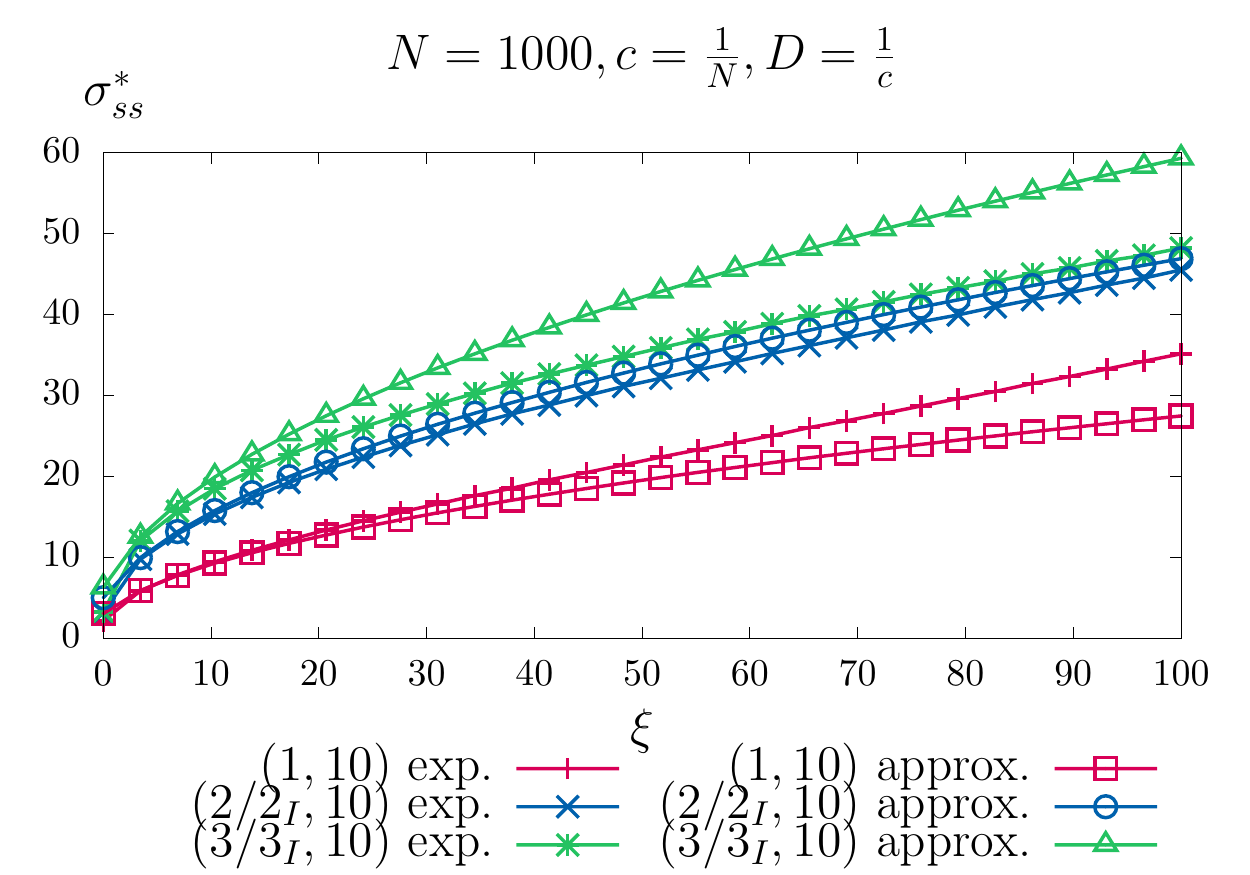}&
    \includegraphics[width=0.35\textwidth]{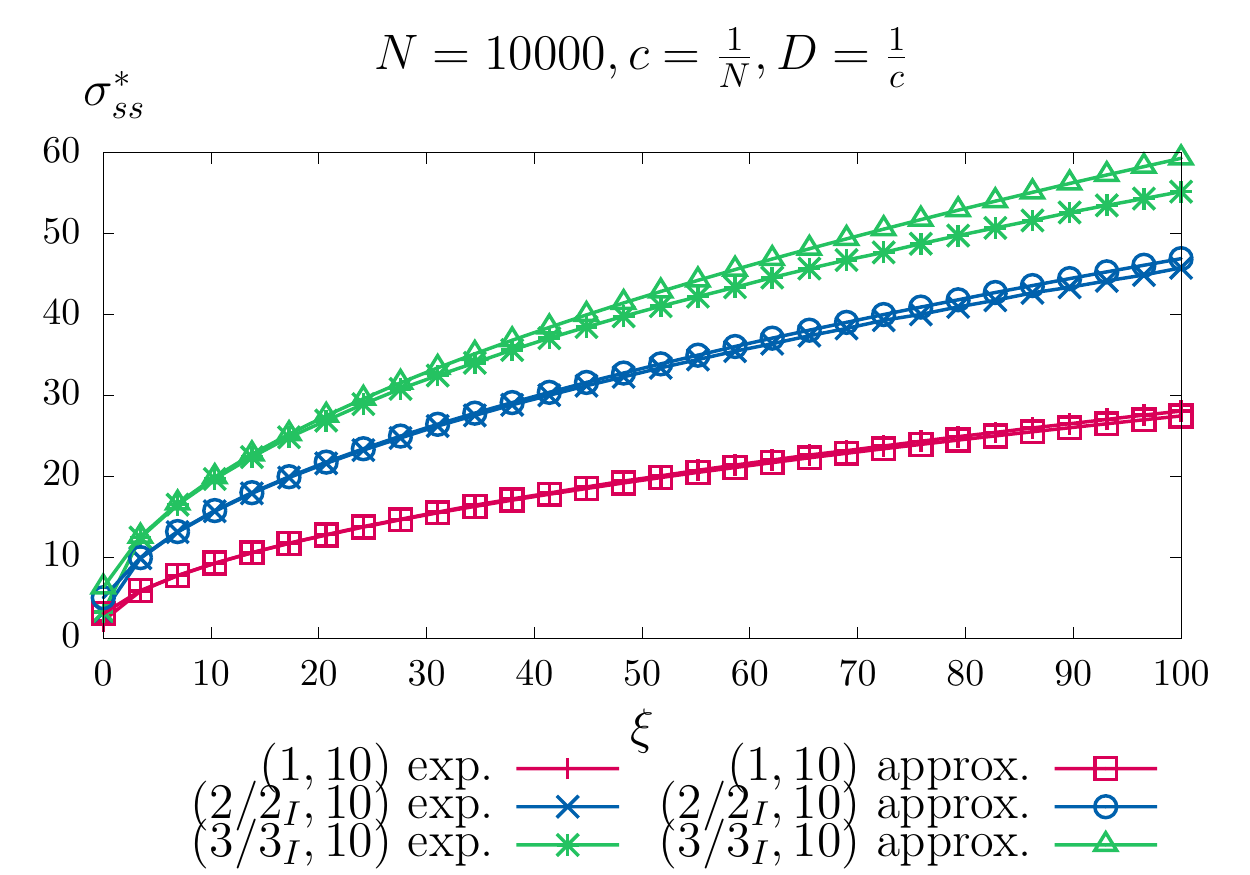}\\
    \includegraphics[width=0.35\textwidth]{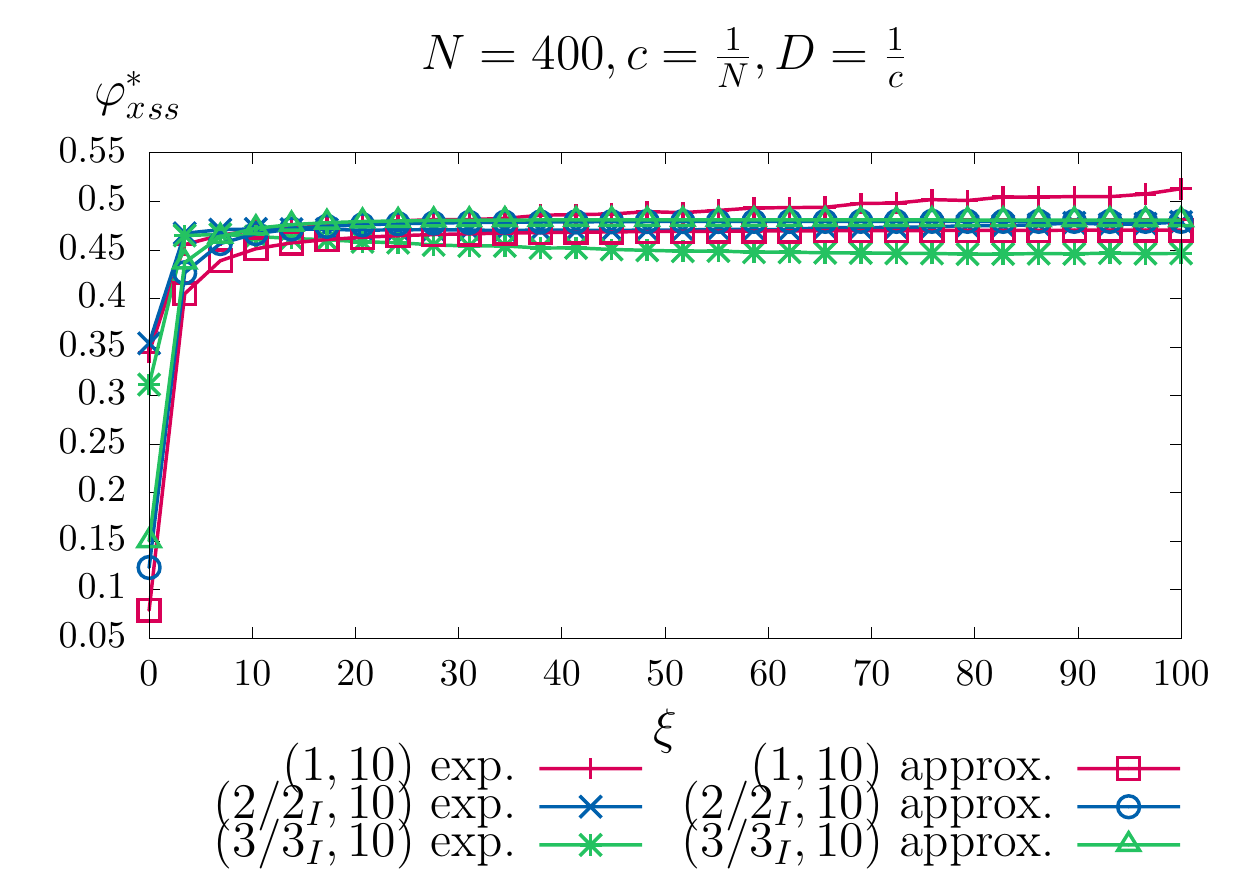}&
    \includegraphics[width=0.35\textwidth]{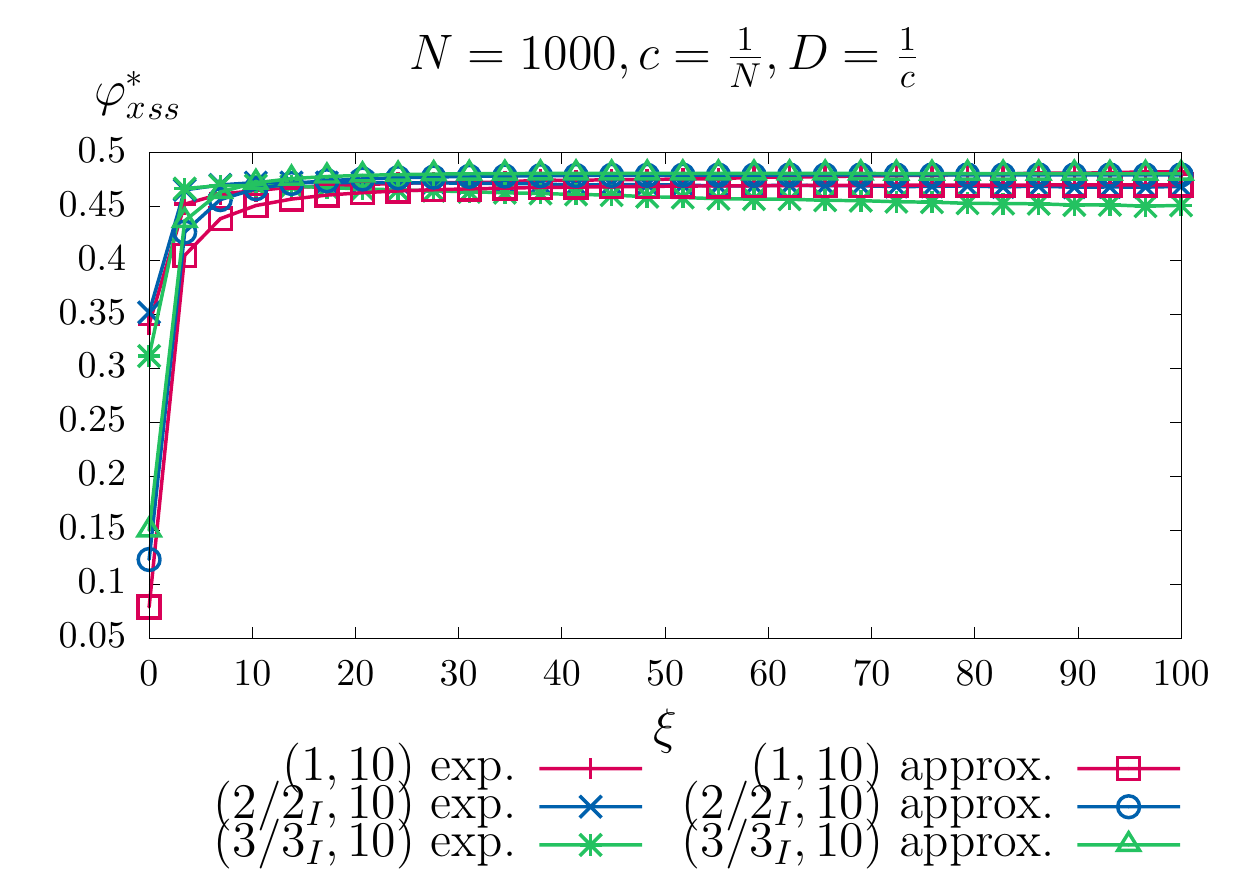}&
    \includegraphics[width=0.35\textwidth]{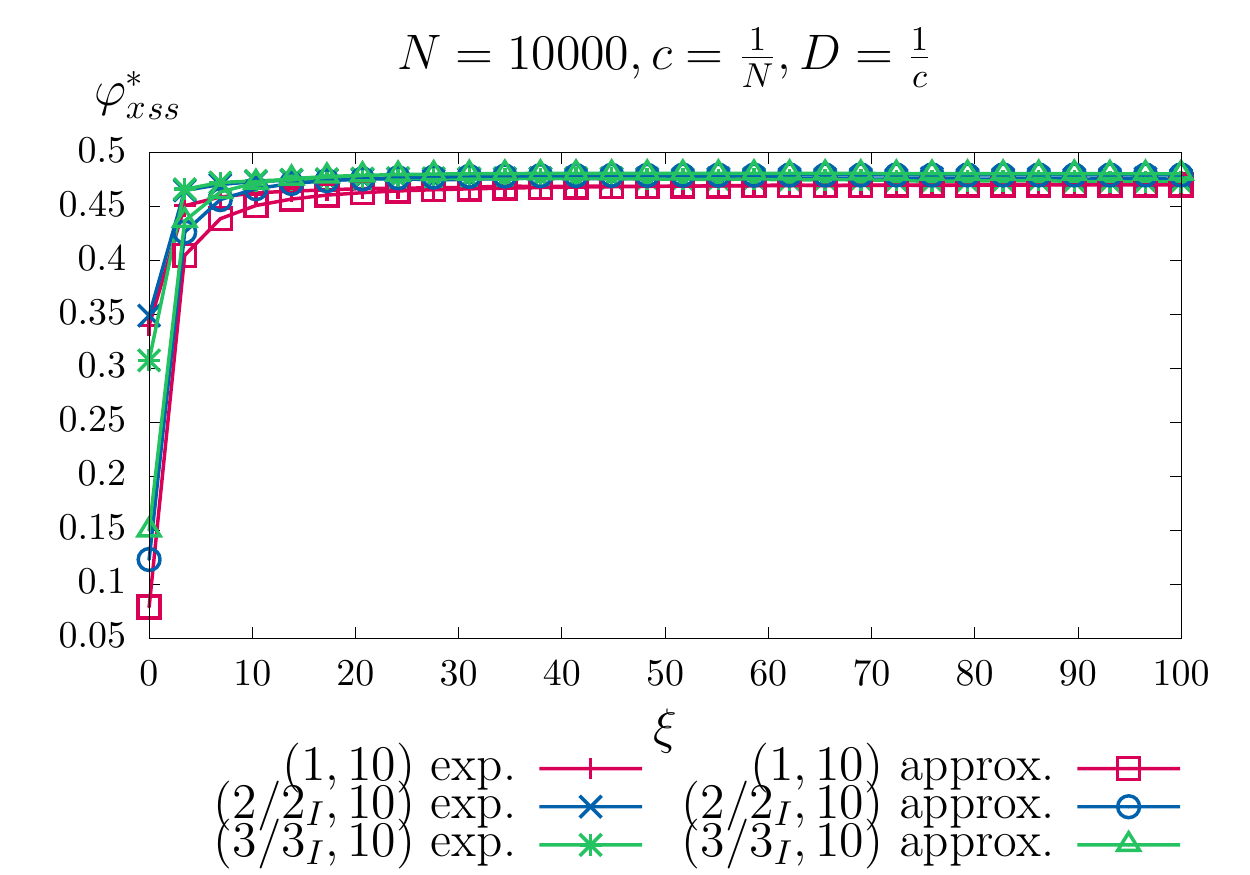}\\
    \includegraphics[width=0.35\textwidth]{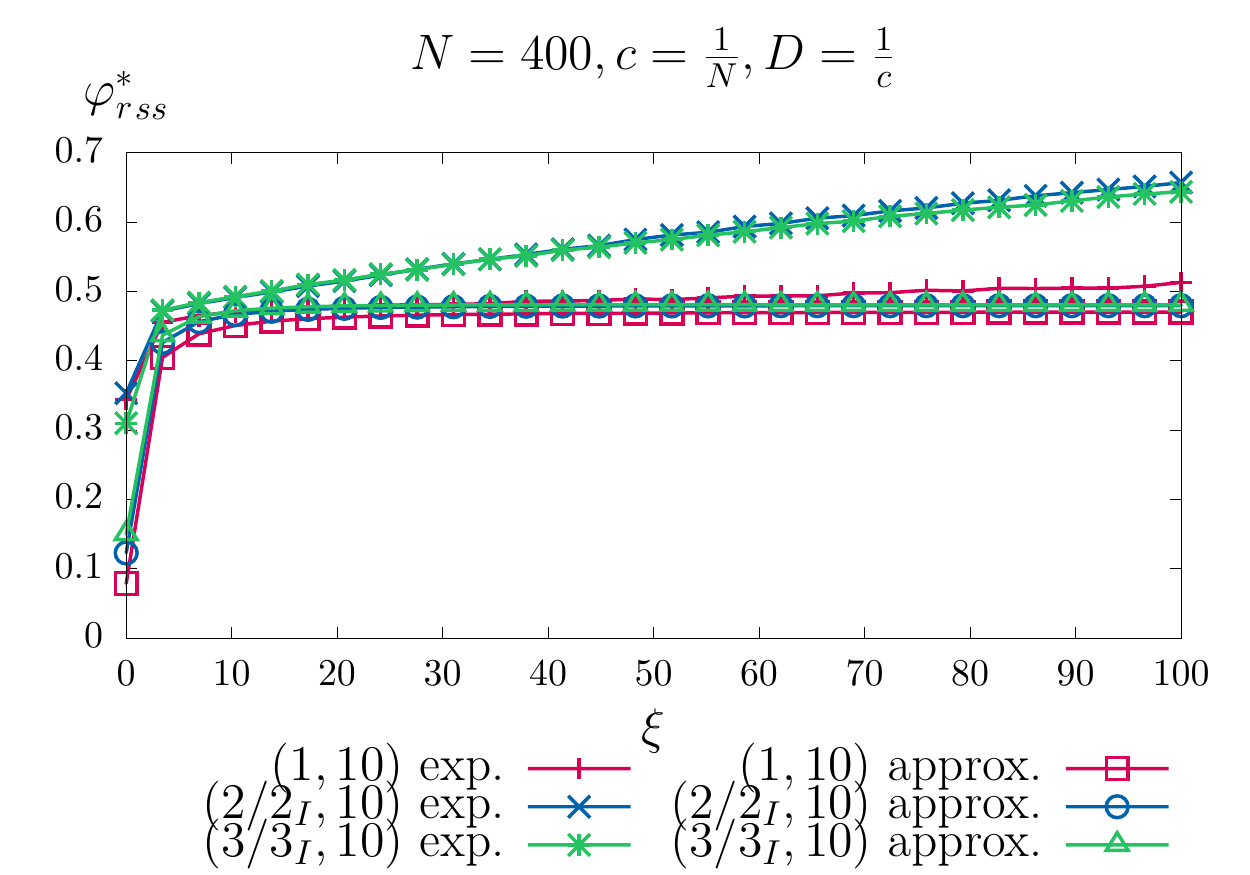}&
    \includegraphics[width=0.35\textwidth]{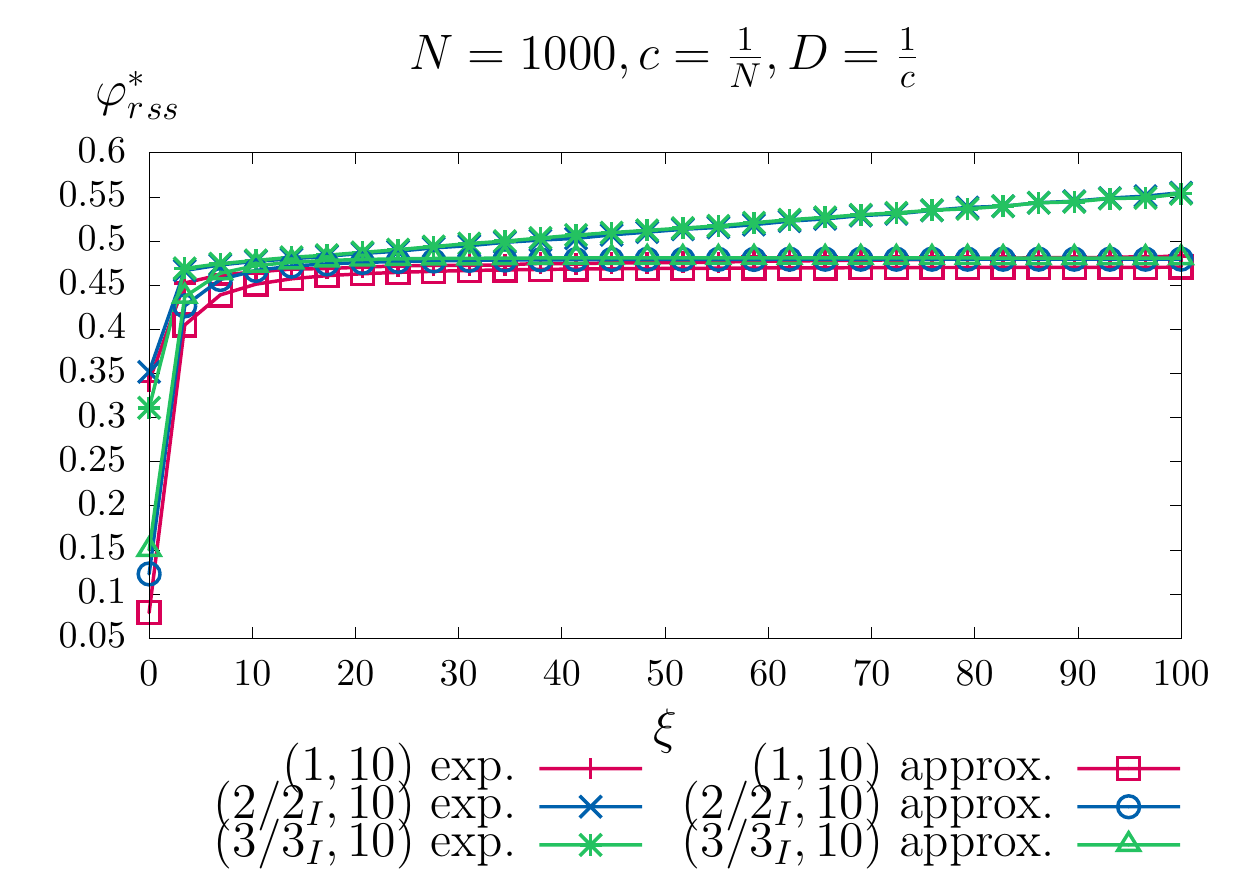}&
    \includegraphics[width=0.35\textwidth]{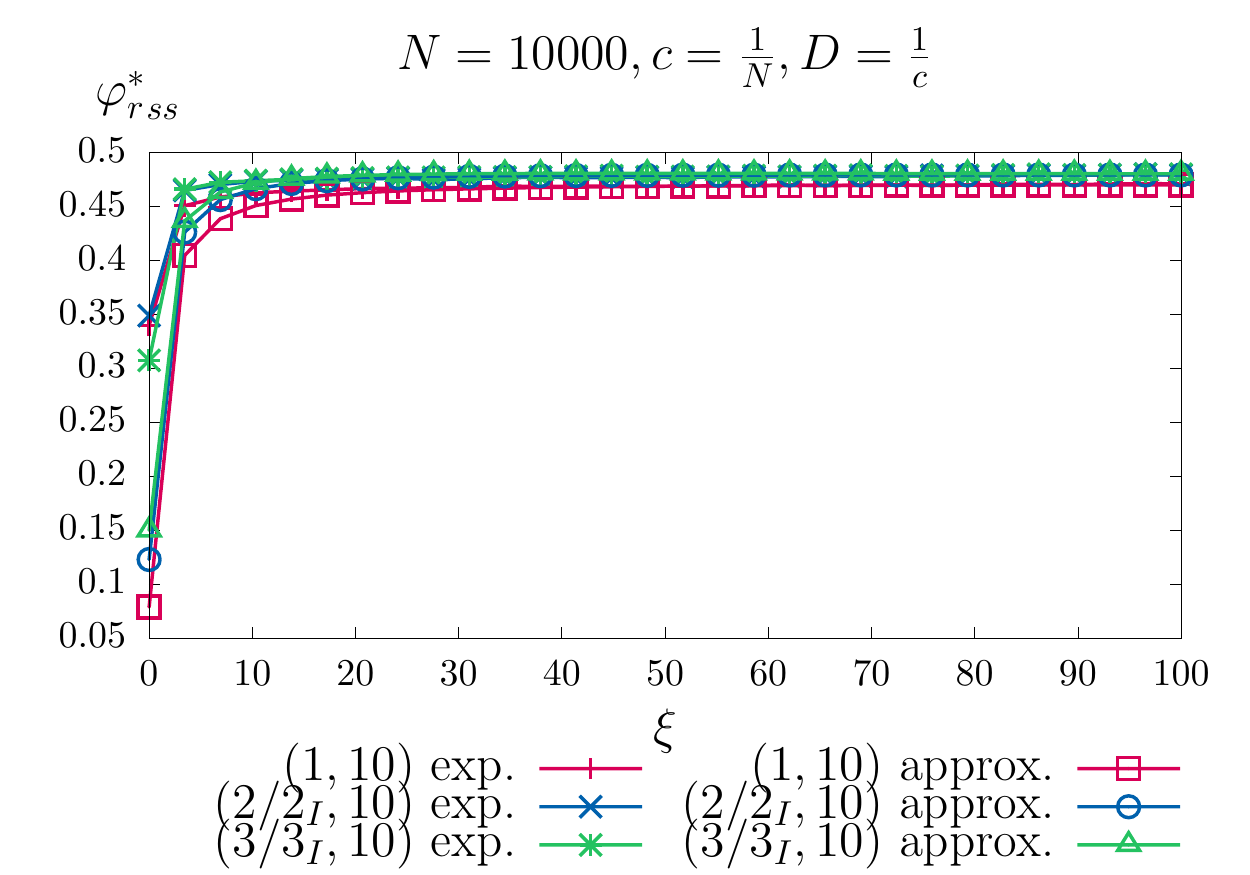}\\
    \includegraphics[width=0.35\textwidth]{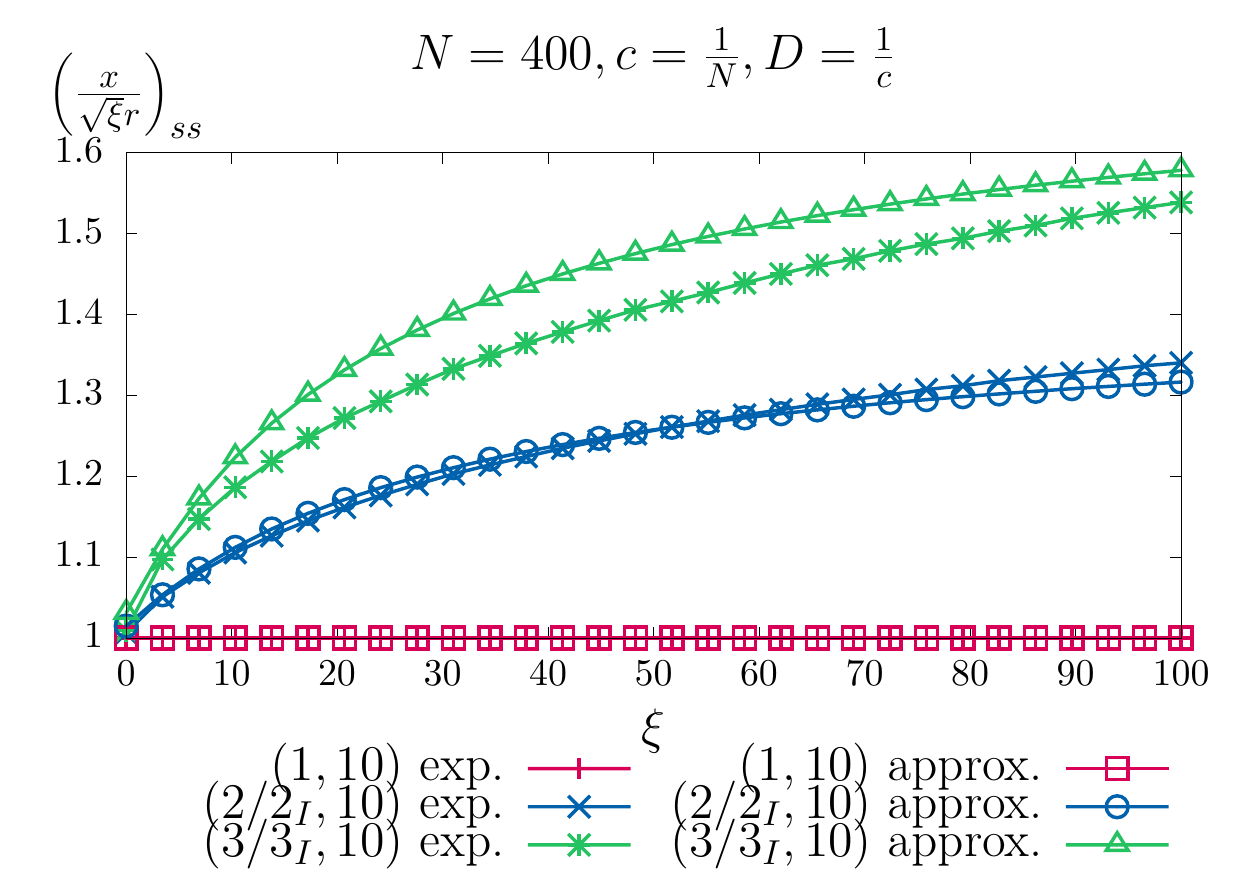}&
    \includegraphics[width=0.35\textwidth]{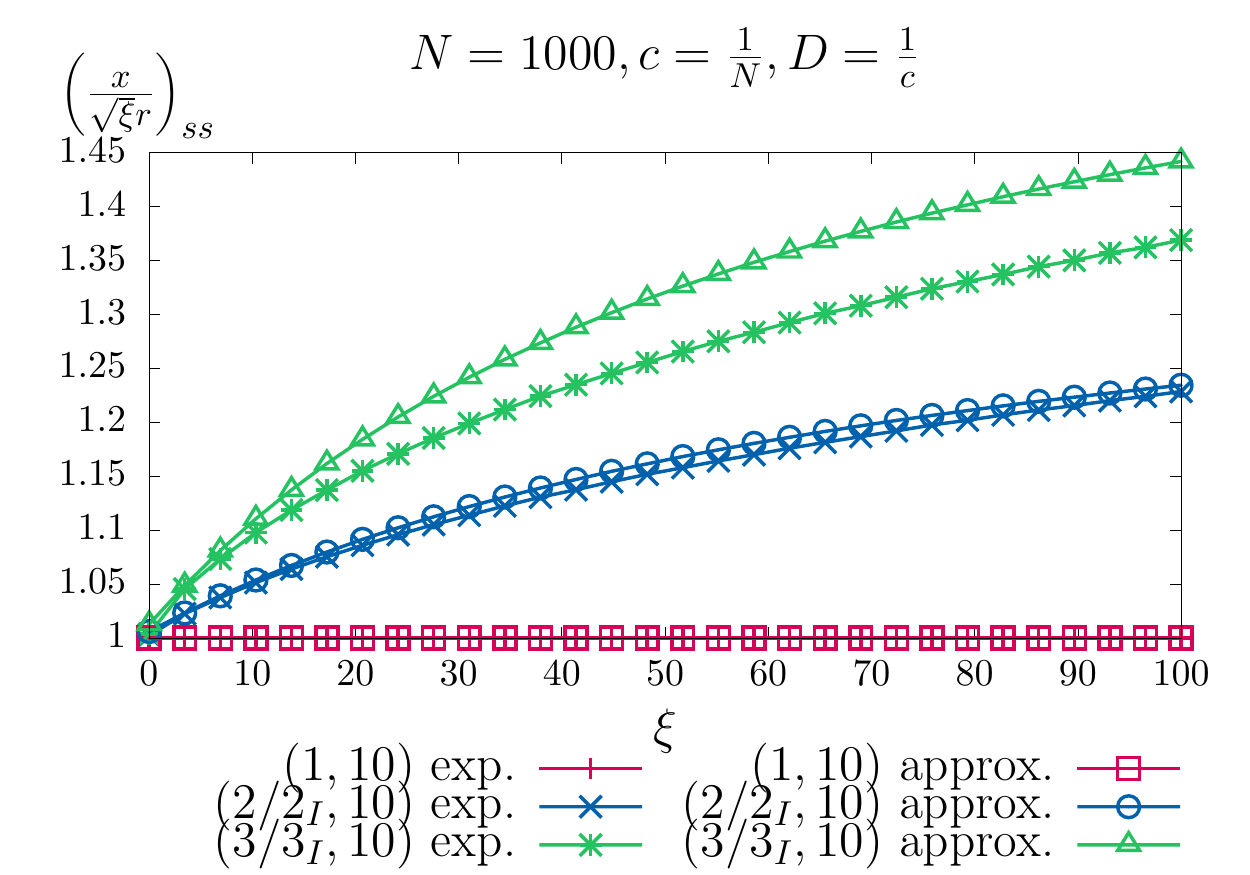}&
    \includegraphics[width=0.35\textwidth]{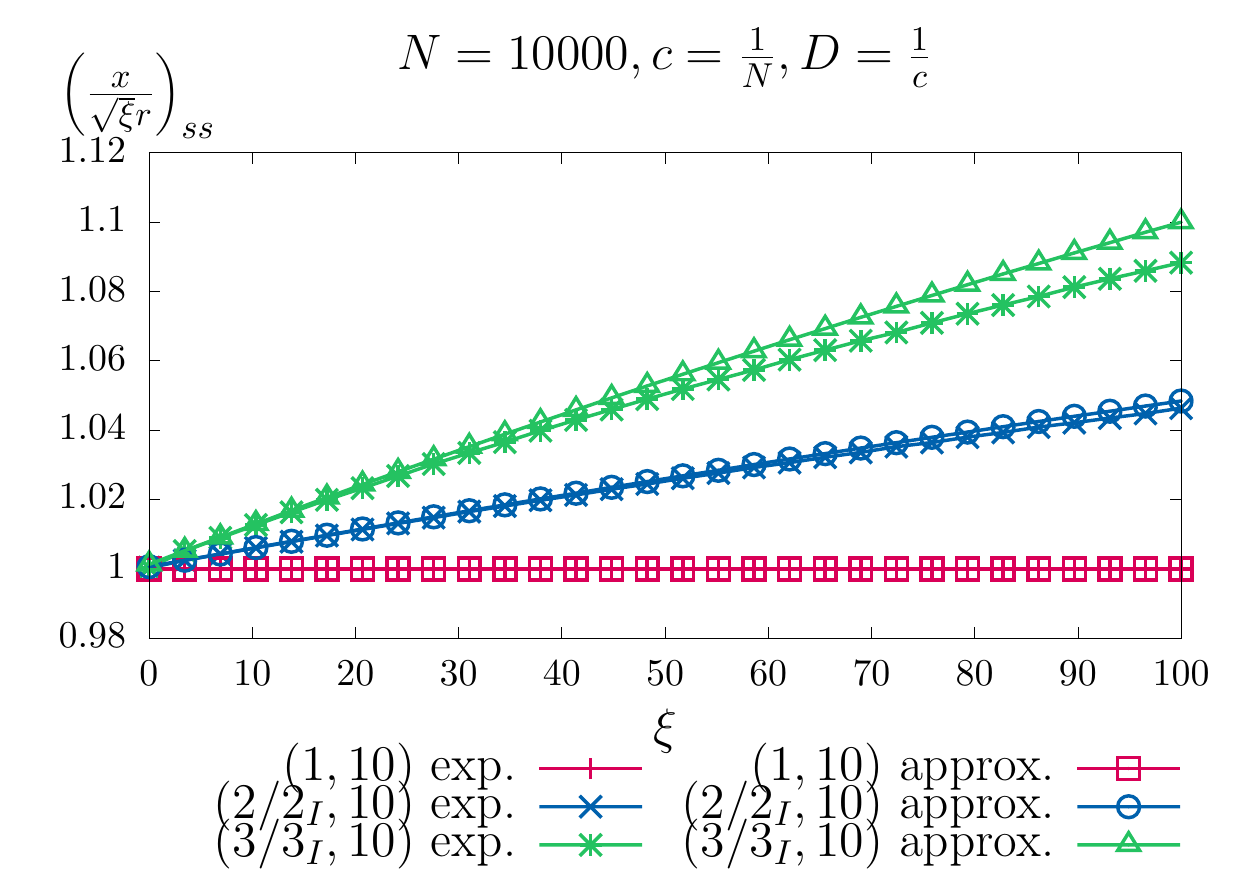}
  \end{tabular}
  \caption[Steady state closed-form approximation and real-run
  comparison of the $(\mu/\mu_I,\lambda)$-CSA-ES
  with repair by projection
  applied to the conically constrained problem.]
  {Steady state closed-form approximation and real-run
  comparison of the $(\mu/\mu_I,\lambda)$-CSA-ES
  with repair by projection
  applied to the conically constrained problem.
  \sppatemp{}}
  \label{sec:theoreticalanalysis:fig:steadystatecomparisonclosedform1divN}
\end{figure}

Neglecting terms already in
\Cref{sec:theoreticalanalysis:eq:sigmass1divNequation4}
is another approach to arrive at a simpler approximate form. Neglecting
${\sigma_{ss}^*}^{-1}\left(\frac{\mu c_{\mu/\mu,\lambda}}{\sqrt{1+\xi}}\right)$
in
\Cref{sec:theoreticalanalysis:eq:sigmass1divNequation4},
results in
\begin{equation}
  \begin{multlined}
    {\sigma_{ss}^*}^2\left(\frac{2+\xi}{(1 + \xi)^2 4 \mu}\right)
    +{\sigma_{ss}^*}\left(\frac{-c_{\mu/\mu,\lambda}}{\sqrt{1+\xi}(1+\xi)}
    \right)
    +\left(\frac{\xi-2\mu \xi c_{\mu/\mu,\lambda}^2}
    {2(1+\xi)}\right)
    =
    0.
  \end{multlined}
  \label{sec:theoreticalanalysis:eq:sigmass1divNequation5}
\end{equation}
Solving
\Cref{sec:theoreticalanalysis:eq:sigmass1divNequation5}
for the positive root yields
\begin{equation}
  \begin{multlined}
    {\sigma_{ss}^*} \approx
    \frac{
    \frac{c_{\mu/\mu,\lambda}}{\sqrt{1+\xi}(1+\xi)}
    +\sqrt{\left(\frac{-c_{\mu/\mu,\lambda}}{\sqrt{1+\xi}(1+\xi)}\right)^2 -
      4\left(\frac{2+\xi}{(1 + \xi)^2 4 \mu}\right)
      \left(\frac{\xi-2\mu \xi c_{\mu/\mu,\lambda}^2}
           {2(1+\xi)}\right)}}
    {2\left(\frac{2+\xi}{(1 + \xi)^2 4 \mu}\right)}.
  \end{multlined}
  \label{sec:theoreticalanalysis:eq:sigmass1divNequationapproxsolution1}
\end{equation}
\Cref{sec:theoreticalanalysis:fig:steadystatecomparisonclosedform1divN1}
shows plots of the steady state computations. Results computed by
\Cref{sec:theoreticalanalysis:eq:sigmass1divNequationapproxsolution1}
have been compared to real ES runs.
\renewcommand{\sppatemp}{The values for the points denoting the approximations
have been determined by computing the normalized steady state
mutation strength $\sigma_{ss}^*$ using
\Cref{sec:theoreticalanalysis:eq:sigmass1divNequationapproxsolution1}
for different values of $\xi$. The results for
$\varphi_x^*$ and $\varphi_r^*$ have been determined by using the
computed steady state $\sigma_{ss}^*$ values with
\Cref{sec:theoreticalanalysis:eq:varphixnormalizedss4}.
The approximations for $\left(\frac{x}{\sqrt{\xi}r}\right)_{ss}$ have been
determined by evaluating
\Cref{sec:theoreticalanalysis:eq:steadystatedist2}.
The values for the points denoting the experiments have been determined by
computing the averages of the particular values in real ES
runs.}\sppatemp{}

\begin{figure}
  \centering
  \begin{tabular}{@{\hspace{-0.025\textwidth}}c@{\hspace{-0.025\textwidth}}c@{\hspace{-0.025\textwidth}}c}
    \includegraphics[width=0.35\textwidth]{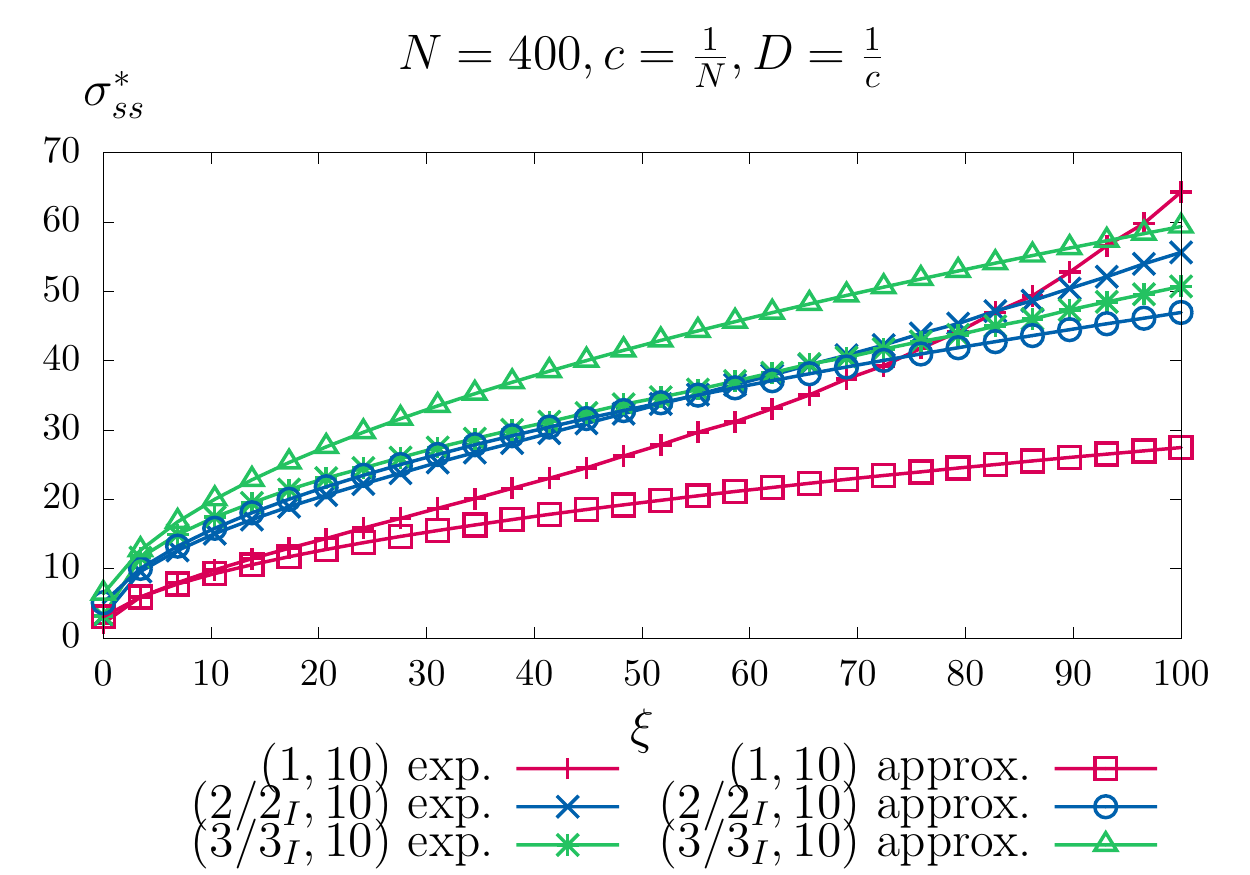}&
    \includegraphics[width=0.35\textwidth]{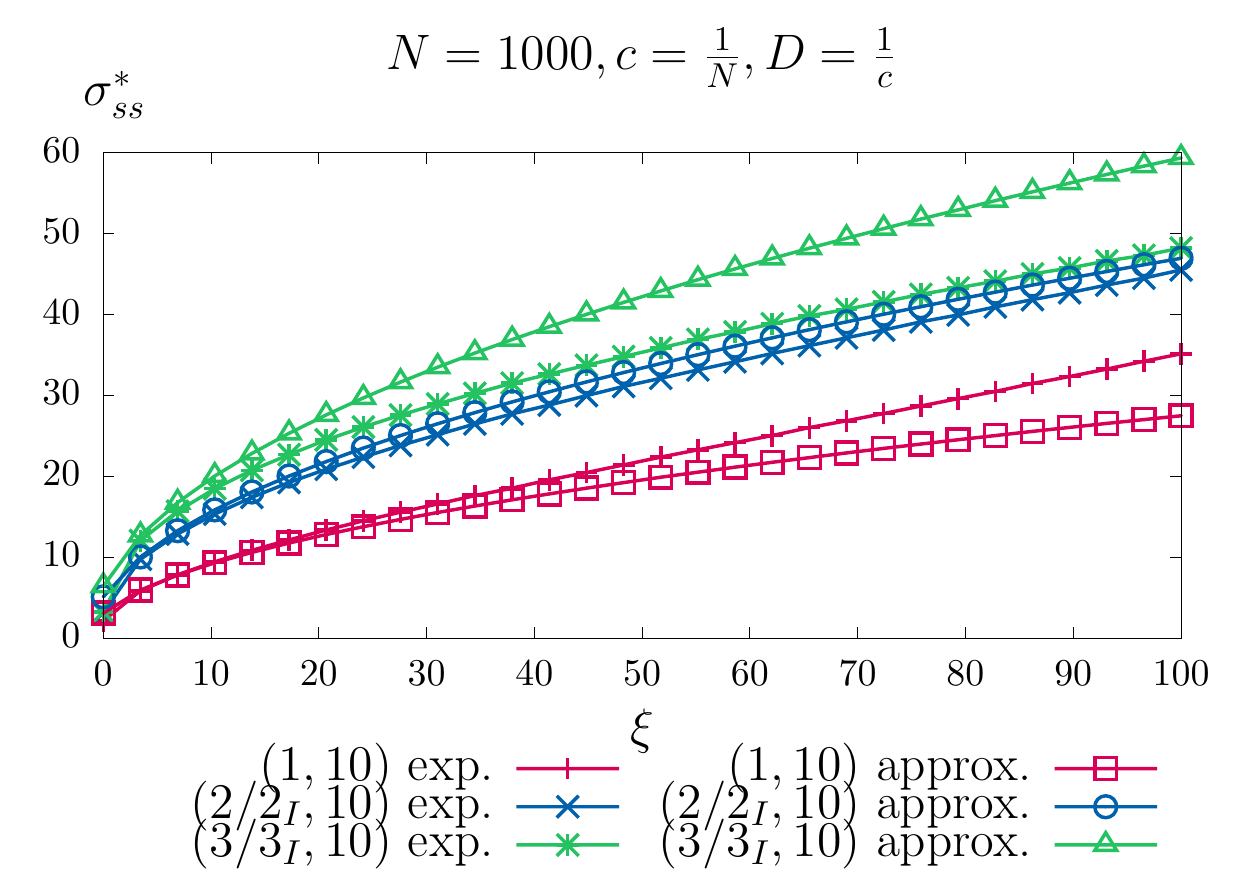}&
    \includegraphics[width=0.35\textwidth]{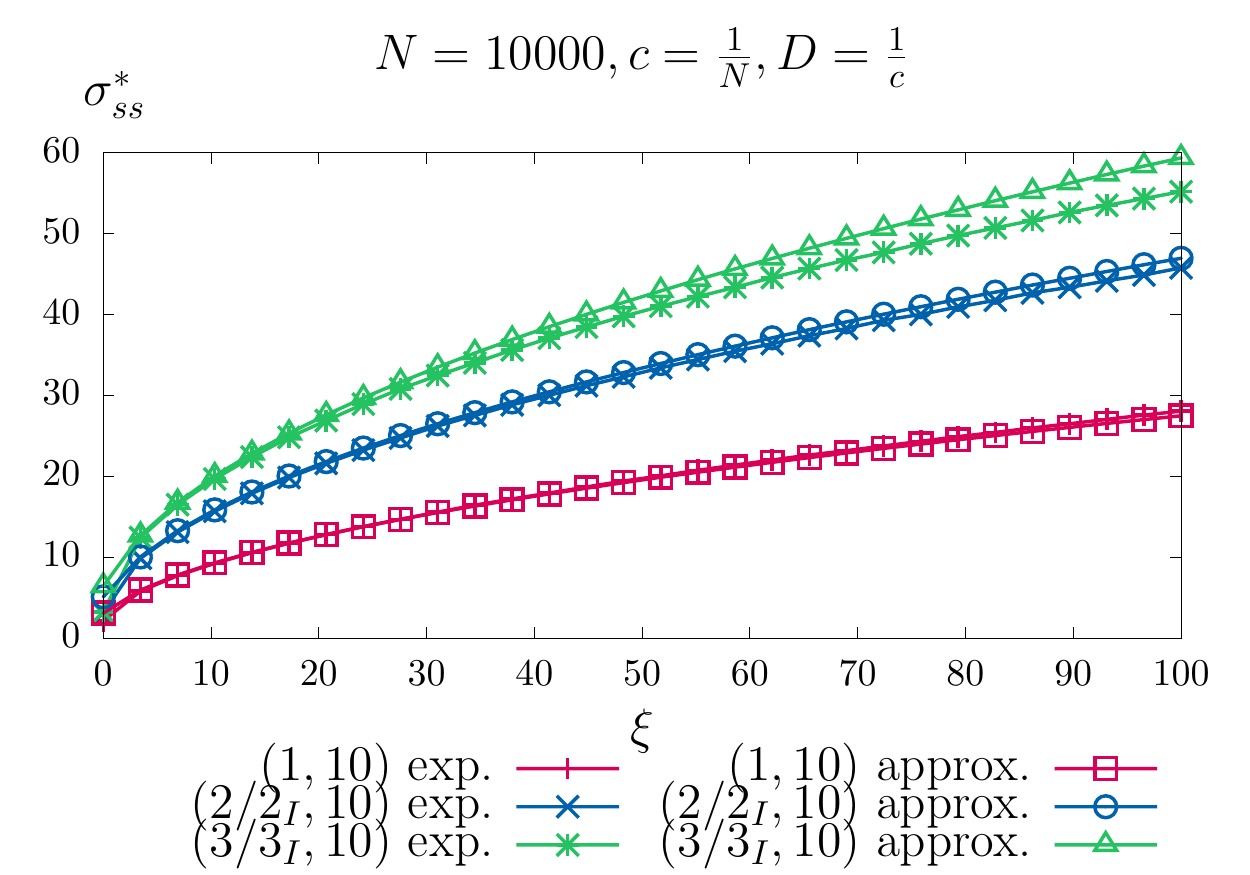}\\
    \includegraphics[width=0.35\textwidth]{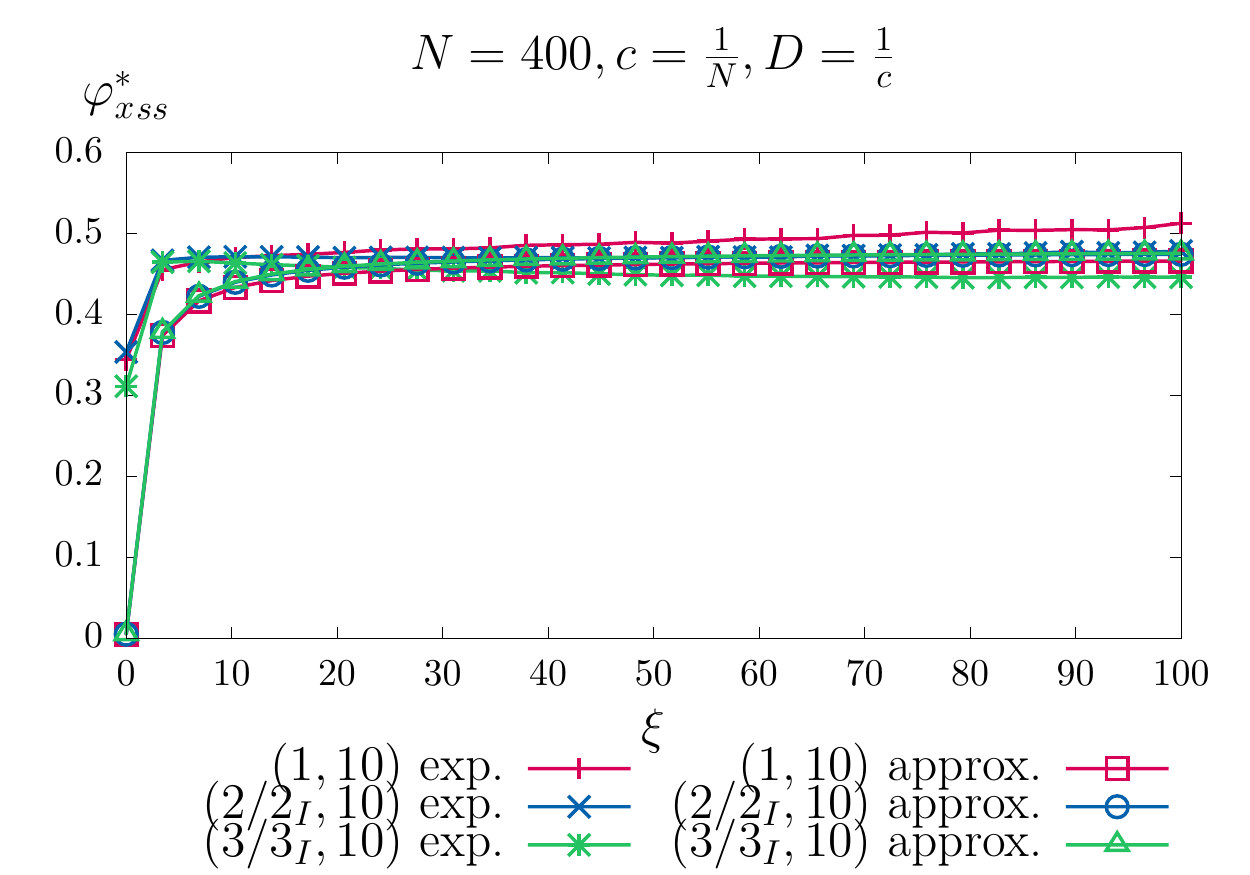}&
    \includegraphics[width=0.35\textwidth]{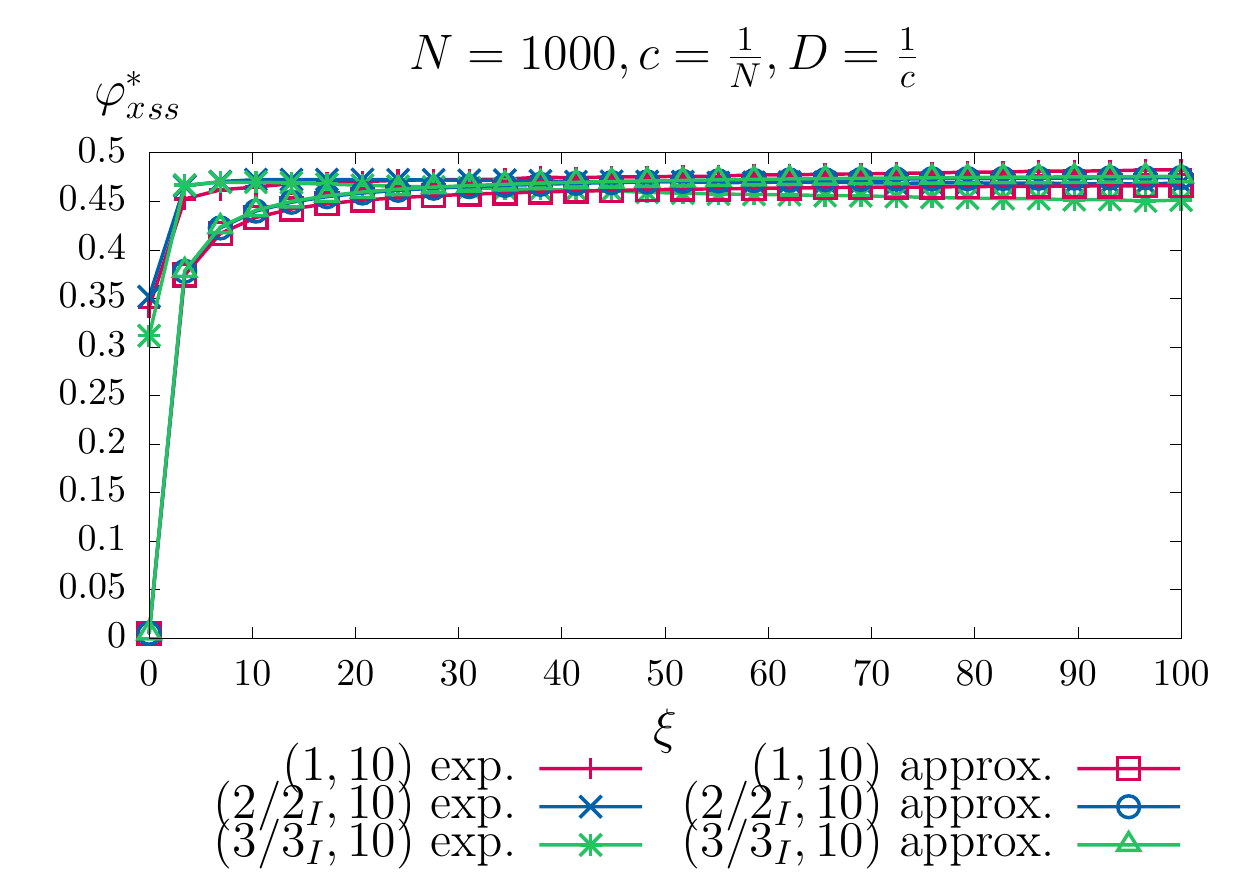}&
    \includegraphics[width=0.35\textwidth]{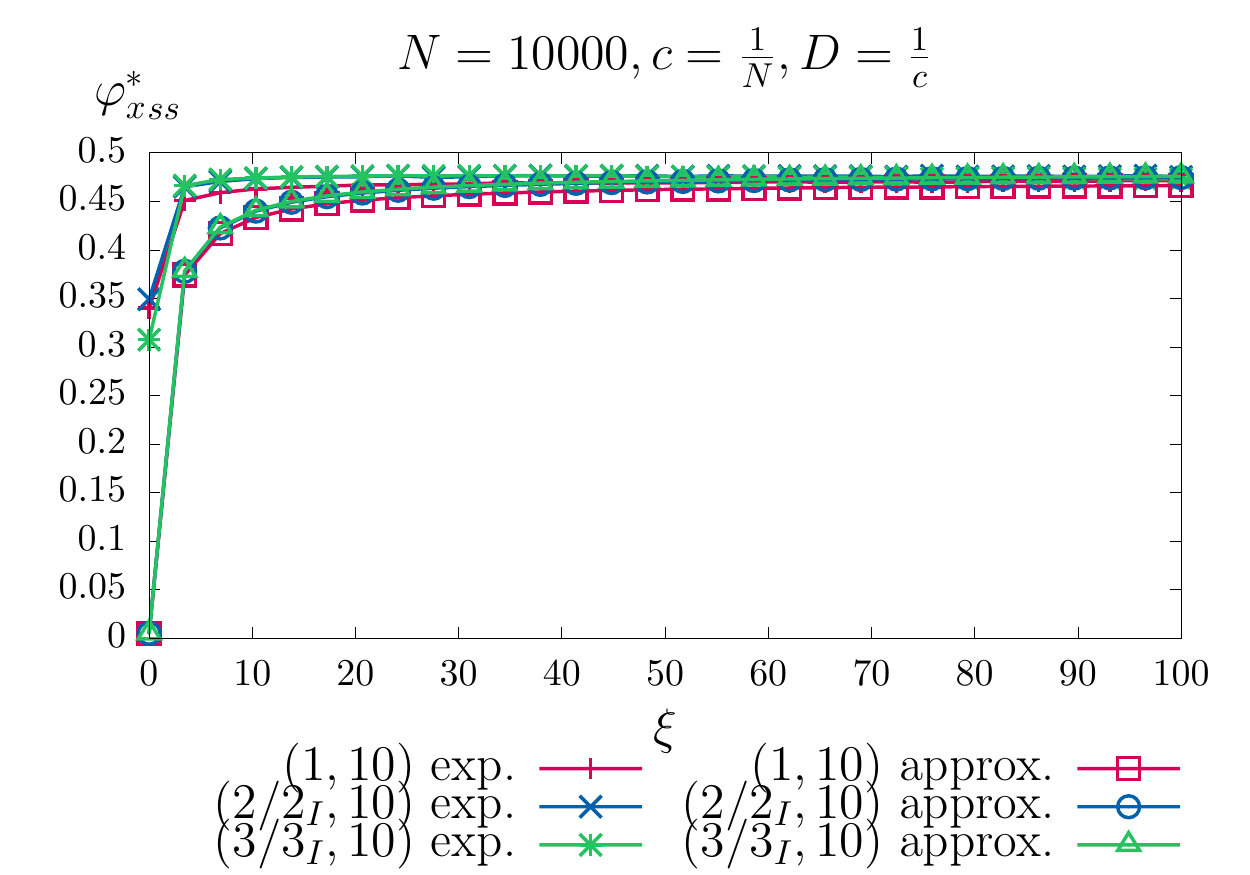}\\
    \includegraphics[width=0.35\textwidth]{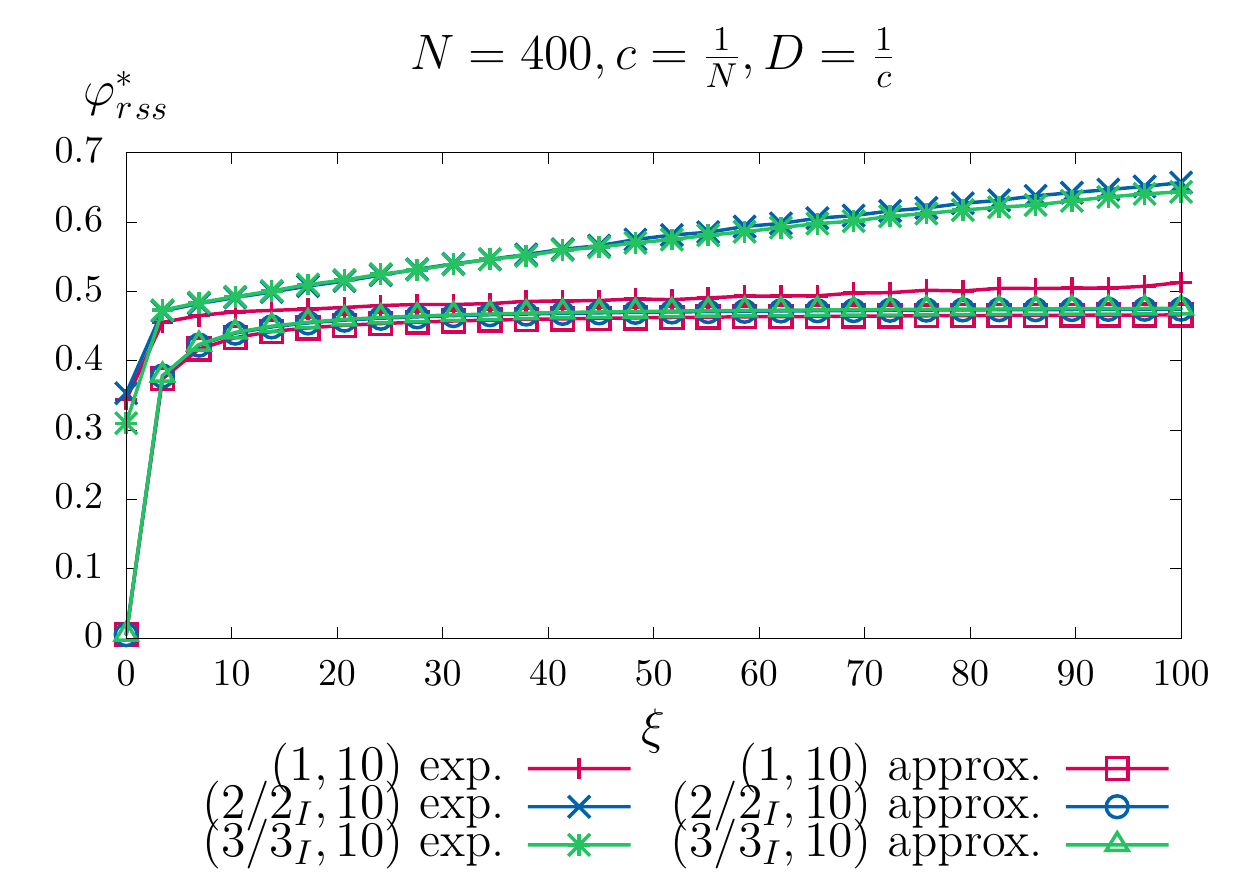}&
    \includegraphics[width=0.35\textwidth]{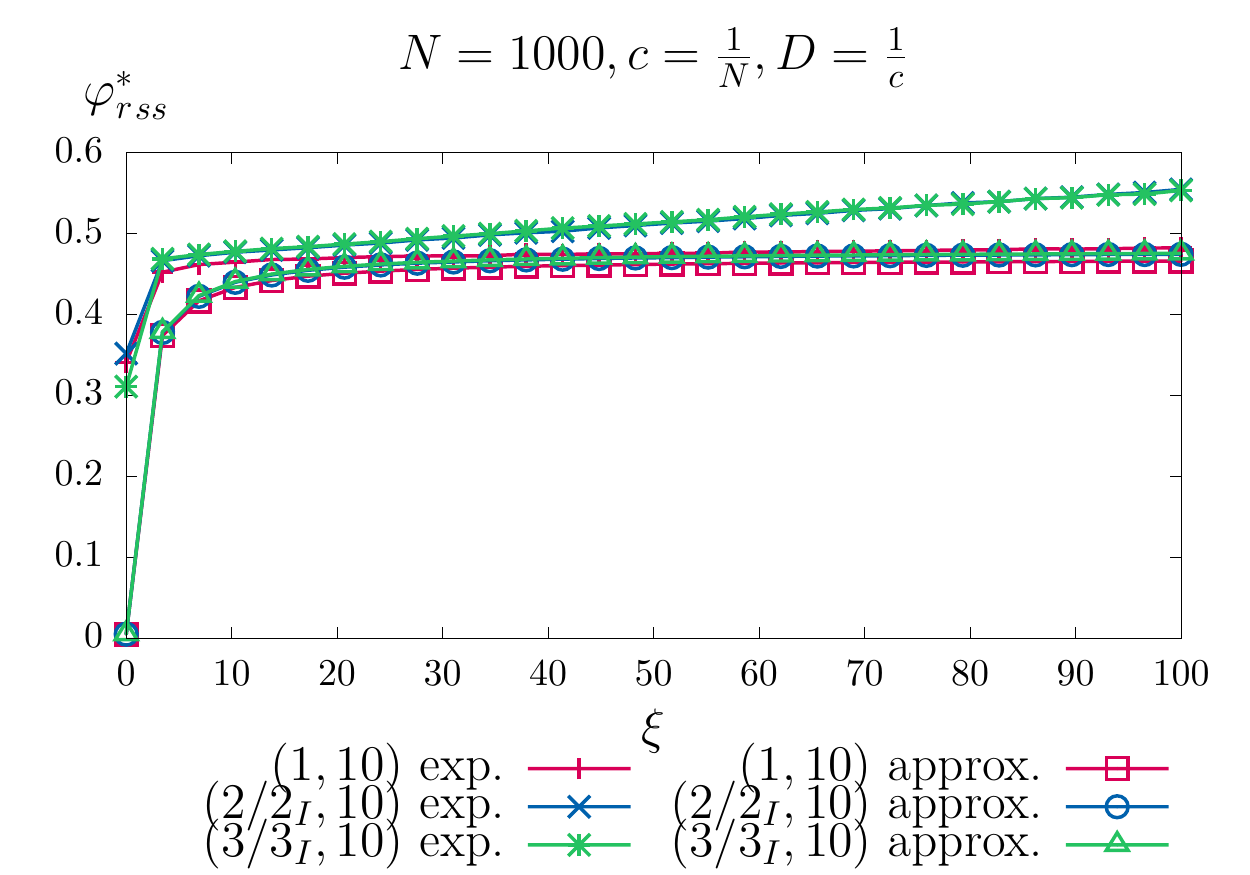}&
    \includegraphics[width=0.35\textwidth]{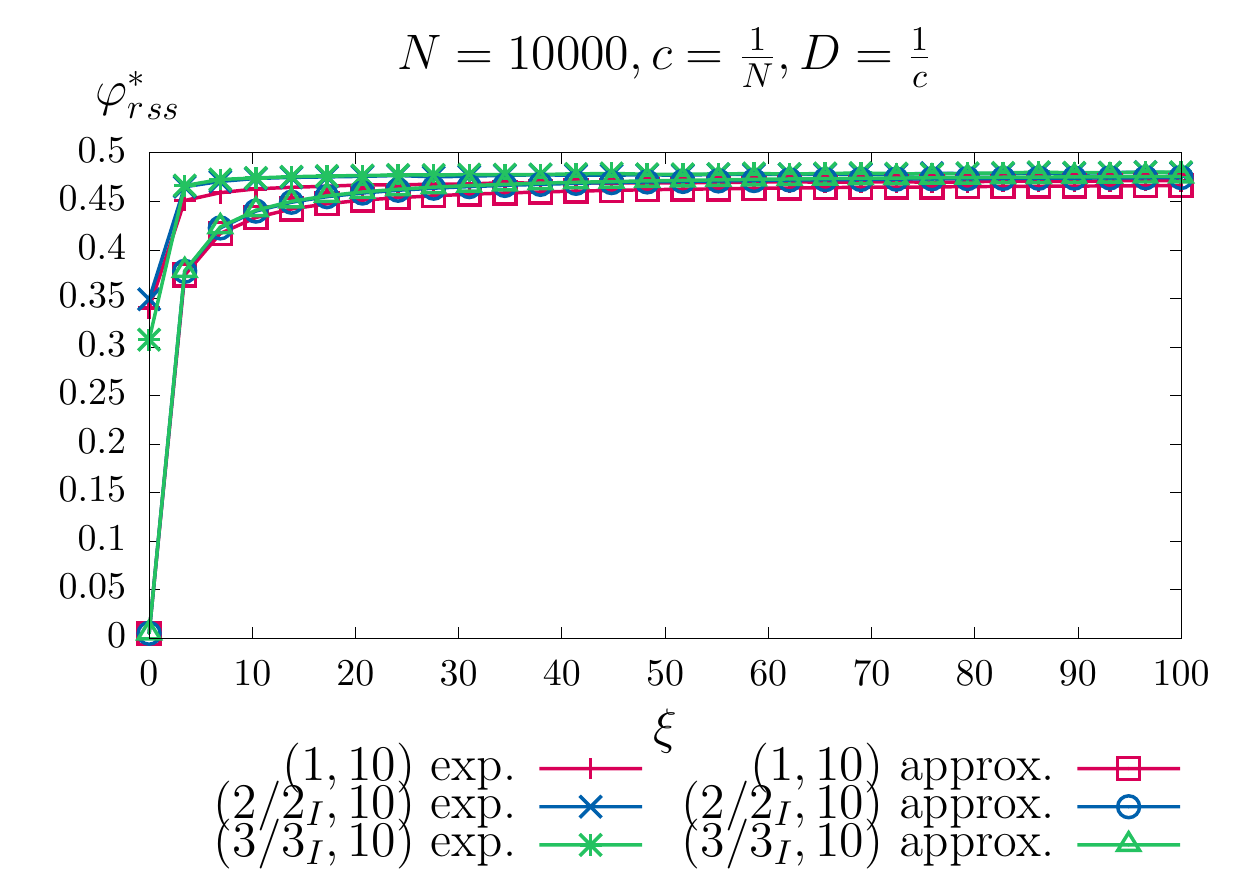}\\
    \includegraphics[width=0.35\textwidth]{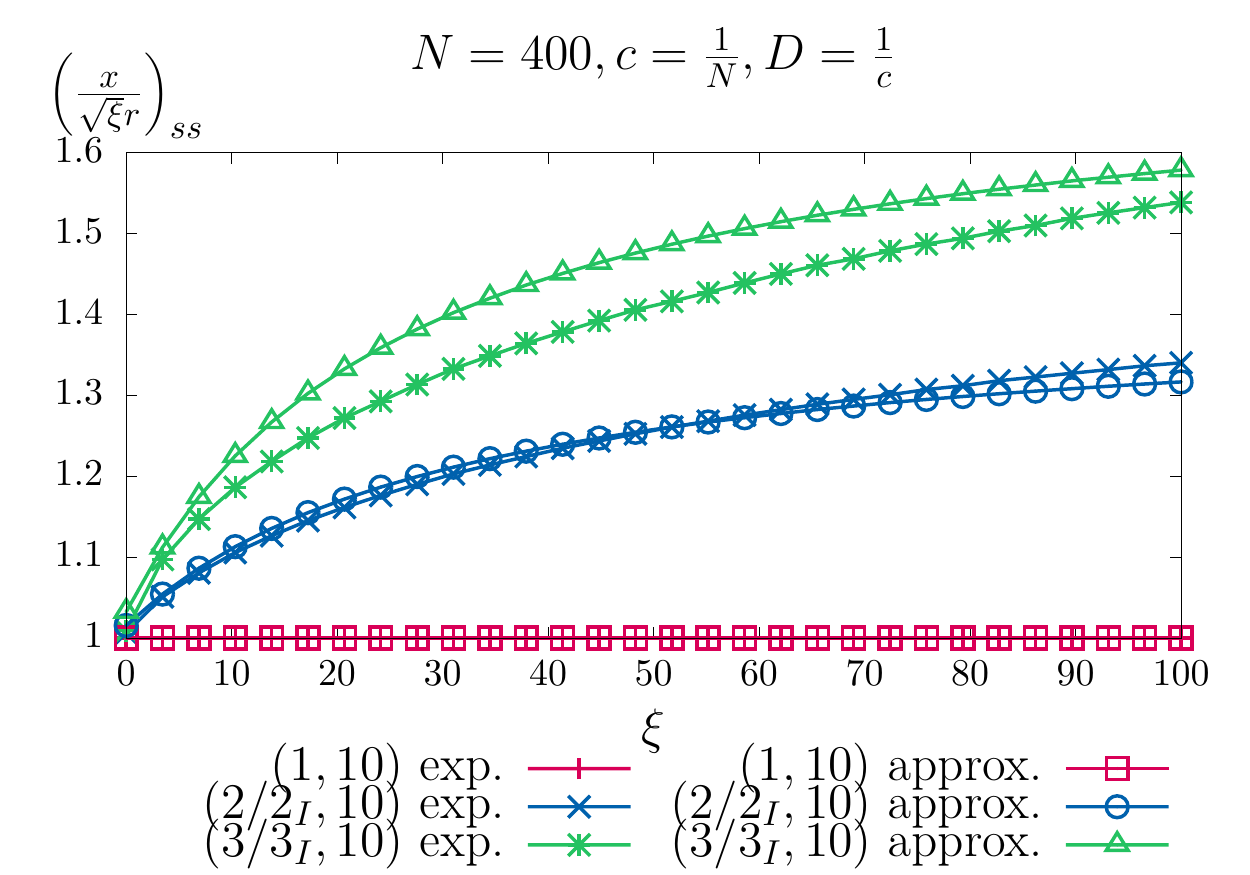}&
    \includegraphics[width=0.35\textwidth]{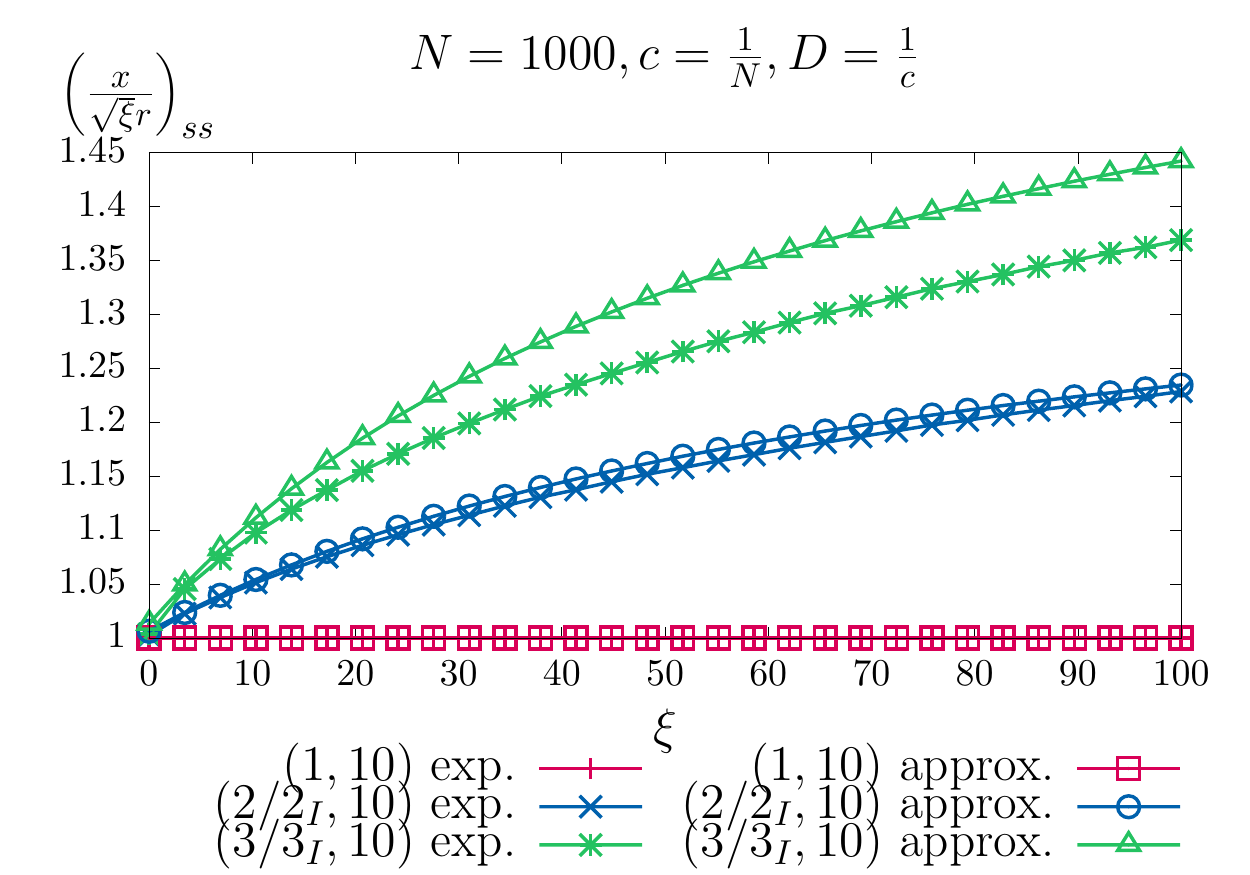}&
    \includegraphics[width=0.35\textwidth]{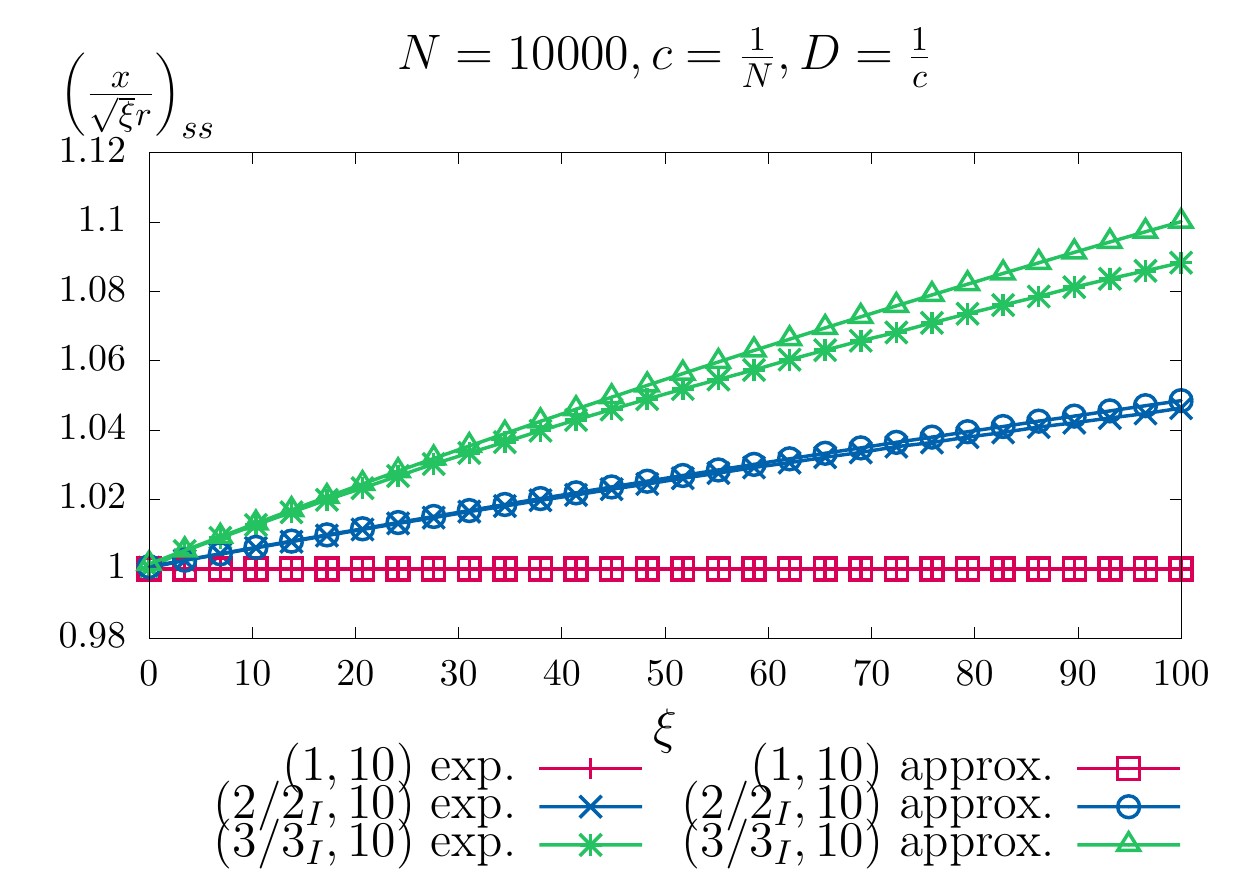}
  \end{tabular}
  \caption[Steady state closed-form approximation and real-run
  comparison of the $(\mu/\mu_I,\lambda)$-CSA-ES
  with repair by projection
  applied to the conically constrained problem.]
  {Steady state closed-form approximation and real-run
  comparison of the $(\mu/\mu_I,\lambda)$-CSA-ES
  with repair by projection
  applied to the conically constrained problem.
  \sppatemp{}}
  \label{sec:theoreticalanalysis:fig:steadystatecomparisonclosedform1divN1}
\end{figure}

\end{appendices}

\end{document}